\documentclass[a4paper,12pt,times,numbered,print,index]{Classes/PhDThesisPSnPDF}

\input{Preamble/preamble}

\title{Generative Models for Novelty Detection}

\subtitle{Applications in abnormal event and situational change detection from data series}

\author{S. Mahdyar Ravanbakhsh}

\dept{Dipartimento di Ingegneria \\navale, elettrica, elettronica e delle telecomunicazioni}

\university{University of Genova}
\crest{\includegraphics[width=0.2\textwidth]{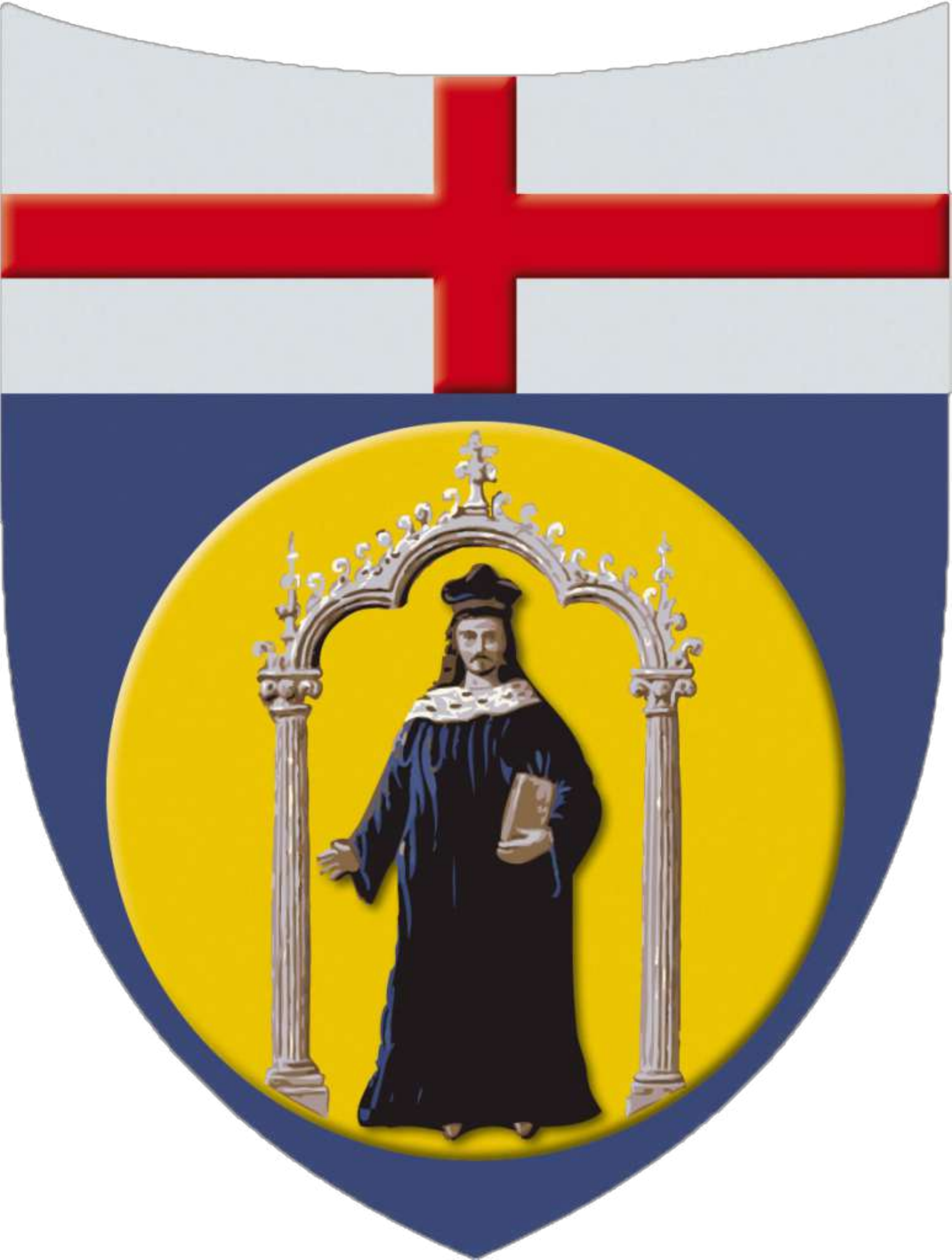}}

\degreetitle{Doctor of Philosophy}

\college{Scuola Politecnica}

\degreedate{February 2019} 

\subject{LaTeX} \keywords{{LaTeX} {PhD Thesis} {Engineering} {University of
Cambridge}}

\ifdefineAbstract
\fi

\ifdefineChapter
\fi

\begin{document}
\frontmatter

\begin{titlepage}
\newgeometry{top=0mm, bottom=0mm, left=-5mm} 
\includegraphics[]{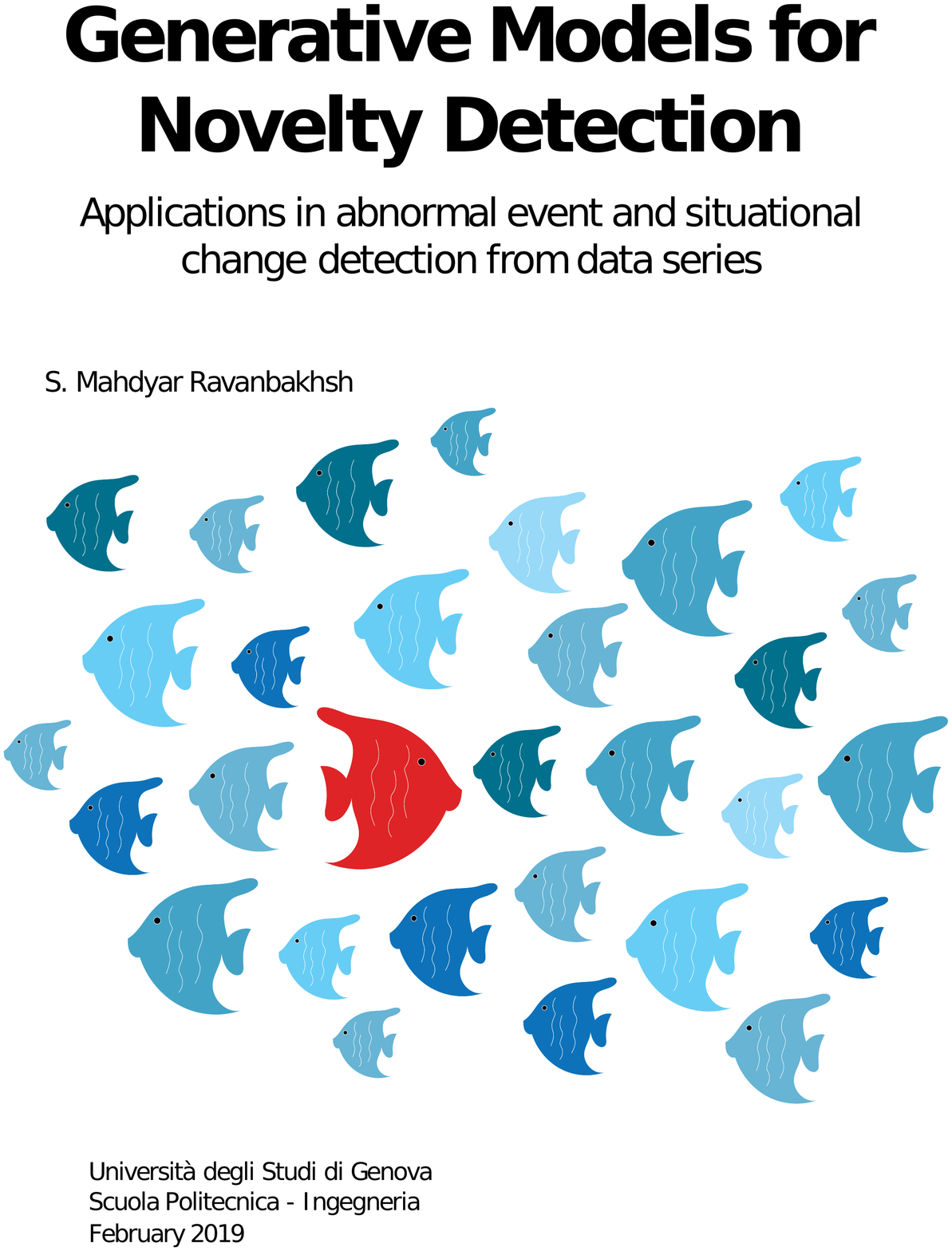}
\restoregeometry
\end{titlepage}

\maketitle


\begin{dedication} 

I would like to dedicate this thesis to my parents and teachers\\ \emph{(without whom nothing would have been possible)} 

\end{dedication}

\newpage
\thispagestyle{plain} 
\mbox{}
\newpage
\thispagestyle{plain} 
\begin{quote}
    \emph{``As far as the laws of mathematics refer to reality, they are not certain; and as far as they are certain, they do not refer to reality.''}\\
    
    -- Albert Einstein
    
    

    

    
\end{quote}

\begin{acknowledgements}      

First of all, I would like to thank my family (Farida, Hamid, Morteza); it is because of their never ending support that I have had the chance to progress in life. Their dedication to my education provided the foundation for my studies.\\ 

\noindent
I started working on computer vision at University of Ljubljana. For this, I am grateful to Professor Ale\v{s} Leonardis; for the past years he has had a critical role in conducting me with computer vision researches.\\

\noindent
I would like to express my deepest gratitude to my supervisors, Professor Carlo Regazzoni and Professor Lucio Marcenaro for providing me with an excellent environment for conducting research, and the continuous support of my Ph.D study and related research. Professor Regazzoni has always amazed me by his intuitive way of thinking and supervisory skills. He has been a great support for me and a nice person to talk to.\\

\noindent
I am especially indebted to Dr. Moin Nabi who has been a great mentor for me during my visit to the Machine Learning Research lab at SAP in Berlin. I am grateful for his support, advice, and friendship. He inspired me by the way he looks at research and life. I also express my sincere gratitude to Professor Nicu Sebe, Dr. Mohammad Rastegari, and Dr. Hossein Mousavi for sharing their experiences and supporting me with their advices. With no doubt, without their support and guidance, I could not come this far.\\

\noindent
I thank my fellow labmates in for the stimulating discussions, for the sleepless nights we were working together before deadlines, and for all the fun we have had in the last three years. In particular, I thank my friends Mohamad Baydoun, Damian Campo, Vahid Bastani, Oscar Urizar, Sebastian Olier, Andrea Toma, and Divya Thekke Kanapram. Also I thank my fellow in the Carlos III University Madrid, Professor David Mart\'{i}n, and Pablo Mar\'{i}n.\\

\thispagestyle{empty}
\noindent
Loving thanks to my friends / colleagues and learning partners in the SAP Berlin ML research lab and the University of Trento. In particular, I offer special thanks to Dr. Tassilo Klein and Dr. Enver Sangineto, who willingly accompanying me on the important parts of this journey of exploration.
\\

\noindent
Last but not least, I’d like to acknowledge funders who supported my PhD studies. The works in this thesis were funded by ISIP40 lab at University of Genova.
\thispagestyle{empty}

\end{acknowledgements}

\begin{abstract}
Novelty detection is a process for distinguishing the observations that differ in some respect from the observations that the model is trained on. Novelty detection is one of the fundamental requirements of a good classification or identification system since sometimes the test data contains observations that were not known at the training time. In other words, the novelty class is often is not presented during the training phase or not well defined.\\

In light of the above, one-class classifiers and generative methods can efficiently model such problems. However, due to the unavailability of data from the novelty class, training an end-to-end model is a challenging task itself. Therefore, detecting the Novel classes in unsupervised and semi-supervised settings is a crucial step in such tasks. \\

In this thesis, we propose several methods to model the novelty detection problem in unsupervised and semi-supervised fashion. The proposed frameworks applied to different related applications of anomaly and outlier detection tasks. The results show the superior of our proposed
methods in compare to the baselines and state-of-the-art methods.

\end{abstract}


\tableofcontents

\listoffigures

\listoftables


\printnomenclature

\mainmatter

\graphicspath{{Chapter1/Figs/}}
\chapter{Introduction}  
\begin{quote}
    
    \emph{``Wonder\footnote{\textbf{wonder} \textipa{/'w2nd@/} (\emph{noun}) a feeling of amazement and admiration, caused by something beautiful, remarkable, or unfamiliar. (\emph{verb}) desire to know something. feel doubt ; Def. Oxford Living Dictionaries.} is the seed of knowledge.''}\\ 
    
    -- Francis Bacon

    
    
    

    
\end{quote}
\newpage
\newpage
\thispagestyle{plain} 
\mbox{}
\newpage

The understanding of surroundings and the capability of dynamically and constantly learn the future (unseen) events are ideas that have been in the minds of people during decades \cite{schmidhuber1987evolutionary,1996incremental}. The possibility of detecting the novel situations/distributions of entities/data in a given environment is an essential component for improving the situational understandings of the the system or preventing undesired situations. A dynamic constant learning system should be able to iteratively observe what the consequences of their actions were in the outside world and adapt/learn themselves dynamically according to a given purpose to be accomplished. 
    
The knowledge of the system for being aware of its capabilities and limitation potentially provides the system with the possibility of adapting its decisions in a more appropriated way. This could be done by redefining its own models, learning new models, and selecting different actions for accomplishing a determined task. In this sense, for modeling such dynamic systems as proposed, it is necessary to learn diverse models that relate the different situations of the system. In order to learn diverse models in an unsupervised fashion, the system first needs to detect the deviated situations itself. This process usually referred as \emph{Novelty detection}.

\begin{figure}[h!]
	\begin{center}
		\includegraphics[width=0.8\linewidth]{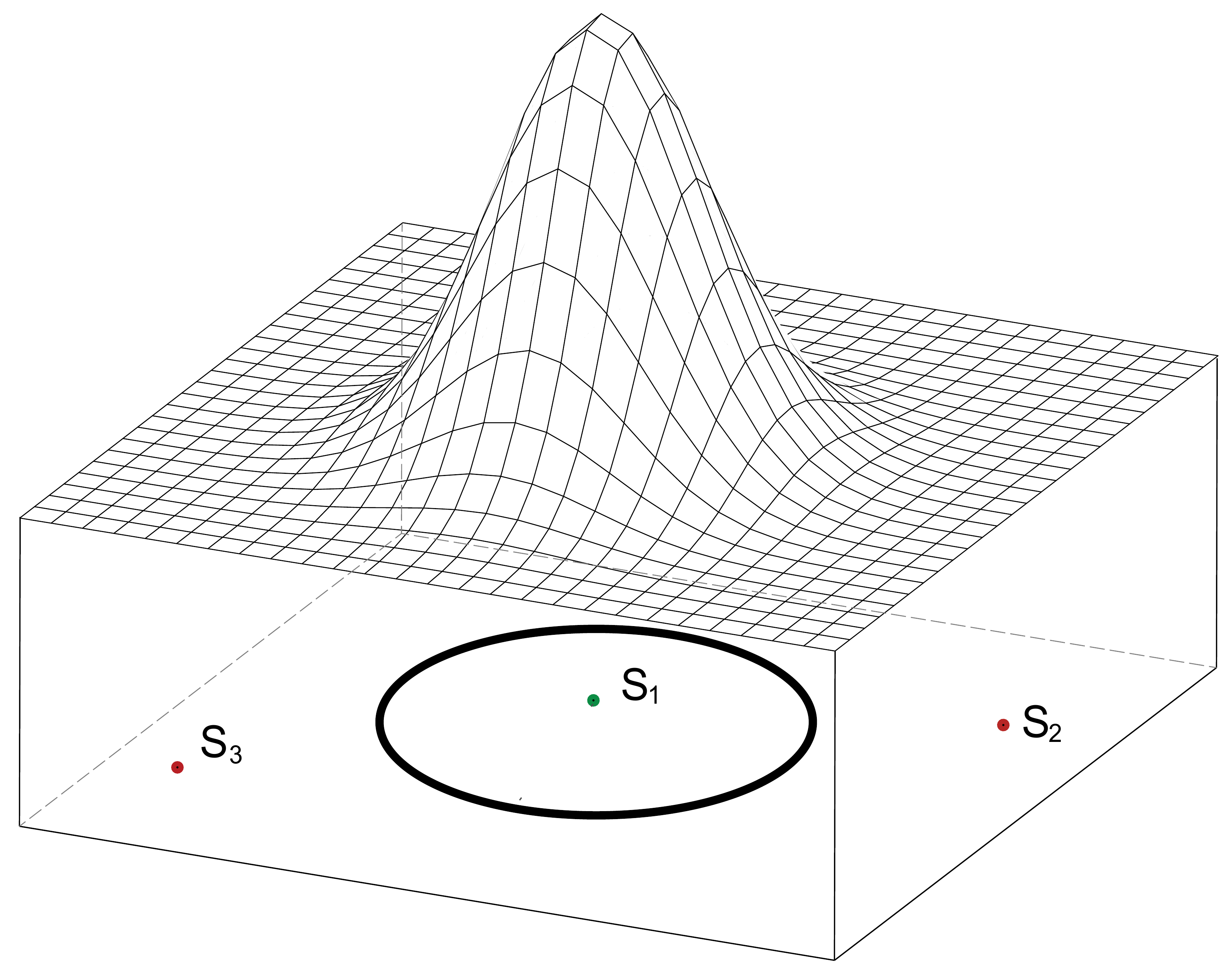}
	\end{center}
	\caption[Visualization of inlier and outlier samples with respect to the learned distribution by a classifier]{Visualization of inlier and outlier samples with respect to the learned distribution by a classifier: As can be seen, $S_1$ enhanced the inlier sample since it placed inside the classifier decision boundary, where $S_2$ and $S_3$ indicates the out-of-distribution samples (outliers).}
	\label{fig:motiv}
\end{figure}

Novelty detection is the process of identifying the new or unexplained set of data to decide whether they are within a previous learned distribution (i.e., inlier in Fig. \ref{fig:motiv}) or outside of it (i.e., outliers in Fig. \ref{fig:motiv}). The \emph{"Novelty"} term is commonly used to refer to the unusual, unseen new observations that do not observed before or is simply different from the other previous observations. Such problems are especially of great interest in computer science studies, as they are very related to outlier detection, anomaly detection tasks, and change detection problems.
    
The purpose of this thesis is to introduce learning models based on the essential knowledge of detecting outliers with respect to learned distributions. Such capability can possibly empower the machine learning strategies to increase the awareness of systems. 

For that end, in this thesis we propose approaches for detecting the abnormal event in unsupervised and semi-supervised fashion. We propose to use Generative Adversarial Nets (GANs) framework, which are trained using {\em normal} samples in order to learn an internal representation of the scene {\em normality}. Since our GANs are trained with only normal data, they are not able to generate abnormal events. At the testing time the real data are compared with both the appearance and the motion representations reconstructed by our GANs and abnormal areas are detected by computing local differences.

Furthermore, we show the capability of learning more complex situations with the proposed approach. The proposed novelty detection-based method potentially can also proved the means of a long-life learning cycle. We propose to create models that define the typical dynamics in a given scene in a semantic way, such that a context and a situation is included in the system's modeling. By applying this way of thinking, it has been proposed a methodology to modeling complex scene situations by learning small simple scene/situations. During the time, the system would be able to learn unseen situation through its situational changes detection capability. The proposed approach not only improves predictions of future events but it can be potentially used for transferring the learned knowledge to other systems. In this context, the situational information related to the environment and contexts perceived by an individual can be moved and interpreted by another body.

\section{Main Contributions}
To form a comprehensive analysis, we study different aspects of novelty detection each evaluated in an aforementioned tasks. For an applicational investigation, we organize this thesis in two parts: \emph{(i)} abnormality detection for crowds, and \emph{(ii)} situational change detection for autonomous agents.

\begin{itemize}
    \item In part \emph{(i)}, we focus on employing the novelty detection methods for abnormality detection in the crowds problem. After an extensive systematic review on state-of-the-art models, we study the outcomes provided by employing our proposed models for the task of abnormality detection. We first investigate on conventional methods
and introduce a tracklet-based approach to tackle this problem. We, next,
study utilizing and learning the deep models for abnormality detection. For this purpose, we have introduced a binary CNN model that encodes the visual semantic patterns into binary codes. Furthermore, we proposed a deep cross-modal generative model to learn the concepts by observing the motion maps and images on the normal situation. after training, our model is able to detect the outliers during the test time with respect to the learned distributions. We evaluated the proposed methods for crowds abnormality detection.

\item In part \emph{(ii)}, we explore the significance of utilizing deep generative model for a complex situational representation in order to detect the possible changes on the observed situation during the test time. For this purpose, we first focus on learning simplest normal situation, then complement it by automatically detect and learn the novel situations information in order to build a richer representation. We specifically introduce a multi-level cross-modal GAN-based representation to model the normal situations for an autonomous agent. This knowledge can be learned through the observing a human operator, while performing the same task. In this sense, any deviated situation from the previously learned models will be recognized as a novel situation (abnormality).
\end{itemize}

The contributions are well described in Chapters \ref{ch:abnormality} and \ref{ch:autonomous_car}, but the main contributions of this thesis is shortened below with a brief description.

\begin{enumerate}
    \item [1.] \textbf{The Binary Tracklets for abnormality detection in crowd:} We proposed an efficient method for crowd abnormal behavior detection and localization. The goal is to show that despite the significant improvements of deep-learning-based methods in this field, but still, they are not fully applicable for the real-time applications. We propose a simple yet effective descriptor based on binary tracklets, containing both orientation and magnitude information in a single feature.
    
     \item [2.] \textbf{The Plug-and-Play Binary CNNs for anomaly detection:} We show that keeping track of the changes in the CNN feature across time can be used to effectively detect local anomalies. Specifically, we propose to measure local abnormality by combining semantic information embedded into a set of binary representation (inherited from existing CNN models) with low-level optical-flow. One of the advantages of this method is that it can be used without the fine-tuning phase. The proposed method is validated on challenging abnormality detection datasets and the results show the superiority of our approach compared with the state-of-the-art methods.
     
     \item [3.] \textbf{Detecting Abnormality with Deep Generative Models:} We propose to use Generative Adversarial Nets (GANs), which are trained using {\em normal} frames and corresponding optical-flow images in order to learn an internal representation of the scene {\em normality}. Since our GANs are trained with only normal data, they are not able to generate abnormal events. At testing time the real data are compared with both the appearance and the motion representations reconstructed by our GANs and abnormal areas are detected by computing local differences.
     
     \item [4.] \textbf{Situational Change Detection for Autonomous Cars:} We introduced a hierarchical model by means of a cross-modal Generative Adversarial Networks (GANs) structured into a several levels. The different levels of the GANs are representing different distributions and constructed by detecting the changes (novel situations) with respect to the previous learned levels in a self-supervised manner. The GANs discriminators decision boundaries used to detect the novel situations. Besides, in this work a dynamic incremental self-awareness (SA) models are proposed that allow experiences done by an agent to be modeled in a hierarchical fashion, starting from more simple situations to more structured ones. Each situation is learned from subsets of private agent perception data as a model capable to predict normal behaviors and detect the novel situations.
     
\end{enumerate}

\subsection{PhD publications}
The contents of this thesis are based on several peer-reviewed papers published during my PhD studies together with some works presented in workshops and partial results of projects that are still active that will be summited later on to review processes in order to be published. Among the published papers the most relevant ones to this thesis are listed as follows. 

\begin{itemize}
    \item "Training Adversarial Discriminators for Cross-channel Abnormal Event Detection in Crowds", M. Ravanbakhsh, M. Nabi, E. Sangineto, N. Sebe, IEEE Winter Conference on Applications of Computer Vision (WACV 2019), Hawaii, USA (2019).
    
    \item "Fast but Not Deep: Efficient Crowd Abnormality Detection with Local Binary Tracklets", M. Ravanbakhsh, H. Mousavi, M. Nabi, L. Marcenaro, C. Regazzoni, IEEE International Conference on Advanced Video and Signal-based Surveillance (AVSS 2018), Auckland, New Zealand (2018).

    \item "Hierarchy of GANs for learning embodied self-awareness model", M. Ravanbakhsh, M. Baydoun, D. Campo, P. Marin, D. Martin, L. Marcenaro, C. S. Regazzoni, IEEE International Conference on Image Processing (ICIP 2018), Athens, Greece (2018).
    
    \item "Plug-and-play cnn for crowd motion analysis: An application in abnormal event detection", M.  Ravanbakhsh, M. Nabi, H. Mousavi, E. Sangineto, N. Sebe, IEEE Winter Conference on Applications of Computer Vision (WACV 2018), Lake Tahoe, USA (2018).
    
    \item "Learning Multi-Modal Self-Awareness Models for Autonomous Vehicles from Human Driving",  M. Ravanbakhsh, M. Baydoun, D. Campo, P. Marin, D. Martin, L. Marcenaro, C. S. Regazzoni, International Conference on Information Fusion (FUSION 2018), Cambridge, UK (2018).
    
    \item "A multi-perspective approach to anomaly detection for self-aware embodied agents", M. Baydoun, M. Ravanbakhsh, D. Campo, P. Marin, D. Martin, L. Marcenaro, A. Cavallaro, C. S. Regazzoni,  IEEE International Conference on Acoustics, Speech and Signal Processing (ICASSP 2018), Calgary, Alberta, Canada (2018)
    
    \item "Detection and localization of crowd behavior using a novel tracklet-based model", Rabiee, H., Mousavi, H., Nabi, M. and Ravanbakhsh, M., International Journal of Machine Learning and Cybernetics (2018).
    
    \item "Abnormal Event Recognition in Crowd Environments", Nabi, M., Mousavi, H., Rabiee, H., Ravanbakhsh, M., Murino, V., Sebe, N, In Applied Cloud Deep Semantic Recognition, Auerbach Publications (2018).
        
    \item "Abnormal Event Detection in Videos using Generative Adversarial Nets", M. Ravanbakhsh, M. Nabi, E. Sangineto, L. Marcenaro, C. Regazzoni, N. Sebe,  IEEE International Conference on Image Processing (ICIP 2017), Beijing, China (2017).
    
    \item "Efficient Convolutional Neural Network with Binary Quantization Layer", M. Ravanbakhsh, H. Mousavi, M. Nabi, L. Marcenaro, C. Regazzoni. Conference on Advances in Neural Information Processing Systems Workshops(NIPSW 2016), Workshop on Efficient Methods for Deep Neural Networks (EMDNN), Barcelona, Spain (2016).
    
    \item "Cnn-aware binary map for general semantic segmentation", M. Ravanbakhsh, H. Mousavi, M. Nabi, M. Rastegari, C. Regazzoni., IEEE International Conference on Image Processing (ICIP 2016), Phoenix, Arizona, USA (2016). 

\end{itemize}
\section{Thesis Overview}
Each chapter of this thesis is designed to be self-contained and their contents are structured in such a way that the document as a whole follows the same line and vision. The rest of this thesis is organized as follows.\\

In Chapter \ref{ch:abnormality}, we briefly review the recent successful approaches for abnormality detection. We also proposed three different methods of unsupervised and semi-supervised outlier detection as a key component of our abnormality detection models. We note that the previous works related to each task (e.g., crowd analysis, deep generative models) is extensively described in the corresponding sections (\ref{sec:binary_tr}, \ref{sec:off_the_shelf}, and \ref{sec:gan_anomaly}).\\

In Chapter \ref{ch:autonomous_car}, we first review self-awareness (SA) models and introduce a two-level SA model for an autonomous driving car. Then, we investigate the proposed models in detail for learning the normal situation representations. We specifically introduce a DBN-based approach to discover and learn novel situations according to the previous learned situations. The main novelty of this proposed approach is a weakly-supervised strategy to divide and solve a complex problem using GANs. The discriminator scores are used to approximate the complexity of a distribution and detecting changes with respect to the learned distribution. As result, the method is able to model highly diverse distributions. Such a learned model can gives the ability of detecting unseen situations to the autonomous agents.\\

Both chapter \ref{ch:abnormality} and chapter \ref{ch:autonomous_car} are supported with experimental results of different tasks on several data sets.





\chapter{Abnormal Event Detection in Crowds}
\label{ch:abnormality}
\graphicspath{{Chapter2/Figs/Vector/}{Chapter2/Figs/}}

\begin{quote}

\emph{``When you're the only sane person, you look like the only insane\footnote{\textbf{insane} \textipa{/Ins'seIn/} (\emph{adjective}) in a state of mind which prevents normal perception, behaviour, or social interaction; Def. Oxford Living Dictionaries.} person.''}\\

-- Criss Jami

\end{quote}
\vfill

\newpage
\newpage
\thispagestyle{plain} 
\mbox{}
\newpage
In recent years, the field of crowd behavior analysis had an outstanding evolution for several problems such as 
motion detection \cite{nabi2012human,nabi2013temporal,rastegari2008multi,ravanbakhsh2015action,2}, tracking \cite{3,he2016connected} and crowd behavior analysis \cite{Biswas2017,direkoglu2017abnormal,rabiee2016novel,rabiee2016crowd,rabiee2017detection,rabiee2016emotion,ravanbakhsh2017training,ravanbakhsh2017abnormal,Turchini2017,XU2017117}. 
However, crowd behavior analysis is still the topic of many studies. 
This is mainly because of both inherent complexity and vast diversity in the crowd scene understanding. In contrast to a low-density crowd, the behavior of each individual in a dense crowd might be affected by different factors such as goals, dynamics, environment, etc. In other words, a dense crowd goes beyond a set of individuals who act independently and show their personal behavioral patterns \cite{ravanbakhsh2016plug,sebe2018abnormal,14}.

Detecting abnormal events in crowd situations plays an important role in public surveillance systems for preventing crimes and providing safe societies. Despite the AI have the potential to revolutionise societies in positive ways, yet there are serious concerns regarding the individuals' privacy and using AI for good \cite{abadi2016deep,geyer2017differentially,zhang2005hiding}. In the crowd analysis field this issues are more critical since always there is a real risk of privacy violations, or using of tools which have a negative impact on the society (cultural/race bias). Hence, its usefulness in real world situations is often questioned due to the lack of adequately trusty trained personnel to monitor a large number of videos captured simultaneously, and to the loss of attention from surveillance operators after a few tens of minutes inspecting the videos. This has attracted immense attention in vision communities to develop techniques to automatically detect abnormal behaviors in video which, in real scenarios, may trigger human attention and help crime prevention. 

Despite the recent improvements in the past years \cite{boiman2007detecting,cong2011sparse,feng2017learning,
DBLP:conf/cvpr/0003CNRD16,
kim2009observe,
li2014anomaly,
lu2013abnormal,
mahadevan2010anomaly,
mehran2009abnormal,
mousavi2015analyzing,
ravanbakhsh2016plug,
ravanbakhsh2017abnormal,sabokrou2016video,7858798,sabokrou2016fully,
sabokrouFFK16,
saligrama2012video,xu2015learning}, abnormality detection in video sequences is still an open problem in computer vision. One of the  main reasons for which abnormality detection is challenging is the relatively small size of the existing datasets with abnormality ground truth. This is emphasized by the fact that the definition of what is an {\em abnormal} event is usually strictly domain-dependent, and most of the abnormality-detection scenarios only provide data describing {\em normal} situations. 
In other words, the abnormalities are basically context dependent and their definitions may vary regarding the crowd characteristics (e.g. dynamic, goal, place, motion flows, date, time, etc.)~\cite{mousavi2015analyzing}. A simple example is individuals running in the Marathon which is a normal activity, while is abnormal if it happens in a train station.

In order to deal with these issues, common abnormality-detection methods focus on learning only the {\em normal} pattern of the crowd, for which only weakly annotated training data are necessary (e.g., videos representing only the normal crowd behaviour in a given scene). Detection is then performed by comparing the the test-frame representation with the previously learned normal pattern (e.g., using a one-class SVM \cite{xu2015learning} or a probabilistic model \cite{DBLP:conf/wacv/PathakSM15,Varadarajan}). However, the lack of sufficiently large datasets provided with abnormality ground truth makes the learning process even more challenging for common fully-supervised training protocols.

In this chapter, we explain how to address the aforementioned hurdles by introducing three approaches based on both classical tracklet-based methods and novel deep learning methods. The key idea of the proposed approaches is under the hypothesis that abnormalities are \emph{outliers} of normal situations. In light of the above, we introduce three different methods for learning the abnormal event detection from the crowd dynamics:
\begin{itemize}
    \item \emph{i)} At the first study, an efficient binary representation for moving patterns is proposed. The crowd dynamics are extracted using analysis of short trajectories (tracklets) through a new video descriptor for detecting abnormal behaviors. In a nutshell, we describe each spatio-temporal window in the video volume using motion trajectories represented by a set of tracklets \cite{22}, which are short trajectories extracted by tracking detected interest points within a short time period. A binary representation of tracklets is introduced to model the movement of a salient point over its corresponding tracklet in both spatial and temporal spaces.
    
    \item \emph{ii)} For the second approach, we introduce a CNN-of-the-shelf technique using a pre-trained fully convolutional neural network followed by a binary quantization layer (BQL). The weights of the former network are borrowed from an already pre-trained network (namely, AlexNet) and the weights of the BQL layer are obtained through an external hashing module and ``plugged'' into the network as an additional convolutional layer. The last layer of the network (BQL) provides the means to quantize the high dimensional deep feature maps into 7-bit binary maps. The training of ITQ is done only once in an ``off-line'' fashion. The plug-and-play nature of our proposed architecture enables our method to work across multiple datasets without specific retraining. We define a histogram-based measurement for abnormal events as irregular events deviated from the normal ones, where the abnormality is measured as the uniformity of the histogram of binary codes.
    
    \item \emph{iii)} At the last study, a deep generative model is proposed to learn the crowd normality in an end-to-end fashion. We propose to use Generative Adversarial Nets (GANs), which are trained to generate only the {\em normal} distribution of the data. During  the adversarial GAN training, a discriminator ($D$) is used as a supervisor for the generator network ($G$) and vice versa. At testing time the real data are compared with both the appearance and the motion representations reconstructed by our GANs and abnormal areas are detected by computing local differences. Since our networks are only learned the ``normality'', any distributions deviated from the learned decision boundary detects as an abnormality. 

\end{itemize}

In the rest of this chapter we review in details the proposed methods for detecting abnormal events in crowded scenes. The proposed binary tracklet-based method in Sec.~\ref{sec:binary_tr}, a novel method for binary quantization of pre-trained deep features is described in Sec.~\ref{sec:off_the_shelf}, finally, we show how to train a deep generative model for outlier detection in Sec.~\ref{sec:gan_anomaly}, and we conclude in Sec.~\ref{sec:anomaly_conclusions}.
\graphicspath{{Chapter2/Figs/Vector/}{Chapter2/Figs/a/}}
\section{Previous Works}
\label{sec:background_anomaly}
There exists a large number of approaches for for abnormality detection in video sequences. In the following some of leading approaches are described. 
The major challenge in abnormality detection is that there is not a clear definition of abnormality, since they are basically context dependent and can be defined as outliers of normal distributions. Based on this widely accepted definition of abnormality, existing approaches for detecting abnormal events in crowds can be generally classified into two main categories: \emph{i)} object-based approaches, and \emph{ii)} holistic techniques.
\subsection{Object-based approaches} 
\begin{figure}[h!]
	\begin{center}
		\includegraphics[width=1\linewidth]{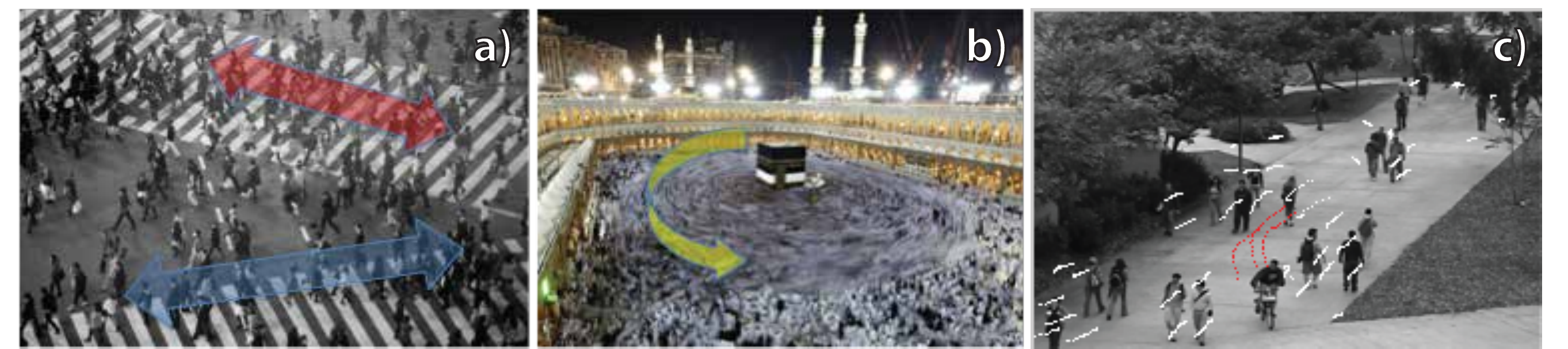}
	\end{center}
	\caption[Samples of crowded scenes]{Samples of crowded scenes: \textbf{(a)} Shibuya crossing (Japan). \textbf{(b)} Mecca (Saudi Arabia). \textbf{(c)} Tracklets extracted from UCSD dataset.}
	\label{fig:long}
\end{figure}

Object-based methods treat a crowd as a set of different objects. The extracted objects are then tracked through the video sequence and the target behavioral patterns are inferred from their motion/interaction models \cite{lan2012discriminative} (e.g. based on trajectories \cite{fu2005similarity}). This class of methods relies on the detection and tracking of people and objects \cite{mointhesis}. Despite promising improvements to address several crowd problems ~\cite{mousavi2016detecting,rabaud2006counting,rittscher05,yuan2015online}, they are limited to low density situations and perform well when high quality videos are provided, which is not the case in real-world situations. In other words, they are not capable of handling high density crowd scenarios due to severe occlusion and clutter which make individuals/objects detecting and tracking intractable~\cite{marques2003tracking,piciarelli2008trajectory}. Fig. \ref{fig:long} (a-b) shows an example of such scenarios. Some works made noticeable efforts to circumvent robustness issues. For instance, Zhao and Nevatia ~\cite{zhao2003bayesian} used 3D human models to detect persons in the observed scene as well as a probabilistic framework for tracking extracted features from the persons. In contrast, some other methods track feature points in the scene using the well-known KLT algorithm \cite{rabaud2006counting,shi1994good}. Then, trajectories are clustered using space proximity. Such a clustering step helps to obtain a one-to-one association between individuals and trajectory clusters, which is quite a strong assumption seldom verified in a crowded scenario. \\
\subsection{Holistic approaches} The holistic approaches, in contract with Object-based approaches, do not aim to separately detect and track each individual/object in a scene. Instead, they treat a crowd as a single entity and try to employ low/medium level visual features extracted from video frames to analyze the crowd scene as a whole \cite{krausz2011analyzing,wang2012abnormal,denseTraj}. The differences among holistic approaches are regarding the way that they represent the scene as well as the way that they detect anomaly. Some of them use statistic (machine learning) techniques in order to learn the “normal” behavior of a crowd in a given environment and define abnormality as those situations having a low probability value under the so constructed probabilistic framework \cite{wang2009unsupervised,wu2014bayesian}. Differently, abnormality can be defined using model based techniques (i.e., methods not involving statistic estimations), for instance dealing with abnormality detection as a saliency detection problem or using ad-hoc rules for finding specific patterns in the optical-flow data. Below the holistic approaches are classified with respect to the way the scene is represented and not the way in which the anomaly is detected.\\
\noindent{\textbf{Optical-flow histogram analysis}.} The simplest way to represent the global movement in a crowded scene is probably using simple statistics extracted from the optical flow data~\cite{wang2012histograms}. Zhong \emph{et. al.} \cite{zhong2004detecting} propose an unsupervised method which use the ``hard to describe'' but ``easy to verify'' property of unusual events without any explicit modeling of the normality. They proposed to use only simple motion models without supervised feature selections, however the method may fail in the crowded scenario and more complex activity patterns. Krausz and Bauckhage in~\cite{krausz2011analyzing,krausz2012loveparade} used the histogram of the optical flow as the basic representation. Simple heuristic rules are then proposed in order to detect specific crowd dangerous behaviors, such as congestion or turbulence situations. For instance, a congestion situation is detected by looking for symmetric pairs of histograms extracted from consecutive frames, which indicate slow lateral oscillations of the people’s upper body.\\
\noindent{\textbf{Spatio-temporal grids}.} Different approaches deal with the complexity of a dynamic scene analysis by partitioning a given video in spatio-temporal volumes \cite{dalal2005histograms,wang2012abnormal}. Each frame is partitioned in a spatial grid of $n \times m$ cells, and the frame sequence is partitioned in $k$ consecutive frames, which altogether brings to a $n \times m \times k$ spatio-temporal volume. In~\cite{kratz2009anomaly,kratz2010tracking} Kratz and Nishino extract spatio-temporal gradients from each pixel of a given frame. Then, the gradients of a spatio-temporal cell are modeled using Spatio-Temporal Motion Pattern Models, which are basically 3D Gaussian clusters of gradients. A simple leader follower on-line clustering algorithm is used to group gradients observed at training time in separate cluster centers (prototypes). At testing time, a single Gaussian cluster is extracted from each cell of the input video and the Kullback-Leibler distance is used in order to select the training prototype with the closest gradient distribution. Finally, a mixed spatio-temporal Hidden Markov Model is used in order to model transition probabilities among prototypes.\\
Mahadevan et al.~\cite{mahadevan2010anomaly} model the observed movement in each spatio-temporal cell using dynamic textures, which can be seen as an extension of PCA-based representations. Whereas PCA spaces only model the appearance of a given patch texture, dynamic textures also represent the statistically valid transitions among textures in a patch. In each cell, all the possible dynamic textures are represented with a Mixture of Dynamic Textures model, which gives the probability of a test patch to be anomalous. In this way, the authors show that not only temporal anomalies but also pure appearance anomalies can be detected. In the same work the authors present also an interesting definition of spatial saliency based on Mutual Information between features and foreground/background classes. In the proposal of Mahadevan et al., only local (cell-level) abnormalities are detected.\\
\noindent{\textbf{Physics inspired models}.} Some research groups exploit mathematical models derived from fluid dynamics or other physics laws in order to model a crowd as an ensemble of moving particles. The Social Force model to describe the behavior of a crowd as the result of interaction of individuals uses the Second Newton’s law to describe the causes of the movement of a set of particles~\cite{helbing1995social}. In~\cite{mehran2009abnormal} the Social Force model is used to detect anomalies and estimate local anomalies by detecting regions in the current frame in which the local optical flow is largely different from the average optical flow computed in the neighboring regions. Randomly selected spatio-temporal volumes of Force Flow are used to model the normal behavior of the crowd and classify frames as normal and abnormal  cases using a bag of words approach. 
The same research group uses Coherent Structures and a fluid dynamics framework to segment optical flow data in dynamically coherent clusters~\cite{Authors17}. Anomalies are detected looking at sharp differences between the segmentation outcomes of consecutive frames. Finally, the Shah’s group in~\cite{solmaz2012identifying} proposes a method to classify the critical points of a continuous dynamical system. They represent the scene as a grid of particles initializing a dynamical system which is defined by the optical flow information. Such  simplification provides a linear approximation of the input complex dynamical system, aiming to identify typical crowd behaviors. In~\cite{wu2010chaotic} a novel method is presented for detecting and localizing anomalies in complicated crowd sequences using a Lagrangian particle dynamics approach, together with chaotic modeling. All these works do not detect and track individuals. Instead, they apply particle advection technique that places particles onto a grid and moves them according to the underlying optical flow field. However, particle advection is not applicable when the camera viewpoints are inappropriate and the resulted occlusions in situations of high pedestrian density.\\
\noindent{\textbf{Segmentation approach.}}  In scenarios with high density crowds, i.e. political rallies,~\cite{solmaz2012identifying} religious festivals and marathons which involve, the large gatherings of people poses significant challenges from the scene monitoring point of view, where current automated surveillance systems fail to deal with such cases. The reason for such failure is the difficulty of detection and tracking of target objects in high density scenes. In such cases, segmenting high density scenes into dynamically and physically meaningful flow segments can be a tractable solution. Such emerging motion patterns are called as ``flow segment''. 
Using instantaneous motions of a video, i.e. The motion flow field is another viable solution presented in ~\cite{Authors22}.
The motion flow field is a union of independent flow vectors computed in different frames and a set of flow vectors representing the instantaneous motions in a video. 
They first use existing optical flow method to compute flow vectors in each frame and then combine them into a global motion field. This flow field may contain thousands of flow vectors, and, therefore, it is computational expensive to obtain the shortest paths based on such a large number of data. Detecting motion patterns in this flow field can therefore be formulated as a clustering problem of the motion flow fields. A hierarchical agglomerative clustering algorithm is applied to group flow vectors into desired motion patterns.\\
\noindent{\textbf{Tracklet-based approach.}} Tracklets are typically short temporal sequences of points, commonly extracted using the KLT method~\cite{shi1994good}. The points to track can be salient points, randomly selected points or densely distributed points on a grid~\cite{denseTraj}. Tracklet-based methods can be seen as a trade-off between object and holistic based approaches. In fact, on the one hand they model the observed scene using trajectory analysis but on the other hand they do not rely on person detection and tracking, being focused on tracking of simple points which is a much simpler task.
Zhou et al.~\cite{Authors19} dealt with a tracklet as a document in a Bag of Word approach. All the points of a given tracklet are associated with words of a codebook, according to their location and velocity directions. Then, tracklets are automatically clustered in “topics”, where each topic describes a semantic region in the scene. Temporal and spatial dependencies between tracklets are modeled using two different MRFs, one for the spatial and one for the temporal dependence respectively. In~\cite{Authors20} the same authors go a step forward and use tracklet clusters in order to classify a test tracklet according to the closest cluster. In the same work, past and future paths of the individual represented by the analyzed tracklet can be simulated.

\subsection{Deep learning-based approaches}
There is a wealth of literature on abnormality detection that is based on hand-crafted features (e.g., Optical-Flow, Tracklets, etc.) to model the normal activity patterns, whereas deep learning-based approaches yet is not studied well. Despite the recent improvements in deep learning on different areas including image classification \cite{alexnet,rastegari2016xnor,russakovsky2015imagenet}, object and scene detection/recognition \cite{sangineto2016self,zhou2014learning}, image segmentation \cite{long2015fully,2016icip}, human action recognition \cite{DBLP:conf/nips/SimonyanZ14,Feichtenhofer_2016_CVPR} vision and language integration \cite{gordon2018iqa,shekhar2017vision,shekhar2017foil}, and human-machine collaboration \cite{abad2017self,abad2017autonomous,russakovsky2015best}, but still the deep learning approaches are not exploited well for abnormality detection task~\cite{fang2016abnormal,ionescu2018object,ravanbakhsh2016plug,7858798,sabokrouFFK16}. 
This is mainly due to the nature of of abnormality detection task and the lack of annotated data. Deep networks are data hungry and training a network with minimal data is a challenging task itself \cite{bach2017learning,pahde2019low,ZHOU2016358}.
Hence, the deep learning-based works for abnormality detection task, mainly use existing Convolutional Neural Network (CNN) models trained for other tasks (e.g., object recognition) which are adapted to the abnormality detection task. For instance, Ravanbakhsh et al.~\cite{ravanbakhsh2016plug} proposed a Binary Quantization Layer, plugged as a final layer on top of a pre-trained CNN, in order to represent patch-based   temporal motion patterns. However, the network proposed in \cite{ravanbakhsh2016plug} is not trained end-to-end and is based on a complex post-processing stage and on a pre-computed codebook of the convolutional feature values. Similarly, in \cite{7858798,sabokrouFFK16}, a fully convolutional neural network is proposed which is a combination of  a pre-trained CNN (i.e., AlexNet \cite{alexnet}) and a new convolutional layer where kernels have been trained from scratch. 
Sabokrou et al. \cite{sabokrou2016video}  introduce a patch-based anomaly detection framework based on an Autoencoder (AE) reconstruction error and sparsity constraints. However this work is  limited to a single modality setup. Stacked Denoising Autoencoders (SDAs) are used by Xu et al.~\cite{xu2015learning} to learn motion and appearance feature representations. The networks used in this work are relatively shallow, since training deep SDAs on small abnormality datasets can be prone to  over-fitting issues and the networks' input is limited to a small image patch.

Moreover, after the SDAs-based features have been learned, multiple one-class SVMs need to be trained on top of these features in order to create the final classifiers, and the learned features may be sub-optimal because they are not jointly optimized with respect to the final abnormality discrimination task.
Feng et al. \cite{feng2017learning} use  3D gradients and a PCANet \cite{PCANet} in order to extract patch-based appearance features whose normal distribution is then modeled using a deep Gaussian Mixture Model  network (deep GMM \cite{deepGMM}). Also in this case the feature extraction process and the normal event modeling are obtained using two separate stages (corresponding to two different networks) and the lack of an end-to-end training which jointly optimizes  both these stages can likely produce sub-optimal representations. Furthermore, the number of Gaussian components in each layer of the deep GMM is a critical hyperparameter which needs to be set using supervised validation data.

The only deep learning based approach proposing a framework which can be fully-trained in an end-to-end fashion we are aware of is the Convolutional AE network proposed in \cite{DBLP:conf/cvpr/0003CNRD16}, where a deep representation is learned by minimizing the AE-based frame reconstruction. At testing time, an anomaly is detected computing the difference between the AE-based frame reconstruction and the real test frame.
\graphicspath{{Chapter2/Figs/a/}}
\section{Crowd Abnormality Detection Datasets}
\label{sec:datasets_anomaly}
In recent years, the number of studies on abnormal crowd behaviour detection has grown rapidly in both academic and commercial fields. This comes with a grown demand for public datasets to use for video surveillance system evaluation, yet there are not many available public datasets for crowd abnormality detection.
\begin{table}[H]
    \centering
    \begin{tabular}{p{1.3cm}p{6.2cm}p{6.5cm}}
    \toprule
         Dataset &  Sequences & Description\\
         \midrule
         UCSD  \cite{mahadevan2010anomaly} & two subsets: \emph{PED1}: 70 videos (34 training, 36 testing), $158 \times 238$ pixels. \emph{PED2}: 30 videos (16 training, 14 testing), $240 \times 360$ pixels. & normal events define as pedestrians in the walkways, and non-pedestrians correspond to abnormal events.\\
         \midrule
         UMN \cite{mehran2009abnormal}& including 11 videos from three different indoor and outdoor scenes. Resolution of $240 \times 320$ pixels. & each sequence starts with a normal scene and ends with crowd dispersion as an abnormal event. \\
         \midrule
         Violent-Flows \cite{26} & 246 videos (123 violence and 123 no violence) with resolution of $240 \times 320$ pixels. & videos are captured from YouTube, and containing violence actions in crowded places.\\
         \midrule
         Action Movies  \cite{nievas2011violence}& 200 videos (100 violence and 100 no violence). & captured scenes from movies. fights are defined as abnormal events.\\
         \midrule
         Hockey Fight  \cite{nievas2011violence} & 1000 videos (500 violence and 500 no violence). Resolution of $ 576 \times 720$ pixels. & violence detection in ice hockey rink (uncrowded scene). \\
         \midrule
         Web dataset  \cite{mehran2009abnormal} & 20 videos (8 for training and 12 for testing).  & normal events define as pedestrians and abnormal events are clash, escape panic, fight, etc.\\
        \midrule
         MED  \cite{rabiee2016novel} &  including 31 videos from three outdoor scenes. Resolution of $ 640 \times 470$ pixels. & fine-grained crowd behaviour analysis. Normal events define as walking pedestrians and abnormal events are fight, congestion, panic, etc.\\
         \bottomrule
    \end{tabular}
    \caption[List of datasets for video surveillance systems]{Crowd abnormality detection datasets for video surveillance systems \cite{BENMABROUK2018480}.}
    \label{tab:crowd_dbs}
\end{table}
The currently available datasets can be categorized into three main types: i) consisting of videos containing often violent actions. ii) videos describing the interaction between a group of persons acting abnormally such as panic escape. iii) fine-grained behaviour recognition datasets. A list of main datasets is presented in Tab. \ref{tab:crowd_dbs}.
\subsection{Evaluation datasets}
\begin{figure}[H]
	\begin{center}
		\includegraphics[width=0.98\linewidth]{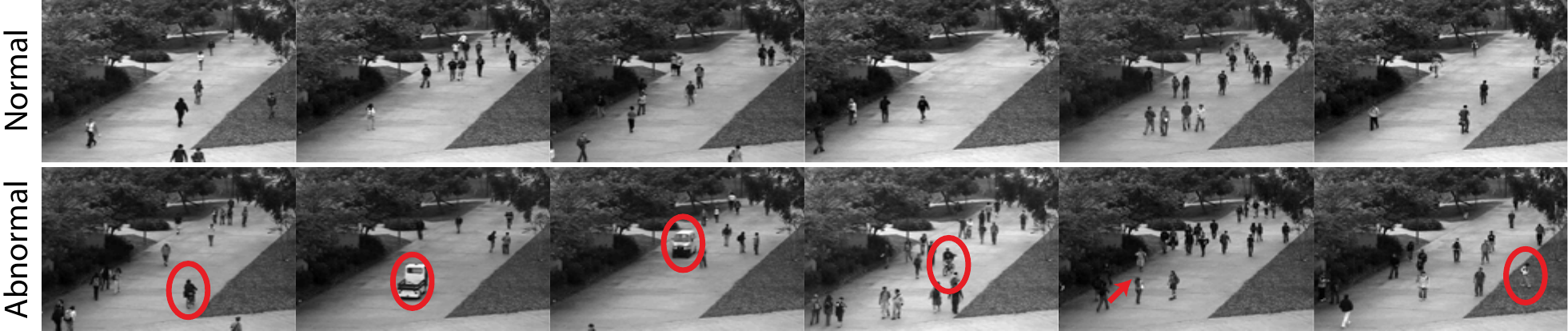} (a)
		\includegraphics[width=0.98\linewidth]{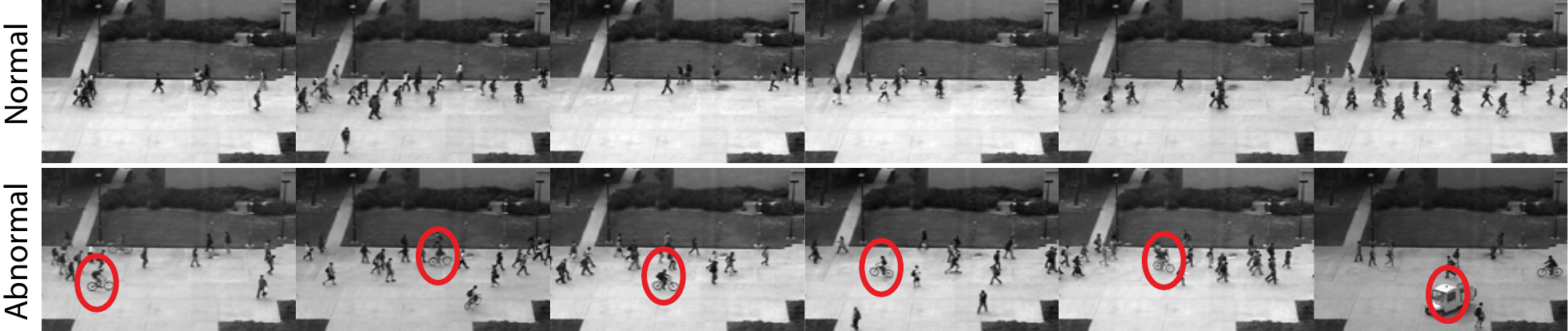} (b)
		\includegraphics[width=0.98\linewidth]{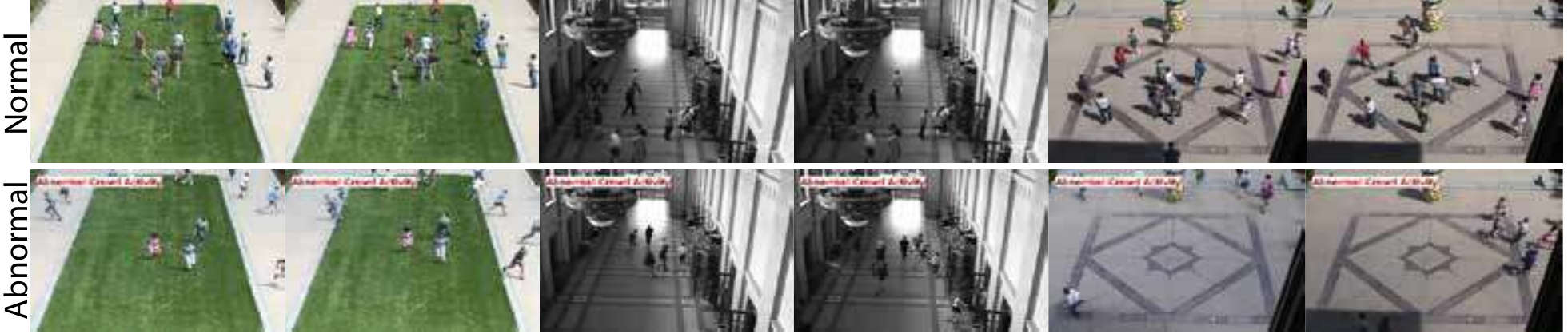} (c)
		\includegraphics[width=0.98\linewidth]{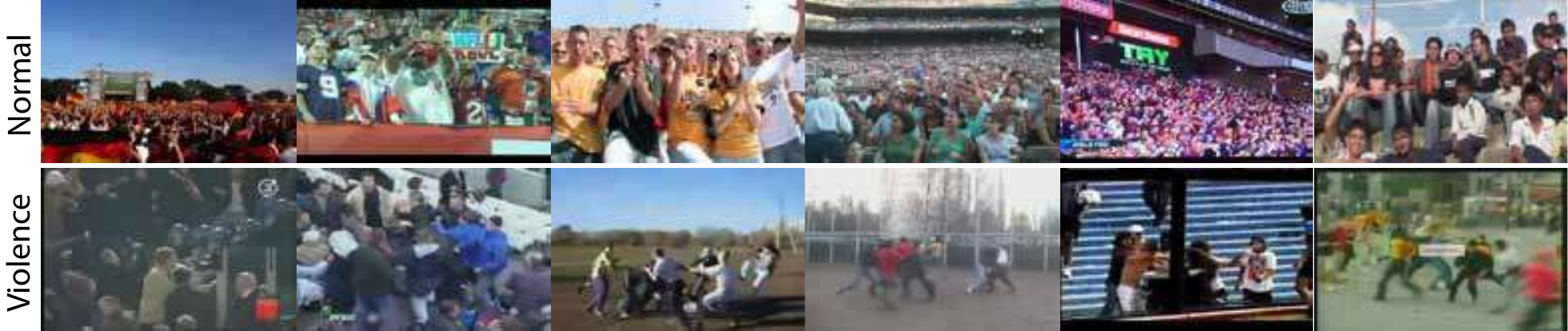} (d)
	\end{center}
	\caption[Sample frames from crowd datasets]{Sample frames from crowd datasets for both normal and abnormal situations: \textbf{(a)} UCSD PED1. \textbf{(b)} UCSD PED2. \textbf{(c)} UMN. \textbf{(d)} Violent-Flows dataset.}
	\label{fig:ab_db}
\end{figure}

In our experiments we used three publicly available datasets for evaluation, including UCSD \cite{mahadevan2010anomaly}, UMN \cite{mehran2009abnormal} and Violence-in-crowds \cite{26}. Examples of video frame scenes is shown in Fig. \ref{fig:ab_db}. In this section we briefly describe the selected datasets and review the major characteristics of each dataset. \\

\noindent\textbf{UCSD dataset.} consists of two datasets captured from different scenes: PED1 and PED2.
\begin{itemize}
    \item [-] PED1 contains 34/16 training/test sequences with frame resolution $238 \times 158$ pixels. Video sequences indicate groups of individuals walking toward and away from the camera, with a certain degree of perspective distortion. The dataset consists of 3,400 abnormal frame samples and 5,500 normal frames.
    \item [-] PED2 includes 16/12 training/test video samples, with about 1,600 abnormal frames and 350 normal samples. This subset indicates a scene where most pedestrians move horizontally. The frames resolution is $360 \times 240$. 
\end{itemize}
This dataset is challenging due to different camera view points, low resolution, different types of moving objects across the scene, presence of one or more anomalies in the frames. The video footage of each scene is divided into two subsets: test and training (only normal conditions).\\
    
\noindent\textbf{UMN dataset.} contains 11 different scenarios in three different indoor and outdoor situations. UMN includes 7700 frames in total with frame resolution of $320 \times 240$ pixels. All sequences start with a normal scene and end with abnormality section.\\

\noindent\textbf{Violent-Flows dataset.} is composed of real-world scenarios and video sequences of crowd violence, along with standard benchmark protocols designed to test violent and nonviolent classification. It is divided into five subsets: half violent crowd behavior and half nonviolent crowd behavior for training.

\subsection{Evaluation metrics}
Among the several metrics to evaluate a video surveillance system, two of them are commonly used, and both criteria are computed from the Receiver Operating Characteristic (ROC) Curve:\\
- Area Under Roc curve (AUC):  The AUC mainly is used for performance comparison in different tasks \cite{BENMABROUK2018480}.\\
- Equal Error Rate (EER): The EER is the point on a ROC curve where the false positive rate (i.e., normal detects as abnormal) is equal to the false negative rate (i.e., abnormal is classified as normal).

The performances of a system can be considered as good if the value of the AUC be as high, while the value of the EER is as small as possible. In our works, we followed these standard evaluation metrics and present our results in terms of ROC curves, AUC and ERR.

\graphicspath{{Chapter2/Figs/Vector/}{Chapter2/Figs/a/}}
\section{Analyzing Local Binary Tracklets for Abnormal Event Detection}
\label{sec:binary_tr}

Abnormal events are usually considered as outliers of normal distributions \cite{ravanbakhsh2016plug,sebe2018abnormal,14}. Under this hypothesis, abnormalities are rare observations, which highly contrast with the normality. Sudden changes in pedestrian directions and their high speed in the presence of non-pedestrian moving objects could be considered as abnormal behaviors. The field of view is another effective parameter in video recording. However, the perspective geometry which introduces distortions and apparent motions do not correspond to the actual movements and are ignored. Another issue to consider is the number of individuals in a crowd scene, which may affect the quality of abnormality detection and localization. 
Moreover, it is also important to detect abnormal behaviors both in space and time domains. This refers to isolate the abnormal frames in a crowd video (namely, frame-level abnormality detection) and to localize the abnormal areas in identified abnormal frames (namely, pixel-level abnormality detection).

To overcome aforementioned challenges, in this section, we propose a novel video descriptor for abnormal behavior detection and localization. In the proposed method, spatio-temporal abnormalities in densely crowded scenes are detected by a new tracklet based descriptor. For this purpose, we first divide the video sequence into a set of non-overlapping spatio-temporal 3D patches. The short trajectories (tracklets) are then extracted by tracking randomly selected points in video frames within a short period of time. Using the orientation and magnitude of extracted tracklets, we compute our proposed binary tracklet based descriptor. In a nutshell, a video sequence is segmented into spatio-temporal patches. Then, using motion trajectories represented by a set of tracklets \cite{22,ravanbakhsh2018fast}, each patch is being described. Unlike most of the standard approaches which describe frames with dense descriptors such as optical flow \cite{23} and interaction force \cite{mehran2009abnormal}, we define spatio-temporal patches and gather the statistics on trajectories that intersect them. More specifically, the orientation and magnitude of such intersecting tracklets are encoded in our proposed Local Binary Tracklets (LBT). Under the assumption that abnormalities are outliers of normal situations and considering the fact that we have only access to normal samples for training, which it seemed to be a realistic assumption in the real world, we employed one-class SVM for behavior classification. By combining our proposed descriptor with the dense optical flow \cite{24} (i.e., LBT+DOF), we also proposed to accurately localize the abnormal behavior region in the abnormal frames. 

Our contributions are as follows: \emph{(i)} We introduce a novel descriptor for abnormality detection, namely Local Binary Tracklets (LBT). It is much more efficient if compared to other state-of-the-art methods and is shown to have competitive results. \emph{(ii)} We present a novel framework by combining Local Binary Tracklets (LBT) with Dense Optical Flow (DOF) \cite{24} model which can localize the abnormal behavior area in a frame. \emph{(iii)} Since abnormal behavior samples are hardly accessible in real-world crowd scenarios and are not sufficiently available at training time, we propose using one-class SVM model which only needs normal samples at training time.

The rest of the section is organized as follows: the proposed Local Binary Tracklets (LBT) model for abnormality detection is presented in Sec. \ref{sec:method}. In Sec. \ref{sec:detect} abnormal behavior detection schemes are described using LBT and LBT-Dense Optical Flow (DOF) models. The experimental results are presented in Sec. \ref{sec:exp}, following with a discussion on the obtained results in Sec.  \ref{sec:disc}.



\begin{figure*}
    \hspace{.5in}
	\begin{minipage}[b]{\linewidth}
		\begin{center}
			\includegraphics[width=0.9\linewidth]{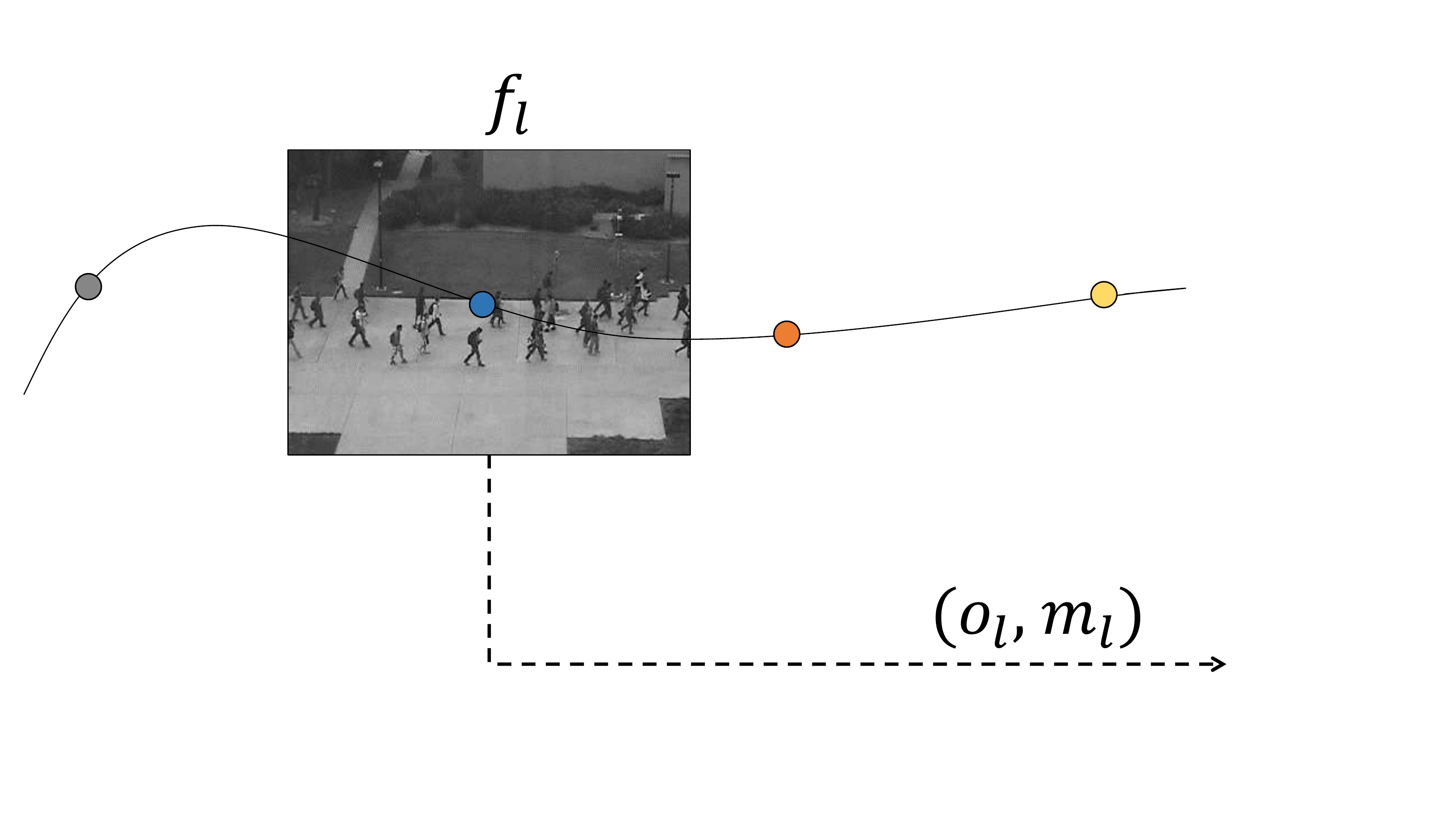}
			\vspace{-.6in}
			\hspace{0.009in} \\
		\hspace{-.9in}	(a) 
		\end{center}
	\end{minipage}
	
	\begin{minipage}[b]{\linewidth}
		\begin{center}
			\includegraphics[width=0.6\linewidth]{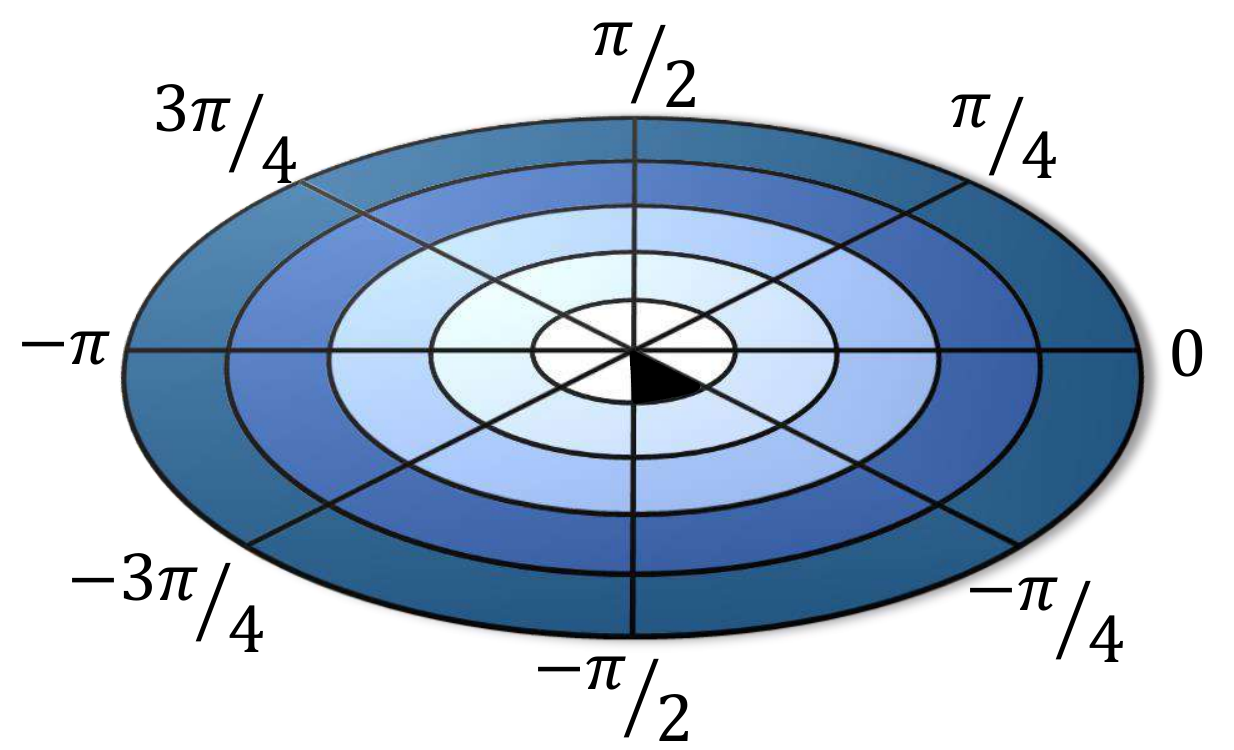}
			\hspace{0.01in}
			\vspace{0.009in}~\\
			(b)
		\end{center}
	\end{minipage}
	
	\begin{minipage}[b]{1\linewidth}
		\begin{center}
			\includegraphics[width=\linewidth]{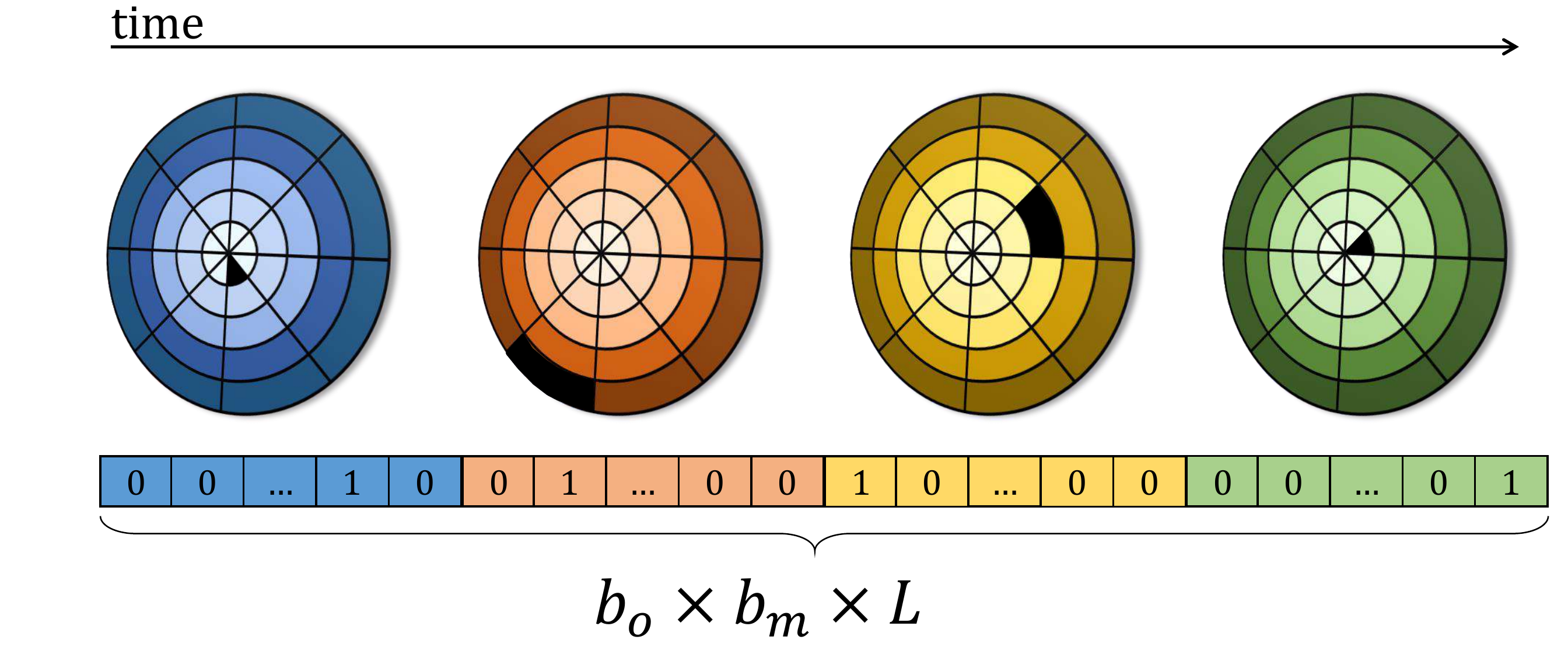}
			(c)
		\end{center}
	\end{minipage}
	\caption[Local Binary Tracklet computation]{\small{Local Binary Tracklet computation: (a) Tracklets are extracted from corresponding salient points tracked over frames. (b) A polar histogram of magnitude and orientation of a salient point at the $l^{th}$ frame of a tracklet (motion pattern). (c) The tracklet binary code is constructed by concatenating a set of motion patterns over $L $ frames. The "$1$" value sectors show with black.}}
	\label{fig:res}
\end{figure*}
\subsection{Local Binary Tracklets (LBT)}
\label{sec:method}

Tracklets \cite{22} are compact spatio-temporal representations of moving rigid objects. They represent fragments of an entire trajectory corresponding to the movement pattern of an individual point, generated by the frame-wise association between point localization results in the neighbor frames. Tracklets capture the evolution of patches and were originally introduced to model human motions for the task of action recognition in video sequences~\cite{mousavi2015abnormality,22}.
More formally, a tracklet is represented as a sequence of points in the spatio-temporal space as: $\mathbf{tr}^{n}= \left ( d_{1},\cdots d_{l},\cdots d_{L} \right )$,
where each $d_{l}$ represents two-dimensional coordinates $\left (  x_{l}, y_{l} \right )$ of the $l^{th}$ point of the tracklet in the $l^{th}$ frame and $L$ indicates the length of each tracklet. Tracklets are formed by selecting regions (or points) of interest via a feature detector and by tracking them over a short period of time. Thus, one can say that tracklets represent a trade-off between optical flow and object tracking. In a given video sequence, all the tracklets are derived using standard OpenCV code\footnote{http://www.ces.clemson.edu/stb/klt/}.
Specifically, the SIFT algorithm is employed to detect possible salient points~\cite{40} in a frame. Then the KLT algorithm ~\cite{23} is applied to track the salient points for $L$ frames - Fig. (\ref{fig:res}-a). The spatial coordinates of the tracked points are used to form the tracklets set $\mathcal{T} =  {\left \{ \mathbf{tr}^n\right \}_{n=1}^{N}}$, where $N$ is the number of all extracted tracklets and $\mathbf{tr}^{n}$ refers to the $n^{th}$ tracklet in the video. 
Each tracklet $tr^{n}$ is characterized by a set of spatial coordinates of its corresponded salient point tracked over $L+1$ frames $\{ ( x_{l}^{n} , y_{l}^{n})  \}_{l=1}^{L+1}$. These spatial coordinates are employed to compute the motion orientation and magnitude of the salient point at $l$-th frame as:
 \begin{eqnarray}
o_{l}^{n} & = & \arctan\frac{(y_{l}^{n}-y_{l-1}^{n})}{(x_{l}^{n}-x_{l-1}^{n})}
\label{eq:eg1} \\
m_{l}^{n} &  = &  \sqrt{\left( x_{l}^{n}-x_{l-1}^{n}\right)^2+\left( y_{l}^{n}-y_{l-1}^{n}\right)^2} 
\label{eq:eg2}
\end{eqnarray}  
where $ 2 \leq l \leq L+1$. This step computes a temporary ordered set of $L$ orientations and magnitudes of the salient point corresponded to $n$-th tracklet $ \{  ( o_{l}^{n}, m_{l}^{n} )  \}_{l=1}^{L} $
, where the orientation and magnitude of $n$-th tracklet at $l$-th frame are indicated as $o_{l}^{n}$ and $m_{l}^{n}$, respectively.
In order to represent the motion patterns, first, a motion pattern is encoded as a polar histogram $h_{l}^{n}$ computed from the orientation and magnitude of the $n$-th tracklet at frame $l$. This can be done by a simple hashing function whose input is $( o_{l}^{n}, m_{l}^{n} )$ and returns a binary polar histogram with only one "$1$" value at sector $( o_{l}^{n}, m_{l}^{n} )$ and zeros for the rest \cite{mousavi2015crowd}. Then, the computed polar histogram is vectorized into a vector of length $ b_{o} \times b_{m} $, where the number of quantized bins for orientation and magnitude are indicated as $b_{o}$ and $b_{m}$, respectively.  This is illustrated in Fig.~(\ref{fig:res}-b), where the color spectrum of each sector indicates the quantized bin of magnitude, and each arc shows the quantized bin of orientation, and the black sector shows the "$1$" value sector.\\
\noindent{\textbf{Tracklet binary code.}}
Given a set of orientations and magnitudes $ \{  ( o_{l}^{n}, m_{l}^{n} )  \}_{l=1}^{L} $, the vectorized motion patterns $ \{ h_{l}^{n}  \}_{l=1}^{L}  $ are computed and concatenated to form a tracklet histogram $ H^{n} = [h_{1}^{n},...,h_{L}^{n}]^{\top} $ of length $ b_{o} \times b_{m} \times L $ ($\top$ is transpose operator). $H$ is referred to as tracklet binary code, which is illustrated in Fig.~(\ref{fig:res}-c).\\
\noindent{\textbf{LBT computation.} }
To compute LBT, each frame $f_t, t=1,2,..,T$, is divided into a set of non-overlapped patches $ \{ p_{s}^{t} \} $, where $s = 1,2,..,S$ indicates the $s$-th patch in the $t$-th frame. $S$ is the total number of paths in a frame (tessellation size). For each patch $p_s^t$, a subset of tracklet binary codes is selected from $\{ H^{n} \}_{n=1}^N$ whose corresponding tracklets spatially pass from patch $p_s^t$, and $p_s^t$ is temporally located at the middle of the selected tracklets (i.e. if the length of a tracklet is $L$, tracklets which start/end $L/2$ frames before/after frame $t$ passing from patch $p_s^t$ are selected). Capturing the temporal context using the middle path is a common strategy \cite{lu2013abnormal,mousavi2015crowd,22}, which provides the means to exploit the dynamics in a short temporal window. Suppose that $N_p$ tracklet motion codes are selected for patch $p_s^t$ denoted by $ \{ H^{n_p} \}_{n_p=1}^{N_p}$. Then, the aggregated tracklet binary code is statistically computed for patch $p_s^t$ as $ \mathcal{H}_s^t = \sum_{n_p = 1}^{N_p} H^{n_p}$. As result, the distribution of local motion patterns of $p_s^t$ is encoded into the aggregated histogram $ \mathcal{H}_s^t$. In a normal crowd, there are no significant changes in direction and speed of individuals and the Local Binary Tracklets (LBT) descriptors are similar to each other. However, when there are sudden changes in directions and speeds of individuals, the corresponding magnitudes and orientations are unusually high and the corresponding LBT descriptors are different from the normal ones. 
The representation of LBT on UCSD crowd dataset is shown in Fig. (\ref{fig:fig4}-b).
\subsection{Abnormality detection with LBT}
\label{sec:detect}
Unlike the usual classification tasks in computer vision, for crowd abnormality detection we hardly can assume that abnormal footage is available at the training step. 
In the context of abnormal behavior detection, previous works employed Latent Dirichlet Allocation - LDA \cite{35} or mixture models \cite{mehran2009abnormal}. In this framework, we employ the one-class SVM to determine what is normal in terms of co-occurrences between the motion pattern features. Given a set of histogram ${\mathcal H}_{s}^{t}$ for each frame $t = 1,2,..., T$, a one-class SVM training corpus $D$ is built using the strategy. In this strategy, Local Binary Tracklets (LBT) derived from all the different 3D patches are concatenated in a single descriptor to preserve the spatial information of each frame:
\begin{equation}
D^{f_t} = \{ \mathcal{H}^{t}_{1} \mathcal{j} \mathcal{H}^{t}_{2}  \mathcal{j} ... \mathcal{j} \mathcal{H}^{t}_{s} \},
\quad D = \{ D^{f_t} \}_{t=1}^{T}
\label{eq:d}
\end{equation}
where $D^{f_t}$ is the computed LBT of $t$-th frame, $s=1,2,..S$ is the total number of paths in the frame, and $D$ is the SVM training corpus.
In this case, one-class SVM captures correlations between motion patterns that occur in different 3D patches of the scene. 
Inspired by \cite{33,kratz2009anomaly,mousavi2015crowd}, we fused the computed LBT with the salient motion map for fine abnormality detection in frames. 
A frame-level detection is formed by considering all patch-level detection feedback. In the pixel-level abnormality detection scheme, we localize the abnormal saliency areas in a frame using proposed combined dense optical flow (DOF) and Local Binary Tracklets (LBT+DOF) model. 
Fig. (\ref{fig:fig4}-b) illustrates the extracted LBTs, and in Fig. (\ref{fig:fig4}-c) the results of early fusion of LBT+DOF in feature space is shown. For the abnormality detection we used the late fusion of LBT+DOF, Fig (\ref{fig:fig4}-e).

\subsection{Experiments}
\label{sec:exp}

\noindent{\textbf{Parameters setting.} }
In our experiments, tracklets are extracted using KLT algorithm \cite{40}. Interest points are selected at first frame and they track over $L = 11$ frames. Interest point re-initialization procedure is done in case of tracking failures. The other parameters are temporal window $W$ , length of the tracklet $L$, tessellation of the frame $S$ and the quantization bins $O$. The tracklets orientation is considered as $O=8$ similar to \cite{mousavi2015analyzing}. Each bin corresponds to a range of orientation angles which placed in $[-\pi , \pi ]$. We equally divide the $[-\pi , \pi ]$ interval into $8$ ranges. The number of quantization levels for magnitude is $M = 5$. Unlike \cite{mousavi2015analyzing}, we changed the temporal window to $W= 11$ frames. The spatial tessellation is considered as $S= 4 \times 6$. We obtained magnitudes and orientations of all the tracklets which intersect each spatio-temporal 3D patch with the size of $S_x \times S_y \times W$ and create corresponding Local Binary Tracklets (LBT).
The value of each bin of histogram is determined by the sum of magnitudes of tracklets that their orientation falls into the range of that bin. Using the extracted histograms for each 3D patch, we finally obtained LBT and combined LBT and DOF by employing the methods mentioned in Sec. \ref{sec:detect}.

\begin{table}
	\begin{center}
		\scalebox{0.99}{
			\begin{tabular}[width=1\textwidth]{lcccccc}
				\toprule
				\multirow{3}{*}{Method} &\multicolumn{2}{c}{Ped1} &\multicolumn{2}{c}{Ped1} &\multicolumn{2}{c}{Ped2}\\
				&\multicolumn{2}{c}{frame level} &\multicolumn{2}{c}{pixel level} &\multicolumn{2}{c}{frame level}\\
				& ERR & AUC & ERR & AUC & ERR & AUC\\
				\midrule
				MDT \cite{mahadevan2010anomaly} 	& 	25\%& 	81.8\% & 		58\%& 	44.1\%& 	25\%& 82.9\%\\
				LMH \cite{42} 	& 	38\%& 	70.1\%& 		80\%& 	37\%& 		40\%& 64.4\%\\
				AMDN\cite{XU2017117} 	& 	22\%&   \textbf{84.9\%} & 		47.1\%& \textbf{57\%}& 		24\% & 81.5\% \\
				CFS~\cite{leyva2017video}&                 \textbf{21.1\%} & 	    82\%    & 		\textbf{39.7\%}  &   \textbf{57\%}    & 		19.2\%  & 84\%      \\
				Turchini et al.~\cite{Turchini2017}&                        24\%    & 	    78.1\%  & 	---    &   ---  & 		\textbf{19\%}    & 80.7\%    \\
				3DConv-AE.~\cite{DBLP:conf/cvpr/0003CNRD16}&                               27.9\%  & 	    81.0\%  & 		---     &   ---     & 		21.5\%  & \textbf{90.0\%}    \\
				\midrule
				LBT 		& 	21.8\% & 81.8\%& 		49\%& 	54\%& 		\textbf{19.0\%}& 83.5\%\\
				\bottomrule
		\end{tabular}}
	\end{center}
	\caption{\small{Equal Error Rates (EER) and AUC on UCSD dataset.}}
	\label{tbl:tab1}
\end{table}

\noindent{\textbf{Evaluation on UCSD dataset.} }Following the standard train/test partitioning proposed in \cite{mahadevan2010anomaly}, our method is evaluated and the results are reported. LBTs are computed on the frame-level using the extracted one-class SVM confidence scores, and after defining a range of thresholds the best results for AUCs (Area Under the ROC Curve) are reported at the pixel-level (localization). The results on Ped1 and Ped2 are presented in Tab. \ref{tbl:tab1}. Results show that in most of the cases the proposed method hit the abnormality correctly in terms of detection and localization. Only in the case of Deep learning based approaches, our measure achieved lower accuracy in abnormality detection. Namely, AMDN \cite{XU2017117} and 3DConv-AE.~\cite{DBLP:conf/cvpr/0003CNRD16} are superior in Ped1 and Ped2 datasets, respectively. However, given the nature of the proposed feature (simplicity, low cost, efficiency) makes it comparable with such CNN-based methods in terms of extraction time, storage requirements, and the computational complexity.

\begin{landscape}
\begin{figure*}
	\begin{center}
		(a) \hspace{2.3 cm} (b) \hspace{2.3 cm} (c)\hspace{2.3 cm} (d)\hspace{2.3 cm} (e)\\
		\includegraphics[width=0.19\linewidth]{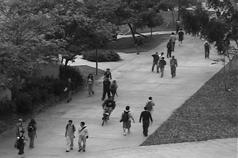}
		\includegraphics[width=0.19\linewidth]{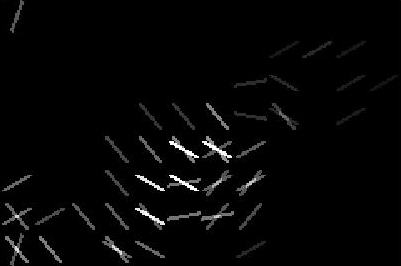}
		\includegraphics[width=0.19\linewidth]{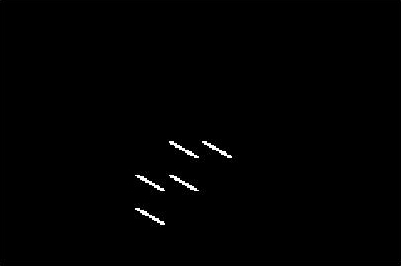}
		\includegraphics[width=0.19\linewidth]{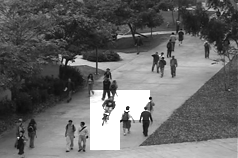}
		\includegraphics[width=0.19\linewidth]{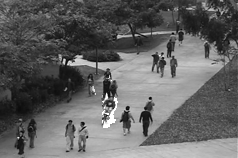}
		
		\includegraphics[width=0.19\linewidth]{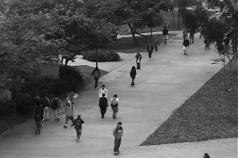}
		\includegraphics[width=0.19\linewidth]{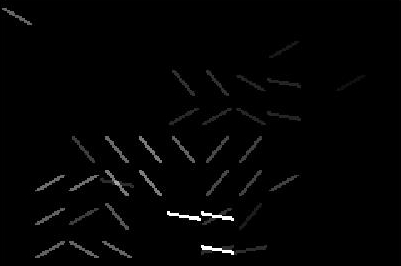}
		\includegraphics[width=0.19\linewidth]{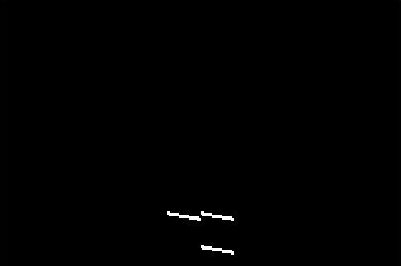}
		\includegraphics[width=0.19\linewidth]{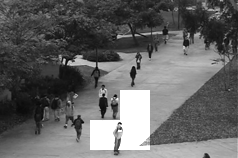}
		\includegraphics[width=0.19\linewidth]{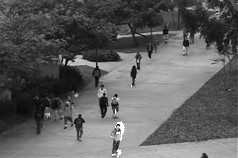}
		
		\includegraphics[width=0.19\linewidth]{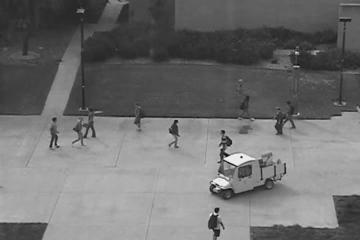}
		\includegraphics[width=0.19\linewidth]{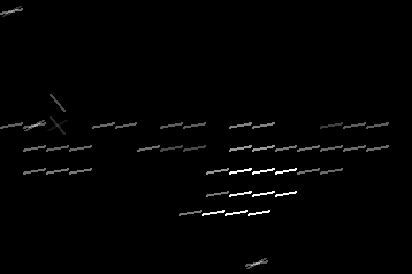}
		\includegraphics[width=0.19\linewidth]{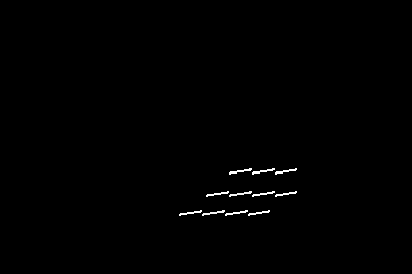}
		\includegraphics[width=0.19\linewidth]{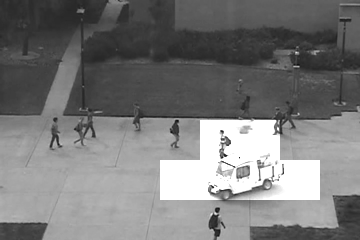}
		\includegraphics[width=0.19\linewidth]{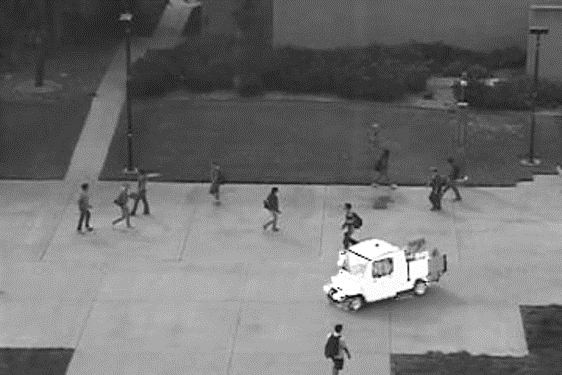}
		
		\includegraphics[width=0.19\linewidth]{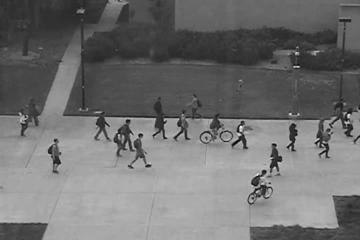}
		\includegraphics[width=0.19\linewidth]{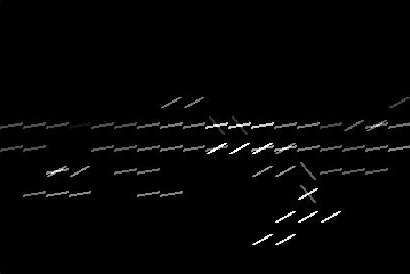}
		\includegraphics[width=0.19\linewidth]{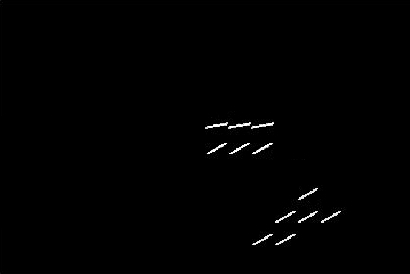}
		\includegraphics[width=0.19\linewidth]{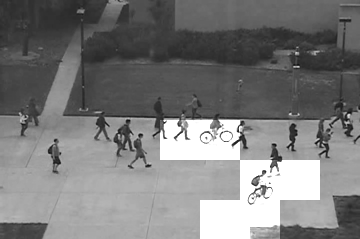}
		\includegraphics[width=0.19\linewidth]{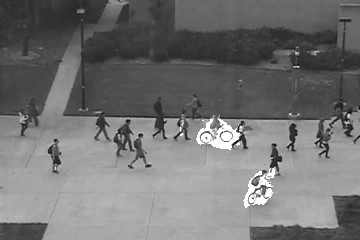}
	\end{center}
	\caption[LBT: Visualization of abnormality localization on UCSD dataset]{\small{Visualization of abnormality localization on UCSD dataset: (a) original frame, (b) visualization of Local Binary Tracklets, (c) visualization of early fusion of LBT+DOF (d) LBT-only abnormality detection, and (e) late fusion of LBT+DOF for abnormality detection filtered out by a threshold.}}
	\label{fig:fig4}
\end{figure*}
\end{landscape}

%
In Fig. \ref{fig:fig4}, the qualitative results for abnormal object localization are presented for a few sample frames of UCSD dataset (Ped1 and Ped2): (a) is illustrated the original frame, (b,c) the visualization of LBT and early fusion LBT+DOF, where the saliency points show abnormal event in the video sequence, (d) shows the detection result for LBT, and (e) illustrating the results for late fusion of LBT+DOF under the best-defined threshold. As reviewed in Sec. \ref{sec:detect}, a combination of LBT and DOF descriptors is used to exploit the benefits of each of them. Fig. \ref{fig:fig4} shows the result of abnormality localization.

\noindent{\textbf{Evaluation on UMN dataset.} }In this experiment, we exploited an evaluation protocol by separating UMN three scenes. The results on each scene and full dataset are presented in Tab. \ref{tbl:tab2} in terms of AUC. On the contrary to earlier approaches which used LDA (Latent Dirichlet Allocation) \cite{mousavi2015analyzing}, we used one-class SVM \cite{Alpher24}. Results are reported in Tab. \ref{tbl:tab2} which is comparable with the state-of-the-art.
\begin{table}
	\begin{center}
		\scalebox{1.2}{
			\small{
				\begin{tabular}[width=\textwidth]{lcccc}
					\toprule
					Method & scene-1 & scene-2 & scene-3 & all scenes \\
					\midrule
					Social force model (SFM) \cite{mehran2009abnormal} 		& 	0.990& 	0.949 & 		0.989		& 0.960\\
					Cem et al.~\cite{direkoglu2017abnormal}& 0.991& 0.948 & 0.977 &	0.964\\
					HMOFP~\cite{7471984}& 0.997& 		0.957 & 			0.986		& 0.980\\
					Biswas et al.~\cite{Biswas2017}& \textbf{0.999}& 		0.971 & 			0.983		& 0.984\\
					Yu et al.~\cite{yu2017abnormal} 		&---& 		--- & 			---		&	0.972\\
					\midrule
					LBT & 	0.988& 	\textbf{0.975} & 		\textbf{0.997}	& \textbf{0.986}\\
					\bottomrule
		\end{tabular}}}
	\end{center}
	\caption{\small{Comparison AUCs on UMN dataset for all the scenes}}
	\label{tbl:tab2}
\end{table}

\noindent{\textbf{Evaluation on Violent in Crowd dataset.} }In this experiment, a one-class SVM with a linear kernel is trained on a set of normal and abnormal video sequences from Violence-in-Crowd dataset across a five-fold cross-validation. The goal is to assign a normal/abnormal label to an input video rather a frame. The video level descriptor $D^v$ of an input video $v$ is simply computed by $D^v = \sum_{f_t \in v} D^{f_t}$, where $D^{f_t}$ is defined in Eq.~\ref{eq:d} as a descriptor for $t$-th frame of $v$.
In order to train the one-class SVM, the training set as ${\cal D} = \{ D^{v} \}_{v=1}^{V}$ are formed where $V$ is the number of training videos. The result of classification accuracy in Tab. \ref{tbl:tab3} shows competitive performance of our method.

\begin{table}
	\begin{center}
		\scalebox{0.99}{
			\begin{tabular}[width=\textwidth]{lcccccc}
				\toprule
				Method      &   LTP    &   HNF    &   VIF    &     OVIF& FF& LBT 		\\
				&   \cite{45}    &   \cite{44}    &   \cite{26}    &   \cite{gao2016violence} & \cite{gracia2015fast}& 		\\
				\midrule
				Accuracy   &   71.53\%           &   56.52\%         &   81.30\%            & 76.80\% &69.4\% & \textbf{81.90\%}\\
				\bottomrule
			\end{tabular}
		}
	\end{center}
	\caption{\small{Classification results on Violence Crowd dataset.}}
	\label{tbl:tab3}
\end{table}
\subsection{Discussions}
\label{sec:disc}

\begin{figure*}[h!]
	\begin{center}
		\includegraphics[width=0.30\linewidth]{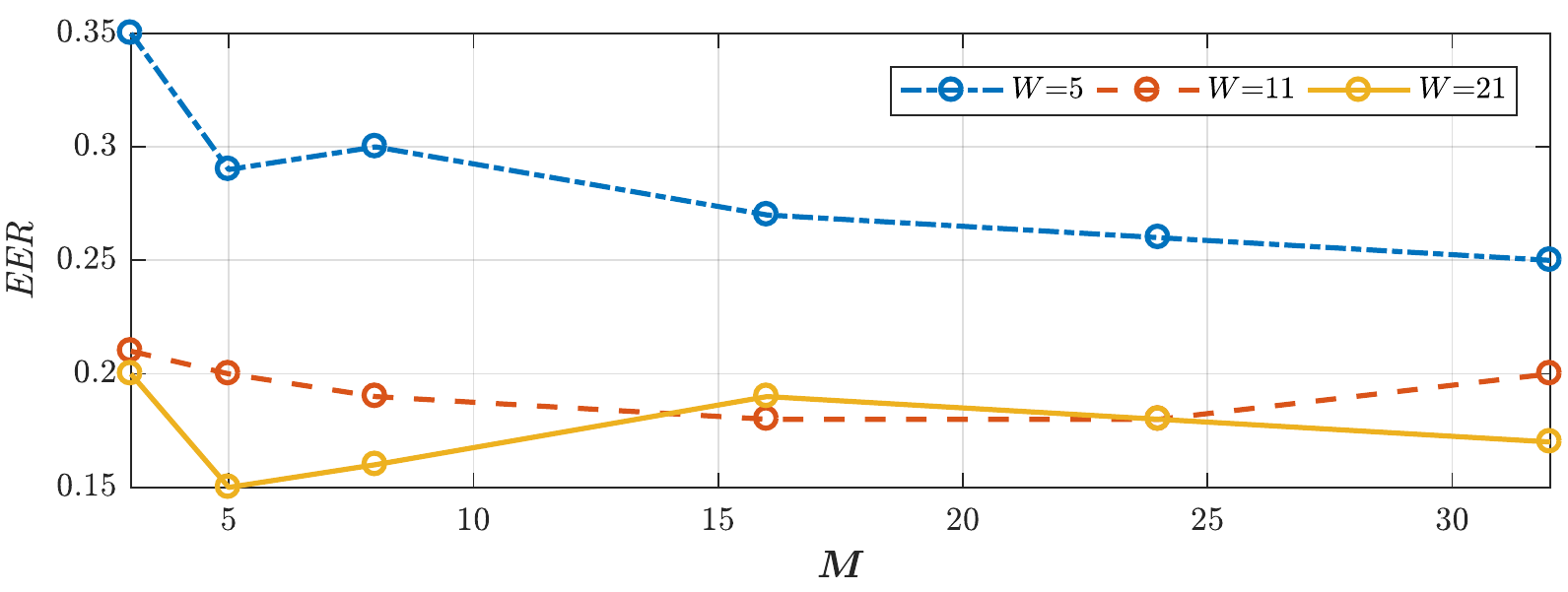}
		\includegraphics[width=0.30\linewidth]{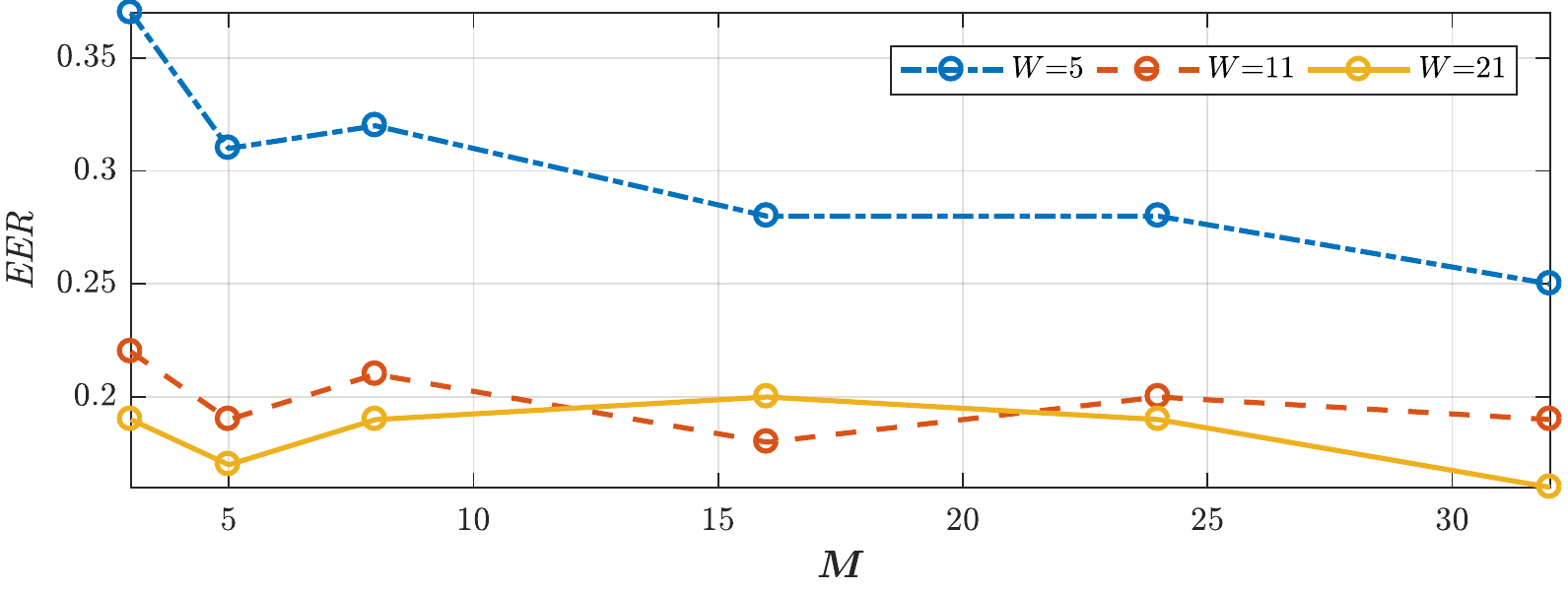}
		\includegraphics[width=0.30\linewidth]{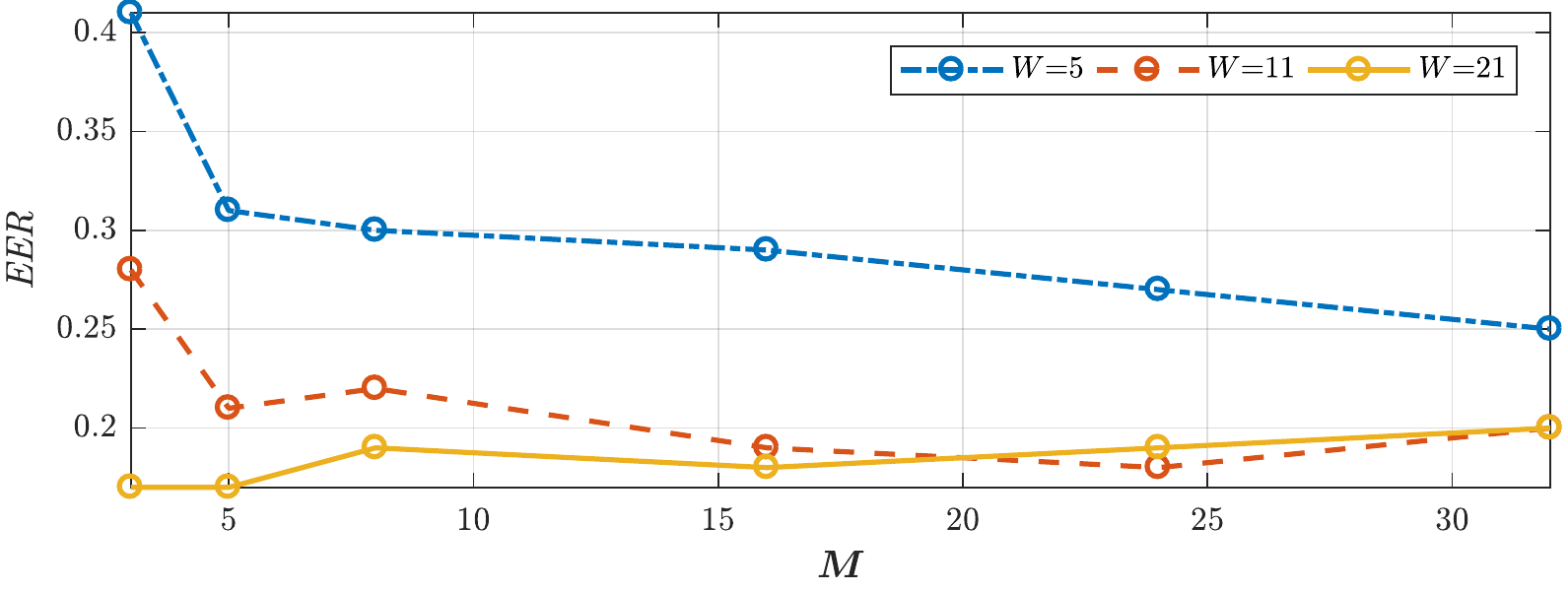}
		
		\hspace{0.3 cm} (a) \hspace{5.1 cm} (b) \hspace{5 cm} (c)
	\end{center}
	\caption[Results for PED2 varying the number of magnitude bins $M$]{Results for PED2 varying the number of magnitude bins $M$, over different temporal window size $W$: (a) large tessellation size , (b) medium tessellation size (c) small tessellation size}
	\label{fig:ab_m}
\end{figure*}

\begin{figure*}
	\begin{center}
		\includegraphics[width=0.30\linewidth]{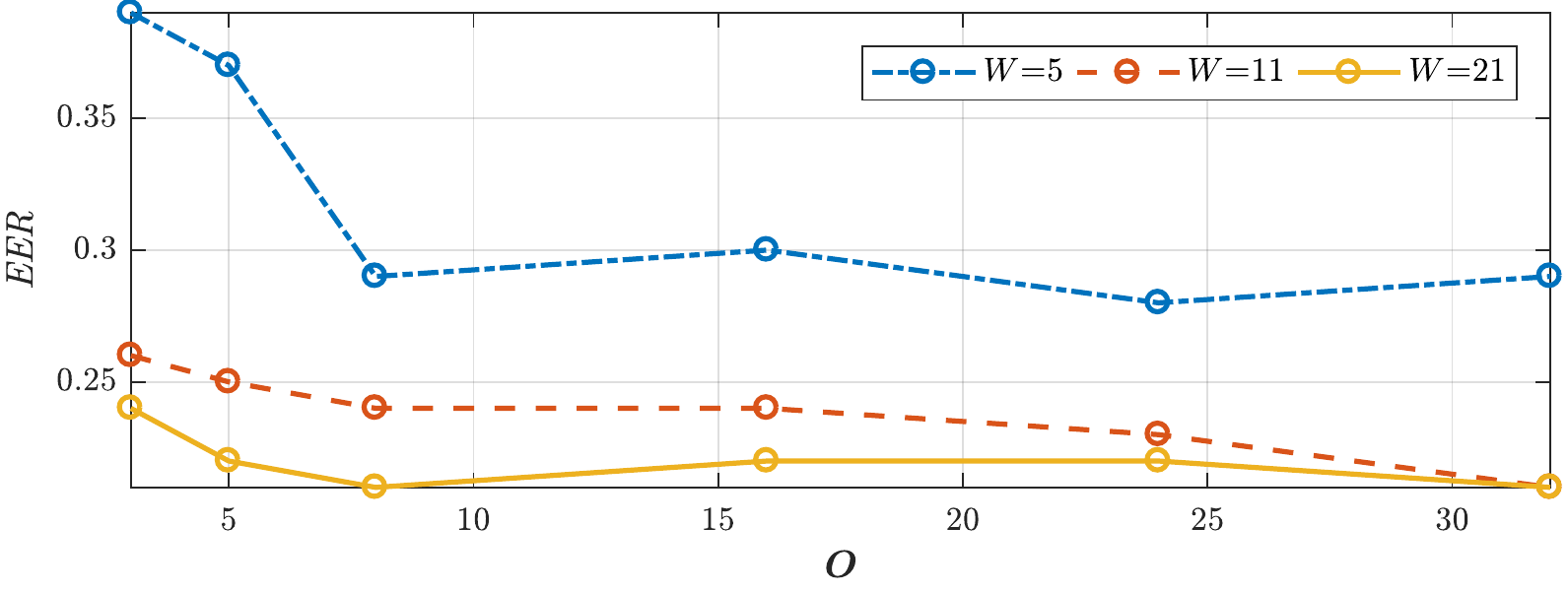}
		\includegraphics[width=0.30\linewidth]{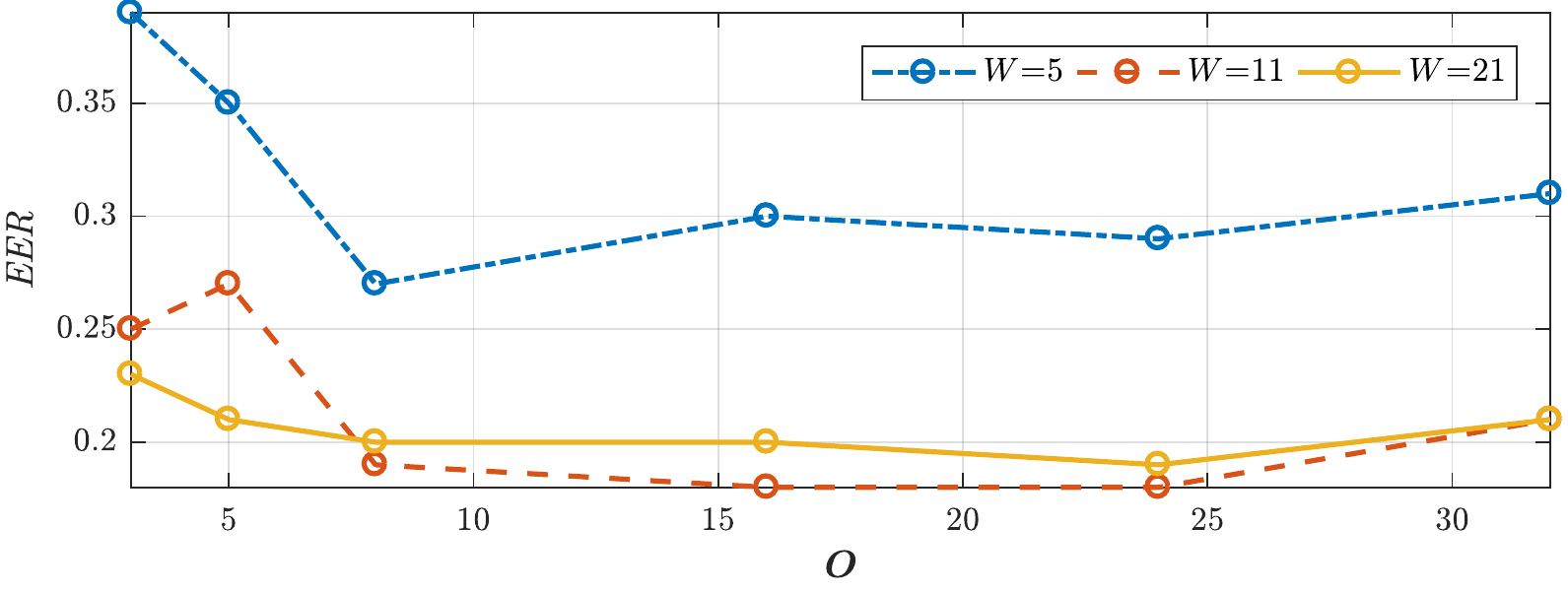}
		\includegraphics[width=0.30\linewidth]{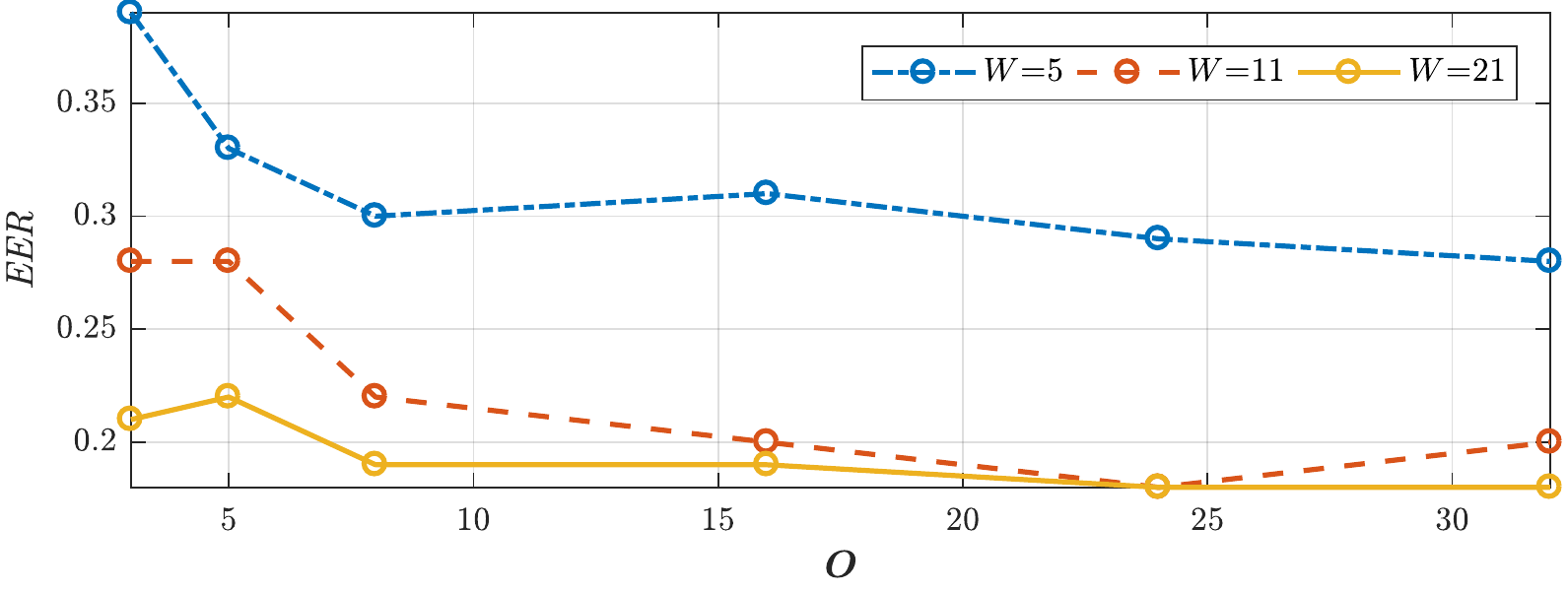}
		
		\hspace{0.3 cm} (a) \hspace{5.1 cm} (b) \hspace{5 cm} (c)
	\end{center}
	\caption[Results for PED2 varying the number of magnitude bins $O$]{Results for PED2 varying the number of magnitude bins $O$, over different temporal window size $W$: (a) large tessellation size , (b) medium tessellation size (c) small tessellation size}
	\label{fig:ab_o}
\end{figure*}

\noindent{\textbf{Parameter sensitivity analysis.} } The main parameters of proposed Local Binary Tracklets (LBT) are including: temporal window $W$, length of the tracklet $L$, tessellation size $S$, the orientation quantization bins $O$, and quantization levels of magnitude $M$. 
In order to study the parameters we fixed some parameters while changing the others. Specifically, two different set of experiments are performed in $PED2$ dataset: 
\emph{i)} First, by setting the magnitude quantization level $M = 5$, the performance of the method is analyzed over the other parameters, reported in Fig. \ref{fig:ab_o}. 
\emph{ii)} Then, by setting the orientation bins $O = 8$, the same procedure followed to analyze the performance over the other parameters, shown in Fig. \ref{fig:ab_m}.
The variational parameters are:
\begin{itemize}
    \item Three level of spatial tessellation sizes: large tessellation size ($S= 2 \times 3$), medium tessellation size ($S= 4 \times 6$), and small tessellation size ($S= 8 \times 12$).
    \item Three different size of window $W \in {5,11,21}$ for short, medium and long temporal window.
\end{itemize}

In Fig. \ref{fig:ab_o} and Fig. \ref{fig:ab_m} the result of this study is reported. The results are presented in terms of EER, where the smaller value in the error rate is desired. In this figures, the vertical axis representing the EER values, and the horizontal axis shows $M$ and $O$ for the number of magnitude quantization levels and orientation bins, respectively.
Clearly, the EER is more stable while $ 5 < M < 24 $ and $ 8 < O < 24 $, with medium and large temporal window size. In general, the medium spatial tessellation size worked slightly better, however this is a parameter that depends on the resolution of the monitoring camera. Such study justifies the choice of selection for $M = 5$ and $O = 8$ and $W = 11$.

\begin{table}
	\begin{center}
		\begin{tabular}[width=\textwidth]{l cc}
			\toprule
			Phase &	Ours	&	AMDN~\cite{XU2017117}\\
			\midrule
			Feature Extraction       &   0.009					& 0.110 \\
			Abnormality Detection    &   0.006					& 0.035 \\
			\midrule
			
			TOTAL 			    &   \textbf{0.015}				& 0.145 \\
			\bottomrule
		\end{tabular}
	\end{center}
	\caption{Average run-time (seconds) for one frame testing.}
	\label{tbl:comp.times}
\end{table}

\noindent{\textbf{Computational costs.} }Local Binary Tracklet is proposed as an efficient descriptor for abnormality detection. To evaluate the computational cost of the proposed method, we computed the computational cost of the LBT, and compared it with a one of the recent deep-learning-based method \cite{XU2017117} for abnormality detection. In Tab.~\ref{tbl:comp.times} we report the corresponding average computation times for one frame. 
The pre-processing operations (i.e., the optical-flow computation) are the same for both methods, while feature extraction, post-processing procedure are the main difference. 
The reported computation times in Tab.~\ref{tbl:comp.times} show at testing time LBT is much faster, while this is obvious that the training time of our method is much less than a deep network.

\graphicspath{{Chapter2/Figs/Vector/}{Chapter2/Figs/b/}}
\section{Plug-and-Play CNNs}
\label{sec:off_the_shelf}

	In the crowd analysis studies, most of the methods \cite{amraee2017anomaly,chaker2017social,del2016discriminative,emonet2011multi,lu2013abnormal,ResnetCrowd2017,mehran2009abnormal, mousavi2015abnormality,rabiee2016crowd,rabiee2016emotion,raghavendra2013anomaly,regazzoni1993real,xu2015learning} mainly rely on complex hand-crafted features to represent the crowd motion and appearance. However, the use of hand-crafted features is a clear limitation, as it implies task-specific a priori knowledge which, in case of a complex video-surveillance scene, can be very difficult to define. 
	
	Recently, Deep Neural Networks have resurfaced as a powerful tool for learning from big data (e.g., ImageNet~\cite{deng2009imagenet} with 1.2M images), providing models with excellent representational capacities. Specifically, Convolutional Neural Networks (CNNs) have been trained via backpropagation through several layers of convolutional filters. It has been shown that such models are not only able to achieve state-of-the-art performance for the visual recognition tasks in which they were trained, but also the learned representation can be readily applied to other relevant tasks~\cite{razavian2014cnn}. These models perform extremely well in domains with large amounts of training data. With limited training data, however, they are prone to overfitting. This limitation arises often in the abnormal event detection task where scarcity of real-world training examples is a major constraint. Besides the insufficiency of data, the lack of a clear definition of abnormality (\emph{i.e.}, the context-dependent nature of the abnormality) induces subjectivity in annotations. Previous work highlighted the fact that the unsupervised measure-based methods may outperform supervised methods, due to the subjective nature of annotations as well as the small size of training data~\cite{mousavi2015crowd,sodemann2012review,xiang2008video,xiong2012energy}.
	
	Attracted by the capability of CNN to produce a general-purpose semantic representation, in this section we investigate how to employ CNN features, trained on large-scale image datasets, to be applied to a crowd dataset with few abnormal event instances. This can alleviate the aforementioned problems of supervised methods for abnormality detection, by leveraging the existing CNN models trained for image classification. Besides, training a CNN with images is much cheaper than with videos; therefore, representing a video by means of features learned with static images represents a major saving of computational cost.
	
	The key idea behind our method is to track the changes in the CNN features across time. We show that even very small consecutive patches may have different CNN features, and this difference captures important properties of video motion. To capture the temporal change in CNN features, we cluster them into a set of binary codes each representing a binary pattern {\em(prototype)}. Intuitively, in a given video block consecutive frames should have similar binary patterns unless they undergo a significant motion. We introduced a simple yet effective statistical measure which captures the local variations of appearance in a video block. 
	We show that combining this measure with traditional optical-flow information, provides the complementary information of both appearance and motion patterns.

	Our method is different from~\cite{cong2011sparse,kim2009observe,lu2013abnormal,mahadevan2010anomaly,ResnetCrowd2017,mehran2009abnormal,mousavi2015analyzing,saligrama2012video}, which focus on learning models on motion and/or appearance features. A key difference compared to these methods is that they employ standard hand-crafted features (\eg, optical-flow, Tracklets, etc.) to model activity patterns, whereas our method proposes using modern deep architectures for this purpose. The advantages of a deep learning framework for anomalous event detection in crowd have been investigated recently in~\cite{sabokrouFFK16,xu2015learning}. Nevertheless, deep networks are data-hungry and need large training datasets. In our work, however, a completely different perspective to abnormality detection is picked out. We specifically propose a measure-based method which allows the integration of semantic information (inherited from existing CNN models) with low-level optical-flow \cite{brox2004high}, with minimum additional training cost. This leads to a more discriminative motion representation while maintaining the method complexity to a manageable level. Most related to our approach is the work by Mousavi \etal~\cite{mousavi2015crowd}, which introduced a similar measure to capture the commotion of a crowd motion for the task of abnormality detection. Instead of capturing the local irregularity of the low-level motion features (e.g., tracklets in~\cite{mousavi2015crowd}) or high-level detectors~\cite{nabi2013temporal}, we propose to represent the crowd motion exploiting the temporal variations of CNN features. This provides the means to jointly employ appearance and motion. Recently Ravanbakhsh \etal~\cite{ravanbakhsh2015action} proposed a complex feature structure on top of CNN features which can capture the temporal variation in a video for the task of activity recognition. However, to our knowledge this is the first work proposing to employ the existing CNN models for motion representation in crowd analysis.
\begin{figure*}
		\begin{center}
			\includegraphics[width=\linewidth]{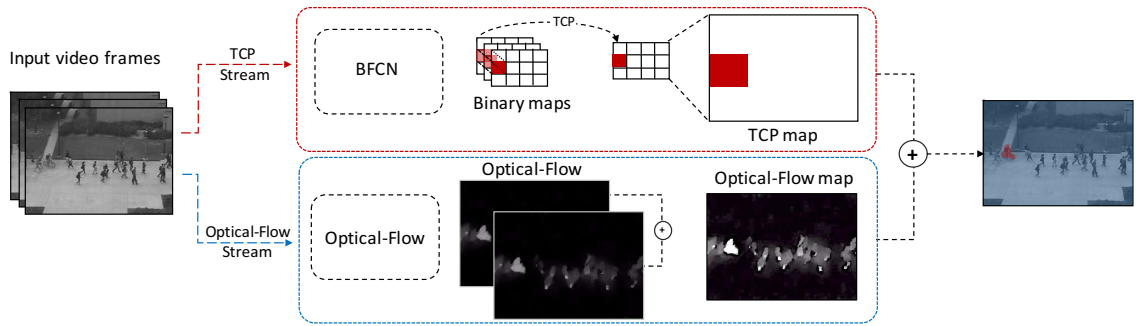}
		\end{center}
		\caption[Plug-and-Play: Overview of the proposed method]{Overview of the proposed method}
		\label{fig:short}
	\end{figure*}
	
	The proposed method (see Fig. \ref{fig:short}) is composed of three main steps: 1) Extract CNN-based binary maps from a sequence of input frames, 2) Compute the Temporal CNN Pattern (TCP) measure using the extracted CNN-binary maps 3) The TCP measure fuse with low-level motion features (optical-flow) to find the refined motion segments. 
	More specifically, all the frames are input to a Fully Convolutional Network (FCN). Then we propose a binary layer plugged on top of the FCN in order to quantize the high-dimensional feature maps into compact binary patterns. The binary quantization layer is a convolutional layer in which the weights are initialized with an external hashing method. The binary layer produces binary patterns for each patch corresponding to the receptive field of the FCN, called {\em binary map}. The output binary maps preserve the spatial relations in the original frame, which is useful for localization tasks. Then, a histogram is computed over the output binary maps for aggregating binary patterns in a spatio-temporal block. In the next step, an \emph{irregularity} measure is computed over these histograms, called TCP measure. 
	Eventually, all the computed measures over all the video blocks are concatenated, up-sampled to the original frame size, and fused with optical-flow information to localize possible abnormalities. In the rest of this section we describe each part in detail.
	
Our major contributions are: \emph{(i)} We introduce a novel Binary Quantization Layer, \emph{(ii)} We propose a Temporal CNN Pattern measure to represent motion in crowd, \emph{(iii)} The proposed method is tested on the most common abnormality detection datasets and the results show that our approach is comparable with the state-of-the-art methods.

	The rest of the section is organized as follows: the Binary Quantization Layer is introduced in Sec.~\ref{sec:net}. In Sec.~\ref{sec:tcp} we show the proposed measure, while our feature fusion is shown in Sec.~\ref{sec:fuse}. The experiments and a discussion on the obtained results is presented in Sec.~\ref{sec:exp}. 
	\subsection{Binary Fully Convolutional Networks (BFCN)}
	\label{sec:net}
	In this section, we present the sequential Fully Convolutional Network (FCN) which creates the binary maps for each video frame. The proposed architecture contains two main modules: {\it 1) the convolutional feature maps}, and {\it 2) binary map representations of local features}. In the following, we describe each part in details.

    \noindent\textbf{Frame-based fully convolutional network.}
	\label{sec:convNet}
	Early layers of convolutions in deep nets present local information about the image, while deeper convolutional layers contain more global information. The last fully connected layers in a typical CNN represent high-level information and usually can be used for classification and recognition tasks. It has been shown that deep net models trained on the ImageNet~\cite{russakovsky2015imagenet} encode semantic information, thus can address a wide range of recognition problems~\cite{donahue2013decaf,rastegari2015computationally,razavian2014cnn}. Since, FCNs do not contain fully-connected layers they preserve a relation between the input-image and the final feature-map coordinates. Hence a feature in the output map corresponds to a large receptive field of the input frame. Moreover, FCNs can process input images of different sizes and return feature maps of different sizes as output. In light of the above, this deep network typology is useful to both extract local and global information from an input image and to preserve spatial relations, which is a big advantage for a localization task.
	
\begin{figure*}[t]
		\begin{center}
			\includegraphics[width=\linewidth]{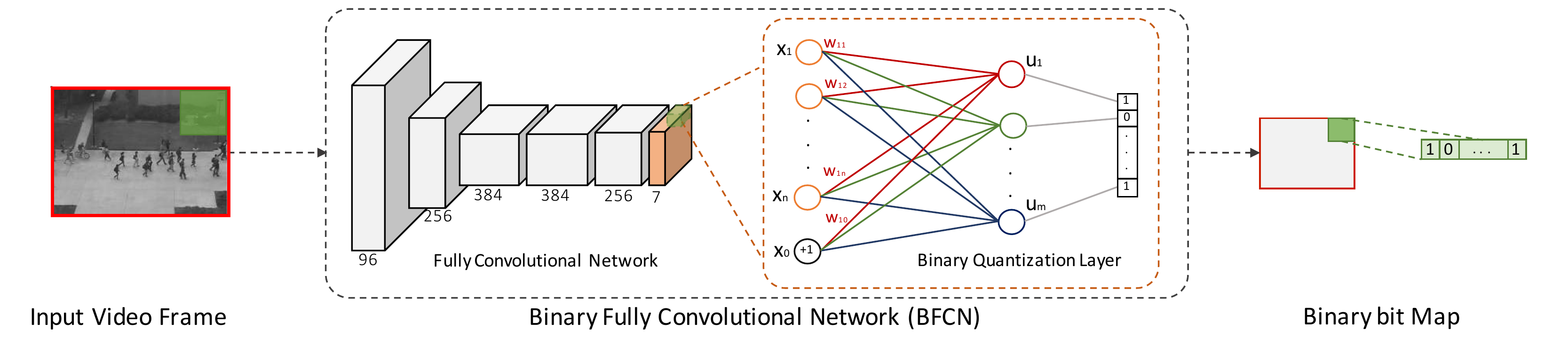}
		\end{center}
		\caption[The Fully Convolutional Net with Binary Quantization Layer]{The Fully Convolutional Net with Binary Quantization Layer, is composed of a fully convolutional neural network followed by a Binary Quantization Layer (BQL). The BQL (shown in orange) is used to quantize the $pool5$ feature maps into 7-bit binary maps.}
		\label{fig:net}
	\end{figure*}

	\noindent\textbf{Convolutional feature maps.}
	To tackle the gap between the raw-pixel representation of an image and its high-level information we choose the output of the last convolutional layer to extract feature maps. These components provide global information about the objects in the scene. To extract convolutional feature maps, we used a pre-trained AlexNet~\cite{alexnet} model. AlexNet contains 5 convolutional layers and two fully connected layers. In order to obtain spatially localizable feature maps, we feed the output feature maps of the last convolutional layer into our binary quantization layer. Fig.~\ref{fig:net} illustrates the layout of our network.
	
	\noindent\textbf{Binary Quantization Layer (BQL).}
	\label{sec:binary}
	In order to generate a joint model for image segments there is a need to cluster feature components. Clustering the extracted high-dimensional feature maps comes with a high computational cost. The other problem with clustering is the need to know a priori the number of cluster centres. One possible approach to avoid extensive computational costs and to obtain reasonable efficiency is clustering high-dimensional features with a hashing technique to generate small binary codes~\cite{itq}. A 24-bits binary code can address $2^{24}$ cluster centres, which is very difficult to be handled by common clustering methods. Moreover, this binary map can be simply represented as a 3-channels RGB image. Dealing with binary codes comes with a lower computational cost and a higher efficiency compared with other clustering methods. The other advantage of using a hashing technique in comparison with clustering is the ability of embedding the pre-trained hash function/weights as a layer inside the network.
	
	Encoding feature maps to binary codes is done using Iterative Quantization Hashing (ITQ)~\cite{itq}, which is a hashing method for binary code unsupervised learning \cite{gong2013iterative}. Training ITQ is the only training cost in the proposed method, which is done only once on a subset of the train data. ITQ projects each high-dimensional feature vector into a binary space. We use the hashing weights, which are learned using ITQ, to build our proposed Binary Encoding Layer (denoted by $hconv6$). 
	Specifically, inspired by~\cite{long2015fully,ravanbakhsh2016efficient,2016icip} we implement this layer as a set of convolutional filters (shown in different colors in Fig. \ref{fig:net}), followed by a sigmoid activation function. The number of these filters is equal to the size of the binary code and the weights are pre-computed through ITQ. Finally, the binarization step has been done externally by thresholding the output of the sigmoid function.
	
	Specifically, if $X=\{x_{1},x_{2},...,x_{n}\}$ is a feature vector represented in $pool5$, the output of $hconv6$ is defined by $hconv6(X) = XW_{i}$, where $W_{i}=\{w_{i1},w_{i2},...,w_{in}\}$ are the weights for the $i^{th}$ neuron. The non-linearity is provided by a sigmoid function $\upsilon= \sigma(hconv6(X))$, and the threshold function is defined by:
	\begin{equation}
	\label{eq:activation}
	g(\upsilon) = \Big\{
	\begin{tabular}{cc}
	0, & $\upsilon \le 0.5$\\
	1, & $\upsilon > 0.5$
	\end{tabular}
	\end{equation}
	Eventually, for any given frame our network returns a binary bitmap, which can be used for localization tasks. 
	
	Such a binary quantization layer can be plugged into the net as a pre-trained module, with the possibility of fine-tuning with back-propagation in an end-to-end fashion.However, in our abnormality task, due to the lack of data, fine-tuning is difficult and can be harmful because of possible overfitting, so all our experiments are obtained without fine-tuning.
	
	\noindent\textbf{Sequential BFCN.} Let $\textbf{v} = \{f_{t}\}_{t=1}^{T}$ be an input video, where $f_{t}$ is the $t$-th frame of the video, and $T$ is the number of frames. The frames of the video are fed into the network sequentially. The output for a single frame $f_{t} \in \textbf{v}$, is an encoded binary bit map (prototype), denoted as $m_{t}$. All the binary maps are stacked and provided as the final representation of a given video, \emph{i.e.}, $\textbf{M} = \{m_{t}\}_{t=1}^{T}$.
	\subsection{Temporal CNN Pattern (TCP)}
	\label{sec:tcp}
	In this section we describe our proposed method to measure abnormality in a video sequence.
	
	\noindent\textbf{Overlapped video blocks.}
	The first step is extracting video blocks from the video clip. As mentioned in Sec.~\ref{sec:net}, for a given video frame $f_{t}$ the output of our FCN is a binary bit map denoted by $m_{t}$, which is smaller in size than the original image. In the binary bit map representation, each pixel describes a corresponding region in the original frame. This feature map partitions the video frame into a certain number of regions, which we called $patch$ denoted by $p_{t}^{i}$ where $t$ is the frame number and $i$ is the $i$-th patch in the frame. $b_{t}^{i}$ is a set of corresponding patches along consecutive frames. The length of a video blocks is fixed, and the middle patch of the video block is selected to indicate the entire video block. If the length of a video block $b_{t}^{i}$ is $L+1$, it starts \nicefrac{L}{2} frames before the frame $t$ and ends \nicefrac{L}{2} frames after that, namely $\{b_{t}^{i}\}=\{p_{l}^{i}\}_{l=t-L/2}^{t+L/2}$. To capture more fine information, the video block $b_{t}^{i}$ has $n$ frames overlapped with the next video block $b_{t+1}^{i}$.
	
	\noindent\textbf{Histogram of binary codes.}
	A video block is represented with a set of binary codes (prototypes), which encode the similarity in appearance. We believe that observing the changes of prototypes over a time interval  is a clue to discover motion patterns. Toward this purpose, for each video block $b_{t}^{i}$ a histogram $h_{t}^{i}$ is computed to represent the distribution of prototypes in the video block.
	
	\noindent\textbf{TCP measure.}
	Similarly to the commotion measure~\cite{mousavi2015crowd}, to obtain the TCP measure for a given video block $b_{t}^{i}$, the \emph{irregularity} of histogram $h_{t}^{i}$ is computed. This is done by considering the fact that, if there is no difference in the appearance, then there is no change in descriptor features and consequently there is no change in the prototype representation. When the pattern of binary bits changes, it means that different appearances are observed in the video block and this information is used to capture motion patterns. The \emph{irregularity} of the histogram is defined as the non-uniformity of the distribution in the video block. 
	
	A uniform distribution of a histogram shows the presence of several visual patterns in a video block. The higher diversity of the prototypes on a video block leads to a low \emph{irregularity} of the histogram. More uniform histograms increase the chance of abnormality. Such \emph{irregularity} in appearance along the video blocks either is generated by noise or is the source of an anomaly. We took advantage of this fact to present our TCP measure. The TCP measure for each video block $b_{t}^{i}$, is computed by summing over the differences between the prototype samples in $h_{t}^{i}$ and the dominant prototype. The dominant prototype is defined as the most frequent binary code in the video block, which has the maximum value (mode) in the histogram $h_{t}^{i}$.
	
	Let $H^{n}$ represent the histogram of binary codes of all patches $\{ p_{t}^{i} \}$ in the video block $b_{t}^{i}$ denoted by $\{ H^{n} \}_{n=1}^{N}$, where $N$ is the number of patches in the video block. The aggregated histogram for block $b_{t}^{i}$ compute as $ \mathcal{H}_{t}^{i} = \sum_{n=1}^{N} H^{n}$. The aggregated histogram $ \mathcal{H}_{t}^{i}$ represents the distribution of the appearance binary codes over the video block $b_{t}^{i}$, and the TCP measure compute as follows:
	\begin{equation}
	\label{eq:coapp}
	tcp(b_{t}^{i}) = \sum\limits_{j=1}^{|\mathcal{H}_{t}^{i}|} {|| \mathcal{H}_{t}^{i}(j) - \mathcal{H}_{t}^{i}(j_{max}) ||_{2}^{2}}
	\end{equation}
	where $|.|$ is the number of bins of the histogram, $||.||_{2}$ is the L2-norm, and the dominant appearance index over the video block is denoted by $j_{max}$ ({\em i.e.}, the mode of $\mathcal{H}_{t}^{i}$).
	
	\noindent\textbf{TCP map.}
	To create a spatial map of the TCP measure $c_{t}$ for any given frame $f_{t}$, the TCP measure is computed for all video blocks $b_{t}^{i}$, and we assign the value of $c_{t}^{i}$ to a patch that is temporally located at the middle of the selected video block. The output $c_{t}$ is a map with the same size as the binary map $m_{t}$ which contains TCP measure values for each patch in the frame $f_{t}$. Finally, the TCP maps are extracted for the entire video footage. We denote the TCP map for frame $f_{t}$ as $c_{t} = \{c_{t}^{i}\}_{i=1}^{I}$, where $I$ is the number of patches in the frame.
	
	\noindent\textbf{Up-sampling TCP maps.}
	Since the frame will pass through several convolution and pooling layers in the network, the final TCP map is smaller than the original video frame. To localize the exact region there is a need to produce a map of the same size as the input frame. For this reason, the TCP value in the map is assigned to all pixels in the corresponding patch of the frame on the up-sampled TCP map.

	\subsection{Motion-aware TCP}
	\label{sec:fuse}
	Since the Up-sampled TCP map can only detect the coarse region of abnormality, we propose to fuse optical-flow with the TCP maps in order to have a more accurate localization.
	The optical-flow \cite{brox2004high} is extracted from each two consecutive frames. However the TCP map is computed for each $L$ frames. To be able to fuse the optical-flow with the corresponding extracted TCP map, an aligned optical-flow map is constructed. Suppose that $f_{t}$ and $f_{t+1}$ are two consecutive frames from video $\textbf{v} = \{f_{t}\}_{t=1}^{T}$, optical-flow map $of_{t}$, with the same resolution of an input frame, represents the optical-flow values transition between the two frames. optical-flow values are extracted for entire video footage $\textbf{v}$ and stacked as optical-flow sequences $\{d_{t}\}_{t=1}^{T-1}$. Finally, similar to the overlapped video block extraction protocol, overlapped optical-flow maps are computed. If the length of a video block $p_{t}^{i}$ is $L+1$, then the corresponding optical-flow map $d_{t}^{i}$ is the sum of all optical-flow values over the corresponding $i$-th region as $d_{t}^{i} = \sum_{l=1}^{L} {d_{t}^{i}(l)}$. The optical-flow map for entire frame $f_{t}$ is described as $d_{t} = \{d_{t}^{i}\}_{i=1}^{I}$.
	
	
	\noindent\textbf{Feature fusion.}
	The extracted optical-flow maps and the computed TCP maps for each video frame are fused together with importance factors $\alpha$ and $\beta$ to create motion segment map: $mseg_{t} = \alpha d_{t} + \beta c_{t} \;, \; mseg = \{mseg_{t}\}_{t=1}^{T}$, where, $\{mseg\}$ is the extracted motion segments along the entire video $\textbf{v}$. The importance factors indicates the influence of each fused map in the final segment motion map, we simply select $0.5$ for both $\alpha$ and $\beta$.
	
	\subsection{Experiments}
	\label{sec:exp}
	In this section, we evaluate our method over two well-known crowd abnormality datasets and compare our results with state of the art. The evaluation has been performed with both a {\em pixel-level} and a {\em frame-level} protocol, under standard setup. The rest of this section is dedicated to describing the evaluation datasets, the experimental setup and the reporting the results quantitatively and qualitatively.
	
	\noindent\textbf{Datasets and experimental setup.}
	In order to evaluate our method two standard datasets: UCSD Anomaly Detection Dataset~\cite{mahadevan2010anomaly} and UMN SocialForce~\cite{mehran2009abnormal}. 
	In our experiments to initialize the weights of $hconv6$ an ITQ is applied on the train set of UCSD pedestrian dataset with a 7-bits binary code representation, which addresses 128 different appearance classes. Video frames are fed to the BFCN sequentially to extract binary bit maps. All video frames are resized to $460 \times 350$, then BFCN for any given frame returns a binary bit map with resolution $8 \times 5$, which splits the frame into a 40-region grid.
	The length of video block extracted from a binary map is fixed to $L=14$ with 13 frames overlapping. The TCP measure is normalized over the entire video block sequence, then a threshold $th<0.1$ is applied for detecting and subtracting the background region in the video.
	
	Optical-flow feature maps are extracted to fuse with our computed features on the TCP measure maps. The fusion importance factor set to $0.5$ equally for both feature sets. These motion segment maps are used to evaluate the performance of our method on detection and localization of anomalous motions during video frames.
	\subsubsection{Quantitative evaluation}
	The evaluation is performed with two different levels: {\em frame level} for anomaly detection, and {\em pixel level} for anomaly localization. We evaluate our method on UCSD abnormality crowd dataset under the original setup~\cite{li2014anomaly}.
	\begin{figure*}
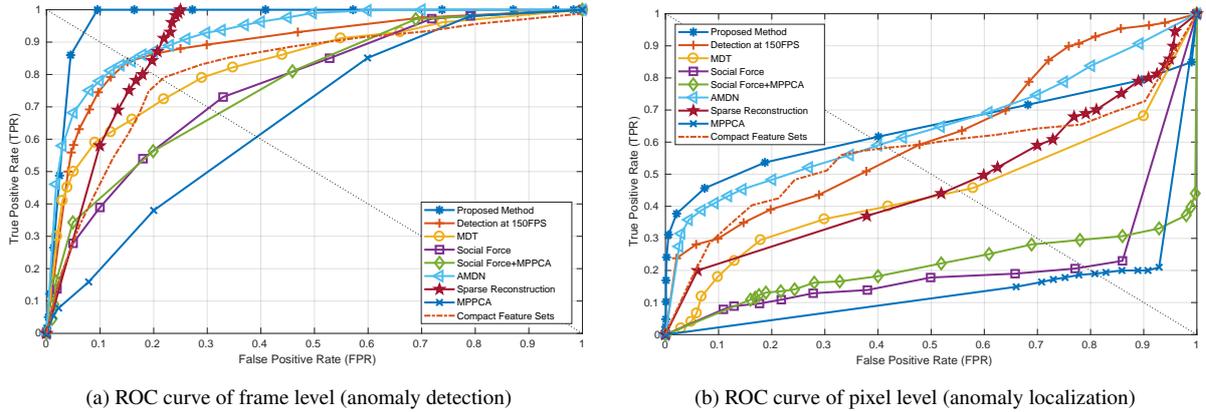

		\begin{center}
			\begin{tabular}{c c}
				\includegraphics[width=0.5\linewidth]{images/ped1_frame_level} & \includegraphics[width=.5\linewidth]{images/ped1_pixel_level} \\ 
				{\scriptsize(a) ROC curve of frame level (anomaly detection)} & {\scriptsize(b) ROC curve of pixel level (anomaly localization)}\\
			\end{tabular}	
		\end{center}
		\caption[Plug-and-Play: Comparison ROC curves of Ped1]{Frame level and Pixel level comparison ROC curves of Ped1 (UCSD dataset).}
		\label{fig:rocfrm_plg}
	\end{figure*}
	\begin{figure*}
			\begin{center}
				\includegraphics[width=0.91\linewidth]{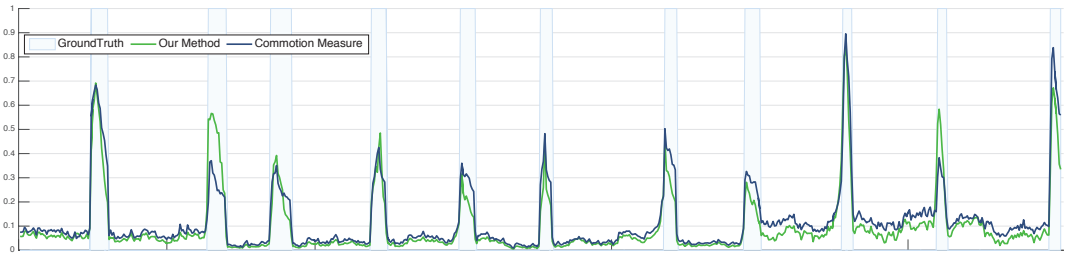}
			\end{center}
			\caption[Plug-and-Play: UMN anomaly detection signal]{Frame-level anomaly detection results on UMN dataset: Our method compares to Commotion Measure~\cite{mousavi2015crowd}. The green and blue signals respectively show the computed TCP by our approach and Commotion Measure over frames of 11 video sequences. The light blue bars indicate the ground truth abnormal frames of each sequence. All the sequences start with normal frames and ends with abnormality.}
			\label{fig:umn}
		\end{figure*}
	
	\noindent\textbf{Frame level anomaly detection.}
	This experiment aims at evaluating the performance of anomaly detection along the video clip. The criterion to detect a frame as abnormal is based on checking if the frame contains at least one abnormal patch. To evaluate the performances the detected frame is compared to ground truth frame label regardless of the location of the anomalous event. The procedure is applied over range of thresholds to build the ROC curve. We compare our method with state-of-the-art in detection performance on UCSD ped1 and ped2 datasets. The result is shown in Table~\ref{tbl:results}, beside the ROC curves on Fig.~\ref{fig:rocfrm_plg}.
	\begin{table}
		\begin{center}
			\begin{tabular}[width=\textwidth]{p{8cm}c}
				\toprule
				Method 														& 	AUC \\
				\midrule
				Optical-Flow~\cite{mehran2009abnormal}	&	0.84 \\
				Social force model (SFM)~\cite{mehran2009abnormal} 				&	0.96\\
				Del Giorno et al.\cite{del2016discriminative} & 0.910 \\
				Marsden et al. \cite{marsden2016holistic} & 0.929\\
				Singh and Mohan~\cite{singh2017graph}& 0.952\\
				Sparse Reconstruction~\cite{cong2011sparse}&0.976\\
				Commotion~\cite{mousavi2015crowd} 		&	\textbf{0.988}\\
				Yu et al.~\cite{yu2017abnormal} 		&	0.972\\
				Cem et al.~\cite{direkoglu2017abnormal} 		&	0.964\\
				Incremental slow feature analysis net.~\cite{5} 	&	0.969\\
				\midrule
				TCP (proposed method) 							&	\textbf{0.988}\\
				\bottomrule
			\end{tabular}
		\end{center}
		\caption[Plug-and-Play: Results on UMN dataset]{Results on UMN dataset. The values of previous methods are reported from~\cite{mousavi2015crowd}.}
		\label{tbl:auc_umn_plg}
	\end{table}
		The proposed method is also evaluated on UMN dataset. A comparison result is shown in Table \ref{tbl:auc_umn_plg}, beside the abnormality signal visualisation in Fig.~\ref{fig:umn}. Fig.~\ref{fig:umn} shows the computed TCP for each frame illustrated as ``detection  signal'' (green). We compared TCP with commotion measure (blue). The overall TCP value for a frame is computed from the sum of TCP measures over the patches in a frame and normalized in $[0,1]$ as an abnormality indicator. In Fig.~\ref{fig:umn}, the horizontal axis represents the time($s$), the vertical axis shows the ``abnormality indicator'', and the light blue bars indicate the ground truth labels for abnormal frames. 
		
		\noindent\textbf{Pixel level anomaly localization.}
	The goal of the pixel level evaluation is to measure the accuracy of anomalous event localization. Following~\cite{li2014anomaly}, detected abnormal pixels are compared to pixel level groundtruth. A true positive prediction should cover at least 40\% of true abnormal pixels over groundtruth, otherwise counted as a false positive detection. 
	Fig.~\ref{fig:rocfrm_plg} shows the ROC curves of the localization accuracy over USDC Ped1 and Ped2. We compare our method with state of the art in accuracy for localization. Result is presented in Table~\ref{tbl:results}.
	
	In our experiments we observed that in most of the cases the proposed method hit the abnormality correctly in terms of detection and localization. Only in some cases our measure achieved slightly lower accuracy in anomaly localization and the anomaly detection performance in compare with the state of the art methods.
		
	\begin{landscape}
		\begin{table}
		\begin{center}
			\begin{tabular}[width=\textwidth]{p{8cm} cc l cc l cc}
			\toprule
			\multirow{2}{*}{Method} &\multicolumn{2}{c}{Ped1 (frame-level)} & &\multicolumn{2}{c}{Ped1 (pixel-level)} & &\multicolumn{2}{c}{Ped2 (frame-level)}\\ \cmidrule{2-3} \cmidrule{5-6} \cmidrule{8-9}
			& EER & AUC & & EER & AUC && EER & AUC\\
				\midrule                                    
				MPPCA~\cite{kim2009observe} & 								40\%    & 	    59.0\%  && 		81\%    &   20.5\%  &&       30\%    & 69.3\%    \\
				Social force(SF)~\cite{mehran2009abnormal} & 				31\%    & 	    67.5\%  && 		79\%    &   19.7\%  & &		42\%    & 55.6\%    \\
				SF+MPPCA~\cite{mahadevan2010anomaly} & 			        32\%    & 	    68.8\%  && 		71\%    &   21.3\%  & &		36\%    & 61.3\%    \\
				SR~\cite{cong2011sparse} & 		                            19\%    &       ---     && 		54\%    &   45.3\%  & &		---     & ---       \\
				MDT~\cite{mahadevan2010anomaly} & 		                25\%    &       81.8\%  & &		58\%    &   44.1\%  & &		25\%    & 82.9\%    \\
				LSA~\cite{saligrama2012video} & 	                        16\%    & 	    92.7\%  & &		---     & 	---     & &		---     & ---       \\
				Detection at 150fps~\cite{lu2013abnormal} & 				15\%    & 	    91.8\%  & &		43\%    &   63.8\%  & &		---     & ---       \\
				AMDN (early fusion)~\cite{xu2015learning} & 				22\%    &       84.9\%  & &	    47.1\%  &   57.8\%  & &		24 \%   & 81.5\%    \\
				AMDN (late fusion)~\cite{xu2015learning} & 				    18\%    & 	    89.1\%  & &		43.6\%  &   62.1\%  & &		19 \%   & 87.3\%    \\
				AMDN (double fusion)~\cite{xu2015learning} & 			    16\%    & 	    92.1\%  & &		40.1\%  &   67.2\%  & &		17 \%   & \textbf{90.8\%}    \\
				SL-HOF+FC~\cite{wang2016anomaly}&                           18\%    & 	    87.45\% & &		\textbf{35\%}    &   64.35\% && 		19\%    & 81.04\%   \\
				Spatiotemporal Autoencoder~\cite{chong2017abnormal}&        12.5\%  & 	    89.9\%  & &		---     &   ---     & &		\textbf{12\%}    & 87.4\%    \\
				Sparse Dictionaries with Saliency~\cite{yu2017abnormal}&    ---     & 	    84.1\%  & &		---     &   ---     & &		---     & 80.9\%    \\
				Compact Feature Sets~\cite{leyva2017video}&                 21.15\% & 	    82\%    & &		39.7\%  &   57\%    & &		19.2\%  & 84\%      \\
                Feng et al.~\cite{feng2017learning}&                        15.1\%  & 	    92.5\%  & &		64.9\%  &   \textbf{69.9\%}  && 		---     & ---       \\
                Turchini et al.~\cite{Turchini2017}&                        24\%    & 	    78.1\%  & &		37\%    &   62.2\%  & &		19\%    & 80.7\%    \\

				Singh and Mohan~\cite{singh2017graph}&                      18\%    & 	    88.71\% && 		28\%    &   75.34\% && 		25\%    & 80.07\%   \\
				Biswas and Vikas~\cite{Biswas2017}&             18\%    & 	    88.71\% && 		28\%    &   75.34\% && 		25\%    & 80.07\%   \\
				Zhou et al. ~\cite{ZHOU2016358}&                            24\%   & 	    85\%    && 		---     &   87\%    & &		24.4\%  & 86\%      \\
				\midrule
				TCP (Proposed Method) &                                     \textbf{8\%}     &       \textbf{95.7\%}  &  &     40.8\%  &   64.5\%  & &      18\%    & 88.4\%    \\
				\bottomrule
			\end{tabular}
		\end{center}
		\caption[Plug-and-Play: Comparison with state-of-the-art on UCSD]{Comparison with state-of-the-art on UCSD dataset: reported ERR (Equal Error Rate) and AUC (Area Under Curve). The values of previous methods are reported from~\cite{xu2015learning}.}
		\label{tbl:results}
	\end{table}
	\end{landscape}
	\begin{landscape}
	\begin{figure*}[t]
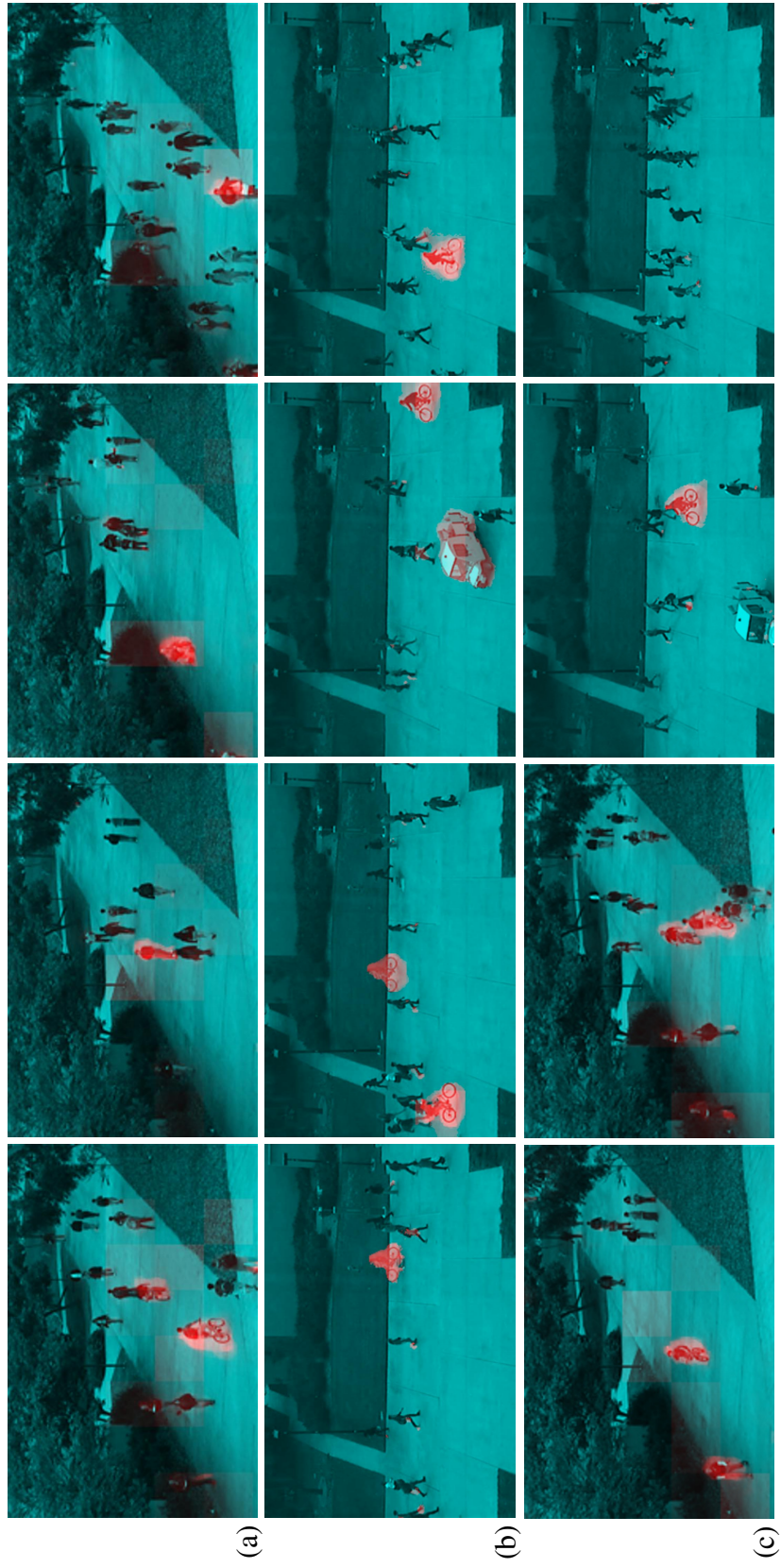

		\begin{center}
		(a)
			\includegraphics[width=.235\linewidth]{images/ped1_1}
			\includegraphics[width=.235\linewidth]{images/ped1_2}
			\includegraphics[width=.235\linewidth]{images/ped1_3}
			\includegraphics[width=.235\linewidth]{images/ped1_4}
			
			(b)	
			\includegraphics[width=.235\linewidth]{images/ped2_3}
			\includegraphics[width=.235\linewidth]{images/ped2_2}
			\includegraphics[width=.235\linewidth]{images/ped2_1}
			\includegraphics[width=.235\linewidth]{images/ped2_4}
			
			(c)
			\includegraphics[width=.235\linewidth]{images/ped1_conf_1}
			\includegraphics[width=.235\linewidth]{images/ped1_conf_2}
			\includegraphics[width=.235\linewidth]{images/ped2_conf_1}
			\includegraphics[width=.235\linewidth]{images/ped2_conf_2}

		\end{center}
		\caption[Plug-and-Play: Sample results of anomaly localization]{Sample results of anomaly localization on UCSD: (a) selected from Ped1, (b) Ped2, and (c) confusion cases from Ped1 and Ped2}
		\label{fig:pedvis_plug}
	\end{figure*}
	\end{landscape}

	 Note that the proposed method is not taking advantage of any kind of learning in comparison with the others. The proposed method can be effectively exploited to detect and localize anomaly with no additional learning costs. Qualitative results on Ped1 and Ped2 are shown in Fig.~\ref{fig:pedvis_plug}. The figure shows we could successfully detect different abnormality sources (like cars, bicycles and skateboards) even in the case in which the object can not be recognized by visual appearance alone (\eg, the skateboard). The last row in Fig.~\ref{fig:pedvis_plug} shows the confusion cases, which not detect the abnormal object (the car) and detect normal as abnormal (the pedestrian). Most of the errors (e.g., miss-detections) are due to the fact that the abnormal object is very small or partially occluded (e.g., the skateboard in the rightmost image) and/or has a ``normal'' motion (i.e., a car moves the same speed of normally moving pedestrians in the scene). 

\subsubsection{Components Analysis}
\label{sec:abl}
 \noindent\textbf{Analysis of the importance of two streams.}
The evaluation of the TCP-only version is performed on ped1 and in two levels: frame-level for anomaly detection, and pixel-level for anomaly localization. In the both cases we unchanged the same experimental setup reviewed in Sec. \ref{sec:exp}.

\begin{figure}
	\begin{center}
		\begin{tabular}{c c}
			\includegraphics[width=0.9\linewidth]{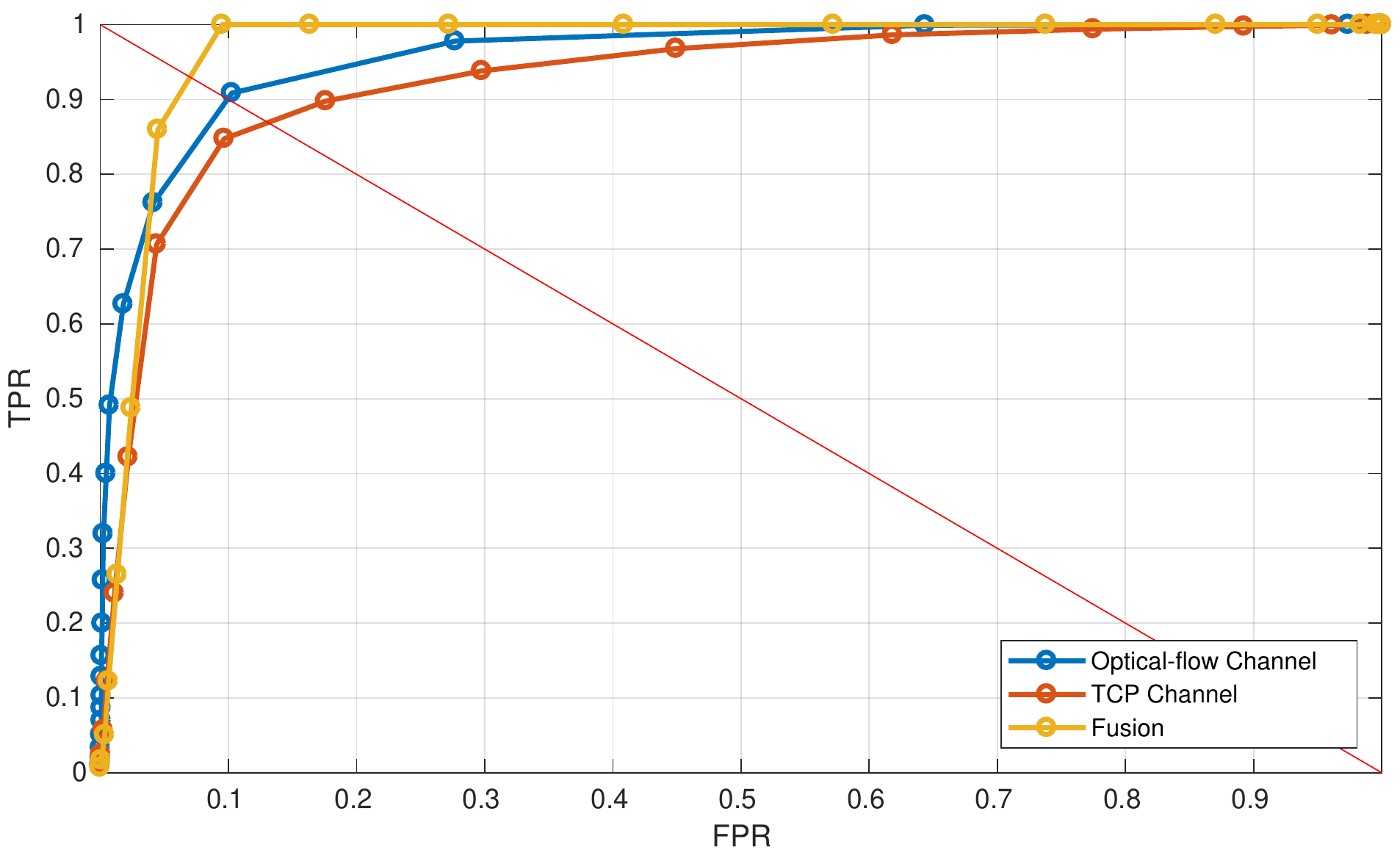} 
		\end{tabular}	
	\end{center}
	\caption[Plug-and-Play: Comparison ROC of different Streams]{Comparison of different Streams in the proposed method on Ped1 frame-level evaluation.}
	\label{fig:rocstreams}
\end{figure}

In the frame-level evaluation, the TCP-only version obtains 93.6\%(AUC), which is slightly lower than 95.7\% of the fused version. In pixel-level evaluation, however, the performance of the TCP-only version dropped 9.3\% with respect to the fused version. This result is still significantly above most of the methods in Tab.1, but this clearly shows the importance of the optical-flow stream for abnormality localization. This is probably due to refining the abnormal segments leveraging the fine motion segments created by the optical-flow map. Hence, fusing appearance and motion can refine the detected area, which leads to a better localization accuracy. Fig. \ref{fig:rocstreams} shows ROCs for the three different states TCP-only, motion-only, and fusion. We simply select equal weights for optical-flow and TCP. 

\noindent\textbf{Binary quantization layer vs. clustering:}
The Binary Quantization Layer ($hconv6$) is a key novelty of our method, which ensures the CNN will work both in the plug-and-play fashion as well as -possibly- being trained end-to-end. In order to evaluate the proposed binary quantization layer, the $hconv6$ removed from the network and a k-means clustering ($k= 2^7$) is performed on the $pool5$ layer of FCN in an offline fashion. Then, the TCP measure is computed on the codebook generated by clustering instead of the binary codes. We evaluated this on UCSD (ped1), obtaining 78.4\% (17.3\% less than our best result on frame-level).

	\subsection{Discussion}
The key idea behind this work is that the consecutive frames should have similar binary patterns, unless they undergo a large semantic change (e.g., abnormal object/motion). The role of TCP measure is to capture such changes across time by computing the irregularity over histogram of binary codes. The proposed TCP measure defines abnormal events as \emph{irregular} events deviated from the normal ones, and the abnormality is measured as the uniformity of the histogram of binary codes. Hence, a flat histogram of binary patterns implies more inconsistency in visual patterns so increases the chance of abnormality. Such particular formulation allows to deal with the context-dependent abnormal events. These characteristics make our method unique in the panorama of the measure-based methods for abnormality detection.

The underlying idea of the proposed approach is to capture the crowd dynamics, by exploiting the temporal variations of CNN features. The CNN network is specifically used to narrow down the semantic gap between low-level pixels and high-level concepts in a crowd scene. The proposed approach provides the means to inject such semantics into model, while maintaining the method complexity in a manageable level. 
	The proposed BFCN is composed of a fully convolutional neural network followed by a binary quantization layer (BQL). The weights of the former network are borrowed from an already pre-trained network and the weights of the BQL layer are obtained through an external hashing module and ``plugged'' into the network as an additional convolutional layer. The last layer of the network (BQL) provides the means to quantize the $pool5$ feature maps into 7-bit binary maps. The training of ITQ is done only once in an off-line fashion and is used for all the experiments without any further fine-tuning. The plug-and-play nature of our proposed architecture enables our method to work across multiple datasets without specific retraining.

\graphicspath{{Chapter2/Figs/Vector/}{Chapter2/Figs/c/}{Chapter2/Figs/d/}}

\section{Abnormality detection with Deep Generative Models (DGMs)}
\label{sec:gan_anomaly}

	As mentioned earlier, there are two main reasons for which abnormality detection is challenging. First, existing datasets with {\em ground truth} abnormality samples are small, where this limitation is particularly significant for deep-learning based methods due to their data-hungry being nature. The second reason is the lack of a clear and objective definition of abnormality. Moreover, these two problems are related to each other, because the abnormality definition subjectivity makes it harder to collect abnormality ground truth. 
	
	In order to deal with these problems, {\em generative} methods for abnormality detection focus on modeling only the {\em normal} pattern of the crowd. The advantage of the generative paradigm lies in the fact that only {\em normal} samples are needed at training time, while detection of what is abnormal is based on measuring the distance from the learned normal pattern.
	However, most of the existing generative approaches rely on hand-crafted features to represent visual information \cite{cong2011sparse,kim2009observe,mahadevan2010anomaly,mehran2009abnormal,mousavi2015analyzing} or use Convolutional Neural Networks (CNNs) trained on external datasets \cite{ravanbakhsh2016plug,sabokrouFFK16}. Recently, Xu et al.~\cite{xu2015learning} proposed to use stacked denoising autoencoders. However, the networks used in their work are shallow and based on small image patches. Moreover, additional one-class SVMs need to be trained on top of the learned representation.
As far as we know, the only other end-to-end deep learning framework for abnormality detection is the recently proposed approach of Hasan et al. \cite{DBLP:conf/cvpr/0003CNRD16}. In \cite{DBLP:conf/cvpr/0003CNRD16} a Convolutional Autoencoder is used to learn the crowd-behaviour normal pattern and used at testing time to {\em generate} the normal scene appearance, 
using the reconstruction error to measure an abnormality score.
The main difference of our approach with \cite{DBLP:conf/cvpr/0003CNRD16} is that we exploit the adversary game between $G$ and $D$ to simultaneously approximate the normal data distribution and train the final classifier.
In Sec.~\ref{sec:exp}-\ref{sec:ablation} we compare our method  with  both \cite{DBLP:conf/cvpr/0003CNRD16} and two strong baselines in which we use the reconstruction error of our generator $G$.
Similarly to \cite{DBLP:conf/cvpr/0003CNRD16}, in \cite{xu2015learning} Stacked Denoising Autoencoders  are used to reconstruct the input image and learn task-specific features using a deep network. However, in \cite{xu2015learning} the final classifier is a one-class SVM which is  trained on top of the learned representations and it is not jointly optimized together with the deep-network-based features.

The second novelty we propose in this work is 
a {\em multi-channel} data representation. Specifically, we use both appearance and motion (optical flow) information: a two-channel approach which has been proved to be empirically important in previous work on abnormality detection \cite{mahadevan2010anomaly, ravanbakhsh2016plug,xu2015learning}.
Moreover, we propose to use a cross-channel approach where, inspired by \cite{Isola_2017_CVPR}, we train two networks which respectively transform raw-pixel images in optical-flow representations and vice versa.
The rationale behind this is that the architecture of  our conditional generators $G$ is  based on an encoder-decoder (see Sec.~\ref{sec:tasks}) 
and we use these channel-transformation tasks in order  
to prevent $G$ learn a trivial identity function and force 
$G$ and $D$  to construct sufficiently informative internal representations.

In this section we propose a generative deep learning method applied to abnormality detection in crowd analysis. More specifically, our goal is to use deep networks to learn a representation of the {\em normal pattern} utilizing only {\em normal} training samples, which are much easier to collect. For this purpose, Generative Adversarial Networks (GANs) \cite{NIPS2014_5423} are used, an emerging approach for training deep networks using only unsupervised data. While GANs are usually used to generate images, we propose to use GANs {\em to learn the normality of the crowd behaviour}.

\begin{figure}
		\centering
		\centerline{\includegraphics[width=0.69\linewidth]{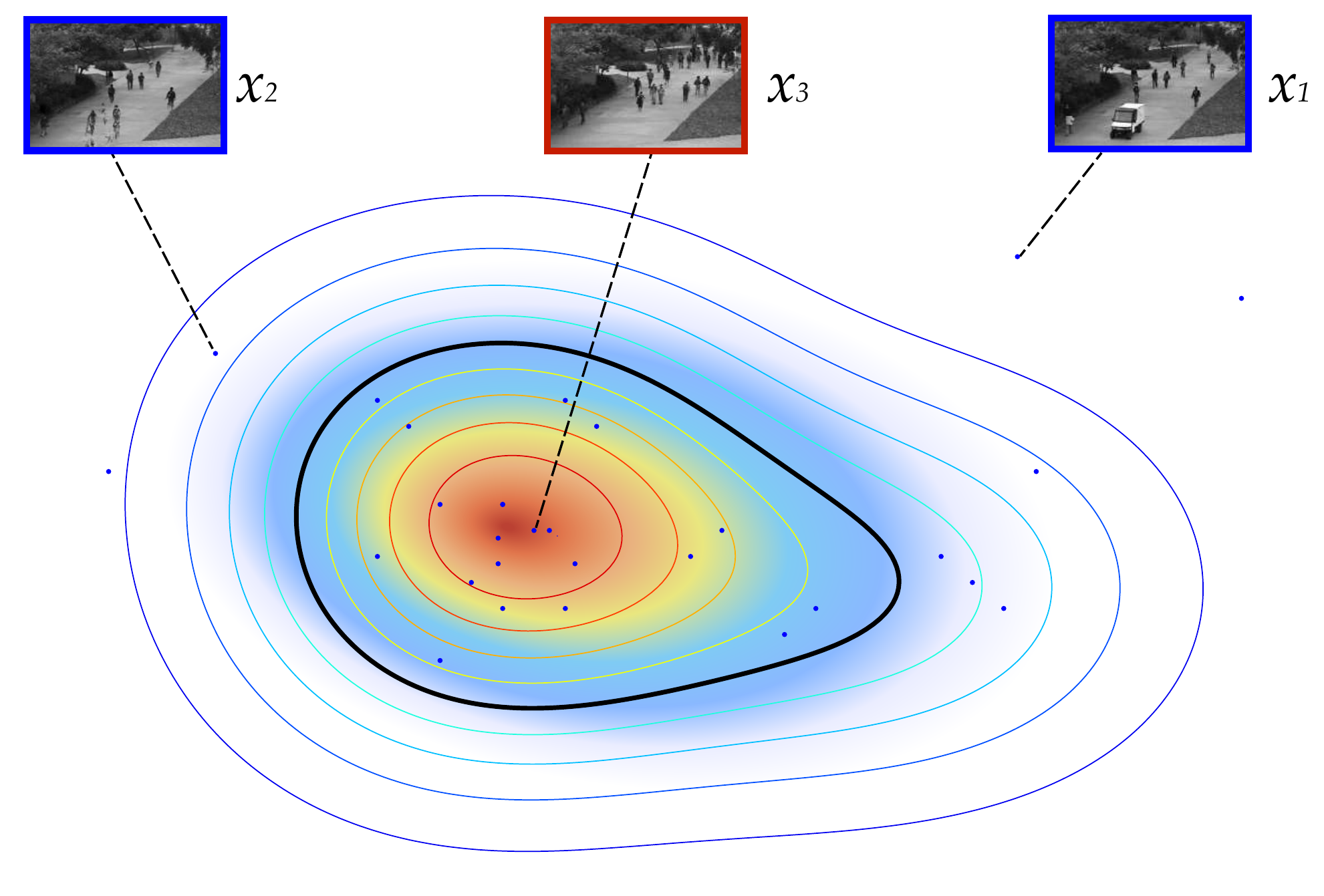}}
	\caption[cGANs: A schematic representation of our Adversarial Discriminator]{A schematic representation of our Adversarial Discriminator. The data distribution is denser in the feature space area corresponding to the only real and ``normal'' data observed by $G$ and $D$ during training. $D$ learns to separate this area from the rest of the feature space. In the figure, the solid black line represents the decision boundary learned by $D$. Outside this boundary lie both non-realistically generated images (e.g., $x_2$) and real but non-normal images (e.g., $x_1$). At testing time we exploit the learned decision boundary in order to detect abnormal events in new images.}
	\label{fig:teaser}
\end{figure}

At testing time two main baselines are considered: 
\begin{itemize}
    \item {\em i)} The trained generator networks $G$ are used to generate appearance and motion information. Since our networks have learned to generate {\em only} what is normal, they are not able to reconstruct appearance and motion information of the possible {\em abnormal} regions of the test frame. Exploiting this intuition, a simple difference between the real test-frame representations and the generated descriptions allows us to easily and robustly detect abnormal areas in the frame.
    \item {\em ii)} The trained discriminator networks $D$ are used to detect possible anomalies in a new scene. During the training process $D$ also learns how to distinguish what is normality from what is not. In this case, the abnormal events are considered as ``outliers'' with respect to the learned decision boundary of $D$ (see Fig.~\ref{fig:teaser}). Since our final goal is a discriminative task, we propose to directly use $D$ after training.
The advantage of this approach is that we do not need to train one-class SVMs or other classifiers  on top of the learned visual representations and we present one of the very first deep learning approaches for abnormality detection which can be trained end-to-end.
\end{itemize}

	Extensive experiments on challenging abnormality detection datasets show the superiority of the proposed approach compared to the state of the art.
	
\subsection{Backgrounds on DGMs}
	\label{sec:RelatedWork_gan}
	In this section we briefly review previous work considering: (1) our application scenario (Abnormality Detection) and (2) our methodology based on GANs.
	
\noindent\textbf{Abnormality detection.} Our method is different from~\cite{cong2011sparse,huang2016crowd,kim2009observe,lu2013abnormal,mahadevan2010anomaly,mehran2009abnormal,mousavi2015abnormality,mousavi2015analyzing,rabiee2016novel,rabiee2016crowd,rabiee2017detection,raghavendra2013anomaly,saligrama2012video}, which also focus on learning generative models on motion and/or appearance features. A key difference compared to these methods is that they employ hand-crafted features (e.g., Optical-flow, Tracklets, etc.) to model normal-activity patterns, whereas our method learns features from raw-pixels using a deep learning based approach. A deep learning-based approach has been investigated also in~\cite{ravanbakhsh2016plug,sabokrouFFK16}. Nevertheless, these works use existing CNN models trained for other tasks (e.g., object recognition) which are adapted to the abnormality detection task. For instance, Ravanbakhsh et al.~\cite{ravanbakhsh2016plug} propose a Binary Quantization Layer plugged as a final layer on top of a CNN, capturing temporal motion patterns in video frames for the task of abnormality segmentation. Differently from~\cite{ravanbakhsh2016plug}, we specifically propose to train a deep generative network \emph{directly} for the task of abnormality detection. 

Most related to our approach is the work of Xu et al.~\cite{xu2015learning}, who propose to learn motion/appearance feature representations using stacked denoising autoencoders. The networks used in their work are relatively shallow, since training deep autoencoders on small abnormality datasets is prone to over-fitting. Moreover, their networks are not end-to-end trained and the learned representation need externally trained classifiers (multiple one-class SVMs) which are not optimized for the learned features. Conversely, we propose to use adversarial training for our representation learning. Intuitively, the adopted conditional GANs provide data augmentation and implicit data 
 supervision thank to the discriminator network. As a result we can train much deeper generative networks on the same small abnormality datasets and we do not need to train external classifiers.
 The only deep learning based approach proposing a framework which can be  fully-trained in an end-to-end fashion we are aware of is the Convolutional AE network proposed in \cite{DBLP:conf/cvpr/0003CNRD16}, where a deep representation is learned by minimizing the AE-based frame reconstruction. At testing time, an anomaly is detected computing the difference between the AE-based frame reconstruction and the real test frame.
We compare with this work 
in Sec.~\ref{sec:exp} and in  Sec.~\ref{sec:ablation} we present two modified versions of our GAN-based approach ({\em Adversarial Generator} and {\em GAN-CNN}) in which, similarly to \cite{DBLP:conf/cvpr/0003CNRD16}, we use the reconstruction errors of our adversarially-trained generators as detection strategy.
Very recently, Ravanbakhsh et al.~\cite{ravanbakhsh2017abnormal} proposed to use the reconstruction errors of the generator networks to detect anomalies at testing time instead of directly using the corresponding discriminators as we propose here. However, their method needs  an externally-trained CNN to capture sufficient semantic information and  a fusion strategy which takes into account  the reconstruction errors of the two-channel generators. Conversely, the discriminator-version proposed in this work is simpler to reproduce and faster to run. Comparison between these two versions is provided in Sec.~\ref{sec:ablation}, together with a detailed ablation study of all the elements of our proposal.

\noindent
{\bf GANs.} \cite{NIPS2014_5423,DBLP:conf/nips/SalimansGZCRCC16,DBLP:journals/corr/RadfordMC15,Isola_2017_CVPR,nguyen2016ppgn} are based on a two-player game between two different networks, both trained with unsupervised data. One network is the {\em generator} ($G$), which aims at generating realistic data (e.g., images). The second network is the {\em discriminator} ($D$), which aims at discriminating real data from data generated from $G$. 
Specifically, the {\em conditional} GANs \cite{NIPS2014_5423}, that we use in our approach,
are trained with a set of data point pairs (with loss of generality, from now on we assume both data points are images): $\{ (x_i, y_i) \}_{i=1,...,N}$, where image $x_i$ and image $y_i$ are somehow each other semantically related.
$G$ takes as input $x_i$ and  random noise $z$  and generates a new image $r_i = G(x_i, z)$. 
$D$ tries to distinguish $y_i$ from $r_i$, while $G$ tries to ``fool'' $D$ producing more and more realistic images which are hard to be distinguished.

It has been shown that GANs having good performance on cross-modal generation and image translation tasks \cite{pahde2018discriminative,Isola_2017_CVPR}. Very recently Isola et al. \cite{Isola_2017_CVPR} proposed an ``image-to-image translation'' framework based on conditional GANs, where both the generator and the discriminator are conditioned on the real data. They show that a ``U-Net'' encoder-decoder with skip connections can be used as the generator architecture together with a patch-based discriminator in order to transform images with respect to different representations.
We adopt this framework in order to generate optical-flow images from raw-pixel frames and vice versa. However, it is worth to highlight that, different from common GAN-based approaches, we do not aim at generating image representations which look realistic, but we use $G$ to learn the normal pattern of an observed crowd scene. At testing time, two strategies are applied:
\begin{enumerate}
    \item \emph{Adversarial Generator)} $G$ is used to generate appearance and motion information of the normal content of the input frame. Comparing this generated content with the real frame allows us to detect the possible abnormal areas of the frame.
    \item \emph{Adversarial Discriminator)} $D$ is directly used to detect abnormal areas using the appearance and the motion of the input frame.
\end{enumerate}

\subsection{Adversarial cross-modal generation}
\label{sec:tasks}
Inspired by Isola et al. \cite{Isola_2017_CVPR}, we built our freamwork to learn the normal behaviour of the crowd in the observed scene. 
We use two channels: appearance (i.e., raw-pixels) and motion (optical flow images) and two cross-channel tasks. In the first task, we generate optical-flow images starting from the original frames, while in the second task we generate appearance information starting from an optical flow image.

Specifically, let $F_t$ be the $t$-th frame of a training video and $O_t$ the optical flow obtained using $F_t$ and $F_{t+1}$. $O_t$ is computed using \cite{brox2004high}.
We train two networks: ${\cal N}^{F \rightarrow O}$, which generates optical-flow from frames (task 1) and 
${\cal N}^{O \rightarrow F}$, which generates frames from optical-flow (task 2).
In both cases, our networks are composed of a conditional generator $G$
and a conditional discriminator $D$. $G$ takes as input an image $x$ and a noise vector $z$ (drawn from a noise distribution ${\cal Z}$) and outputs an image $r = G(x,z)$ 
of the same dimensions of $x$ 
but represented in a different channel. 
For instance, in case of ${\cal N}^{F \rightarrow O}$, $x$ is a frame ($x  = F_t$) and $r$ is {\em the reconstruction} of its corresponding optical-flow image $y  = O_t$. On the other hand,
$D$ takes as input two images: $x$ and $u$ (where $u$ is either $y$ or $r$) and outputs a scalar representing the probability that both its input images came from the real data.

Both $G$ and $D$ are fully-convolutional networks, composed of convolutional layers, batch-normalization layers and ReLU nonlinearities.
In case of $G$ we adopt the U-Net architecture~\cite{ronneberger2015u}, which is  an encoder-decoder, where the input $x$ is passed through a series of progressively downsampling layers until a bottleneck layer, at which point the forwarded information is upsampled. Downsampling and upsampling layers in a symmetric position with respect to the bottleneck layer are connected by {\em skip connections} which help preserving important local information. The noise vector $z$ is implicitly provided to $G$ using dropout, applied to multiple layers.

The two input images $x$ and $u$ of $D$ are concatenated and passed through 5 convolutional layers. In more detail, $F_t$ is represented using the standard RGB representation, while 
$O_t$ is represented using the horizontal, the vertical and the magnitude components. Thus, in both tasks, the input of $D$ is composed of 6 components (i.e., 6 2D images), whose relative order depends on the specific task. 
All the images are rescaled to $256 \times 256$.
We use the popular {\em PatchGAN} discriminator~\cite{li2016precomputed}, which is
based on a ``small'' fully-convolutional discriminator $\hat{D}$.
$\hat{D}$ is
applied to a $30 \times 30$ grid, where each position of the grid corresponds to a $70 \times 70$ patch $p_x$ in $x$ and a corresponding patch $p_u$ in $u$. The output of $\hat{D}(p_x,p_u)$ is a score representing the probability 
that $p_x$ and $p_u$ are  both real. 
During training, the output of $\hat{D}$ over all the grid positions  is averaged
and this provides the final score of $D$ with respect to $x$ and $u$.
Conversely, at testing time we directly use $\hat{D}$ as a ``detector'' which is run over the grid to spatially localize the possible abnormal regions in the input frame (see Sec.~\ref{sec:Detection}).

	\subsection{Learning distribution on normal events}
	\label{sec:learing}
	
	We use the framework proposed by Isola et al. \cite{Isola_2017_CVPR} to learn the normal behaviour of the observed scene. Specifically, let $F_t$ be the $t$-th frame of a training video and $O_t$ the optical-flow obtained using $F_t$ and $F_{t+1}$. $O_t$ is computed using \cite{brox2004high}.
	We train two networks: ${\cal N}^{F \rightarrow O}$, which generates optical-flow from frames and 
	${\cal N}^{O \rightarrow F}$, which generates frames from optical-flow.
	In both cases, inspired by \cite{Isola_2017_CVPR}, our networks are composed of a conditional generator $G$
	and a conditional discriminator $D$ (we refer to \cite{Isola_2017_CVPR} for the architectural details of $G$ and $D$). $G$ takes as input an image $x$ and a noise vector $z$ (drawn from a noise distribution ${\cal Z}$) and outputs an image $p = G(x,z)$ of the same dimensions 
	of $x$ 
	but represented in a different channel. 
	For instance, in case of ${\cal N}^{F \rightarrow O}$, $x$ is a frame ($x = F_t$) and $p$ is {\em the reconstruction} of its corresponding optical-flow image $y = O_t$. On the other hand,
	$D$ takes as input two images (either $(x,y)$ or $(x,p)$) and outputs a scalar representing the probability that both its input images came from the real data.

	$G$ and $D$ are trained using both a conditional GAN loss ${\cal L}_{cGAN}$ and a reconstruction loss
	${\cal L}_{L1}$. In case of ${\cal N}^{F \rightarrow O}$, the training set is composed of pairs of frame-optical flow images
	${\cal X} = \{ (F_t, O_t) \}$, where $O_t$ is represented using a standard three-channels representation of the horizontal, the vertical and the magnitude components. 
	${\cal L}_{L1}$ is given by:
	\begin{equation}
	{\cal L}_{L1}(x,y) = ||y - G(x,z) ||_1
	\end{equation}
	\noindent
	while the conditional adversarial loss ${\cal L}_{cGAN}$ is:
	\begin{align}
	{\cal L}_{cGAN}(G,D)= 
	\mathbb{E}_{(x,y) \in {\cal X}} [\log D(x,y)] + \\
	\mathbb{E}_{x \in \{ F_t \}, z \in {\cal Z}} [\log ( 1 - D(x,G(x,z)) )]
	\end{align}
	Conversely, in case of ${\cal N}^{O \rightarrow F}$, we use ${\cal X} = \{ (O_t, F_t) \}$. We refer to \cite{Isola_2017_CVPR} for more details about the training procedure. 
	What is important to highlight here is that both $ \{ F_t \}$ and $\{ O_t \}$ are collected 
	using the frames of the only {\em normal} videos of the training dataset. 
	The fact that we do not need videos showing abnormal events at training time makes it possible to train our networks with potentially very large datasets without the need of ground truth samples describing abnormality: $G$ acts as an implicit supervision for $D$ (and vice versa). 
	
	During training the  generators of the two tasks ($G^{F \rightarrow O}$ and $G^{O \rightarrow F}$) observe only normal scenes. As a consequence, after training they are not able to reconstruct an abnormal event. 
For instance, in Fig.~\ref{fig:qual2-O-F} a frame $F$ containing a vehicle unusually moving in a University campus
is input to $G^{F \rightarrow O}$ and in the generated optical flow image ($r_O = G^{F \rightarrow O}(F)$) the abnormal area corresponding to that vehicle is not properly reconstructed. Similarly, when the real optical flow ($O$) associated with $F$ is input to $G^{O \rightarrow F}$, the network tries to reconstruct the area corresponding to the vehicle but the output is a set of unstructured blobs (Fig.~\ref{fig:qual2-O-F}-b).
On the other hand, the two corresponding discriminators $D^{F \rightarrow O}$ and $D^{O \rightarrow F}$ during training have learned to distinguish what is plausibly 
real in the given scenario from what is not and we will exploit this learned discrimination capacity at testing time.

\begin{landscape}

\begin{figure*}
	\begin{center}
		\scriptsize{(a)}
		\includegraphics[width=.23\linewidth]{images/ped1_5}
		\includegraphics[width=.23\linewidth]{images/ped2_5}
		\includegraphics[width=.23\linewidth]{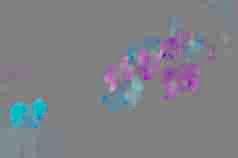}
		\includegraphics[width=.23\linewidth]{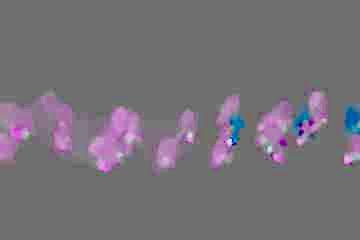}
		
		\scriptsize{(b)}
		\includegraphics[width=.23\linewidth]{images/ped1_2}
		\includegraphics[width=.23\linewidth]{images/ped2_2}
		\includegraphics[width=.23\linewidth]{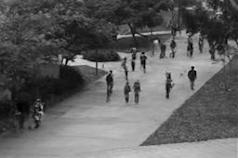}
		\includegraphics[width=.23\linewidth]{images/ped2_conf_2}
		
		\scriptsize{(c)}
		\includegraphics[width=.23\linewidth]{images/ped1_1}
		\includegraphics[width=.23\linewidth]{images/ped2_1}
		\includegraphics[width=.23\linewidth]{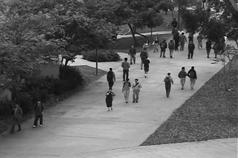}
		\includegraphics[width=.23\linewidth]{images/ped2_conf_1}
	\end{center}
	\caption[cGANs: Images generated by $G^{O \rightarrow F}$]{Images generated by $G^{O \rightarrow F}$ after training is completed: (a) the input optical-flow images, (b) the corresponding generated  frames,
		(c) the real frames corresponding to (a). The first two columns represent an abnormal scene, while the other two columns depict a normal situation. Note that the abnormal objects in the first two samples are completely missing in the corresponding reconstructions. In the first image (from left) there is a large moving area (produced by a vehicle) which has not been reconstructed: $G^{O \rightarrow F}$ simply does not know how to ``draw'' a vehicle, hence the vehicle  ``disappeared'' from the reconstruction and was replaced by  some pedestrian-like blobs. Similarly, in the second image, two fast moving bicycles ``disappeared' from the frame reconstruction.}
	\label{fig:qual2-O-F}
\end{figure*}

\begin{figure*}
	\begin{center}
		\scriptsize{(a)}
		\includegraphics[width=.23\linewidth]{images/ped1_1}
		\includegraphics[width=.23\linewidth]{images/ped2_1}
		\includegraphics[width=.23\linewidth]{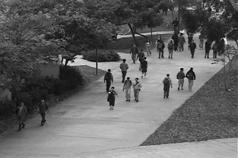}
		\includegraphics[width=.23\linewidth]{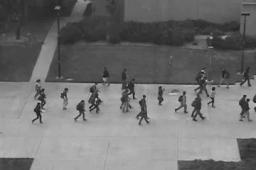}
		
		\scriptsize{(b)}
		\includegraphics[width=.23\linewidth]{images/ped1_3}
		\includegraphics[width=.23\linewidth]{images/ped2_3}
		\includegraphics[width=.23\linewidth]{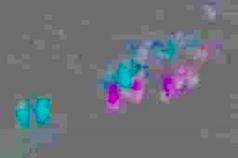}
		\includegraphics[width=.23\linewidth]{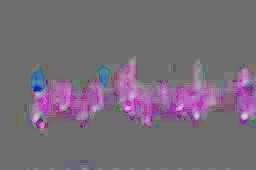}
		
		\scriptsize{(c)}
		\includegraphics[width=.23\linewidth]{images/ped1_5}
		\includegraphics[width=.23\linewidth]{images/ped2_5}
		\includegraphics[width=.23\linewidth]{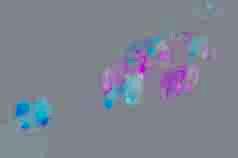}
		\includegraphics[width=.23\linewidth]{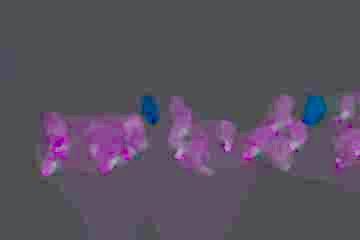}
	\end{center}
	\caption[cGANs: Images generated by $G^{F \rightarrow O}$]{A few qualitative results showing the images generated by $G^{F \rightarrow O}$ after training is completed: (a) the real input frames, (b) the corresponding generated optical flow images,
		(c) the real optical flow images corresponding to (a). The first two columns represent an abnormal scene, while the other two columns depict a normal situation. Note that the motion of the vehicle in the first sample and of the bicycle in the bottom of the second sample have not been reconstructed.}
	\label{fig:qual1-F-O}
\end{figure*}
\end{landscape}


Note that, even if a global optimum can be theoretically reached in a GAN-based training, in which 
the  data distribution and the generative distribution totally overlap each other
\cite{NIPS2014_5423}, in practice the generator is very rarely able to generate fully-realistic images. For instance, in Fig.~\ref{fig:qual2-O-F} the high-resolution details of the generated pedestrians (``normal'' objects)  are quite smooth and the human body is approximated with a blob-like structure. As a consequence, at the end of the training process, the discriminator has learned to separate real data from artifacts.
This situation is schematically represented in Fig.~\ref{fig:qual2-O-F}. 
 The discriminator is represented by the decision boundary on the learned feature space which separates the densest area of this distribution from the rest of the space. Outside this area lie both non-realistic generated images (e.g. $x_2$) and real, abnormal events (e.g., $x_1$).
Our hypothesis is that the latter lie outside the discriminator's decision boundaries because they represent situations never observed during training and hence treated by $D$ as outliers.

\begin{figure}
	\centering
		\begin{minipage}{0.95\linewidth}
			
			\centerline{\includegraphics[width=\linewidth]{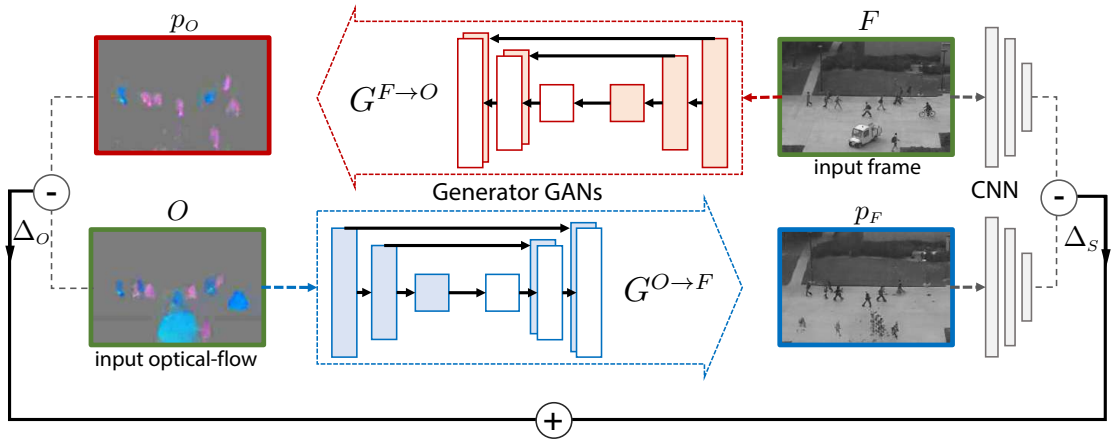}}
		\end{minipage}
		\caption[cGANs: Method overview with adversarial generator networks]{{\small Adversarial generators approach. Top: a generator network takes as input a frame and produces a corresponding optical-flow image. Bottom: a second generator network is fed with a real optical-flow image and outputs an appearance reconstruction.}
		}
		\label{fig:overview}
	\end{figure}

	Considering our two main baselines, at testing time we used both discriminators and the generators corresponding to the trained networks.
	In the adversarial generator baseline, since $G^{F \rightarrow O}$ and $G^{O \rightarrow F}$ have observed only normal scenes during training, they are not able to reconstruct an abnormal event. For instance, in Fig.~\ref{fig:overview} (top) a frame $F$, containing a vehicle unusually moving on a University campus, is input to $G^{F \rightarrow O}$ and in the generated optical flow image ($p_O$) the abnormal area corresponding to that vehicle is not correctly reconstructed. Similarly, when the real optical flow ($O$) associated with $F$ is input to $G^{O \rightarrow F}$, the network tries to reconstruct the area corresponding to the vehicle but the output is a set of unstructured blobs (Fig.~\ref{fig:overview}, bottom). We exploit this {\em inability} of our networks to reliably reconstruct abnormality to detect possible anomalies as explained in the next section.
	For the adversarial discriminator baseline we use the $D^{F \rightarrow O}$ and $D^{O \rightarrow F}$ discriminator's learned decision boundaries in order to detect $x_1$-like events as explained in the next section.
	
	\subsection{Abnormality detection with out-of-distribution detection}
	\label{sec:Detection}
	\noindent\textbf{Adversarial generator approach.}
	At testing time we input $G^{F \rightarrow O}$ and $G^{O \rightarrow F}$ using each frame $F$ of the test video and its corresponding optical-flow image $O$, respectively.
	Note that the random noise vector $z$ is internally produced by the two networks using dropout \cite{Isola_2017_CVPR}, and 
	in the following 
	we drop $z$ to simplify our notation.
	Using $F$, an optical-flow reconstruction can be obtained: $p_O= G^{F \rightarrow O}(F)$, which is compared with $O$ using a simple pixel-by-pixel difference, obtaining $\Delta_O = O - p_O$ (see Fig.~\ref{fig:overview}). $\Delta_O$ highlights the (local) differences between the real optical flow and its reconstruction and these differences are higher in correspondence of those areas in which $G^{F \rightarrow O}$ was not able to generate the abnormal behaviour. 
	
	Similarly, we obtain the appearance reconstruction $p_F = G^{O \rightarrow F}(O)$.
	As shown in Fig.~\ref{fig:overview} (bottom), the network generates ''blobs'' in the abnormal areas of 
	$p_F$. Even if these blobs have an appearance completely different from the corresponding area in the real image $F$,
	we empirically observed that a simple pixel-by-pixel difference between $F$ and $p_F$ is less informative than the difference computed in the optical-flow channel. For this reason, a ''semantic'' difference is computed using another network, pre-trained on ImageNet \cite{russakovsky2015imagenet}.
	Specifically, we use AlexNet \cite{alexnet}. Note that AlexNet is trained using supervised data which are pairs of images and object-labels contained in ImageNet. However, no supervision about crowd abnormal behaviour is contained in ImageNet and the network is trained to recognize generic objects.
	Let $h(F)$ be the $conv_5$ representation of $F$ in this network and $h(p_F)$ the corresponding representation of the appearance reconstruction. The fifth convolutional layer of AlexNet (before pooling) is chosen because it represents the input information in a sufficiently abstract space and is the last layer preserving geometric information. 
	We can now compute a semantics-based difference between $F$ and $p_F$: $\Delta_S = h(F) - h(p_F)$.
	
	Finally, $\Delta_S$ and $\Delta_O$ are fused in order to obtain a unique abnormality map. Specifically, we first upsample $\Delta_S$ in order to obtain $\Delta_S'$ with the same resolution as $\Delta_O$. Then, both $\Delta_S'$ and $\Delta_O$ are normalized with respect to their corresponding channel-value range as follows. For each test video $V$
	we compute the maximum value $m_O$ of all the elements of $\Delta_O$ over all the input frames of $V$. The normalized optical-flow difference map is given by:
	\begin{equation}
	N_O(i,j) = 1/m_O \Delta_O(i,j).
	\end{equation}
	\noindent
	Similarly, the normalized semantic difference map $N_S$ is obtained using $m_S$ computed over all the elements of $\Delta_S'$ in all the frames of $V$:
	\begin{equation}
	N_S(i,j) = 1/m_S \Delta_S'(i,j).
	\end{equation}
	The final abnormality map is obtained by summing $N_S$ and $N_O$: $A = N_S + \lambda N_O$. 
	In all our experiments we use $\lambda = 2$. $A$ is our final abnormality heatmap.
	
\begin{figure}
	\begin{minipage}[b]{0.99\linewidth}
		\centering
		\centerline{\includegraphics[width=\linewidth]{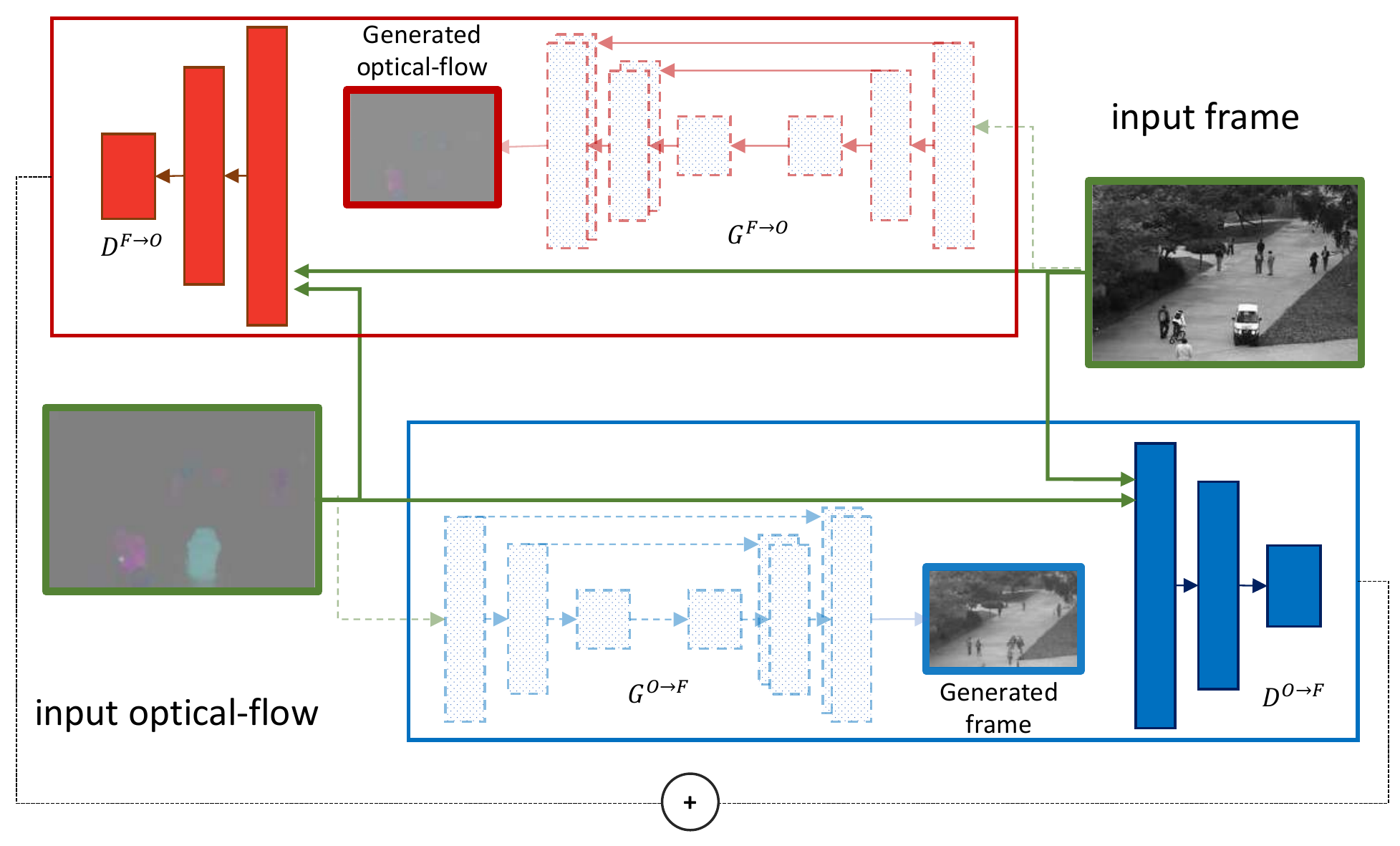}}
	\end{minipage}
	\caption[cGANs: Method overview with adversarial discriminator networks]{Adversarial discriminator approach: a schematic representation of our proposed detection method. }
	\label{fig:over_temp}
\end{figure}
	\noindent\textbf{Adversarial discriminator approach. }
	At testing time only the discriminators are used. More specifically, 
let $\hat{D}^{F \rightarrow O}$ and $\hat{D}^{O \rightarrow F}$ be the patch-based discriminators trained using the two channel-transformation tasks (see Sec.~\ref{sec:tasks}). 
Given a test frame $F$ and its corresponding optical-flow image $O$, we apply the two patch-based discriminators on the same $30 \times 30$ grid used for training. This results in two $30 \times 30$ score maps: $S^O$ and $S^F$ for the first and the second task, respectively. 
Note that we do not need to produce the reconstruction images to use the discriminators. For instance, for a given position on the grid, $\hat{D}^{F \rightarrow O}$ takes as input a patch 
$p_F$ on $F$ and a corresponding patch 
$p_O$ on $O$. 
A possible abnormal area in $p_F$ and/or in $p_O$ (e.g., an unusual object or an unusual movement) corresponds to an outlier with respect to the  distribution learned by $\hat{D}^{F \rightarrow O}$ during training and results in a low value of  $\hat{D}^{F \rightarrow O}(p_F,p_O)$. By setting a threshold on this value we obtain a decision boundary (see Fig.~\ref{fig:teaser}). However, following a common practice, we first fuse the channel-specific score maps and then we apply a range of confidence thresholds on the final abnormality map in order to obtain different ROC points (see Fig \ref{fig:over_temp} and Sec.~\ref{sec:exp}). Below we show how the final abnormality map is constructed.

The two score maps are summed with equal weights: $S= S^O + S^F$.
The values in $S$ are normalized in the range $[0, 1]$. In more detail, for each test video $V$
we compute the maximum value $m_s$ of all the elements of $S$ over all the input frames of $V$. 
For each frame the normalized  score map is given by:
\begin{equation}
N(i,j) = 1/m_s S(i,j), i,j \in \{1, ..., 30\}
\end{equation}
Finally, we upsample $N$  to the original frame size 
($N'$)
and the previously computed optical-flow is used to filter out non-motion areas, obtaining the final abnormality map:
\begin{equation}
\label{eq.filtering}
A(i,j) = \left\{%
\begin{array}{ll}
1 - N'(i,j)  & \mbox{if $O(i,j) > 0$}   \\
0  & \mbox{otherwise.}    \\
\end{array}
\right.
\end{equation}
Note that all the post-processing steps (upsampling, normalization, motion-based filtering) are quite common strategies for abnormal-detection systems \cite{xu2015learning} and we do not use any hyper-parameter or ad-hoc heuristic which need to be tuned on a specific dataset.

\subsection{Experiments}
	\label{sec:exp}


	
	In this section we compare the proposed method
against the state of the art   
using  common benchmarks for crowd-behaviour abnormality detection. The evaluation is performed using both a {\em pixel-level} and a {\em frame-level} protocol and the  evaluation setup proposed in \cite{li2014anomaly}. The rest of this section describes the datasets, the experimental protocols and the obtained results.\\
\noindent\textbf{Implementation details.}
${\cal N}^{F \rightarrow O}$ and ${\cal N}^{O \rightarrow F}$ are trained using  the training sequences of the UCSD dataset (containing only ``normal'' events). 
All frames are resized to $256 \times 256$ pixels (see Sec.~\ref{sec:tasks}). Training is based on stochastic gradient descent with momentum 0.5 and batch size 1. 
We train our networks for 10 epochs each.
All the  GAN-specific hyper-parameter values have been set following the suggestions in \cite{Isola_2017_CVPR}, 
while in our approach there is no dataset-specific hyper-parameter which needs to be tuned. This makes the proposed method particularly robust, especially in a weakly-supervised scenario in which ground-truth validation data with abnormal frames are not given. 
All the results presented in this section but ours are taken from \cite{mousavi2015crowd,xu2015learning} 
which report the best results  achieved by each method independently tuning the method-specific hyper-parameter values.

Full-training of one network (10 epochs) takes on average less than half an hour with 6,800 training samples.
At testing time, one frame is processed in 0.53 seconds (the whole processing pipeline, optical-flow computation and post-processing included). These computational times have been computed using a single GPU (Tesla K40).
		\begin{figure*}
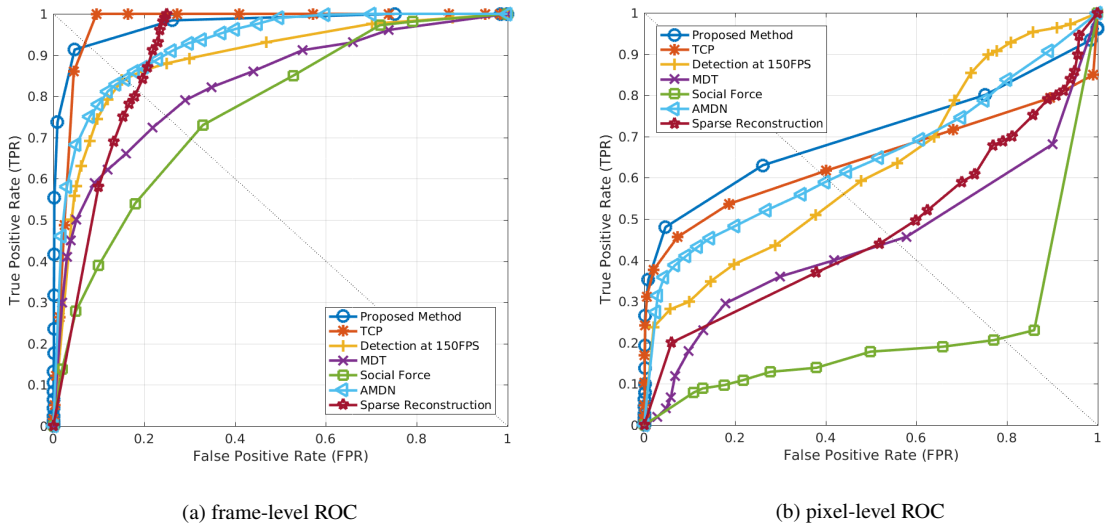

	\begin{minipage}[b]{.5\linewidth}
		\centering
		\centerline{\includegraphics[width=1\textwidth]{images/ped1_frame_level}}
		\centerline{\scriptsize(a) frame-level ROC}\medskip
	\end{minipage}
	\begin{minipage}[b]{0.5\linewidth}
		\centering
		\centerline{\includegraphics[width=1\textwidth]{images/ped1_pixel_level}}
		\centerline{\scriptsize(b) pixel-level ROC}\medskip
	\end{minipage}
	\caption[cGANs: ROC curves for Ped1]{ROC curves for Ped1 (UCSD dataset).}
	\label{fig:rocfrm}
\end{figure*}
	\begin{table*}
	\begin{center}
		\resizebox{1\textwidth}{!}{
		\begin{tabular}[width=1.2\textwidth]{l cc l cc l cc}
			\toprule
			\multirow{2}{*}{Method} &\multicolumn{2}{c}{Ped1 (frame-level)} & &\multicolumn{2}{c}{Ped1 (pixel-level)} & &\multicolumn{2}{c}{Ped2 (frame-level)}\\ \cmidrule{2-3} \cmidrule{5-6} \cmidrule{8-9}
			& EER & AUC & & EER & AUC && EER & AUC\\
			\midrule
			MPPCA~\cite{kim2009observe} & 		            40\%& 	59.0\% & 	&	81\%& 20.5\%& 	&	    30\%& 69.3\%\\
			Social force (SFM)~\cite{mehran2009abnormal} &    31\% & 	 67.5\% & 	&	79\%& 19.7\%& 	&	    42\%& 55.6\%\\
			SF+MPPCA~\cite{mahadevan2010anomaly} & 		32\% & 	 68.8\%& 	&	71\%& 21.3\%& 	&		36\%& 61.3\%\\
			Sparse Reconstruction~\cite{cong2011sparse} & 		                19\%&   ---     & 	&	54\%& 45.3\%& 	&	    --- & --- \\
			MDT~\cite{mahadevan2010anomaly} & 		    25\% &  81.8\% & 	&	58\%& 44.1\%& 	&		25\%& 82.9\%\\
			Detection at 150fps~\cite{lu2013abnormal} & 	15\%& 	91.8\% & 	&	43\%& 63.8\%& 	&		--- & ---\\
			TCP~\cite{ravanbakhsh2016plug} &  8\% &   95.7\%&     &   40.8\%& 64.5\%& &       18\%& 88.4\%\\
			AMDN (double fusion)~\cite{xu2015learning} & 	16\% & 	 92.1\%& 	&	40.1\%& 67.2\%& &		17\%& 90.8\%\\
			Convolutional AE~\cite{DBLP:conf/cvpr/0003CNRD16} & 	27.9\% & 	 81\%& 	&	--- & --- & &		21.7\% & 90\%\\
			PCANet-deep GMM~\cite{feng2017learning} & 	15.1\% & 	 92.5\%& 	&	35.1\% & 69.9\% & &		--- & ---\\
			\midrule
			Adversarial Generator (GAN-CNN) &               \textbf{8\%} & \textbf{97.4\%} &   &\textbf{35\%}& \textbf{70.3\%}& &\textbf{14\%}& \textbf{93.5\%}\\
			Adversarial Discriminator &               \textbf{7\%} & \textbf{96.8\%} &   &\textbf{34\%}& \textbf{70.8\%}& &\textbf{11\%}& \textbf{95.5\%}\\
			\bottomrule
		\end{tabular}
		}
	\end{center}
	\caption[cGANS: Comparison of different methods on UCSD dataset]{UCSD dataset. Comparison of different methods. The results of {\em PCANet-deep GMM} are taken from \cite{feng2017learning}. The other results but ours are taken  from~\cite{xu2015learning}.}
	\label{tbl:UCSD_gan}
\end{table*}

	\noindent\textbf{Datasets and experimental setup.}
	We use two standard datasets: the UCSD Anomaly Detection Dataset~\cite{mahadevan2010anomaly} and the UMN SocialForce~\cite{mehran2009abnormal}. 
	In our experiments, the calculated difference maps are resized from $256 \times 256$ to the groundtruth frames size, in the motion stream optical-flow obtaining the distances would be straight forward by simply computing distances over generated optical-flow and the groundtruth, while in the appearance stream the frames (original and generated) forward to alexnet\cite{alexnet}, and the feature maps from $conv5$ are extracted to compute distances between groundtruth and the generated frame. Afterwards, these computed maps are resize to the groundtruth frames size.
	In the fusion phase, the output optical-flow maps from motion channel are fused with computed distance maps on appearance stream. 
	Finally, a dynamic background subtraction applied to remove the background individually for each video frame, by clipping optical-flow maps to refine the computed segments. These segment maps are used to evaluate the performance of our method on detection and localization of anomalous motions during video frames.


\begin{table}
	\begin{center}
		\begin{tabular}[width=\textwidth]{p{8cm} c c}
			\toprule
			Method 							&			& 	AUC \\
			\midrule
			Optical-flow~\cite{mehran2009abnormal}	  &  &	0.84 \\
			Social force model (SFM)~\cite{mehran2009abnormal} 		&		&	0.96\\
			incremental slow feature analysis~\cite{5} 		&		&	0.96\\
						Chaotic Invariants~\cite{wu2010chaotic}		&&	0.99\\
			Sparse Reconstruction~\cite{cong2011sparse} & &  0.97\\
			Commotion Measure~\cite{mousavi2015crowd} 	&	    &	0.98\\
			TCP~\cite{ravanbakhsh2016plug}&&	0.98\\
			\midrule
				Adversarial Generator (GAN-CNN)  							&&	\textbf{0.99}\\
				
			Adversarial Discriminator 		&					&	\textbf{0.99}\\
			\bottomrule
			
		\end{tabular}
	\end{center}
	\caption[cGANs: Comparison of different methods on UMN]{UMN dataset. Comparison of different methods. All but our results are taken  from~\cite{mousavi2015crowd}.}
	\label{tbl:umn_gan}
\end{table}

\noindent\textbf{Frame-level abnormality detection.}
	The frame-level abnormality detection criterion is based on checking if the frame contains at least one predicted abnormal pixel: in this case the abnormal label is assigned to the whole frame. The procedure is applied over a range of thresholds to build a ROC curve. We compare our method with the state of the art. Quantitative results using both EER (Equal Error Rate) and AUC (Area Under Curve) are shown in Tab.~\ref{tbl:UCSD_gan}, and the ROC curves in Fig.~\ref{fig:rocfrm}.
	The proposed method is also evaluated on UMN dataset frame level evaluation (Tab.~\ref{tbl:umn_gan}). 
	Our methods are called {\em Adversarial Generator (GAN-CNN)} and {\em Adversarial Discriminator}.

	\noindent\textbf{Pixel-level abnormality localization.}
	The goal of the pixel-level evaluation is to measure the accuracy of the abnormality {\em localization}. Following~\cite{li2014anomaly}, 
	a true positive prediction should cover at least 40\% the ground truth abnormal pixels, otherwise the frame is counted as a false positive. 
	Fig.~\ref{fig:rocfrm} shows the ROC curves of the localization accuracy over USDC, and Tab.~\ref{tbl:UCSD_gan} reports a quantitative comparison with the state of the art. The results reported in Tab.~\ref{tbl:UCSD_gan}-\ref{tbl:umn_gan} show that the proposed approach overcomes all the other compared methods.	
	
	\noindent\textbf{Information fusion analysis.}
	In order to analyze the impact on the accuracy provided by each network, ${\cal N}^{O \rightarrow F}$ and ${\cal N}^{F \rightarrow O}$, we perform a set of experiments on UCSD Ped1. In the frame-level evaluation, ${\cal N}^{O \rightarrow F}$ obtains 84.1\% AUC and ${\cal N}^{F \rightarrow O}$ 95.3\% AUC, which are lower than the 97.4\% obtained by the fused version. In the pixel-level evaluation, however, the performance of ${\cal N}^{O \rightarrow F}$ dropped to 30.1\%, while the ${\cal N}^{F \rightarrow O}$ is 66.2\%. 
We believe this is due to the low resolution of $\Delta_S$ (computed over the results obtained using
${\cal N}^{O \rightarrow F}$), which makes the pixel-level localization a hard task.
By fusing appearance and motion we can refine the detected area, which leads to a better localization accuracy.


%
\begin{figure}
		\begin{center}
			\small{\hspace{1.4cm}real frame  \hspace{1.5cm} generated frame \hspace{1.5cm} generated OF \hspace{0.8cm} abnormality heatmap}
			
			\scriptsize{(a)}
			\includegraphics[width=.24\linewidth]{images/ped1_1}
			\includegraphics[width=.24\linewidth]{images/ped1_2}
			\includegraphics[width=.24\linewidth]{images/ped1_3}
			\includegraphics[width=.24\linewidth]{images/ped1_4}

			\scriptsize{(b)}
			\includegraphics[width=.24\linewidth]{images/ped2_1}
			\includegraphics[width=.24\linewidth]{images/ped2_2}
			\includegraphics[width=.24\linewidth]{images/ped2_3}
			\includegraphics[width=.24\linewidth]{images/ped2_4}
			
			\scriptsize{(c)}
			\includegraphics[width=.24\linewidth]{images/ped2_conf_1}
			\includegraphics[width=.24\linewidth]{images/ped2_conf_2}
			\includegraphics[width=.24\linewidth]{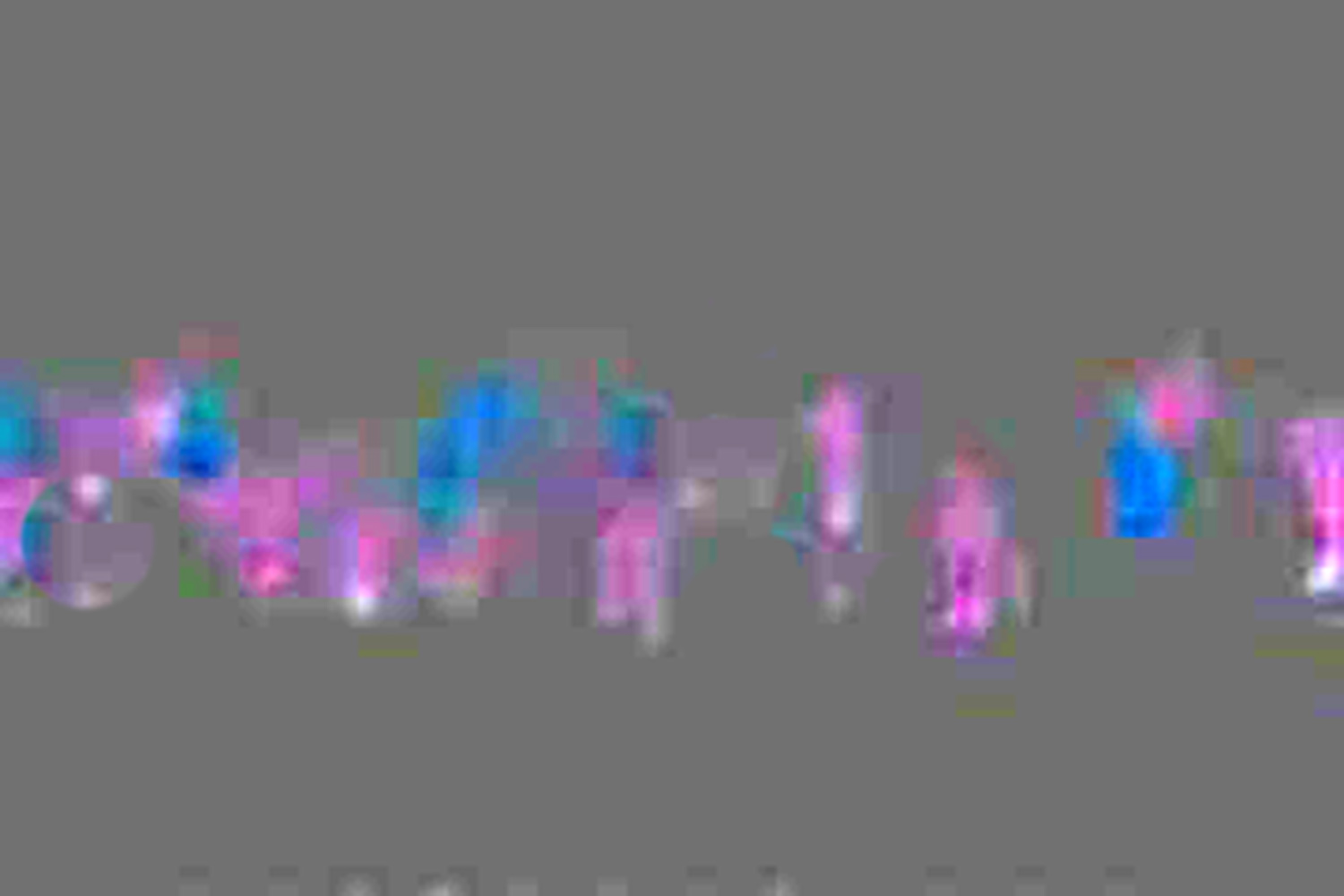}
			\includegraphics[width=.24\linewidth]{images/ped2_conf_4}
			
		\end{center}
		\caption[cGANs: Examples of abnormality localizatio]{Some examples of abnormality localization of our adversarial generator approach on UCSD. Visualizing the abnormality score heat-maps. (a) Ped1, (b) Ped2 (c) error case.}
		\label{fig:pedvis}
	\end{figure}
\noindent\textbf{Qualitative results for adversarial generator baseline (GAN-CNN).}
Fig.~\ref{fig:pedvis} shows some results using the standard visualization protocol for abnormality localization (red pixels represent abnormal areas). The figure shows that our approach can successfully localize different abnormality types. Moreover, since the generator learned a spatial distribution of the normal motion in the scene, common perspective issues are automatically alleviated. Fig.~\ref{fig:pedvis} also shows the intuition behind our approach. Normal objects and events (e.g., walking pedestrians) are generated with a sufficient accuracy. However, the generators are not able to reproduce abnormal objects and events (e.g., a vehicle in the first row) and this inability in reproducing abnormalities is what we exploit in order to detect abnormal areas. 
The last row in Fig.~\ref{fig:pedvis} shows a failure case, miss detecting the abnormal object (a skateboard). The failure is probably due to the fact that the skateboard is very small, has a ``normal'' motion (the same speed of normal pedestrians), and is partially occluded. 

\graphicspath{{Chapter2/Figs/Vector/}{Chapter2/Figs/d/}}
\begin{figure*}
	\begin{center}
		\scriptsize{(a)}
		\includegraphics[width=.23\linewidth]{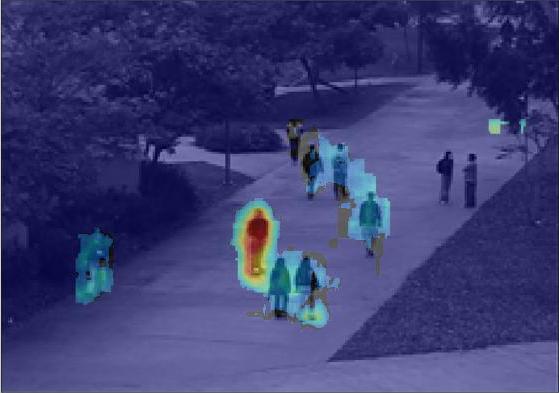}
		\includegraphics[width=.23\linewidth]{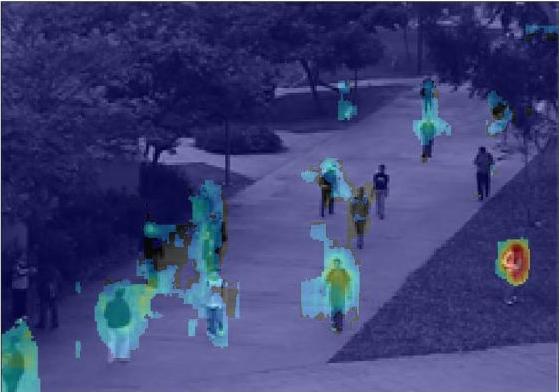}
		\includegraphics[width=.23\linewidth]{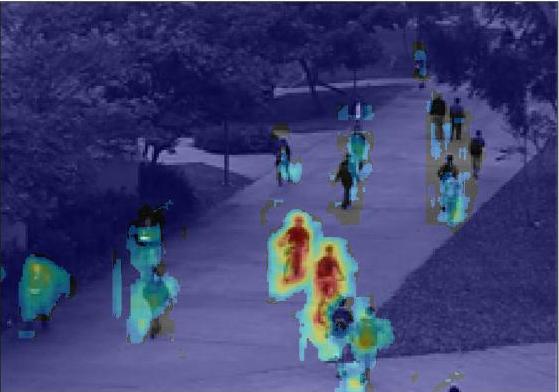}
		\includegraphics[width=.23\linewidth]{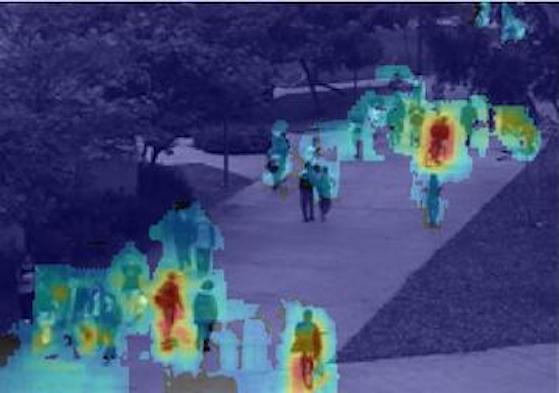}

		\scriptsize{(b)}
		\includegraphics[width=.23\linewidth]{images/ped2_4}
		\includegraphics[width=.23\linewidth]{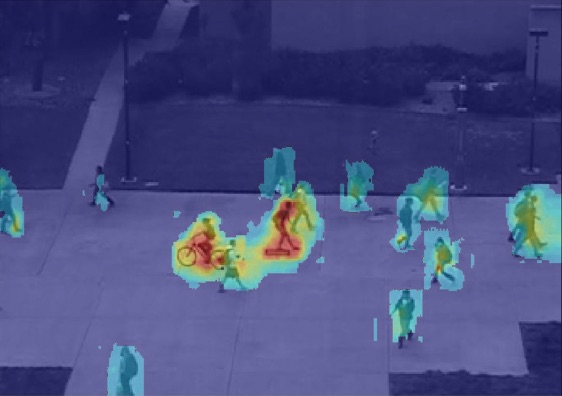}
		\includegraphics[width=.23\linewidth]{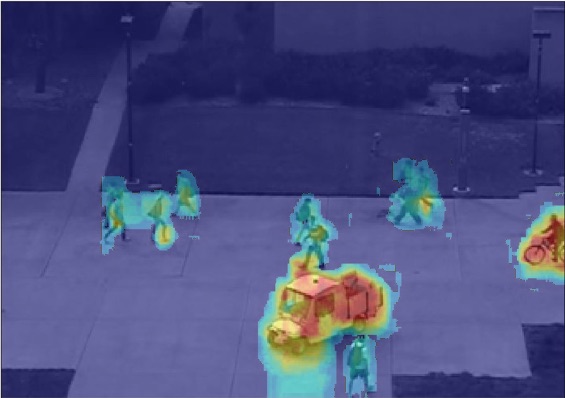}
		\includegraphics[width=.23\linewidth]{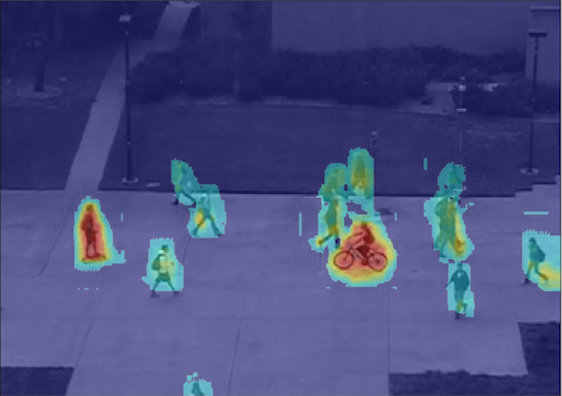}

		\scriptsize{(c)}
\includegraphics[width=.23\linewidth]{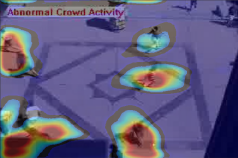}
\includegraphics[width=.23\linewidth]{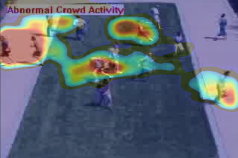}
\includegraphics[width=.23\linewidth]{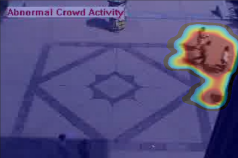}
\includegraphics[width=.23\linewidth]{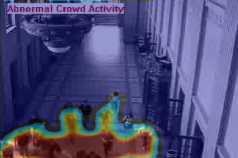}

\scriptsize{(d)}
\includegraphics[width=.23\linewidth]{images/ped2_conf_4}
		\includegraphics[width=.23\linewidth]{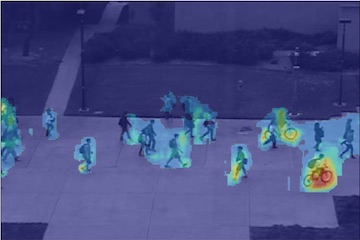}
		\includegraphics[width=.23\linewidth]{images/ped1_conf_4}
		\includegraphics[width=.23\linewidth]{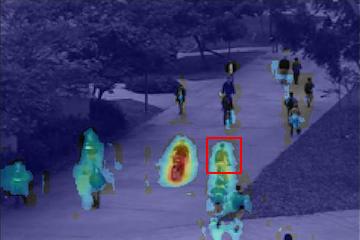}
		
	\end{center}
	\caption[cGANs: Examples of pixel-level detections]{Examples of pixel-level detections of adversarial discriminator. The abnormality heat-maps: (a) Ped1, (b) Ped2, (c) UMN, (d) errors (red rectangles highlight the errors).} 
	\label{fig:qual-det-succ}
\end{figure*}

\noindent\textbf{Qualitative results for adversarial discriminator baseline.}
In this section we show some qualitative results of our generators 
$G^{F \rightarrow O}$ and $G^{O \rightarrow F}$ (Fig.~\ref{fig:qual2-O-F}) and some detection visualizations of the {\em Adversarial Discriminator} output. Fig.~\ref{fig:qual2-O-F} show that the generators are pretty 
good in generating normal scenes. However, high-resolution structures of the pedestrians are not accurately reproduced. This confirms that the  data distribution and the generative distribution do not completely overlap each other (similar results have been observed in many other previous work using GANs
\cite{NIPS2014_5423,Isola_2017_CVPR,nguyen2016ppgn,DBLP:journals/corr/RadfordMC15,DBLP:conf/nips/SalimansGZCRCC16}). On the other hand, abnormal objects or fast movements  are completely missing from the reconstructions:  the generators simply cannot reconstruct what they have never observed during training. This inability of the generators in reconstructing anomalies is directly exploited by both {\em Adversarial Generator} and {\em GAN-CNN} (Sec.~\ref{sec:ablation}) and 
intuitively  confirms our hypothesis that anomalies are treated as outliers of the data distribution 
(Sec.~\ref{sec:intro},\ref{sec:learing}).
Fig.~\ref{fig:qual-det-succ} shows a few pixel-level 
detection examples of the {\em Adversarial Discriminator} in different
situations.
Fig.~\ref{fig:qual-det-succ}-(d) shows some detection errors. Most of the errors (e.g., miss-detections) are due to the fact that the abnormal object is very small or partially occluded 
(e.g., the second bicycle)
and/or has a ``normal'' motion (i.e., the same speed of normally moving pedestrians in the scene). 
The other sample shows a false-positive example (the two side-by-side pedestrians in the bottom), which is probably due to the fact that their bodies are severely  truncated and the visible body parts appear to be larger than normal  due to perspective effects. 

%
%
\begin{figure*}
		\centering
		\includegraphics[width=0.87\linewidth]{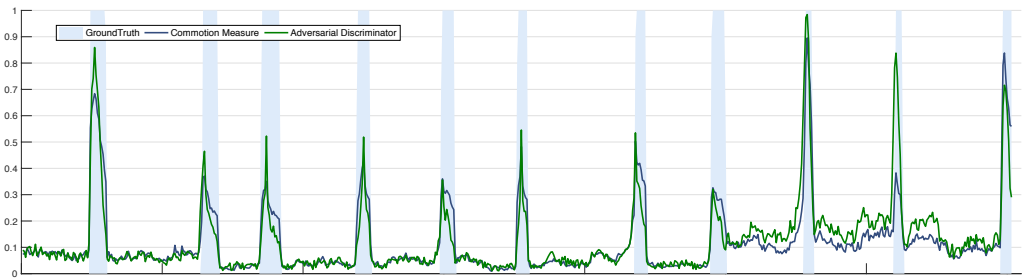}
	\caption[cGANs: Visualization of the ``detection  signal'' on UMN]{Visualization of the ``detection  signal'' as proposed in \cite{mousavi2015crowd,Ionescu_2017_ICCV}.
	}
	\label{fig:signal}
\end{figure*}
\noindent\textbf{Abnormality detection signal. } 
To further study the behaviour of adversarial discriminator for frame-level abnormality detection, the ``detection signal'' is visualized in Fig.~\ref{fig:signal} on the sequences of UMN dataset using the visualization methodology proposed in \cite{mousavi2015crowd,Ionescu_2017_ICCV}. For this purpose, all the scores of all the pixel-level prediction values in a given test frame are averaged and normalized in $[0,1]$, then plotted in a timely fashion. 
In this figure, the horizontal axis shows the time ($s$) and the vertical axis representing the normalized abnormality scores. Note that the ground truth abnormal events label (represented in light blue), is highly correlated with the abnormality scores generated by our proposed adversarial discriminator.

\begin{table}
	\begin{center}
		\begin{tabular}[width=\textwidth]{l cc}
			\toprule
			Baseline &  EER & AUC \\
			\midrule
			Adversarial Generator & 15.6\% & 93.4\%\\
			Same-Channel Discriminator & 17\%& 88.7\%\\
			Adversarial Discriminator F & 24.9\%& 81.6\%\\
			Adversarial Discriminator O &13.2\% &90.1\% \\
			Adversarial Discriminator &\textbf{11\%}& \textbf{95.5\%}\\
			GAN-CNN &\textbf{11\%}& 95.3\%\\
			\toprule
		\end{tabular}
	\end{center}
	\caption[cGANs: Results of the ablation analysis on the UCSD]{Results of the ablation analysis on the UCSD dataset, Ped2 (frame-level evaluation).}
	\label{tbl:ablation}
\end{table}

\subsection{Analysis of the method: Ablation studies}
\label{sec:ablation}

In this section we analyse the main aspects of the proposed method, which are: the use of the discriminators trained by our conditional GANs as the final classifiers, the importance of the cross-channel tasks and the influence of the multiple-channel approach (i.e., the importance of fusing appearance and motion information).
For this purpose we use the UCSD Ped2  dataset (frame-level evaluation)
and we test different strong baselines obtained by amputating important aspects of our method.

The first baseline, called {\em Adversarial Generator}, is obtained using the reconstruction error of $G^{F \rightarrow O}$ and $G^{O \rightarrow F}$, which are the  generators trained as in Sec.~\ref{sec:tasks}. In more detail,  at testing time we use $G^{F \rightarrow O}$ and $G^{O \rightarrow F}$ to {\em generate} a channel transformation of the input frame $F$ and its corresponding optical-flow image $O$. Let $r_O = G^{F \rightarrow O}(F)$ and $r_F = G^{O \rightarrow F}(O)$. Then, similarly to Hasan et al. \cite{DBLP:conf/cvpr/0003CNRD16}, we compute the appearance reconstruction error using: $e_F = |F - r_F|$ and the motion reconstruction error using: $e_O = |O - r_O|$. When an anomaly is present in $F$ and/or in $O$, $G^{F \rightarrow O}$ and $G^{O \rightarrow F}$ are not able to accurately reconstruct the corresponding area (see Sec.~\ref{sec:exp} and Fig.~\ref{fig:qual2-O-F}). Hence, we expect that, in correspondence with  these abnormal areas, $e_F$ and/or $e_O$ have higher values than 
the average values computed when using normal test frames. 
The final abnormality map is obtained by applying the same post-processing steps described in Sec.~\ref{sec:Detection}:
(1) we upsample the reconstruction errors, 
(2) we normalize the
the two  errors with respect to all the  frames in the test video $V$ and in each channel independently of the other channel,
(3) we fuse the normalized maps and (4) we use optical-flow to filter-out non-motion areas. The only difference with respect to the corresponding post-processing stages adopted in case of {\em Adversarial Discriminator} and described in Sec.~\ref{sec:Detection} is a weighted fusion of the channel-dependent maps by weighting the importance of $e_O$ twice as the importance of $e_F$. 

In the second strong baseline we do use the discriminators but we do not use a cross-channel strategy. In more detail, we train two networks ${\cal N}^{F \rightarrow F}$ and ${\cal N}^{O \rightarrow O}$ as described in Sec.~\ref{sec:tasks} but we use two ``same-channel'' tasks, using the generators of the two networks for autoencoding tasks, similarly to \cite{DBLP:conf/cvpr/0003CNRD16} and \cite{xu2015learning}. Note that the noise ($z$) provided by dropout (Sec.~\ref{sec:tasks}) acts as a form of input perturbation. At testing time we use the corresponding patch-based discriminators $\hat{D}^{F \rightarrow F}$ and $\hat{D}^{O \rightarrow O}$. The rest of the pipeline (training, post-processing, etc.) is the same as in {\em Adversarial Discriminator}. We call this baseline {\em Same-Channel Discriminator}.

In {\em Adversarial Discriminator F} we use only $\hat{D}^{O \rightarrow F}$ and in 
{\em Adversarial Discriminator O} we use only $\hat{D}^{F \rightarrow O}$. These two baselines show the importance of channel-fusion. 

The results are shown in  Tab.~\ref{tbl:ablation}. 
 It is clear that {\em Adversarial Generator} achieves a very high accuracy: Comparing {\em Adversarial Generator} with all the methods in Tab.~\ref{tbl:UCSD_gan}
(except our {\em Adversarial Discriminator}), it is the state-of-the-art approach.
Conversely, the overall accuracy of {\em Same-Channel Discriminator} drops significantly with respect to {\em Adversarial Discriminator} and is also clearly worse than {\em Adversarial Discriminator O}. This shows the importance of the cross-channel tasks. However, comparing {\em Same-Channel Discriminator} with  the values in 
Tab.~\ref{tbl:ablation}, also this baseline outperforms or is very close to the best performing systems on this dataset, showing that the discriminator-based strategy can be highly effective even without cross-channel training. 

Finally, the worst performance was obtained by {\em Adversarial Discriminator F}, with values much worse than {\em Adversarial Discriminator O}. We believe this is due to the fact that {\em Adversarial Discriminator O} takes as input a real frame which contains  much more detailed information with respect to the optical-flow input of {\em Adversarial Discriminator F}. However, the fusion of these two detectors is crucial in boosting the performance of the proposed method {\em Adversarial Discriminator}.

It is also interesting to compare our {\em Adversarial Generator}
with the Convolutional 
Autoencoder proposed in \cite{DBLP:conf/cvpr/0003CNRD16},
being both based on the reconstruction error (see Sec.~\ref{sec:intro}).
The results of the Convolutional 
Autoencoder on the same dataset are:
$21.7\%$ and $90\%$ EER and AUC, respectively (Tab.~\ref{tbl:UCSD_gan}), which are significantly 
worse than our  baseline based on GANs.


%
%



\begin{table}
	\begin{center}
		\begin{tabular}[width=\textwidth]{l cc}
			\toprule
			&	Adversarial Discriminator	&	GAN-CNN\\
			\midrule
			net feed-forward &   0.0134						& 0.120 \\
			post-processing &   0.0153						& 0.07 \\
			total 			&   \textbf{0.0287}				& 0.190 \\
			\toprule
		\end{tabular}
	\end{center}
	\caption[cGANs: Average run-time (seconds) for one frame testing]{Average run-time (seconds) for one frame testing using two different versions of our approach.}
	\label{tbl:comp.times}
\end{table}

Finally, in the last row of  Tab.~\ref{tbl:ablation} we report the results published in \cite{ravanbakhsh2017abnormal}, where the authors adopted a strategy similar to the {\em Adversarial Generator} baseline above mentioned. The main difference between {\em GAN-CNN} \cite{ravanbakhsh2017abnormal} and {\em Adversarial Generator} is the use of an additional AlexNet-like CNN \cite{alexnet}, externally trained on ImageNet (and not fine-tuned) which takes as input both $F$ and the appearance generation produced by $G^{O \rightarrow F}(O)$ and computes a ``semantic'' difference  between the two images.
The accuracy results of {\em GAN-CNN} are basically on par with respect to the results obtained by the {\em Adversarial Discriminator} proposed in this section. However, in {\em GAN-CNN} a fusion strategy needs to be implemented in order to take into account both the semantic-based and the pixel-level reconstruction errors, 
while the testing pipeline of
{\em Adversarial Discriminator} is very simple. Moreover, even if the training computation time of the two methods is the same, at testing time {\em Adversarial Discriminator} is much 
faster because  
$G^{O \rightarrow F}$, $G^{F \rightarrow O}$ and the semantic network are not used. In Tab.~\ref{tbl:comp.times} we report the corresponding average computation times for one frame using a Tesla K40 GPU. The pre-processing operations (i.e., the optical-flow computation) are the same for both methods, while post-processing in  {\em GAN-CNN} includes passing both $r_F = G^{O \rightarrow F}(O)$ and $F$ through the AlexNet and computing the last-layer-based difference.

\section{Conclusions}
\label{sec:anomaly_conclusions}
Abnormal crowd behaviour detection attracts  a large interest due to its importance in video surveillance scenarios. However, the ambiguity and the lack of sufficient {\em abnormal} ground truth data makes end-to-end training of large deep networks hard in this domain. In this section, we introduced three methods to address the task of abnormality detection in crowded scenes.
\begin{itemize}
    \item We introduced the Local Binary Tracklets descriptor for the task of abnormality detection. A binary representation of tracklets is used to model the movement of a salient point over its corresponding tracklet in both spatial and temporal spaces. Such representation models the statistics of a tracklet at each frame in terms of magnitude and orientation. In this scheme, abnormal behaviors are detected at different levels: detection and localization. Since the abnormal samples are hardly accessible in real-world, we employed the generative one-class SVM to learn our models. We showed that our proposed descriptor, despite its simplicity, is competitive with the state-of-the-art.
    
    \item We proposed using a Fully Convolutional Network as a pre-trained model and plugged an effective binary quantization layer as the final layer to the net. Our method provides both spatial consistency as well as low dimensional semantic embedding. We then introduced a simple yet effective unsupervised measure to capture temporal CNN patterns in video frames. We showed that combining this simple measure with traditional optical-flow provides us with the complementary information of both appearance and motion patterns. 
    
    \item We presented a GAN-based approach for abnormality detection. We propose to use Generative Adversarial Nets (GANs), which are trained to generate only the {\em normal} distribution of the data. During  the adversarial GAN training, a discriminator ($D$) is used as a supervisor for the generator network ($G$) and vice versa. At testing time we use $D$ to solve our  {\em discriminative} task (abnormality detection),  where $D$  has been trained without the need of manually-annotated  abnormal  data. Moreover, in order to prevent $G$ learn a trivial identity function, we use a cross-channel approach, forcing $G$ to transform raw-pixel data in motion information and vice versa.\\
    
\end{itemize}

The qualitative and quantitative results on the most common abnormality detection benchmarks and a set of challenging datasets show that our method is comparable to the state-of-the-art methods. 

We performed a detailed ablation analysis of the proposed methods in order to show the contribution of each of the main components. Specifically, we compared the proposed approaches with several strong baselines (i.e, reconstruction-based baselines and same-channel encoding/decoding tasks) showing the overall accuracy and computational advantages of the proposed method.\\

	As future work, we will study plugging a TCP measure layer and fine-tuning this layer with back-propagation in a fully binary network \cite{rastegari2016xnor} by the means of obtaining an efficient compressed network for on-device running models \cite{kim2015compression,salama2018pruning,abdollah2019icassp,yu2018nisp}. Such models can be compressed and embedded into smartphones, autonomous cameras and drones for crowd analysis task and other privacy-sensitive use cases. Moreover, exploring the use of \cite{juefei2017local,rastegari2016xnor,wang2018source} as alternatives to Binary Fully Convolutional Net (BFCN) for end-to-end training of abnormality detection would be a potential direction. Also as a future work we will investigate the use of Dynamic Images \cite{Bilen16a} as an alternative to optical-flow in order to represent motion information collected from more than one frame.

\graphicspath{{Chapter3/Figs/a/}}
\chapter{Situational Change Detection for Autonomous Agents}
\label{ch:autonomous_car}

\begin{quote}
    \emph{``The whole of life, from the moment you are born to the moment you die, is a process of learning.''}\\
    
    -- Jiddu Krishnamurti
    
    

    

    
\end{quote}
\blfootnote{This Chapter was in collaboration with Damian Campo and Mohamad Baydoun}
\newpage
\newpage
\thispagestyle{plain} 
\mbox{}
\newpage
In recent years, developing autonomous systems that are able to learn constantly for assisting humans in everyday tasks has been studied in computer science. The continual learning problem is the matter of learning new concepts, knowledge or tasks while not forgetting old knowledge \cite{oleksycvpr2019,parisi2019continual,rolnick2018experience,schmidhuber1996general,shin2017continual,tessler2017deep,thrun2012learning}. The ultimate ideal goal is to provide a generalization from old experience to learn new concepts faster. A real autonomous agent, by its nature, needs to interact with a continuous dynamic permanent changing environment and continuously learns the novel unseen concepts. Such environment is not often available to train the agent on it, so the agent should have an understanding about the capacity and limitations of itself. Such understanding often called Self-awareness (SA). 

Self-awareness models make it possible for an agent to evaluate whether faced situations at a given time correspond to previous experiences. Self-aware computational models have been studied and several architectures have been introduced \cite{fusion18_damian,icassp2018,lewis2016self,fusion18_ravan}. Such models have to provide a framework where autonomous decisions and/or teleoperation by a human can be integrated as a capability of the device itself to dynamically evaluate the contextual situation \cite{Bellman2017,lewis2016self,schmidhuber1996general,wortsman2018learning}. An exemplar schematic representation of a self-aware agent is shown in \ref{fig:sa_driver}. Using such models, the agent gains the ability to either predict the future evolution of a situation or to detect situations potentially unmanageable. This ``sense of the limit'' is considered as a level of self-awareness and allows an agent predicting potential changes with respect to the previous experiences to involve a human operator for support in due time. In this sense, the capability of detecting the novel situations is an important feature included in self-awareness models as it can allow autonomous systems to anticipate in time their situation/contextual awareness about the effectiveness of the decision-making sub-modules \cite{Campo2017,Olier2017a}.
\begin{figure}[t] 
	\centering
	\includegraphics[width=\linewidth]{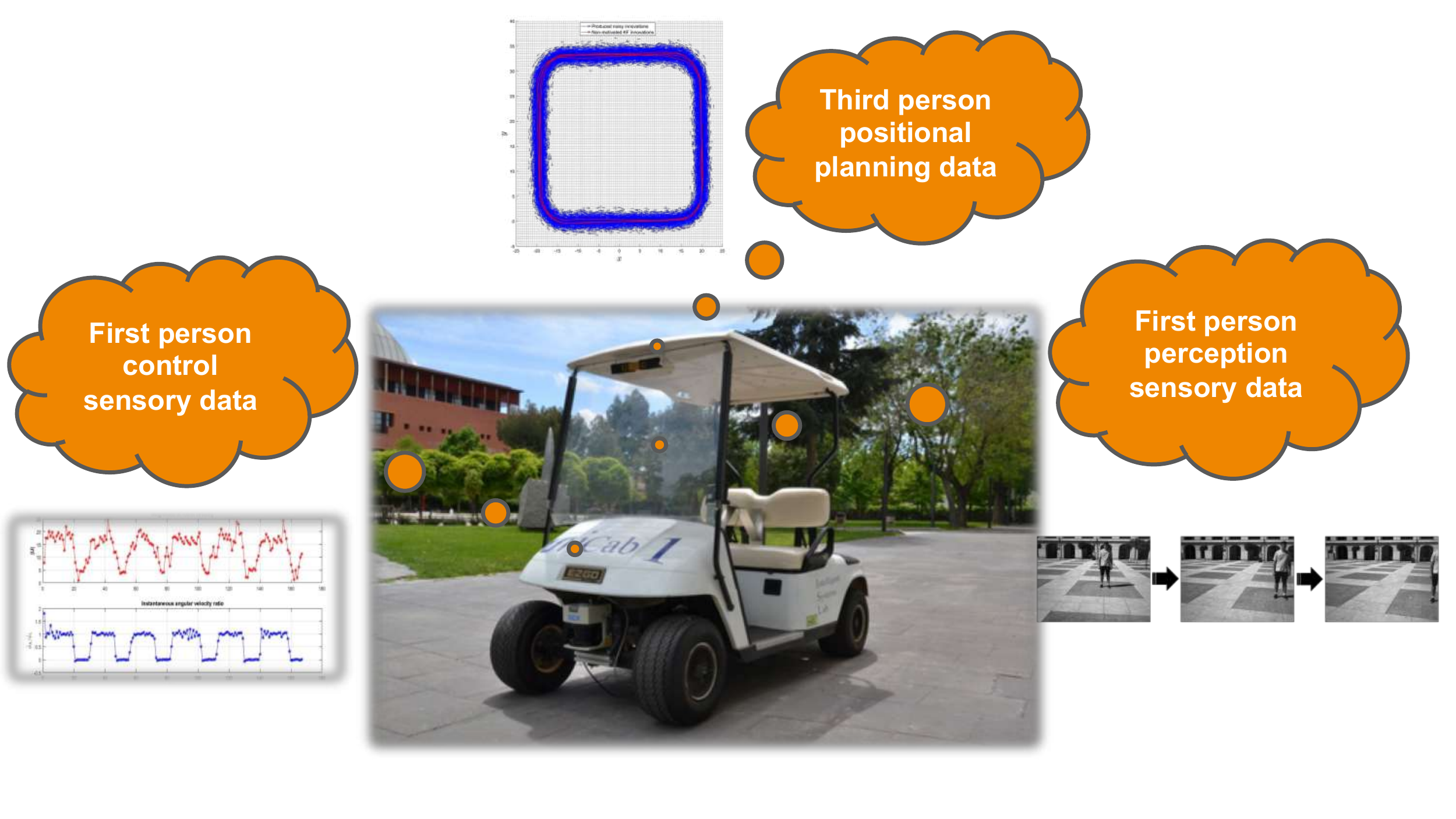}
	\caption{An autonomous agent can use different sensory data to capturing its perception}
	\label{fig:i_cub_sens} 
\end{figure}
In this chapter, dynamic incremental self-awareness (SA) models are proposed that allow experiences done by an agent to be learned. The agent starting to learn from more simple situations to more structured ones in a self-paced fashion \cite{jiang2015self,pahde2018self,sangineto2016self}. Each situation is learned from subsets of private agent perception data (See Fig. \ref{fig:i_cub_sens}) as a model capable to predict normal behaviors and detect abnormalities. Abnormalities are defined as the deviated situations from the previously learned model.

The contributions of our proposed method is well described in this chapters, however the main contributions is shortened below with a brief description:

\begin{itemize}
    \item \emph{i)} Proposing a multi-layer Self-awareness modeling to perceive the situations through different sensorial modalities and can be integrated in order to build a uniform structure of cross-modal self-awareness for an agent.
    
    \item \emph{ii)} A hierarchical model is introduced by means of a cross-modal Generative Adversarial Networks (GANs) processing high dimensional visual data. Different levels of the GANs are detected in a ``self-supervised'' manner using GANs discriminators decision boundaries.
    
    \item \emph{iii)} In addition, a probabilistic framework based on switching dynamic system is proposed to learn the SA layers from different sources. Such network potentially provides us with a robust self-aware model through a cross-correlation between the layers using multi-modal DBNs.
    
\end{itemize}
The rest of this chapter is organized as follow: Sec. \ref{sec:intro} describes a general overview on SA models, the procedure of training and testing our proposed SA model is discussed in Sec. \ref{sec:dynamics_modeling} and Sec. \ref{sec:online_testing_sa}, finally, in Sec. \ref{sec:exp_sa} results of evaluations are reported and discussed.

Finally, section \ref{ch:concl_future} draws some concluding remarks and summarizes the most valuable results obtained in this work, as well as future perspectives.

\begin{landscape}
\begin{figure}[t] 
	\centering
	\includegraphics[width=\linewidth]{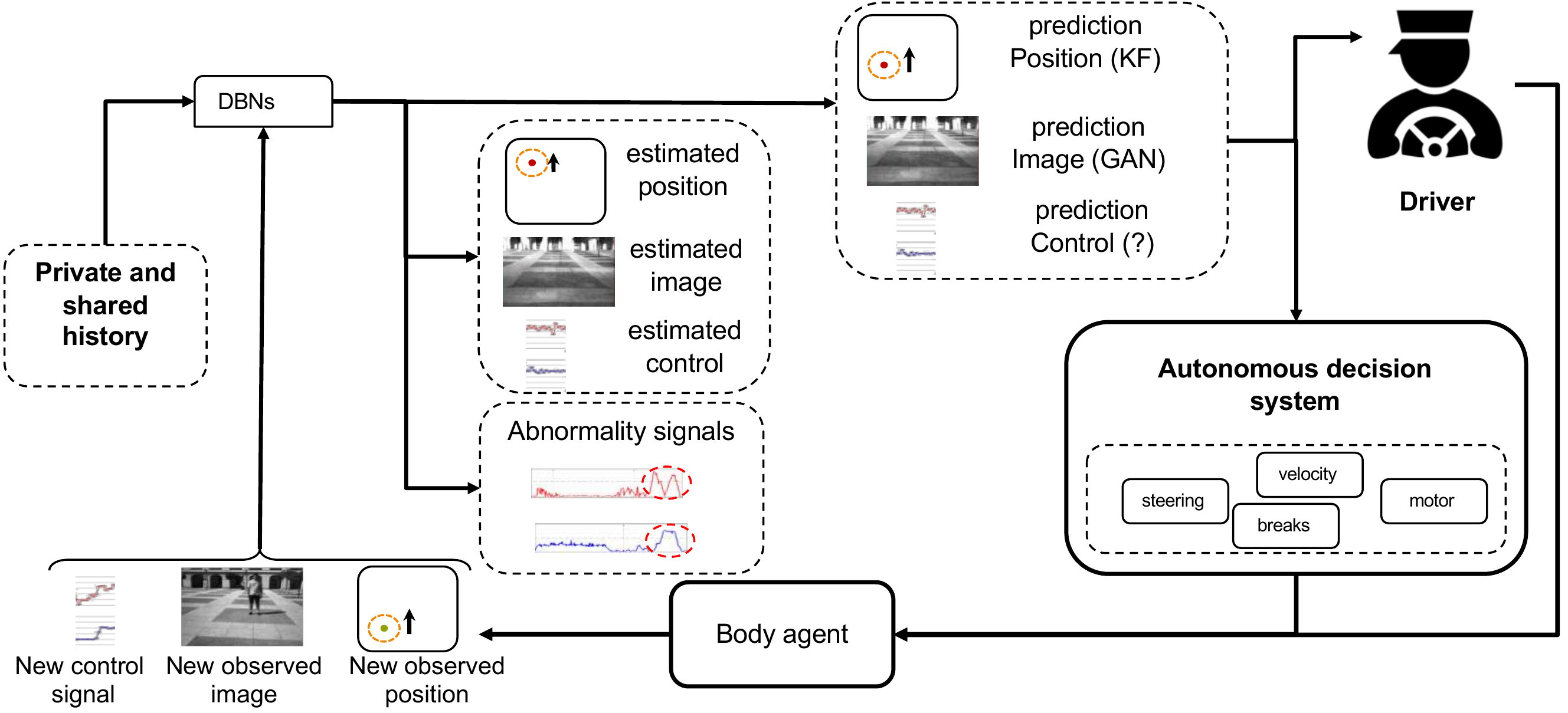}
	\caption[SA: Representation of Self-aware agent proposal]{Representation of Self-aware agent proposal: A Self-awareness models make it possible for an agent, in this case an autonomous car, to evaluate whether the current observed situations correspond to previous learned experiences. The knowledge stored as learning history modeled by probabilistic dynamic switching system. In this sense, the agent predicts potential changes with respect to the previous experiences to either provide a direct feedback to the autonomous decision system or involve a human operator for support in due time.}
	\label{fig:sa_driver} 
\end{figure}
\end{landscape}

\graphicspath{{Chapter3/Figs/a/}}

\section{Background}
\label{sec:intro}
Self-awareness refers to system capability to recognize and predict its own states, possible actions and the result of these actions on the system itself and its environment \cite{Schlatow2017}. Recent developments in signal processing and machine learning techniques can be useful to design autonomous systems equipped with a self-awareness module that facilitates the similarities between current realizations and previous experiences related to a given executed task. The capability of predicting the task's evolution in normal conditions (\emph{i.e.}, when the task follows the rules learned in previous experiences) and jointly detecting abnormal situations allows autonomous systems to increase their situational awareness and the effectiveness of the decision making sub-modules \cite{icassp2018,Campo2017}. Models of different self-awareness layers can be integrated to build up a structured multi-modal self-aware behavior for an agent.

In \cite{icassp2018,ravanbakhsh2018hierarchy} a self-awareness model was introduced that consists of two layers: Shared Level (SL) and Private Layer (PL). The analysis of observed moving agents for learning the models of normal/abnormal dynamics in a given scene from an external viewpoint, represents an emerging research field \cite{Bastani2016,lin2017tube,Morris2008,nabi2013temporal,Moll2010,rabiee2016crowd,ravanbakhsh2016plug,sabokrou2016fully,sebe2018abnormal}. The planned activity of an entity is one of such models that can be defined as the sequence of organized state changes (actions) that an entity has to perform in a specific context to achieve a task. This set of actions can be learned from examples and clustered into sequential discrete patterns of motions. The availability of a plan that associates the current state with an action class makes it possible to detect normal/abnormal situations in future repetitions of the same task. In general, computational models for abnormality detection are trained on a set of observations corresponding to standard behaviors. Accordingly, abnormalities can be defined as observations that do not match with the patterns previously learned as regular, \emph{i.e.}, behaviors that have not been observed before\cite{Ramik2014}.

An active self-aware action plan can be considered as a dynamical filter that predicts and estimates state behaviors using linear and non-linear dynamic and observation models. Switching Dynamical Systems (SDSs) are well-known Probabilistic Graphical Models (PGMs) that are capable of managing discrete and continuous dynamic variables jointly in dynamical filters. Accordingly, we propose to learn a switching Dynamic Bayesian Network (DBN) that facilitates the prediction and estimation of future and present states at continuous and discrete levels dynamically. SDSs have been used successfully to improve decision making and tracking capabilities \cite{Bastani2016}. In SDSs, each dynamical model linking together continuous state variables in successive time instants is associated with a discrete random variable representing higher level motivations of the dynamic model. Shareable observations related to actions that are a function of agent positions (plan) can be used for learning active probabilistic plan models including hidden continuous and discrete states.  Markov Jump Linear Systems (MJLS) \cite{Doucet2000} is one of the most used algorithms for taking advantage of learned hierarchical probabilistic knowledge in the online SDS phase. MJLS uses a combination of Kalman filter (KF) and particle filter (PF) to predict and update the joint continuous and discrete state space posterior probabilities. In this work, an MJLS is used to exploit learned knowledge for the SL that includes the capability to self-detect abnormality situations.

However, an agent can also infer dynamic models with respect to inner variables that it can observe from a First Person viewpoint while performing the same task; this private knowledge is only directly accessible to the agent itself. Detecting abnormalities by using a self-awareness model learned starting from private multisensorial first person data acquired by the agent can be possible, while doing the same task for which the self-awareness SL model has been obtained. Such a model can be defined as the Private Layer (PL) of self-awareness. An external observer has no access to such information, so not being able to directly detect PL abnormalities, while it can still evaluate SL abnormalities using third person models including shared variables. A PL model can allow an agent to be able to evaluate abnormalities related to PL and SL models, as it was shown in \cite{icassp2018}. However, previous works mostly rely on a high level of supervision to learn PL self-awareness models \cite{Olier2017a,Ramik2014}, while in this work, we propose a weakly-supervised method based on a hierarchy of Cross-modal Generative Adversarial Networks (GANs) \cite{NIPS2014_5423} for estimating PL models. Weakly supervised PL models not can also provide a level of information to boost the SL model as well as they can be used to provide a joint self aware multisensorial modality to cross-predict heterogeneous multimodal anomalies related to the same task execution. This work describes a novel method to learn the PL model using an incremental hierarchy of GANs \cite{NIPS2014_5423,ravanbakhsh2017abnormal}. The private camera views acquired by first-person images taken by an agent during task execution can be used together with related optical-flows to learn models using Generative Adversarial Networks (GANs), more appropriate for reducing high dimensionality for the visual modalities.

\section{Cognitive Dynamic Modeling}
\label{sec:dynamics_modeling}
As reviewed in Sec. \ref{sec:intro}, the objective is to incrementally learning new dynamic equilibrium models from data acquired along agent experiences as a part of incremental self-awareness (SA) model. For example in this work we will consider action of a human driver along different driving experiences as observed by an agent corresponding to the driven car (that play the role of "First Person" in this case). The car can record both video sequences and estimated position data during the experience by its own endo/esosensors. We propose an incremental adaptive process that makes it possible the car agent learns switching DBN models from recorded data. Such learned DBN models not only can predict (i.e., to generate) the new perturbed situations occurring according to different stationary probabilistic rules with respect to reference ones, but also they capable to adaptively estimate the current states by filtering data with respect to the most fitting model (i.e., to discriminate). In other words, the learned DBN models allowing both prediction and estimation for situations different from the reference dynamic equilibrium.

\begin{figure*}[h!]
	\includegraphics[width=0.96\linewidth]{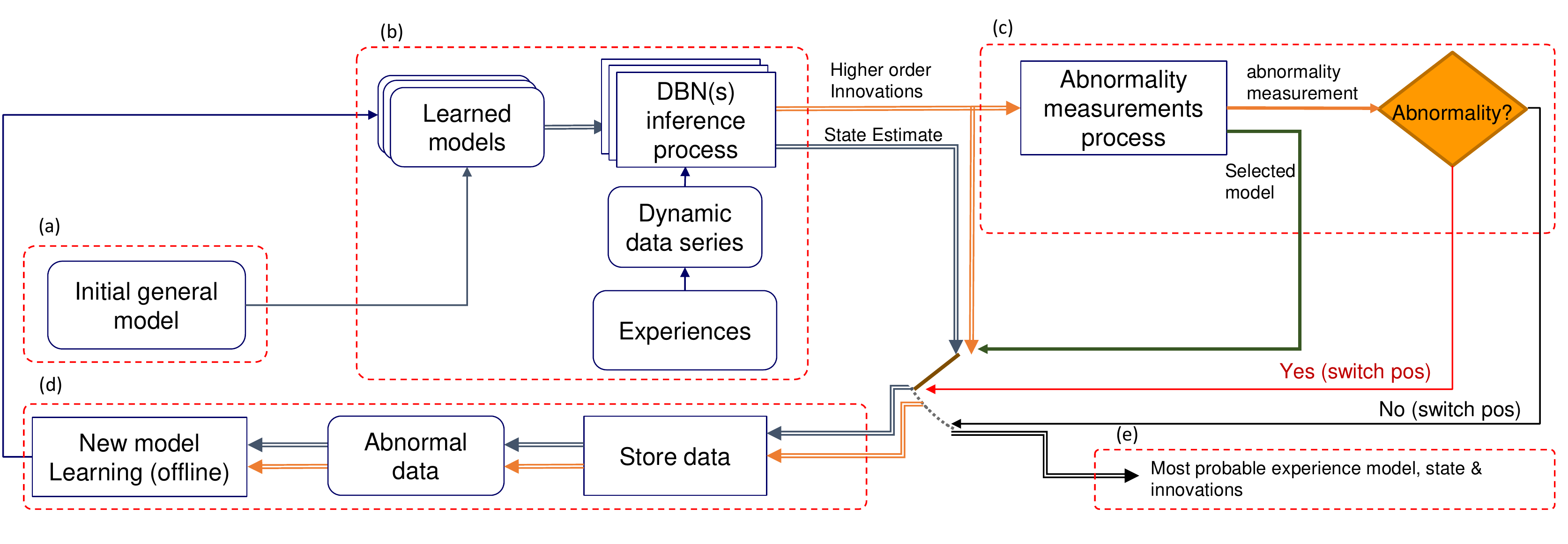}
	\caption[SA: Generic block diagram of learning process]{Generic block diagram of learning process: state DBN Learning starts with an initial general model, during an experience the new observed situation with respect to the general model detects as abnormality. Detected abnormalities are stored and they can be used to learn a new model, once a learned model is available different than the initial general model the process can be iterated.}
	\label{fig:main_bd}
\end{figure*}

In this work, multi-modal self-awareness models, and their learning process are proposed. The proposed self-awareness (SA) model consists of two levels: The private layer (PL), and the shared level (SL). Each level of self-awareness includes different modalities, where for each modality an unified incremental learning procedure is designed (See Fig. \ref{fig:main_bd}). The learning process is considered as a differential incremental process adding generative and discriminative knowledge to a reference model that describes a very general dynamic equilibrium situation between the agent and the environment. A statistically significant deviations from such a dynamic equilibrium can be considered as abnormal situations whose characteristics have to be captured by the new learned models. An agent can so take advantage of new experiences perturbating the reference dynamic equilibrium situation to increase its models by adding new definition of alternate possible dynamic equilibrium models. All together the availability and the capability to use such a set of models allow the agent to be provided of self-awareness capability that make it possible to understand the different interactions it can perform with the environment when doing a task as well as to decide when available models are not sufficient to predict task evolution in a safe way. This latter decision is shown can be used to incrementally enrich SA models by successive learning steps. In our experiments private level and shared level are learned based on visual perception (first person vision data available to the agent only) and localization (third-person vision) that can be also observed by other agents in the environment, respectively. A probabilistic framework based on a set of switching dynamical models \cite{fusion18_damian,Doucet2000} is used to incrementally learn the SL filters models. An incremental bank of cross-modal Generative Adversarial Networks (GANs) is used to learn the PL models. The rest of this section is dedicated to introduce the generic procedure of learning dynamics models from a set of observed experiences, and detailed explanation of learnings for each level of SA.

\begin{figure*}
	\includegraphics[width=0.96\linewidth]{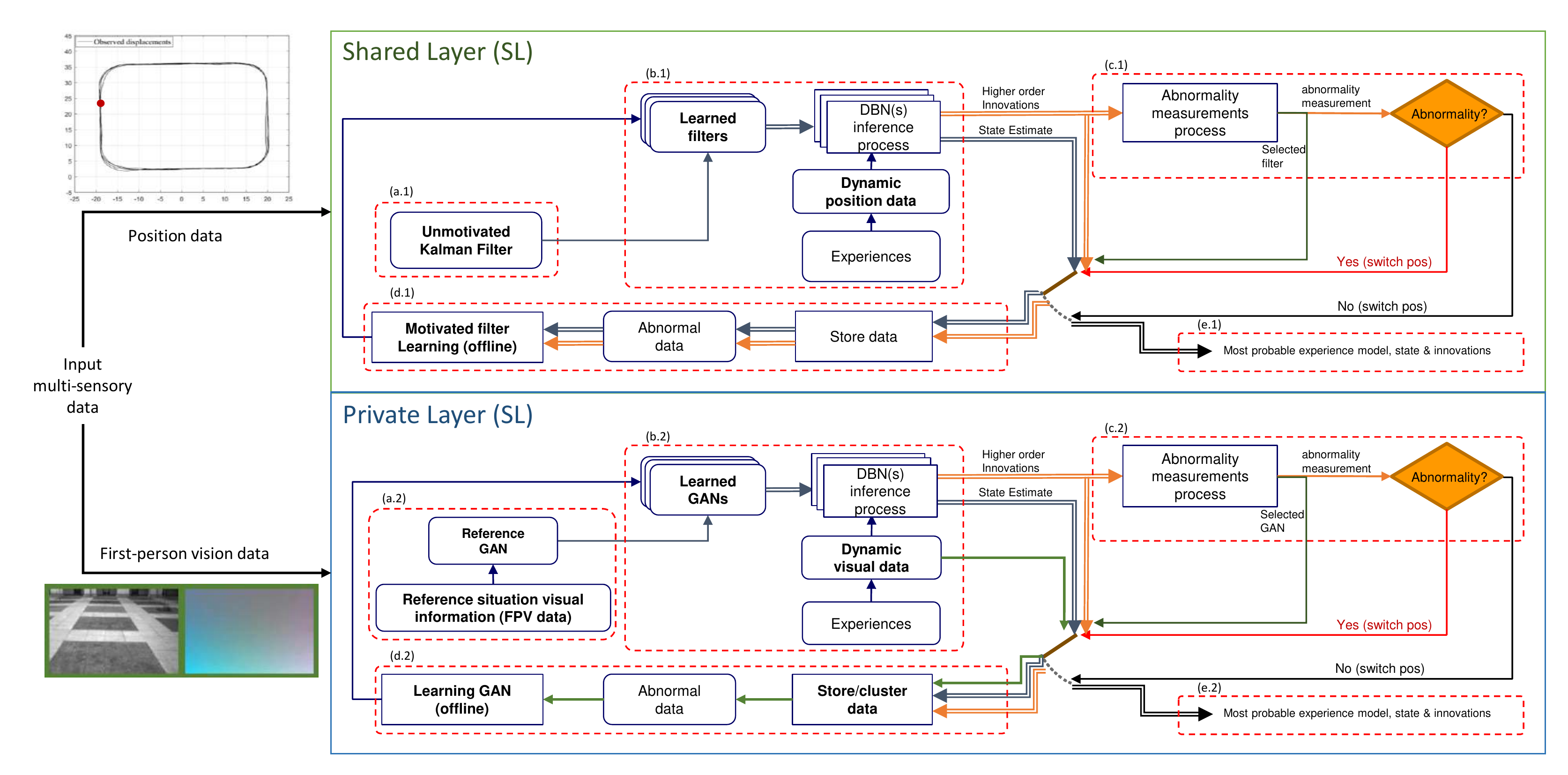}
	\caption[SA: Learning Process in SL and PL]{Learning Process in SL and PL: for each individual experience the position information input to the SL, while the first person vision data are used as a source of information for PL. Both layers structured based on the generic incremental adaptive training process, so they would be able to increase self-awareness capabilities by learning the multi-perspective observations acquired in new experiences.}
	\label{fig:learning_bd}
\end{figure*}

A generic adaptive learning process flow is shown in Fig. \ref{fig:main_bd}, allowing an agent to incrementally learn the additional knowledge necessary to describe the new dynamic equilibrium situation by encoding states into a new switching model that increases the self-awareness of the agent. Learning of a new switching model starts by observing a dynamic data series related to an experience (Fig. \ref{fig:main_bd}-b). Such data is filtered by using the initial reference model and by keeping trace of deviations w.r.t the associated dynamic model (Fig. \ref{fig:main_bd}-a). The reference model consists of a simple filter in its category whose dynamic model describes a basic dynamic equilibrium situation w.r.t the type of data considered. For example, in the case of SL where low dimensional data like position and velocity are considered, the initial filter can correspond to a Kalman filter with the simplest dynamic model, assuming the agent remain static with null speed. Such a filter would make reasonable predictions if no force is present in the environment. In that case, agent motion can be due only to random noise oscillations of the state. In particular, on the shared level of SA position observations, that can be done both by agent and by third person external entities, consist of a set noisy data series of sparse measurements. In this case, filtering an agent experience by applying the reference Kalman filter on the related data series corresponds to hypothesize that a null force is deriving from the interaction between the agent and the environment. Therefore, we define it as the Unmotivated-agent Kalman Filter (UKF). Such a reference filter is illustrated in Fig. \ref{fig:learning_bd}-a.1. UKF assumes no force motivates the agent, then $\dot{x} = \omega$, where $\dot{x}$ is the derivative of state $x$ and $\omega$ is the perturbation error. When UKF applied along a (motivated) experience, it produces a set of errors due to the fact that the object in that case is motivated (\emph{i.e.}, it is causally determined to move according to forces depending on the environment). In case of private layer of SA a pre-trained reference GAN is used as an initial general model (see Fig. \ref{fig:learning_bd}-a.2), where it trained on an experience where the agent is moving straight on an empty path. Similar to the UKF general model, the reference GAN assumes a dynamic equilibrium is present between the environment and the agent, but on different data type(\emph{i.e.}, first person video data). In this case the reference equilibrium corresponds to the agent visual data as they are dynamically modified by a stationary force that attracts the agent during a linear straight motion towards a fixed motivation point. If another forces present in the environment the dynamics of visual data is changed with respect to the one experienced along  straight movements, and consequently, a set of prediction mismatches between the GAN filter predictions and updating new observations are produced as errors. Considering the set of errors (here defined as innovations, using terminology from Kalman Filters) obtained from UKF and the reference GAN, cumulative probabilistic tests can be designed for both SL and PL to evaluate if a data series corresponding to an experience can be considered abnormal. Moreover, such innovations can be used to decide when collecting data to learn a new filter. The new filter learns a new stationary dynamic equilibrium condition, where this new condition is caused due to different but stationary forces. This process is shown in Fig. \ref{fig:main_bd}-c, where different type of probabilistic abnormality measurements of the switching models are estimated and thresholded. In the abnormality measurement process block a set of related abnormality tests can be considered to detect the possible anomalies with respect to the already learned dynamic equilibrium conditions. This process not only ranks the innovations and computes the abnormality measurements, but also it can be used to select the most probable model among those learned from previous experiences. The most probable model could be a SDS in case of SL or a couple of cross-modal GANs for PL, and abnormality measurements can be seen as comparable evaluations that can drive a soft decision process \cite{mazzu2016cognitive}. Abnormality measurments input to the abnormality detection block (see Fig. \ref{fig:main_bd}-d), where it compare with a threshold in order to detect possible deviations from the learned normalities.

Such procedure allows defining an incremental process similar to Dirichlet \cite{antoniak1974mixtures} (stick-breaking processes \cite{sethuraman1994constructive}, Chinese restaurant \cite{aldous1985exchangeability},...) where the abnormality measurements are the variables that determine the choice whether a new experience will to sit into an already available experience table and can be still described by the currently learned dynamic equilibrium conditions embedded into DBN (Fig. \ref{fig:main_bd}-e) or there is a need to learn a new one (Fig. \ref{fig:main_bd}-d), opening a new table. In fact, the new experiences data can be structured into multiple partitions of the state-innovation, where the correlations between instantaneous states and innovations allows clustering the new data into classes characterized by different parameters, forming a new learned model. Fig. \ref{fig:main_bd}-d shows the procedure of learning new models from state-innovation pairs. Such data couple establish a relationship between the states and an error measurement (innovations) obtained through the initial models, and can be clustered through the new learning model process. 
In particular, in SL the new learning model process results a set of regions which segment the state space motion depending on innovations (See Fig. \ref{fig:learning_bd}-d.1), and equivalently on the PL it can be considered as clustering multidimensional consecutive images and their corresponding optical flows (change) maps based on a similarity measurement (\emph{i.e.}, local innovation), shown in Fig. \ref{fig:learning_bd}-d.2. Eventually, detected regions/clusters can form a explicit (in case of SDS) or implicit (in case of GANs) vocabulary of switching variables such that the new learned model can address adaptively different models when it will have to evaluate new states produced by new experiences. Note that, in case of PL to train new set of GANs, there is a need for accessing dynamic visual information. Hence, as it shows in Fig. \ref{fig:learning_bd}-d.2, for new detected clusters based on state-innovation pairs the original data is used directly for training the new GANs.
In this sense we can say that learned models in case of PLs are all related to different effects of forces different from the one producing a straight motion onto a video sequence. For example a curve in two different orientation can generate two different GANs.

\subsection{Shared-level self-awareness modeling}
\label{subsec:learn_SL}
This level of self-awareness focuses on the analysis of observed moving agents for understanding their dynamics in a given scene. This model is said to be shared as corresponding measurements can be directly observed by the agent itself or external observers.

As mentioned before, the initial general model is based on a simple dynamic model. A Kalman Filter (KF) based on an ``unmotivated model'' in Fig. \ref{fig:learning_bd}-a.1 is learned for tracking agents' motions. Such dynamic model is a simple filter describing a basic dynamic equilibrium situation, where an observed object moves according to a random dynamic model considered as: 
\begin{equation}\label{eq1.0}
X_{k+1} = AX_{k} + w_{k},
\end{equation} 
where $X_{k}$ represents the agent's state composed of its coordinate positions and velocities at a time instant $k$, such that $X_k = [\boldsymbol{x} \hspace{0.2cm} \boldsymbol{\dot{x}}]^\intercal$. $\boldsymbol{x} \in \mathbb{R}^d$ and $\boldsymbol{\dot{x}} \in \mathbb{R}^d$. $d$ represents the dimensionality of the environment. $A = [A_1 \hspace{0.2cm} A_2]$ is a dynamic model matrix: $A_1 = [I_d \hspace{0.2cm}  0_{d,d}]^\intercal$ and $A_2 = 0_{2d,d}$. $I_n$ represents a square identity matrix of size $n$ and $0_{l,m}$ is a $l \times m$ null matrix. $w_k$ represents the prediction noise which is here assumed to be zero-mean Gaussian for all variables in $X_k$ with a covariance matrix $Q$, such that $w_k \sim \mathcal{N}(0,Q)$. The filter is called unmotivated and assumes that the agent is under a simple dynamic equilibrium where there is no forces on the agent, so it remains fixed. In other words, it moves only due to random noisy fluctuations associated with $w_k$. Let $Z_K$ be the dynamic position data of an object in a given experiences in Fig. \ref{fig:learning_bd}-b.1. The linear relationship between measurements and the state of individuals is defined as:
\begin{equation}
\label{eq1.1}
Z_{k} = HX_{k} + v_k,
\end{equation}
where $H = [I_d \hspace{0.2cm} 0_{d,d}]$ is an observation matrix that maps states onto observations and $v_k$ represents the measurement noise produced by the sensor device which is here assumed to be zero-mean Gaussian with a covariance matrix $R$, such that, $v_k \sim \mathcal{N}(0,R)$. By considering the dynamical filter in equation \eqref{eq1.0}, it is possible to estimate the velocity  by using KFs' innovations, such that:
\begin{equation}
\label{eq1.2}
v_k= \frac{Z_k - HX_{k+1|k}}{\Delta k}
\end{equation}
where $X_{k+1|k}$ is the state space estimation of the time instant $k+1$ given observations until time $k$. $\Delta k$ is the sampling time.
By considering a 2-dimensional reference system, i.e., $d=2$, the state of an agent can be described as: $X_k = [x_k,y_k,\dot{x}_k,\dot{y}_k]^\intercal$. As explained before the fig. \ref{fig:main_bd}-c, it is possible to obtain Abnormal data  that used for learning a new model, where in this case it is a Motivated Kalman Filter (MKF) in Fig. \ref{fig:learning_bd}-d.1.

This work expresses the evolution in time of object's state based on a quasi-constant velocity models. Such type of motion can be defined as a MKF, such that:
\begin{equation}\label{eq1.3}
X_{k+1} = AX_{k} + BU_{S_{k}} + w_k,
\end{equation}   
where $B = [I_2\Delta k \hspace{0.2cm} I_2]^\intercal$ is a control input model.  $U_{S_{k}} = [\dot{x}_k$, $\dot{y}_k]^T$ is a control vector that encodes the expected entity's velocity when its state falls in a discrete region $S_k$, $k$ indexes the time, $\Delta k$ is the sampling time and $w_{k}$ is the process noise. 
 
In order to discretize the agents’ states $X_k$ produced by the unmotivated filter, we employed a Self Organizing Maps (SOM) \cite{Kohonen2001} that receives $X_k$ and produces a set of learned identified neurons $\boldsymbol{S}$ where similar information (quasi-constant velocities) are valid, such that:
\begin{equation}\label{eq1.4}
\boldsymbol{S} = \{S_1,S_2,\dots,S_{L}\},
\end{equation}
where  $S_k \in \boldsymbol{S}$	and $L$ is the total number of neurons.
The clustering process prioritizes similar velocities (actions) based on a weighted distance. Accordingly, the following distance function that uses the weights $\beta$ and $\alpha$ is used for training SOM, where $\beta + \alpha = 1$ and  $\alpha > \beta$ to favor clustering of patterns with smaller differences in velocity, such that:

\begin{equation}\label{eq1.5}
{d}(\mathcal{X},\mathcal{Y}) =\sqrt{(\mathcal{X}-\mathcal{Y})^\intercal D(\mathcal{X}-\mathcal{Y})},
\end{equation} 
where $D = [\mathcal{B} \hspace{0.2cm} \mathcal{A}]$. $\mathcal{B} = [\beta I_2 \hspace{0.2cm} 0_{2,2}]^\intercal$, $\mathcal{A}  = [0_{2,2} \hspace{0.2cm} \alpha I_2]^\intercal$. $\mathcal{X}$ and $\mathcal{Y}$ are both 4-dimensional vectors of the form $[x \hspace{0.2cm} y \hspace{0.2cm} \dot{x} \hspace{0.2cm} \dot{y}]^\intercal$. 

The main idea consists in learning a switching  DBN, see Fig.  \ref{fig:dbn_pl} (b) for tracking and predicting the dynamical system over time. Accordingly, arrows represent conditional probabilities between the involved variables. Vertical arrows facilitate to describe causalities between both, continuous and discrete levels of inference and observed measurements.
Trained neurons represent a set of zones that segment the continuous state space  $X_k$, into a set of regions here called Superstates $S_k \in \boldsymbol{S}$. Each superstate is employed to learn a MKF (Eq. \eqref{eq1.3}) for modeling motivators’ effects acting on agents in particular environment’s locations. Accordingly, it is possible to estimate the evolution of agents’ states $P(X_k|X_{k-1})$. Additionally, each $S_k$ is represented by the variables $\xi_{S_k}$, $Q_{S_k}$ and $\psi_{S_k}$, which encode respectively the mean value, the covariance matrix of clustered states and a range of validity of the proposed superstates, i.e., a threshold value where linear continuous models associated to obtained superstates are valid. Such threshold is defined as:
\begin{equation}\label{eq1.6}
\psi_{S_k} = E(d_{S_k}) + 3\sqrt{(V(d_{S_k}))},
\end{equation}  
where $d_{S_k}$ represents a vector that contains all distances between adjacent neurons, $E(\cdot)$ receives a vector of data and calculates its mean and $V(\cdot)$ its variance. By considering the threshold value $\psi{S_k}$ as a distance from the mean superstates values $\xi_{S_k}$, it is possible to define a certainty boundary where the proposed models are valid that means valid neurons. Accordingly, if they are valid, they use the dynamic model in Eq.~\eqref{eq1.3}, instead when they are not valid (empty neuron), they will use a random filter where  $U_{S_{k}} = 0_{2,1}$ i.e. a random walk filter with no control term keeping memory of previous experiences.     

By analyzing the activated superstates over time while executing a certain activity, it is possible to obtain a set of temporal transition matrices $C_{t}$ which encode the time-variant transition probabilities of passing or staying between superstates, given that the agent has spent a duration $t$ in the current superstate. Such set of matrices facilitates the inference of the next superstate, given the current one, $p(S_k|S_{k-1},t)$.

\colorlet{rectangle edge}{blue!50}
\colorlet{rectangle area}{red!20}

\tikzset{filled/.style={fill=rectangle area, draw=rectangle edge, thick},
    outline/.style={draw=rectangle edge, thick}}

\begin{figure}
\begin{minipage}{.45\linewidth}
\scalebox{0.45}{
\begin{tikzpicture}[->,
roundnode/.style={circle, draw=black!60, fill=green!5, very thick, minimum size=1.3cm},
imaginarynode/.style={circle, very thick, minimum size=1mm},
rectanglenode/.style={rectangle, draw=black!60, fill=blue!5, very thick, minimum size=1cm}
]
\filldraw[color=red!60, fill=red!5, very thick, rounded corners=15pt](-2,-0.8) rectangle (5.5,3.2) node[rotate=-90]at (6,1.8) {\textbf{HMM}};
\draw[color=blue!60, very thick, rounded corners=15pt](-2,0.8) rectangle (5.5,-5.8) node[rotate=-90]at (6,-3) {\textbf{GANs}};
\node[roundnode]        (xk_1)                                  {${\cal X}_{k-1}$};
\node[roundnode]        (sk_1)                [above=of xk_1]   {$C_{k-1}$};
\node[roundnode]        (xpk_1)               [below=of xk_1]   {${\cal X}^{\cal P}_{k-1}$};
\node[roundnode]        (zk_1)                [below=of xpk_1] {${\cal Z}_{k-1}$};

\node[imaginarynode]    (sk_2)       [left=of sk_1]     {};
\node[imaginarynode]    (xk_2)       [left=of xk_1]     {};
\node[imaginarynode]    (xpk_2)      [left=of xpk_1]    {};
\node[imaginarynode]    (zk_2)       [left=of zk_1]     {};

\node[roundnode]    (xk)                [right=0cm and 2cm of xk_1]     {${\cal X}_{k}$};
\node[roundnode]    (xpk)               [right=0cm and 2cm of xpk_1]    {${\cal X}^{\cal P}_{k}$};
\node[roundnode]    (sk)                [above=of xk]                   {$C_{k}$};
\node[roundnode]    (zk)                [below=of xpk]                   {${\cal Z}_{k}$};

\node[imaginarynode]    (sk_n)       [right=of sk]  {};
\node[imaginarynode]    (xk_n)       [right=of xk]  {};
\node[imaginarynode]    (xpk_n)      [right=of xpk] {};
\node[imaginarynode]    (zk_n)       [right=of zk]  {};

\draw[dashed] (sk_2.east) -- (sk_1.west);
\draw[dashed] (xk_2.east) -- (xk_1.west);
\draw[dashed] (xpk_2.east) -- (xpk_1.west);
\draw[dashed] (zk_2.east) -- (zk_1.west);

\draw[->] (sk_1.south) -- (xk_1.north);
\draw[->] (xk_1.south) -- (xpk_1.north);
\draw[->] (xpk_1.south) -- (zk_1.north);

\draw[->] (sk_1.east) -- (sk.west)  node[midway, above] {$p(C_k|C_{k-1})$};
\draw[->] (xk_1.east) -- (xk.west) node[midway, above] (of_k) {\scriptsize{$p({\cal X}_k={\cal X}_{k-1})$}};

\draw[->] (sk.south) -- (xk.north) node[midway, right] {$p({\cal X}_k|C_k)$};
\draw[->] (xk.south) -- (xpk.north) node[midway, right] {$p({\cal X}_k|{\cal X}^{\cal P}_{k})$};
\draw[->] (xpk.south) -- (zk.north) node[midway, right] {$p({\cal X}^{\cal P}_{k}|{\cal Z}_{k})$};
\draw [->] (xk.south west) to [out=250,in=110] (zk.west);
\node[imaginarynode]    (1)       [below=of of_k]     {$p({\cal X}_{k}|{\cal Z}_{k})$};

\draw[dashed] (sk.east) -- (sk_n.west);
\draw[dashed] (xk.east) -- (xk_n.west);
\draw[dashed] (xpk.east) -- (xpk_n.west);
\draw[dashed] (zk.east) -- (zk_n.west);


\end{tikzpicture}}
\centering
(a)
\end{minipage}
\begin{minipage}{.45\linewidth}
	\scalebox{0.5}{
\begin{tikzpicture}[
roundnode/.style={circle, draw=black!60, fill=green!5, very thick, minimum size=1.3cm},
imaginarynode/.style={circle, very thick, minimum size=1mm},
]
\filldraw[color=red!60, fill=red!5, very thick, rounded corners=15pt](-2,-0.8) rectangle (5.5,3.2) node[rotate=-90]at (6,1.8) {\textbf{Particle Filter}};
\draw[color=blue!60, very thick, rounded corners=15pt](-2,0.8) rectangle (5.5,-3.2) node[rotate=-90]at (6,-1) {\textbf{Kalman Filter}};
\node[roundnode]        (xk_1)                              {$X_{k-1}$};
\node[roundnode]        (sk_1)                [above=of xk_1] {$S_{k-1}$};
\node[roundnode]        (zk_1)                [below=of xk_1] {$Z_{k-1}$};

\node[imaginarynode]    (sk_2)       [left=of sk_1] {};
\node[imaginarynode]    (xk_2)       [left=of xk_1] {};
\node[imaginarynode]    (zk_2)       [left=of zk_1] {};

\node[roundnode]    (xk)                [right=0cm and 2cm of xk_1] {$X_{k}$};
\node[roundnode]    (sk)                [above=of xk] {$S_{k}$};
\node[roundnode]    (zk)                [below=of xk] {$Z_{k}$};

\node[imaginarynode]    (sk_n)       [right=of sk] {};
\node[imaginarynode]    (xk_n)       [right=of xk] {};
\node[imaginarynode]    (zk_n)       [right=of zk] {};

\draw[dashed] (sk_2.east) -- (sk_1.west);
\draw[dashed] (xk_2.east) -- (xk_1.west);
\draw[dashed] (zk_2.east) -- (zk_1.west);

\draw[->] (sk_1.south) -- (xk_1.north);
\draw[->] (xk_1.south) -- (zk_1.north);

\draw[->] (xk_1.east) -- (xk.west) node[midway, above] {$p(X_k|X_{k-1})$};
\draw[->] (sk_1.east) -- (sk.west)  node[midway, above] {$p(S_k|S_{k-1})$};

\draw[->] (sk.south) -- (xk.north) node[midway, left] {$p(X_k|S_k)$};
\draw[->] (xk.south) -- (zk.north) node[midway, left] {$p(Z_k|X_k)$};

\draw[dashed] (sk.east) -- (sk_n.west);
\draw[dashed] (xk.east) -- (xk_n.west);
\draw[dashed] (zk.east) -- (zk_n.west);


\end{tikzpicture}}
\centering
(b)
\end{minipage}
\caption{Proposed DBN switching models for (a) private layer, (b) shared layer}
\label{fig:dbn_pl}
\end{figure}
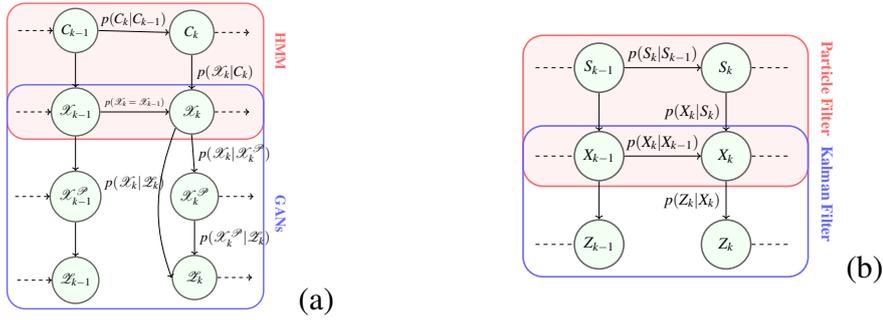
\subsection{Private-level self-awareness modeling}
\label{subsec:learn_PL}
\noindent{\textbf{State estimation in private layer.}}
The private layer deals with the high-dimensional visual information observed by the agent. Namely, a sequence of images (frames) ${\cal I}$, and their corresponding optical-flow maps (motion) ${\cal O}$. To model the PL of self-awareness, a set of cross-modal GANs \cite{NIPS2014_5423} is trained to learn the normality from this set of visual data. In general, generative models try to maximize likelihood by minimizing the Kullback-Leibler (KL) distance between a given data distribution and the generator's distribution \cite{arjovsky2017towards}
, where GANs learn it during the training process by an adversarial game between two networks: a generator ($G$) and a discriminator ($D$). 

Generator networks learn to preform the prediction task. In our case, the prediction task includes generating both the next image (frame) and the next optical-flow (motion map), this could be seen as hidden state ${\cal X}^{\cal P}_{k}$ (see Fig. \ref{fig:dbn_pl}). The update task includes having the likelihood and the prediction. In the case of GAN the likelihood is learned and approximated by the Discriminator networks. The outputs of the Discriminators are the encoded version of optical-flow ${\cal D}^{\cal O}$ and encoded version of image ${\cal D}^{\cal I}$. In our approach, the Discriminator networks are used to approximate the distance of the probabilities between the likelihood and the prediction, so if they are not completely overlapped then it leads to a higher error in the output encoded version.

Intuitively, the encoded version of image can be seen as the state in the SL, while the encoded version of optical-flow represents the state derivative (motion). Accordingly, the error ${\cal E}$ can be seen as the distances between the two encoded versions from prediction and observation. Following the same intuition applied in SL, we cluster the encoded version of images, motion and the error into a set of superstates. In light of the above, the generalized states here can be defined as a function of image, motion, and the error $f([{\cal D}^{\cal I},{\cal D}^{\cal O},{\cal E}])$ for any model (superstate).

The rest of this section is dedicated to describe our PL representation for self-awareness in details. We propose a hierarchical structure of cross-modal GANs \cite{NIPS2014_5423}, where they are trained to learn the normality using a sequence of observed images (${\cal I}$) synchronously collected from a first person viewpoint paired with their corresponding optical-flow maps (${\cal O}$) as the representation of the agent's motion. 
In order to understand the relation between these two modalities, the hierarchy of cross-modal GANs is adopted and trained in a weakly-supervised manner. The only supervision here is provided by a subset of normal data to train the first level of the hierarchy that we called $Base GANs$. The $Base GANs$ provides a reference for the next levels of the hierarchy, in which all the further levels are trained in a self-supervised manner. The source of such self-supervision is the criteria provided by the $Base GANs$. The rest of this section is dedicated to explain the procedure of learning a single cross-modal GAN, constructing the hierarchy of GANs, and finally the online application of the learned model for prediction and anomaly detection.

\noindent{\textbf{Learning the cross-modal representation.}}
GANs are deep networks commonly used to generate data (e.g., images) and are trained using only unsupervised data. The supervisory information in a GAN is indirectly provided by an adversarial game between two independent networks: a generator ($G$) and a discriminator ($D$). During training, $G$ generates new data and $D$ tries to distinguish whether its input is real (i.e., it is a training image) or it was generated by $G$. This competition between $G$ and $D$ is helpful in boosting the ability of both $G$ and $D$. To learn the normal pattern two channels are used as the observation representation: appearance (i.e., raw-pixels) and motion (optical-flow images) for two cross-channel tasks. In the first task, optical-flow images are generated from the original frames, while in the second task appearance information is estimated from an optical flow image.
Specifically, let ${\cal I}_k$ be the $k$-th frame of a training video and ${\cal O}_k$ the optical-flow obtained using ${\cal I}_k$ and ${\cal I}_{k+1}$. ${\cal O}_k$ is computed using \cite{brox2004high}.
Two networks are trained: ${\cal N}^{{\cal I} \rightarrow {\cal O}}$, which is trained to generate optical-flow from frames (task 1) and ${\cal N}^{{\cal O} \rightarrow {\cal I}}$, which generates frames from optical-flow (task 2).
In both cases, inspired by \cite{Isola_2017_CVPR,ravanbakhsh2017abnormal}, our architecture is composed by two fully-convolutional networks: the conditional generator $G$ and the conditional discriminator $D$. 
The $G$ network is the U-Net architecture \cite{Isola_2017_CVPR}, which is an encoder-decoder following with {\em skip connections} helping to preserve important local information. 
For $D$ the {\em PatchGAN} discriminator \cite{Isola_2017_CVPR} is proposed, which is based on a ``small'' fully-convolutional discriminator.

\noindent{\textbf{Hierarchy of cross-modal GANs.}}
As described in Sec.\ref{sec:intro}, the assumption is that the distribution of the normality patterns is under a high degree of diversity. In order to learn such distribution we suggest a hierarchical strategy for high-diversity areas by encoding the different distributions into the different hierarchical levels, in which, each subset of train data is used to train a different GAN. To construct the proposed hierarchy of GANs, a recursive procedure is adopted. As shown in Alg. \ref{alg:hgan} the inputs of the procedure are represented by two sets: ${\cal Z}$ could be seen as the set of observation vectors which includes all the observations from the normal sequence of training data. Specifically, ${\cal Z}$ includes a set of coupled Frame-Motion maps, where ${\cal Z} = \{ [{\cal I}_k, {\cal O}_k] \}_{k=1,...,N}$, and $N$ is the number of total train samples. Besides, the input ${\cal V}_l$ is a subset of ${\cal Z}$, provided to train GANs for each individual level of the hierarchy. For instance, in case of the first level GANs, the initial set ${\cal V}_0$ is used to train two cross-modal networks ${\cal N}^{{\cal I} \rightarrow {\cal O}}_0$, and ${\cal N}^{{\cal O} \rightarrow {\cal I}}_0$. Note that, the only supervision here is the initial ${\cal V}_0$ to train the first level of the hierarchy, and the next levels are built accordingly using the supervision provided by the first level of GANs.
After training ${\cal N}^{{\cal I} \rightarrow {\cal O}}_0$, and ${\cal N}^{{\cal O} \rightarrow {\cal I}}_0$, we input $G^{{\cal I} \rightarrow {\cal O}}_0$ and $G^{{\cal O} \rightarrow {\cal I}}_0$ using each frame ${\cal I}$ of the entire set ${\cal Z}$ and its corresponding optical-flow image ${\cal O}$, respectively. The generators predict Frame-Motion couples as:
\begin{equation}\label{eq:pred_gan}
\begin{split}
\quad {\cal X}^{\cal P} = \{ [{\cal P}^{\cal I}_k , {\cal P}^{\cal O}_k]\}_{k=1,...,N}\\ 
{\cal P}^{\cal I}_k = G^{{\cal O} \rightarrow {\cal I}}_0 ( {\cal O}_k),\quad
{\cal P}^{\cal O}_k = G^{{\cal I} \rightarrow {\cal O}}_0 ( {\cal I}_k)
\end{split}
\end{equation}
where ${\cal P}^{\cal I}_k$ and ${\cal P}^{\cal O}_k$ are $k$-th predicted image and predicted optical-flow, respectively.
Then the encoded versions of observations ${\cal Z}$ and is computed by the discriminator networks $D$:
\begin{equation}\label{eq:dist_gan_z}
\begin{split}
\quad{\cal D}^{\cal I} = \{D^{{\cal O} \rightarrow {\cal I}}_0 ( {\cal I}_k,  {\cal O}_k)\}_{k=1,...,N} ,\\ 
{\cal D}^{\cal O} = \{D^{{\cal I} \rightarrow {\cal O}}_0 ( {\cal O}_k,  {\cal I}_k)\}_{k=1,...,N} \quad 
\end{split}
\end{equation}
where ${\cal D}^{\cal I}$ and ${\cal D}^{\cal O}$ are the encoded version of observed image and observed optical-flow, respectively.
Similarly, the encoded distance maps ${\cal E}$ between the observations $\cal Z$ and the predictions $\cal P$ for both channel are computed as:
\begin{equation}\label{eq:dist_gan}
\begin{split}
\quad \quad  {\cal E} = \{ [{\cal E}^{\cal I}_k , {\cal E}^{\cal O}_k]\}_{k=1,...,N}\\ 
\quad{\cal E}^{\cal I}_k = D^{{\cal O} \rightarrow {\cal I}}_0 ( {\cal I}_k,  {\cal O}_k) -  
D^{{\cal O} \rightarrow {\cal I}}_0 ( {\cal P}^{\cal I},  {\cal O}_k) ,\\\quad 
{\cal E}^{\cal O}_k = D^{{\cal I} \rightarrow {\cal O}}_0 ( {\cal O}_k,  {\cal I}_k) -  
D^{{\cal I} \rightarrow {\cal O}}_0 ( {\cal P}^{\cal O}_k,  {\cal I}_k)\quad 
\end{split}
\end{equation}
 The distance maps ${\cal E}$ can be seen as the error in the coupled image-motion states representation, The joint states $\{ [{\cal D}^{\cal I}_k , {\cal D}^{\cal O}_k, {\cal E}_k]\}_{k=1,...,N}$ input to a self-organizing map (SOM) \cite{Kohonen2001} in order to cluster similar appearance-motion information. Similar to clustering position-velocity information in the shared layer, here the clustering is done to discretize the appearance-motion representations into a set of super-states. Specifically, the SOM's output is a set of neurons encoding the state information into a set of prototypes. Detected prototypes (clusters) provide the means of discretization for representing a set of super-states $\{{C_k}\}_{k=1,...,M_0}$, where $M_0$ is the number of detected clusters in the first level of the hierarchy.

It is expected that the clusters which are containing the training data should obtain lower distance score, since the innovation between the prediction and observation is lower. This is the criteria to detect the new distributions for learning new GANs, in which the clusters with high average scores are considered as a new distributions. The new detected distributions build the new subsets ${\cal V}_{l}$ to train new networks ${\cal N}^{{\cal I} \rightarrow {\cal O}}_l$, and ${\cal N}^{{\cal O} \rightarrow {\cal I}}_l$ for the $l$-th level of the hierarchy, where $l=0,1,...,L$ is the level in the hierarchy. This procedure continues until no new distribution detects. Then GANs and detected super-states in each level are stacked incrementally for constructing the entire hierarchical structure of GANs ${\cal H}_{l}$. Such incremental nature of the proposed method makes it a powerful model to learn a very complex distribution of data in a self-supervised manner.

\begin{algorithm}
\caption{Constructing the hierarchy of GANs}\label{euclid}
\label{alg:hgan}
\begin{algorithmic}[1]
\Require
\State $\theta:  \text{  Threshold parameter for train a new GAN}$
\State ${\cal Z}:$ \text{ Entire training sequences} ${{\cal Z} = \{ ({\cal I}_k, {\cal O}_k) \}_{k=1,...,N}}$ 
\State ${\cal V}_0: \text{  Subset of } {\cal Z}$
\State $l=0: \text{  Counter of hierarchy level }$
\Ensure
\State $[{\cal H}_l] \text{ Hierarchy of GANs}$
\Procedure{TRAINING HIERARCHY OF GANS}{}\label{marker}
\Label $\texttt{train:}$
\State $\text{Train networks } {\cal N}^{{\cal I}\rightarrow {\cal O}}, {\cal N}^{{\cal O} \rightarrow {\cal I}}\text{, with } {\cal V}_l$
\State $[{\cal H}_l] \gets \text{Trained networks }{\cal N}^{{\cal I}\rightarrow {\cal O}}, {\cal N}^{{\cal O} \rightarrow {\cal I}} $
\State ${\cal X}^{\cal P} \gets {G} ({\cal Z}) \text{: predictions}$
\State ${\cal D} \gets {D} ({\cal Z}) \text{: encoded observation}$
\State ${\cal E} \gets ||{D} ({\cal Z}) - {D} ({\cal X}^{\cal P})||_1 \text{: error}$
\State ${\cal X} \gets [{\cal D}, {\cal E}] \text{: states}$
\State $\text{Clustering states: }SOM({\cal X}) \text{: super-states} \{C_k\}$

\For{$\text{each identified cluster}$}
\State $\mu \gets \text{Average score maps in this cluster}$ 

\If{$\mu \ge \theta $}
\State $ l= l +1 $
\State ${\cal V}_l \gets \text{Samples from this cluster in }{\cal Z}$
\State \Goto \texttt{train}
\EndIf
\EndFor
\State \textbf{return} $[{\cal H}_l]$
\EndProcedure
\end{algorithmic}
\end{algorithm}

 In our experiments, we use a \emph{Hierarchy of GANs}, which consists of two levels: $Level0$ or \emph{base GAN}, which is trained with a provided initialization subset ${\cal V}_0$ for representing the empty straight path, and $Level1$ which is trained over the high-scored cluster (in our case this cluster semantically representing the curves). Note that, selecting the straight empty path (in normal situation) as initialization subset is considered for the sake of simplicity, since it is the most efficient way to reach a target and so the variation with respect to this behavior could be described as differential actions with respect to going straight. Similarly, the discriminator's learned decision boundaries are also used to detect the abnormal events at testing time that is explained in the next section.

\graphicspath{{Chapter3/Figs/a/}}
\section{On-line Situational Change Detection}
\label{sec:online_testing_sa}
\subsection{Shared-level change detection: a DBN framework}
\label{subsubsec:MJM}
\begin{figure}[t]
	\centering
	\includegraphics[width=1\linewidth]{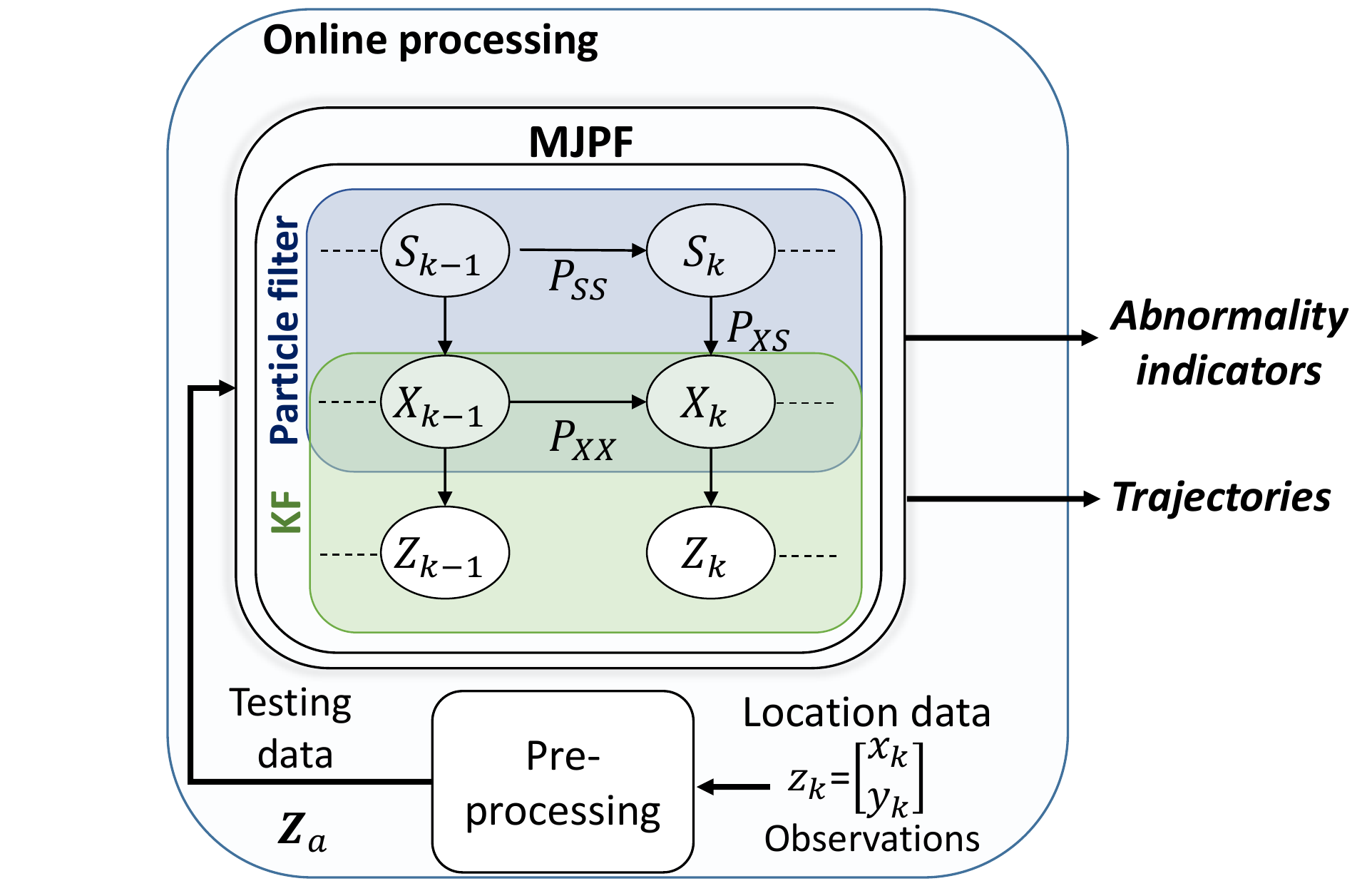}
	\caption{Online phase: A Markov Jump
Particle Filter (MJPF) is employed to make inference on the DBN}
	\label{fig:10}
\end{figure}

The SL model can be learned by means of a DBN switching model see Fig.  \ref{fig:dbn_pl} (b). Such model includes a discrete set of state regions subspaces corresponding to the switching variables. A set quasi-constant velocity model described before (Eq. \eqref{eq1.3}) is associated to one of such regions that describe a possible alternative relation between consecutive temporal states. A further learning step facilitates to obtain temporal transition matrices between superstates. A Markov Jump Particle Filter (MJPF) is used to infer posterior probabilities on discrete and continuous states iteratively see Fig.  \ref{fig:10}. MJPF essentially consists in a particle filter (PF) working at discrete level, embedding in each particle a Kalman filter (KF). Consequently, for each particle has attached a KF which depends on the superstate $S_K$ (see Eq. \eqref{eq1.3}). Such filter is used to obtain the prediction for the continuous state associated with a particle's superstate $S_k^*$, that is $p(X_k|X_{k-1},S_{k-1}^*)$; and the posterior probability $p(X_k,S_k^*|Z_{k})$ is estimated according to current observation $Z_k$.

\noindent{\textbf{Anomaly detection.} }Abnormalities can be seen as deviations from predictions that the MJPF can do using the learned models embedded in its switching model  and new observed trajectories where interaction with the environment differ from training data.
Since a probabilistic filtering approach is considered, two main moments can be distinguished: $i)$ Prediction: which corresponds to an estimation of future states at a give time $k$. $ii)$ Update: computation of the generalized state posterior probability based on the comparison between predicted states and new measurements. Accordingly, abnormality behaviors can be measured in the update phase, i.e., when predicted probabilities are far from to observations. As it is well known, innovations in KFs are defined as:
\begin{equation}\label{eq4}
\epsilon_{k,n} = Z_k - H\hat{X}^n_{k|k-1},
\end{equation}
where $\epsilon_{k,n}$ is the innovation generated in the zone $n$ where the agent is located at a time $k$. $Z_k$ represents observed spatial data and $\hat{X}^n_{k|k-1}$ is the KF estimation of the agent's location at the future time $k$ calculated in the time instant $k-1$  Eq. \eqref{eq1.3}. Additionally, $H$ is the observation model that maps measurements into states, such that $Z_k = H X_k + v$ where $v \sim \mathcal{N}(0,R)$ and $R$ is the covariance observation noise.

Abnormalities can be seen as moments when a tracking system fails to predict subsequent observations, so that new models are necessary to explain new observed situations. A weighted norm of innovations is employed for detecting abnormalities, such that:
\begin{equation}\label{eq5}
\mathcal{Y}_k = \mathbf{d}(Z_k,H\hat{X}_{k|k-1}).
\end{equation}
In the MJPF, the expression shown \eqref{eq5} is computed for each particle and the median of such values is used as a global anomaly measurement of the filter. Further details about the implemented method can be found in \cite{fusion18_damian}.

\graphicspath{{Chapter3/Figs/a/}}
\subsection{Private-level change detection: an adversarial framework}
\label{subsubsec:Online}

Once the GANs hierarchy $\{{\cal H}_l\}$ is trained, it can be use for online prediction and anomaly detection task. Here we describe the testing phase for state/label estimation and detecting the possible abnormalities.

\noindent{\textbf{Label estimation.} }At testing time we aim to estimate the state and detect the possible abnormality with respect to the training set. More specifically, input the test sample into the first level in the hierarchy of GANs, let ${D}^{{\cal I} \rightarrow {\cal O}}_0$ and ${D}^{{\cal O} \rightarrow {\cal I}}_0$ be the patch-based discriminators trained using the two channel-transformation tasks. Given a test frame ${\cal I}_k$ and its corresponding optical-flow image ${\cal O}_k$, we first produce the reconstructed ${\cal P}^{\cal O}_k$ and  ${\cal P}^{\cal I}_k$ using the first level generators ${G}^{{\cal I} \rightarrow {\cal O}}_0$ and ${G}^{{\cal O} \rightarrow {\cal I}}_0$, respectively. Then, the pairs of patch-based discriminators ${D}^{{\cal I} \rightarrow {\cal O}}$ and ${D}^{{\cal O} \rightarrow {\cal I}}$, are applied for the first and the second task, respectively. 
This operation results in two scores maps for the observation: $D^{{\cal I} \rightarrow {\cal O}}_0 ( {\cal O}_k,  {\cal I}_k)$ and $D^{{\cal O} \rightarrow {\cal I}}_0 ( {\cal I}_k,  {\cal O}_k)$, and two score maps for the prediction (the reconstructed data): $D^{{\cal I} \rightarrow {\cal O}}_0 ( {\cal P}^{\cal O}_k,  {\cal I}_k)$ and $D^{{\cal O} \rightarrow {\cal I}}_0 ( {\cal P}^{\cal I}_k,  {\cal O}_k)$. In order to estimate the state, we used Eq. \ref{eq:dist_gan} to generate the joint representation ${\cal X}_k = [{\cal X}^{\cal I}_k , {\cal X}^{\cal O}_k]$, where:
\begin{equation}\label{eq:state_gan}
\begin{split}
\quad{\cal X}^{\cal I}_k = D^{{\cal O} \rightarrow {\cal I}}_0 ( {\cal I}_k,  {\cal O}_k) -  
D^{{\cal O} \rightarrow {\cal I}}_0 ( {\cal P}^{\cal I},  {\cal O}_k) ,\\
{\cal X}^{\cal O}_k = D^{{\cal I} \rightarrow {\cal O}}_0 ( {\cal O}_k,  {\cal I}_k) -  
D^{{\cal I} \rightarrow {\cal O}}_0 ( {\cal P}^{\cal O}_k,  {\cal I}_k)\quad 
\end{split}
\end{equation}
Accordingly, to estimate the current super-state we use ${\cal X}_k$ to find closest SOM's detected prototypes. This procedure repeat for all the $l=0,1,..,L$ levels in the hierarchy $[{\cal H}_l]$.
This model can be seen as a switching model (see Fig. \ref{fig:dbn_pl}), where in continues level a hierarchy of GANs estimating the states and the discrete space modeled by HMM.

\noindent{\textbf{Anomaly detection.} }Note that, a possible abnormality in the observation (e.g., an unusual object or an unusual movement) corresponds to an outlier with respect to the data distribution learned by ${\cal N}^{{\cal I} \rightarrow {\cal O}}_l$ and ${\cal N}^{{\cal O} \rightarrow {\cal I}}_l$ during training. The presence of the anomaly, results in a low value of $D^{{\cal O} \rightarrow {\cal I}}_l ( {\cal P}^{\cal I},  {\cal O}_k)$ and $D^{{\cal I} \rightarrow {\cal O}}_l ( {\cal P}^{\cal O}_k,  {\cal I}_k)$ (predictions), but a high value of $D^{{\cal O} \rightarrow {\cal I}}_l ( {\cal I}_k,  {\cal O}_k)$ and $D^{{\cal I} \rightarrow {\cal O}}_l ( {\cal O}_k,  {\cal I}_k)$ (observation). 
Hence, in order to decide whether an observation is normal or abnormal with respect to the scores from the current hierarchy level of GANs, we simply calculate the average value of the innovations between prediction and observation maps for both modalities, which is obtained from:
\begin{equation}\label{eq:dist}
\tilde{Y}_k = \overline{{\cal X}^{\cal I}_k} + \overline{{\cal X}^{\cal O}_k}
\end{equation}

The final representation of private layer for an observation ${\cal Z}_k = ({\cal I}_k , {\cal O}_k )$ consists of the computed $\tilde{Y}_k$ and estimated super-state $C_k$. We defined an error threshold $\tilde{Y}_{th}$ to detect the abnormal events: when all the levels in the hierarchy of GANs tag the sample as abnormal (e.g., dummy super-state) and the measurement $\tilde{Y}$ is higher than this threshold, current measurement is considered as an abnormality. Note that, the process is aligned closely to the one followed with SL layer, despite GAN are more powerful as they allow to deal with strong multidimensional inputs as well as with not linear dynamic models at the continuous level. This complexity is required by video variables involved in PL as different from low dimensional positional variable involved in SL.

\graphicspath{{Chapter3/Figs/a/}}
\section{Experiments}
\label{sec:exp_sa}
\begin{figure}[t]
	\vspace{-0.25cm}
	\centering
	\begin{minipage}[t]{0.45\textwidth}
		\centering
		\includegraphics[width=6cm, height=4cm]{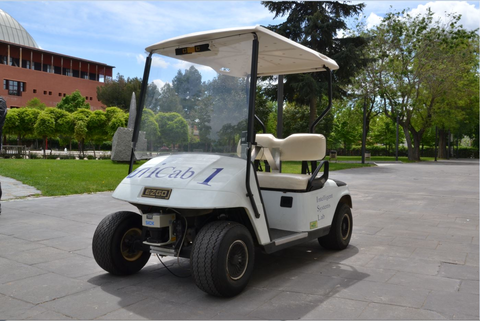}
		{Autonomous vehicle ``iCab''}
		
	\end{minipage}%
	~ 
	\begin{minipage}[t]{0.45\textwidth}
		\centering
		\includegraphics[width=5.2cm, height=4cm]{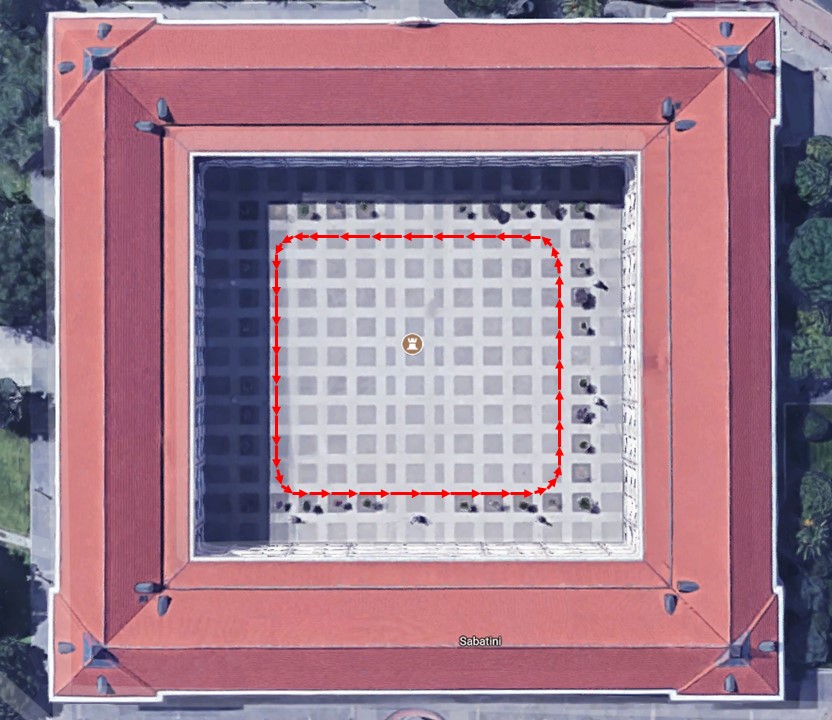}
		Infrastructure
		\label{fig:Environment}
	\end{minipage}
	\caption{Proposed moving entity and closed scene}
	\label{fig:iCab}
\end{figure}
\subsection{Proposed dataset}
\label{sec:datasets}
In our experiments an iCab vehicle \cite{Marin2016} drove by a human operator is used to collect the dataset (see Fig. \ref{fig:iCab});  we obtained the vehicle's position mapped into Cartesian coordinates from the odometry manager~\cite{Marin2016}, as well as captured video footage from a first person vision acquired with a built-in camera of the vehicle. 
The observations are generated by taking state space position of iCab and estimating its flow components over the time. Furthermore, for the cross-modal GANs in the PL self-awareness model, we input the captured video frames and their corresponding optical-flow maps.

\begin{figure*}[t]
    \begin{center}
    \begin{minipage}[t]{0.32\textwidth}
		\centering
		\includegraphics[width=\textwidth]{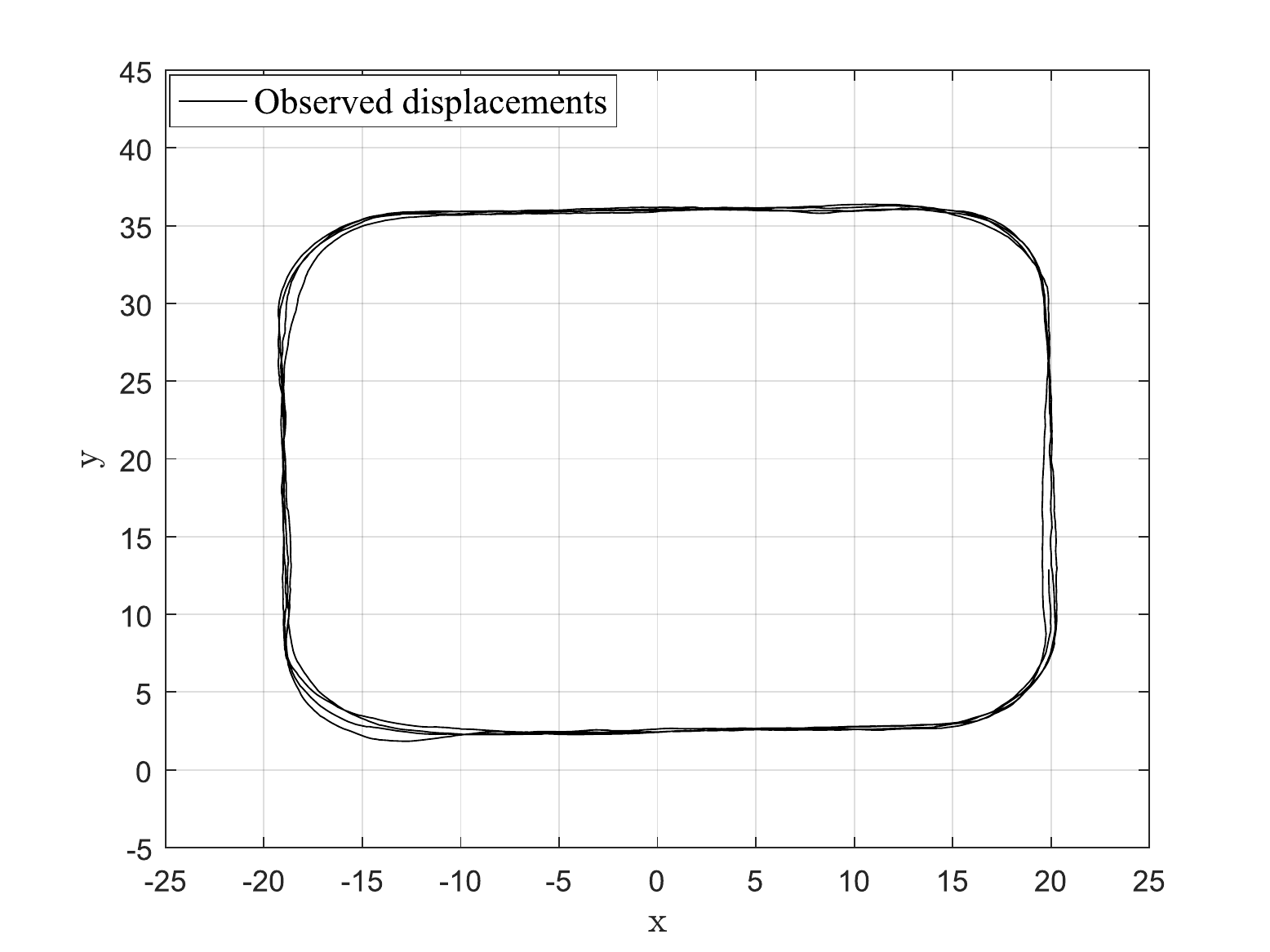}
		(a)
	\end{minipage}
	\begin{minipage}[t]{0.32\textwidth}
		\centering
		\includegraphics[width=\textwidth]{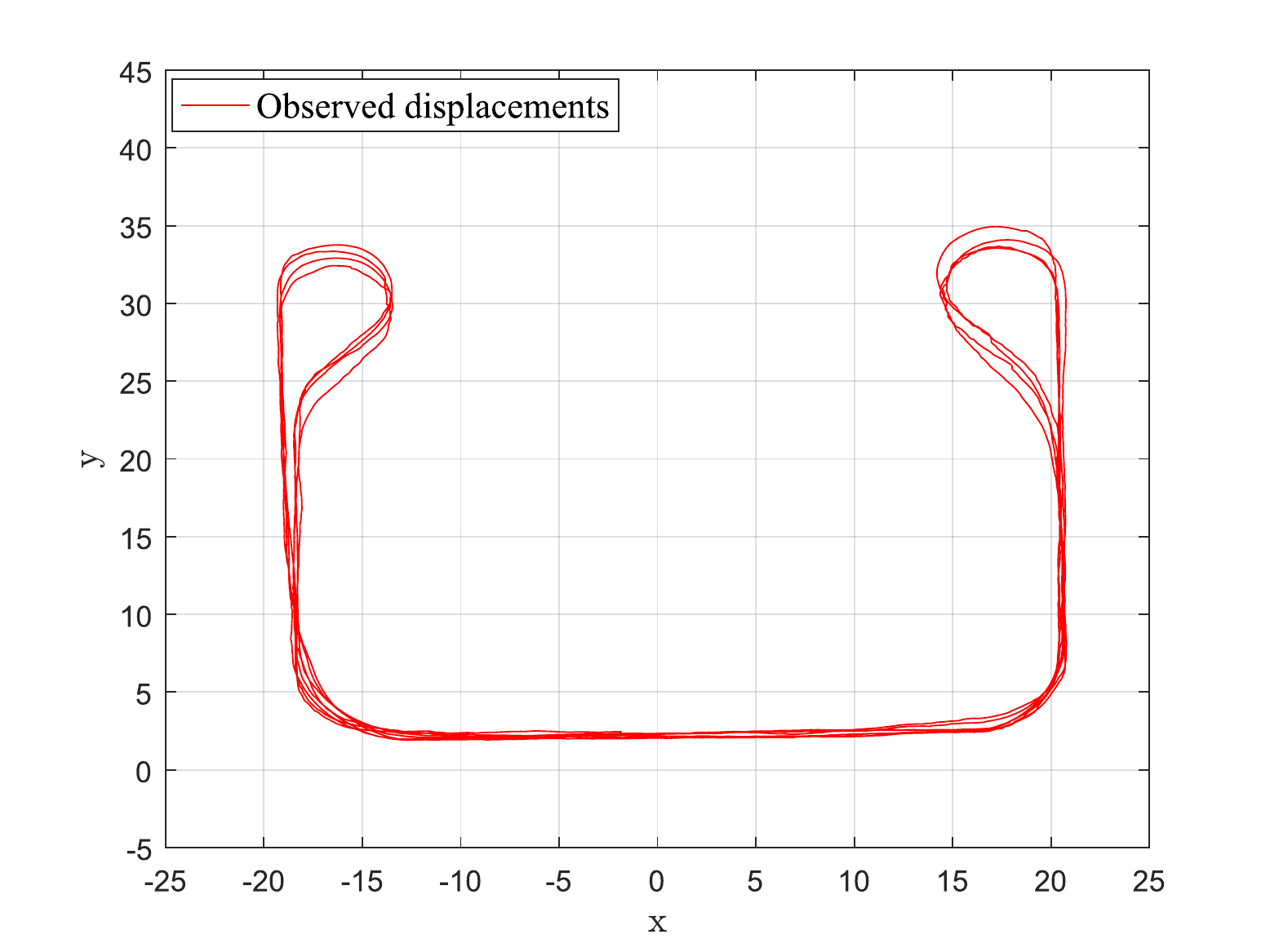}
		(b)
	\end{minipage}
	\begin{minipage}[t]{0.32\textwidth}
		\centering
		\includegraphics[width=\textwidth]{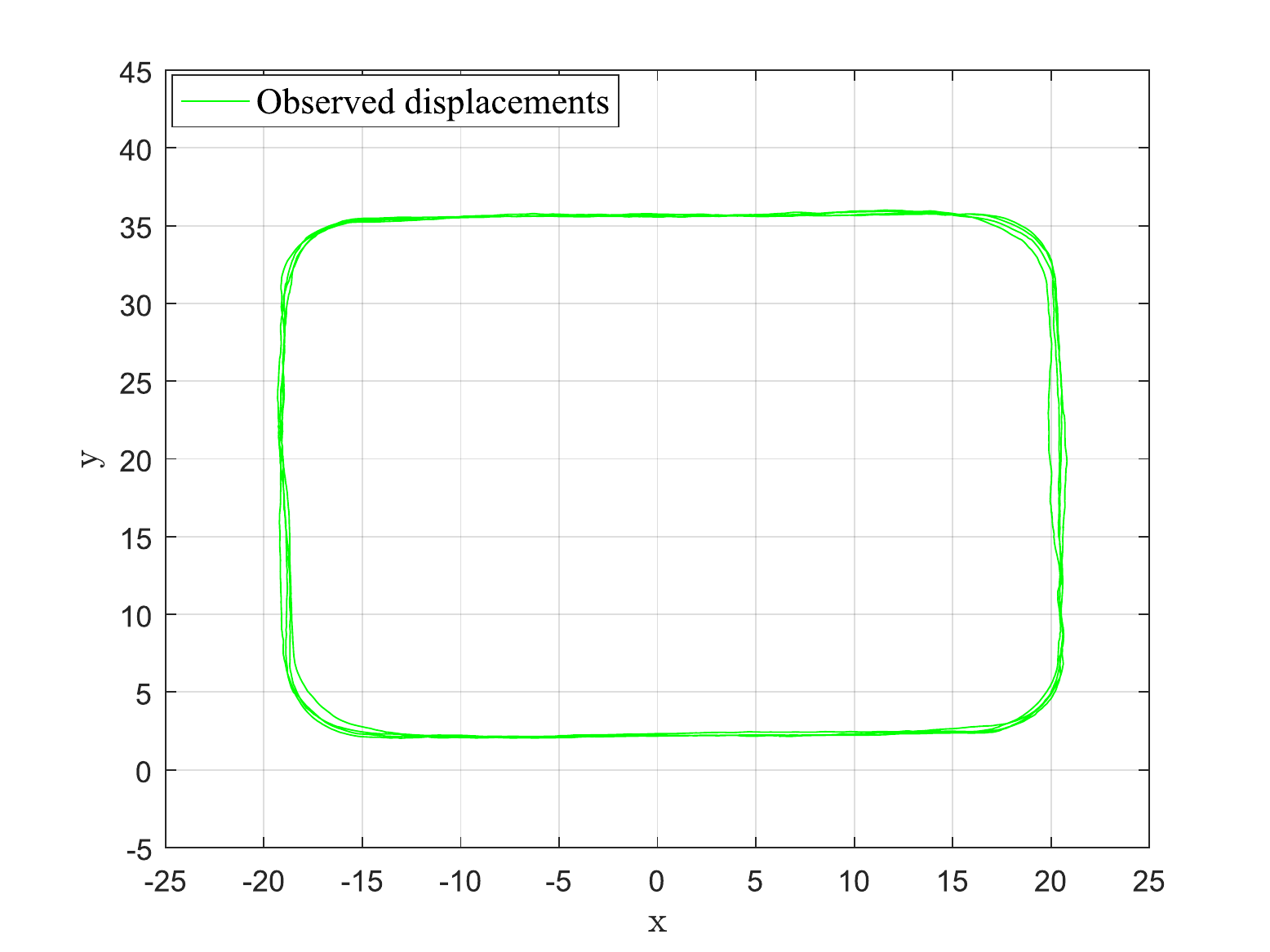}
		(c)
		\label{fig:Stop}
	\end{minipage}
	\end{center}
	\caption[SA: Training and testing scenarios]{Three different action scenarios: (a) perimeter monitoring under the normal situation (training set), and performing perimeter monitoring task in presence of abnormality (test sets): (b) U-turn, and (c) emergency stop.}
	\label{fig:plans}
\end{figure*}

We aim to detect dynamics that have not been seen previously based on the normal situation (Scenario I) learned with the proposed method. Scenarios II and III includes unseen manoeuvres caused by the presence of pedestrians while the vehicle performs a perimeter control task. 
Accordingly, 3 situations (experiments) are considered in this work: \emph{Scenario I)} or normal perimeter monitoring, where the vehicle follows a rectangular trajectory around a building (see Fig.~\ref{fig:plans}-a). \emph{Scenario II)} or U-turn, where the vehicle performs a perimeter monitoring and is faced with a pedestrian, so it makes a U-turn to continue the task in the opposite direction (see Fig.~\ref{fig:plans}-b). \emph{Scenario III)} or emergency stop, where the vehicle encounters with pedestrians crossing its path and needs to stop until the pedestrian leaves its field of view (see Fig.~\ref{fig:plans}-c).

\begin{figure}[t]
	\vspace{-0.25cm}
	\centering
	\begin{minipage}[t]{\textwidth}
		\centering
		\includegraphics[height=3cm]{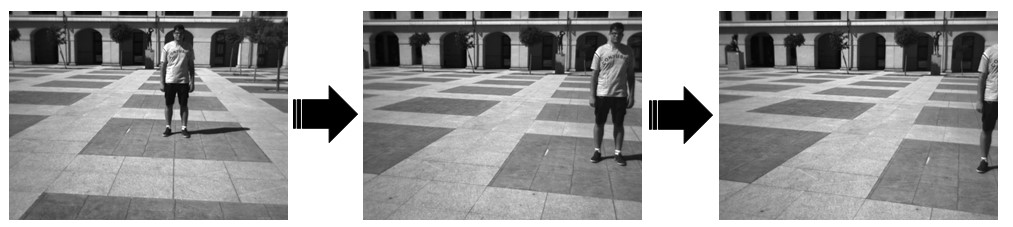}
		
		{scenario II ``U-turn manuver''}
		
	\end{minipage}%
	\\
	\begin{minipage}[t]{\textwidth}
		\centering
		\includegraphics[height=3cm]{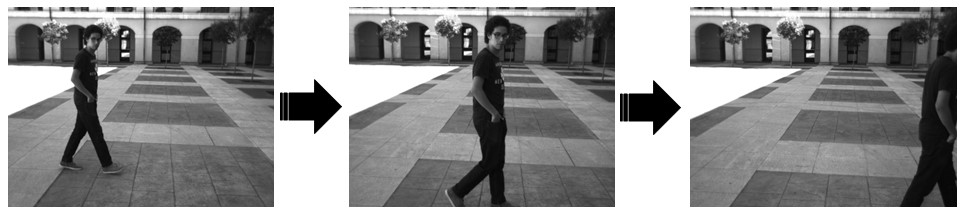}
		
		scenario III ``emergency stop''
		\label{fig:Environment}
	\end{minipage}
	\caption{Final testing scenarios: unseen ``U-turn manuver'' and ``emergency stop'' situations}
	\label{fig:pablo_snap}
\end{figure}

Situations II and III can be seen as deviations of the perimeter monitoring dynamics (see Fig. \ref{fig:pablo_snap}). 
 When an observation falls outside the superstate, as the learned model are not applicable, a {\em  dummy neuron} is used to represent the unavailability of an action from the learned experience and random filter where $U = 0_{2,1}$ in Eq.~\eqref{eq1.3} is considered for prediction to represent the uncertainty over state derivatives.

\graphicspath{{Chapter3/Figs/a/}}
\subsection{Experimental setup for learning normality}
 The training phase of our models has been performed over a normal scenario dataset. For each level we used different modalities to train the models. For SL we used spacial positional information, while for the PL the first person visual data is used. In the rest of this part we show the training process and output of each layer.\\
 

\begin{figure}[H]
	\flushright(a)\includegraphics[width=0.94\linewidth]{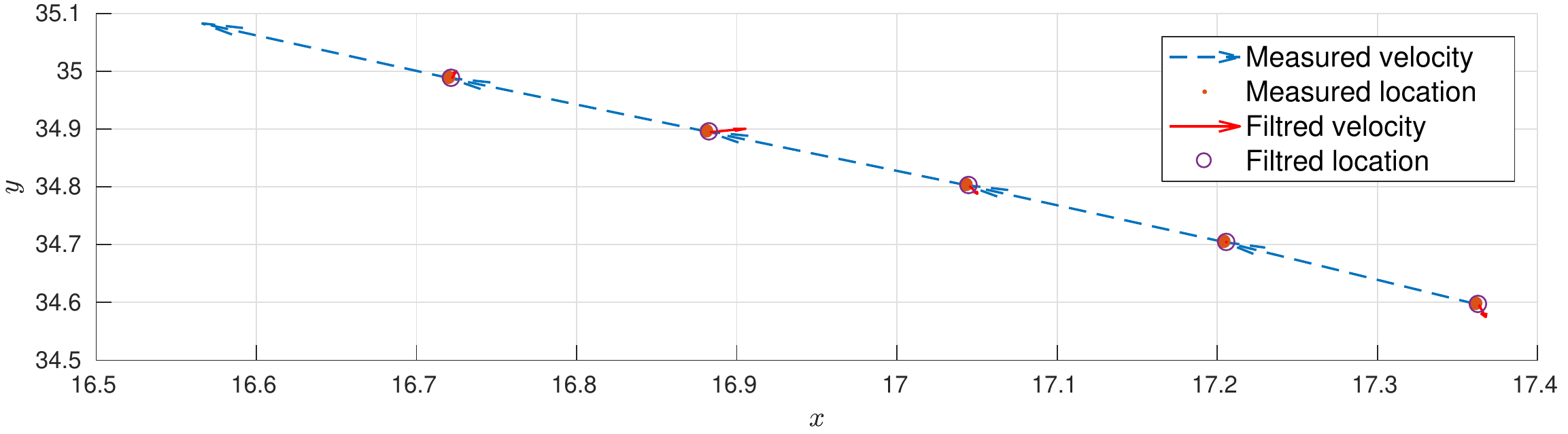}
	\flushright\includegraphics[width=0.92\linewidth]{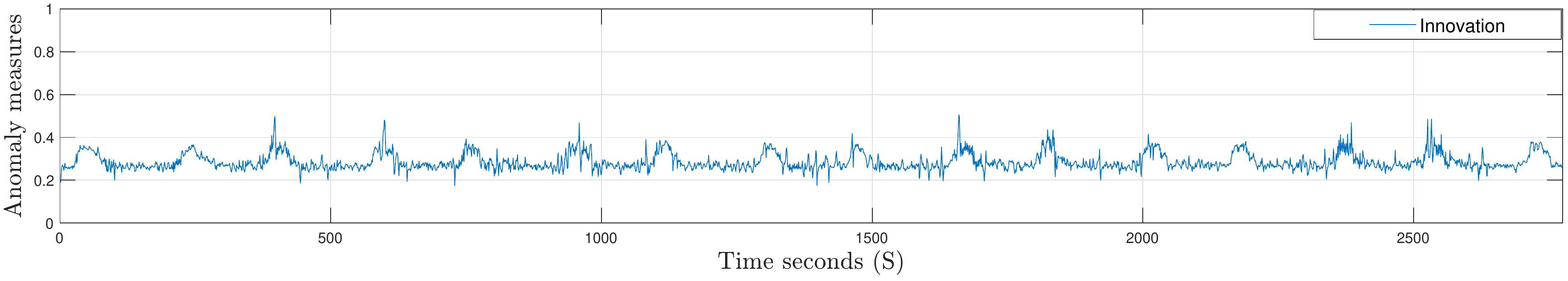}
	
	\flushright(b)\includegraphics[width=0.94\linewidth]{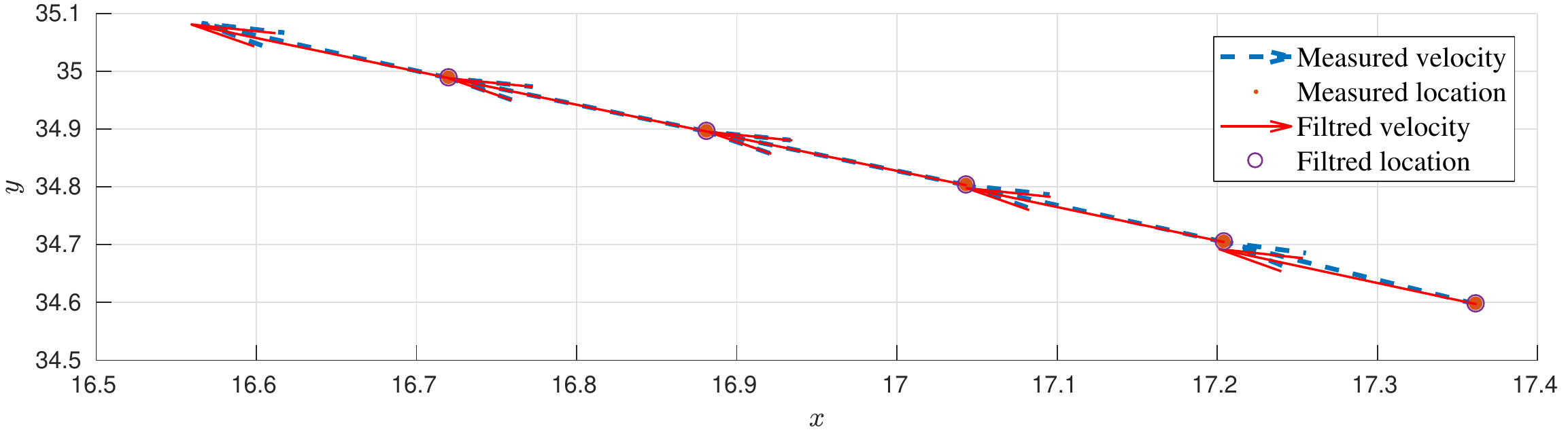}
	\flushright\includegraphics[width=0.92\linewidth]{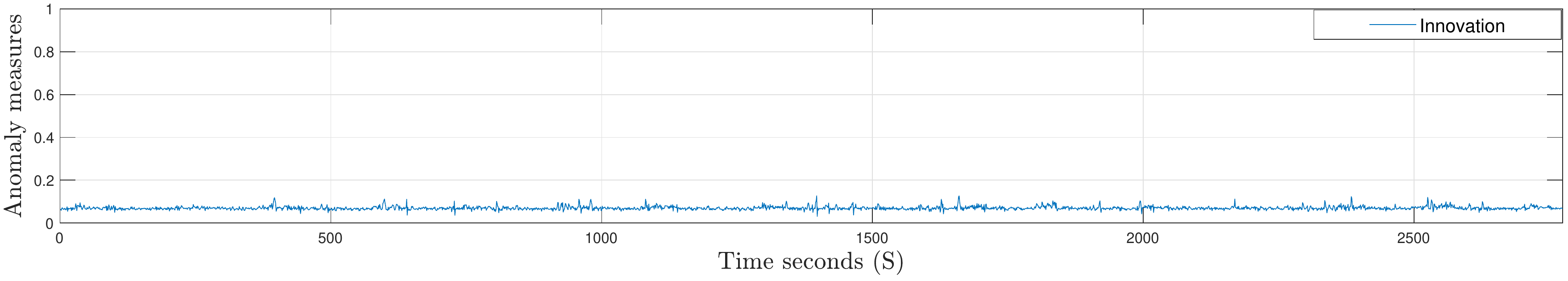}
	\caption[SA: Share layer state estimation]{Share layer state estimation: in (a) the estimated state from UKF is compared with the ground truth, in (b) the state estimation from the learned filters is shown. the second row shows the output abnormality signals corresponding to each filter. The horizontal axis represents the sample number, and the vertical axis shows innovation values (abnormality signal).}
	\label{fig:ukf_training}
\end{figure}

\noindent{\textbf{Training process of SL.} }
As we discussed in Sec. \ref{sec:method}, the reference filter for the shared layer is an UKF (see Fig. \ref{fig:learning_bd}-(a.1)). UKF assumes a simple dynamic equilibrium where the agent is not moving. A sequence of state estimation samples of UKF is shown in Fig. \ref{fig:ukf_training}-a. In this figure UKF always predicts the agent is keep staying in a still position, including a negligible perturbation error. However, this assumption is not true and leads to a huge innovation value. This set of innovation values are used on the next training iteration to incrementally learn new set of filters (see Fig. \ref{fig:learning_bd}-(d.1)). This new set of learned filters model different dynamic equilibriums, in which they will be able to perform estimations more accurate with respect to the simple UKF. The estimation results of this set of trained filters are shown in Fig. \ref{fig:ukf_training}-b, where the predictions are closer to the observations with small amount of error and consequently lower innovation values. 



\begin{figure*}[t]
	\centering
	(a)\includegraphics[width=0.45\linewidth,height=0.1\linewidth]{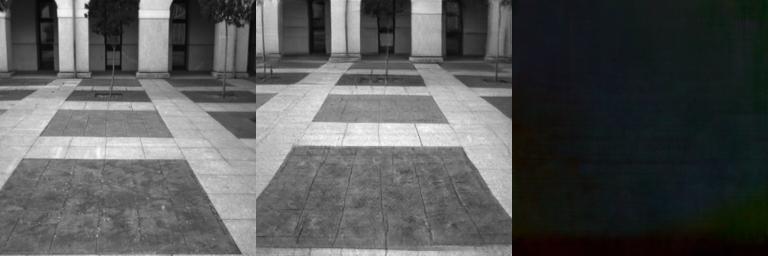}
	(b)\includegraphics[width=0.45\linewidth,height=0.1\linewidth]{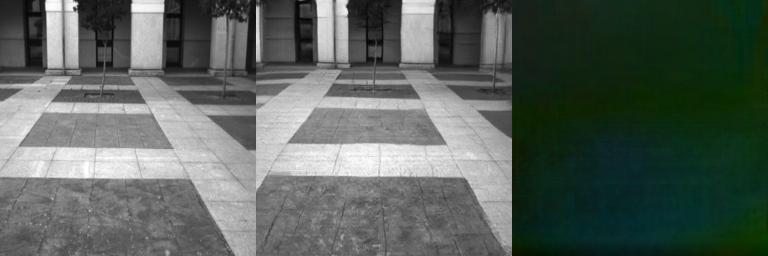}
	(c)\includegraphics[width=0.45\linewidth,height=0.1\linewidth]{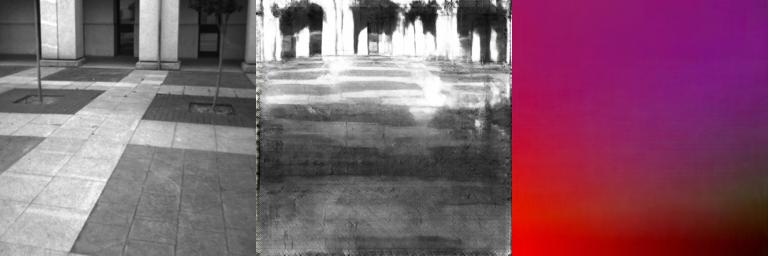}
	(d)\includegraphics[width=0.45\linewidth,height=0.1\linewidth]{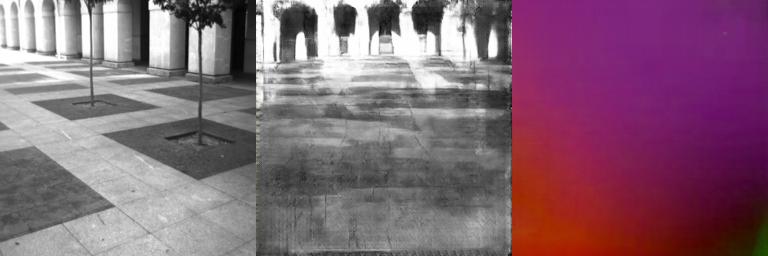}
	(e)\includegraphics[width=0.45\linewidth,height=0.1\linewidth]{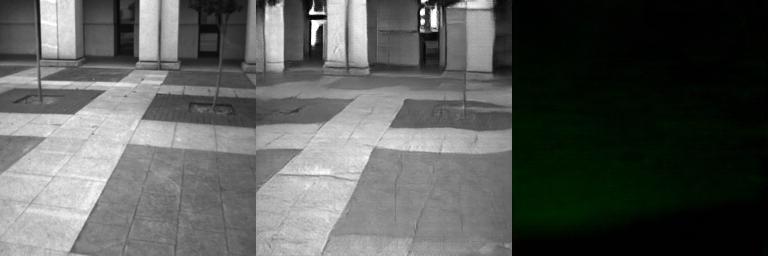}
	(f)\includegraphics[width=0.45\linewidth,height=0.1\linewidth]{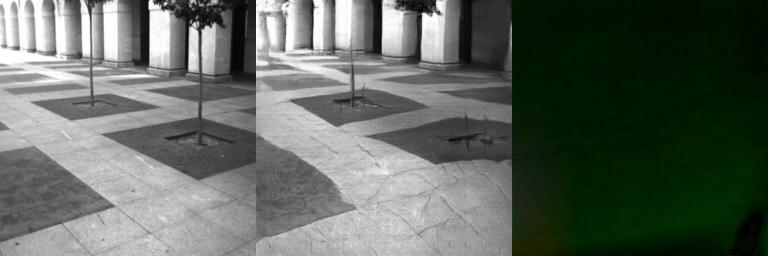}
	\caption[SA: Private layer state estimation]{Private layer state estimation: (a) and (b) are the estimated video frame and optical-flow motion map from the reference GAN ($baseGAN$) while the vehicle moves toward a straight path. (c) and (d) are the estimations from the $baseGAN$ while the vehicle curving, and finally (e) and (f) are the result of second level of GANs for the similar situation (curving). The first image is the real observation, the second image is the predicted frame, and the final image is the difference between prediction optical-flow map and the ground truth, where the black color means lower difference.}
	\label{fig:gan_training_estimation}
\end{figure*}

\noindent{\textbf{Training process of PL.} } As reviewed in Sec. \ref{subsec:learn_PL} constructing the GAN hierarchy is done based on distance of discriminators scores between the predictions and the real observations. The first level of GANs ($base GANs$) is trained on a selected subset of normal samples from \emph{perimeter monitoring} sequence. This subset represents the captured sequences while the vehicle moves on a straight path in a normal situation, i.e. when the road is empty and the expected behavior is the vehicle moving straight. The hypothesis is that this subset only represents one of the distributions of motion and appearance in a highly diverse data condition. As result, when the pair of $base GANs$ detects an abnormality in the corresponding set on which is trained, therefore, it is expected that the corresponding observations can be considered as outliers. This hypothesis is confirmed by testing the $base GANs$ over the entire sequences of \emph{perimeter monitoring}, by observing the discriminators scores distances between the prediction and the observation.
Fig. \ref{fig:train} shows the results of training reference GAN ($base GANs$). Our hypothesis regarding the complexity of distributions is confirmed on Fig. \ref{fig:train} (a), where the test is performed using only the $base GANs$. $base GANs$ can detect the straight path (white background area) perfectly, while when the vehicle curves (green bars) it failed and recognized curving as an abnormal event. The discriminator scores distances between the prediction and the observation (abnormality signal) is higher over the curving areas, which was expected. However, after training the second level GANs using this subset of data and applying entire hierarchy, the model can recognize entire training sequence as normal. This happened because the different distribution of samples tagged as abnormal by the $base GANs$ input to the second level of the hierarchy, where they are recognized as normal samples.

The estimation of optical-flow and frame for each level (iteration) of the training process is shown in Fig. \ref{fig:gan_training_estimation}. For each case, it can be identified an image triplet where the left image is the ground truth observed frame, the central image shows the predicted frame, and the right image is the difference between the observed optical-flow motion map and the predicted optical-flow. The lower the distance between predicted motion and observation the blacker (values are near to ``$0$") the right image. In this figure, (a), (b), (c), and (d) show the output estimation for the first iteration ($base GANs$). As can be seen, straight motions displayed in (a) and (b) are correctly estimated, see the low error, i.e., black pixels, in the right frames of their triplets. Nonetheless, the initial model is unable to predict curve motions shown in (c) and (d), see the high error, i.e., colorful pixels, in the right frames of their triplets. Fig. \ref{fig:gan_training_estimation} (e) and (f) show the estimation from the hierarchy of GANs after a full training phase in case of curving. It can be observed this time how the GANs estimate the curving motion with high accuracy, see the amount of black pixels in the triplet's right frames. 

As the $base GANs$ is trained on the reference situation which is the straight movements (see Fig. \ref{fig:learning_bd}-(a.2)), therefore is it expected to having a good estimation while the vehicle moving straight (Fig. \ref{fig:gan_training_estimation}-a-b). However, this filter fails to estimate curves (Fig. \ref{fig:gan_training_estimation}-c and d). The incrementally nature of the proposed method is demonstrated in Fig. \ref{fig:learning_bd}-(d.2), where low estimation errors are obtained by using the second level GANs for predicting curve motions, see \ref{fig:gan_training_estimation}-e-f. 

 \begin{figure}[t]
	\centering
	\includegraphics[width=0.8\linewidth]{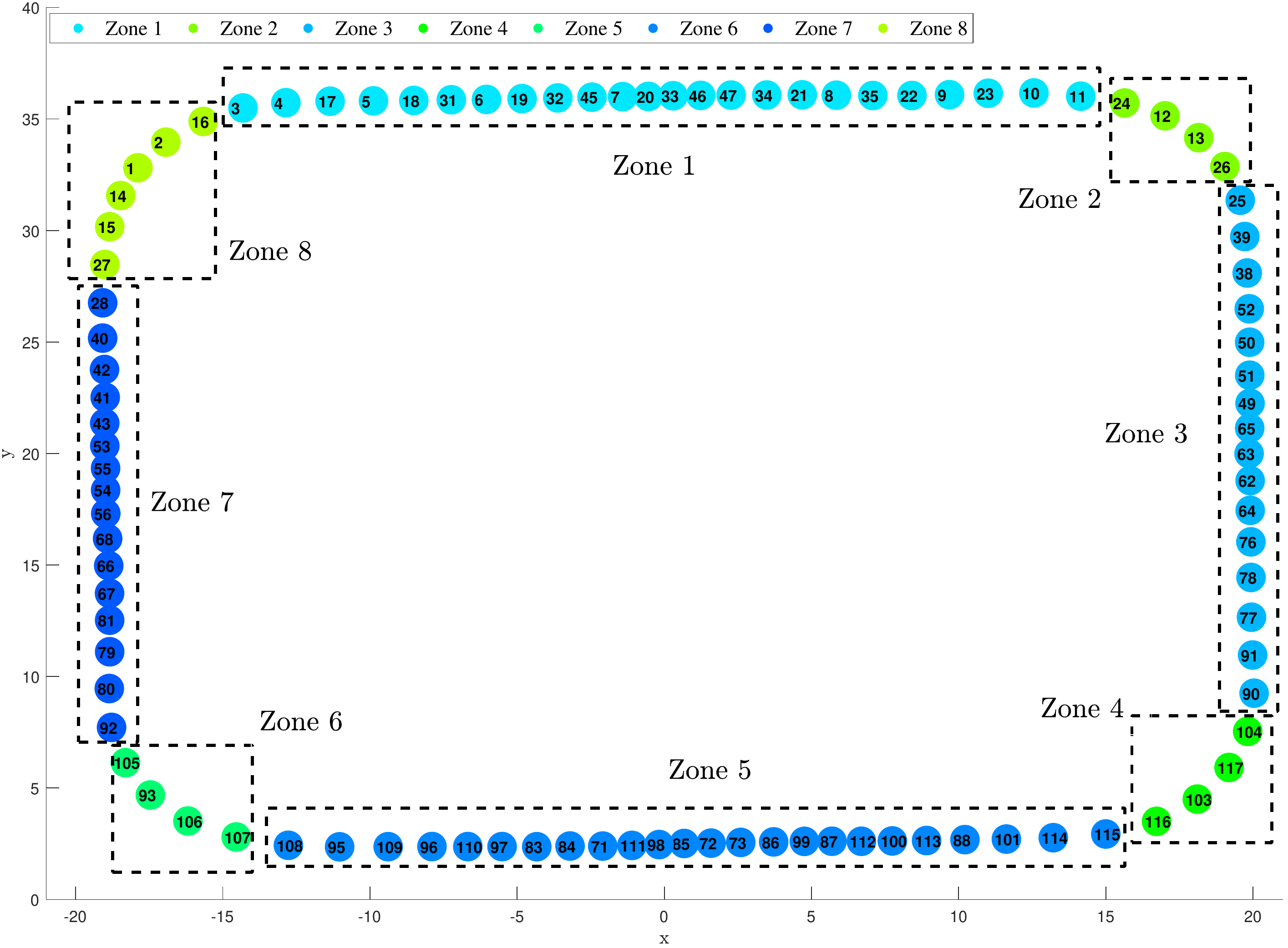}
	\caption{Color-coded zones from SL and PL.}
		\label{fig:zones}
\end{figure}
\begin{figure}[htb]
\begin{minipage}[b]{\linewidth}
  \centering
  \setlength\tabcolsep{0pt}
  \begin{tabular}{p{0.05\linewidth} p{0.95\linewidth}}
     (a)&  \includegraphics[width=\linewidth,trim={4.55cm 0.6cm 3.7cm 0},clip]{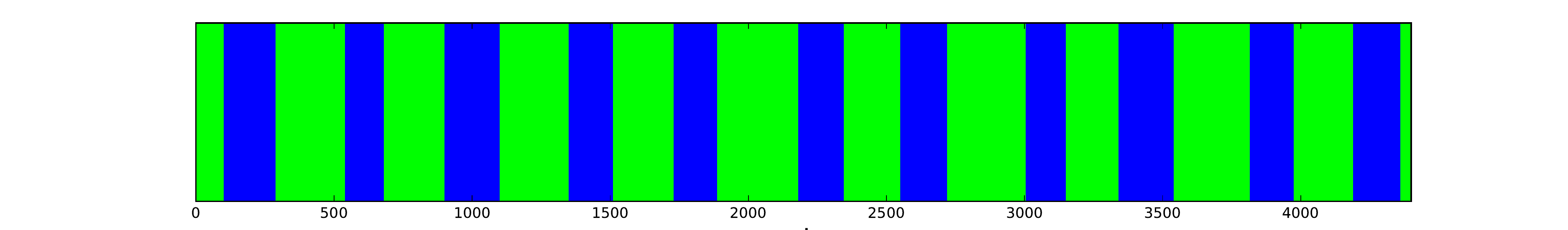}\\
     (b)&  \includegraphics[width=\linewidth,trim={4.55cm 0 3.7cm 0},clip]{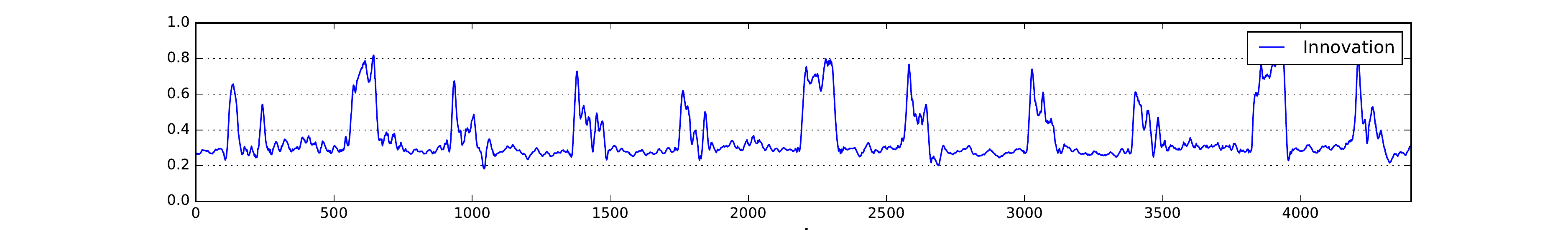}\\
     (c)&  \includegraphics[width=\linewidth,trim={4.55cm 0 3.7cm 0},clip]{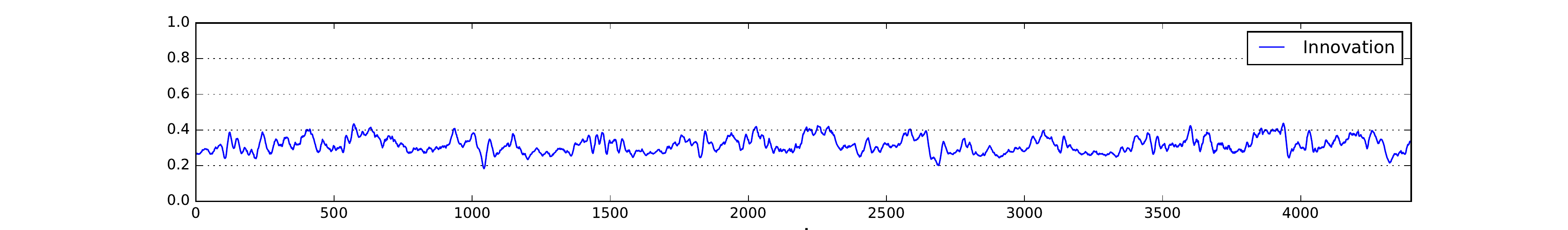}\\
  \end{tabular}
\end{minipage}

\caption[SA: Training hierarchy of GANs]{Training hierarchy of GANs: (a)  ground truth labels, the green background means the vehicle moves on a straight line, while the blue bars indicate curving. (b) and (c) showing the signal of the averaged score distance values between prediction and observation (innovation) for the first level of GANs and the hierarchy of GANs, respectively. The horizontal axis represent the time, and the vertical axis shows the innovation values.}
\label{fig:train}
\end{figure}

\begin{table*}[]
	\centering
	\begin{tabular}{p{1cm}p{1.5cm}p{0.8cm}p{1.8cm}p{1cm}p{1cm}p{1.5cm}p{1.8cm}p{1.2cm}}
		\hline
		\#Zone         &  1                                                                                  &  2         &  3                                                         &  4   &  5 &  6  &  7 &  8  \\ \hline
		SL ($\{S_k\}$) & 3-11, 17-23, 31-35, 45-47 & 12, 13, 24, 26 & 25, 38, 39, 49-52, 62-65, 76-78, 90, 91 & 103, 104 &  111-115    & 93, 105-107        &  28, 40-43, 53-56, 66-68, 79-81, 92      & 1, 2, 14-16, 27        \\
		PL ($\{C_k\}$) & 1-3  & 4-7        & 1-3  & 4-7  & 1-3  & 4-7 & 1-3  & 4-7 \\ \hline
	\end{tabular}
	
	\caption[SA: List of corresponding detected super-states from PL and SL for the normal scenario]{List of corresponding detected super-states from PL and SL for the normal scenario: for each individual zone the number of color-coded super-states sequences from PL ($\{C_k\}$) and SL ($\{S_k\}$) are shown.}
	\label{tab:pri_ss}
\end{table*}

\begin{figure}[]
	\centerline{\scriptsize{(a) }\includegraphics[width=0.96\linewidth,trim={4.4cm 0.6cm 3.2cm 0},clip]{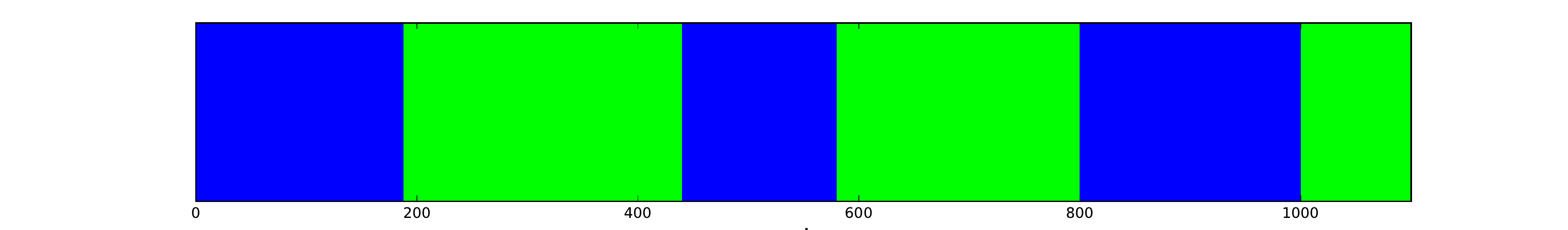}}
	\centerline{\scriptsize{(b) }\includegraphics[width=0.96\linewidth,trim={4.4cm 0 3.2cm 0},clip]{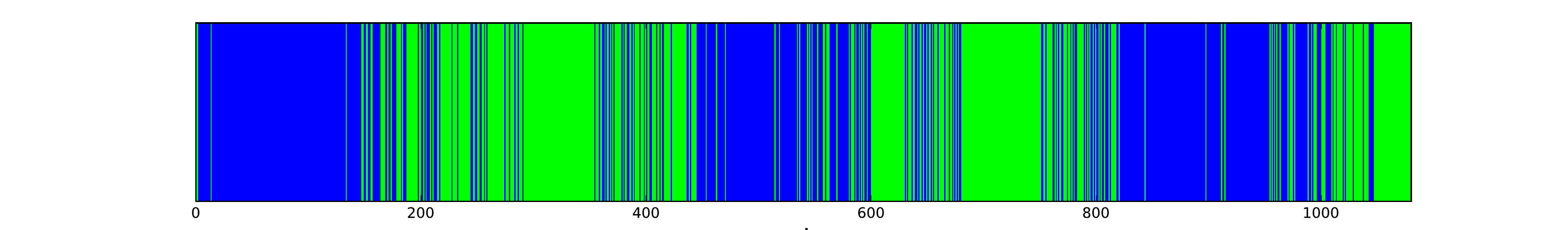}}
	\centerline{\scriptsize{(c)  }\includegraphics[width=0.935\linewidth,trim={0.55cm 0 0cm 0},clip]{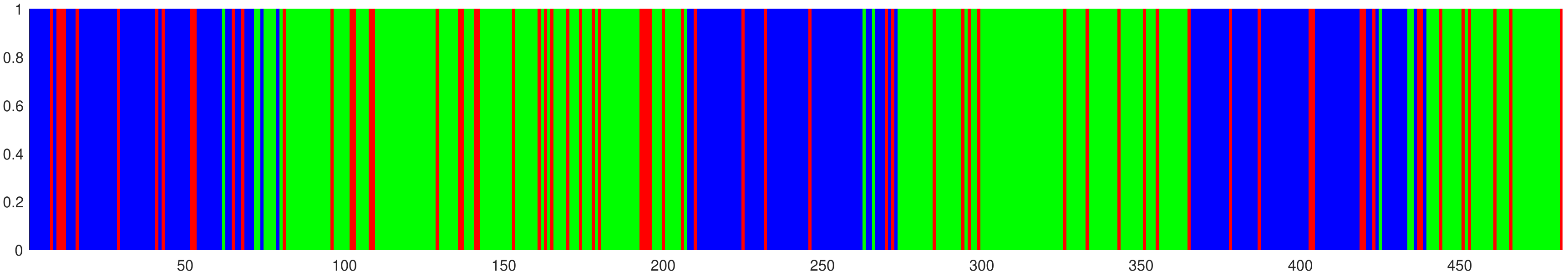}}
	\centerline{\scriptsize{(d)}\includegraphics[width=0.96\linewidth,trim={4cm 0 3.2cm 0},clip]{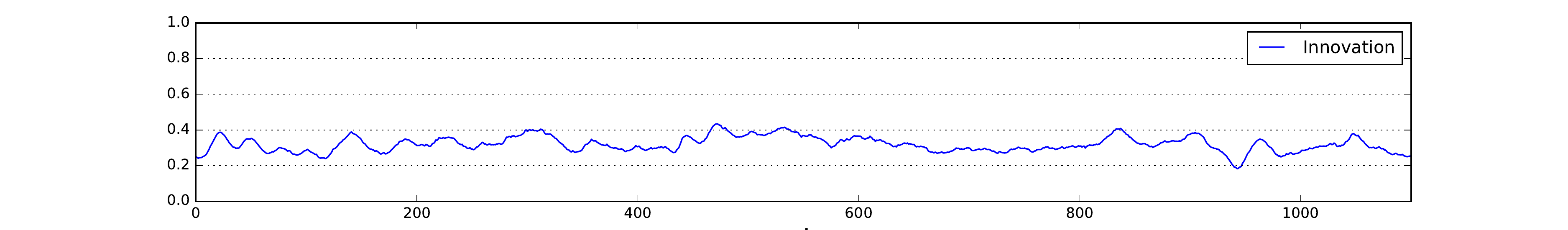}}
	\centerline{\scriptsize{(e)}\includegraphics[width=0.96\linewidth,trim={4.8cm 0 3.5cm 0},clip]{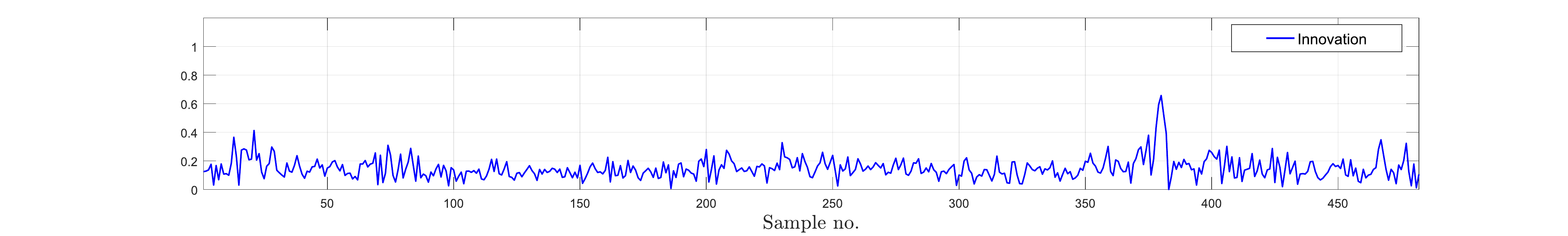}}
	\caption[SA: Normality representations from PL and SL]{Normality representations from PL and SL: in (a) the ground truth labels are shown, moving straight is green and blue bars represent the curving. Color-coded super-states sequences $\{C_k\}$ and $\{S_k\}$ are shown in (b) and (c), respectively. They are highly correlated with the agent's real status (a). (d) and (e) show the abnormality signals from PL and SL, respectively. The horizontal axis represents the sample number, and the vertical axis shows the innovation values (abnormality signal).}
	\label{fig:pri_states}
\end{figure}

\noindent{\textbf{Representation of normality from SL and PL.} }
 The two levels of the proposed self-awareness model, including the shared layer (modeled by MJPF), and the private layer (modeled as a hierarchy of GANs), are able to learn the normality. In our experiments this is defined as Scenario I (Fig.~\ref{fig:plans}-a) and it is used to learn both models. As reviewed in Sec. \ref{sec:method} both SL and PL, represent their situation awareness by a set of super states following with abnormality signals. We select a period of normal perimeter monitoring task (see Fig. \ref{fig:subplans}-a) as a test scenario. The result for PL and SL is shown in Fig. \ref{fig:pri_states}, which simply visualizes the learned normality representations. The ground truth label is shown in Fig. \ref{fig:pri_states}-a, and the color-coded detected super states from PL $\{{C_t}\}$ and SL $\{S_k\}$ are illustrated in Fig. \ref{fig:pri_states}-b and Fig. \ref{fig:pri_states}-c, respectively. It clearly shows not only the pattern of superstates are repetitive and highly-correlated with the ground truth, but also there is a strong correlation between the sequence of PL and SL super states.
 
Study of the cross-correlation between the SL and PL is beyond the scope of this paper, but it is also interesting to demonstrate such relation. In order to show the correlation of two learned models (SL and PL), we divided the environment into eight meaningful zones, including curves and straight path. This semantical partitioning of state-space is shown Fig.~\ref{fig:zones}. For the training scenario (normal situation) the color-coded super-states from SL and PL are visualized over the environment plane. As it shown in Tab. \ref{tab:pri_ss} for the SL we have detected 115 super-states each corresponding to an individual filter, while for PL we have detected 7 different super-states corresponding to two GAN filters. 

Note that the abnormality signals are stable for both PL and SL, while in case of an abnormal situation we expect to observe identical spikes over the signals. In order to study such abnormal situations we apply the trained self-awareness model over unseen test sequences. Two different scenarios are selected, in which the moving vehicle performing the perimeter monitoring task has to face abnormal events. In each scenario the agent performs different actions in order to solve the abnormal situation. The goal of this set of experiments is to evaluate the performance of the proposed self-awareness model (SL and PL) for detecting the abnormalities.

\graphicspath{{Chapter3/Figs/a/}}
\subsection{Experiments on the novel situations}
In this phase we performed an online testing setup to evaluate the performance of our models. This procedure of this phase for the SL is shown in Fig. \ref{fig:learning_bd}-(b.1), (c.1) and (d.1) and for the PL is shown in Fig. \ref{fig:learning_bd}-(b.1), (c.1) and (d.1).

\begin{figure*}[t]
	\begin{center}
		\begin{minipage}[t]{0.325\textwidth}
			\centering
			\includegraphics[width=\textwidth,trim={0.89cm 0.18cm 1.25cm 0.68cm},clip]{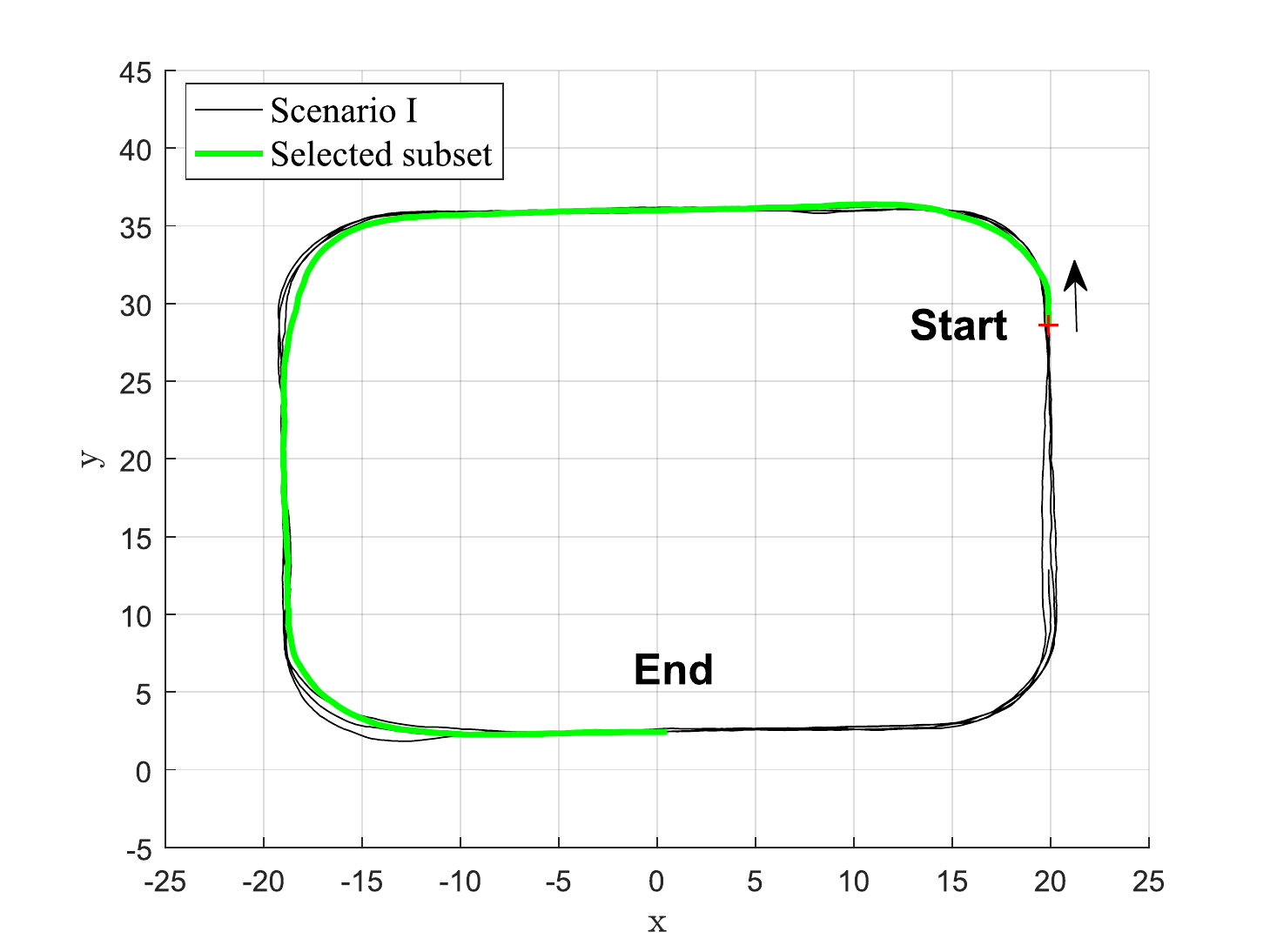}
			\scriptsize{(a) perimeter monitoring}
		\end{minipage}
		\begin{minipage}[t]{0.325\textwidth}
			\centering
			\includegraphics[width=\textwidth,trim={0.9cm 0.2cm 1.25cm 0.7cm},clip]{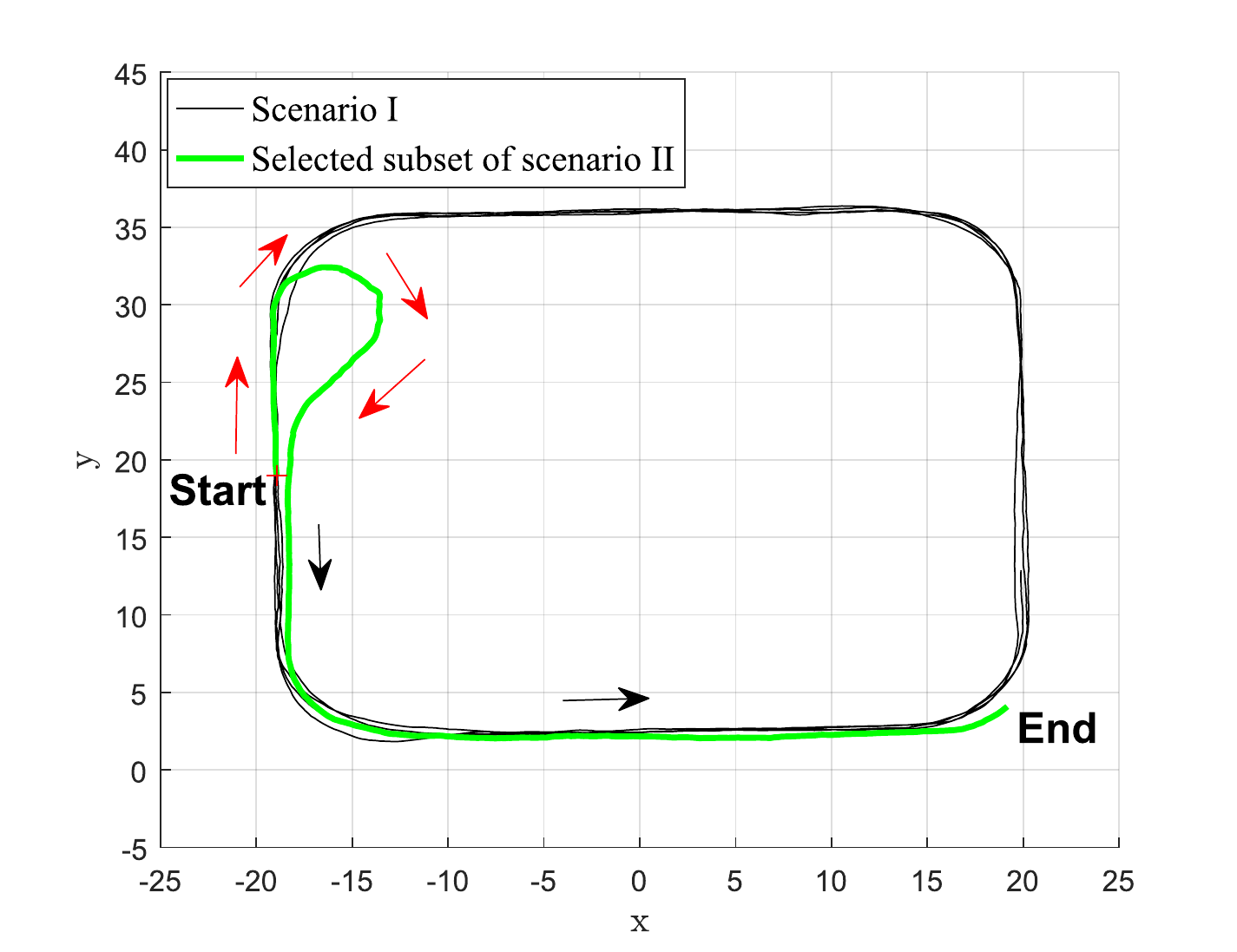}
			\scriptsize{(b) U-turn}
		\end{minipage}
		\begin{minipage}[t]{0.325\textwidth}
			\centering
			\includegraphics[width=\textwidth,trim={0.9cm 0.2cm 1.25cm 0.7cm},clip]{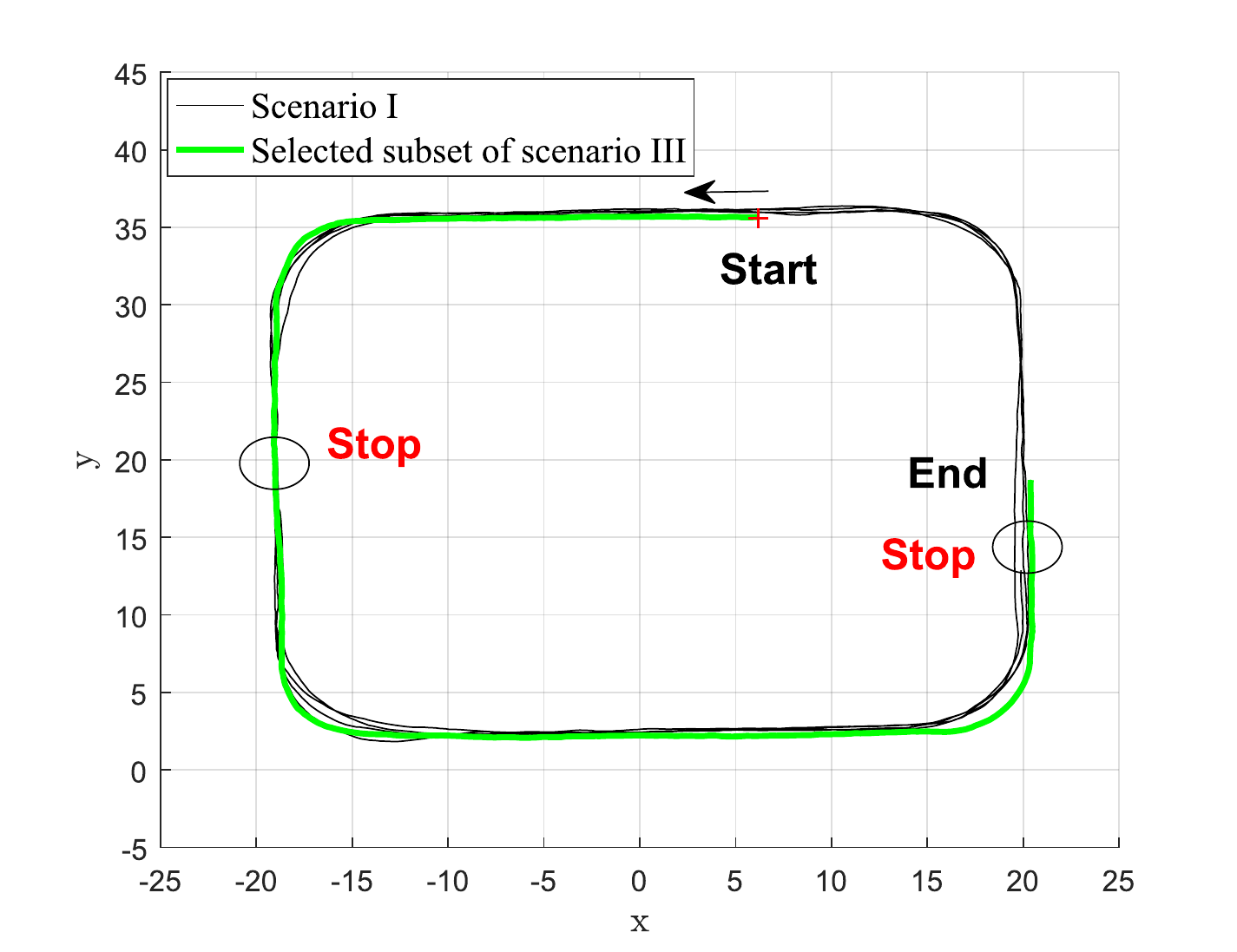}
			\scriptsize{(c) emergency stop}
			\label{fig:Stop}
		\end{minipage}
	\end{center}
	\caption{Sub-sequence examples from testing scenarios reported in the experimental results.}
	\label{fig:subplans}
\end{figure*}

\begin{figure}[H]
	\centerline{\scriptsize{(a)}\includegraphics[width=0.96\linewidth,trim={4.35cm 0.6cm 3.2cm 0},clip]{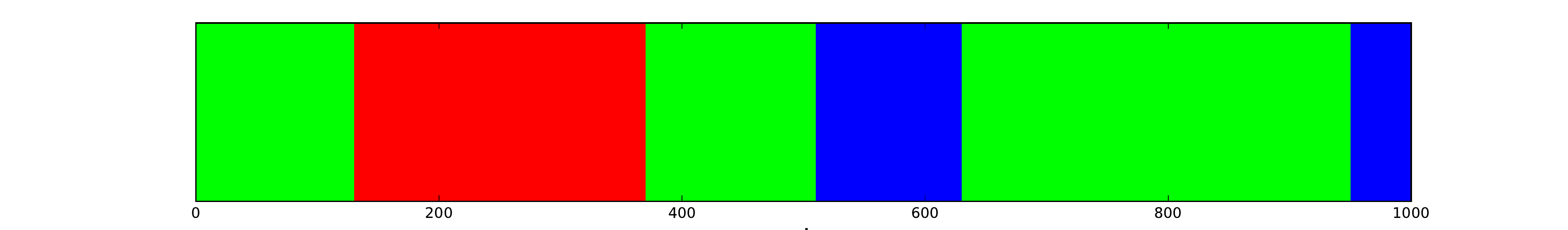}}
	\centerline{\scriptsize{(b)}\includegraphics[width=0.96\linewidth,trim={4.35cm 0 3.2cm 0},clip]{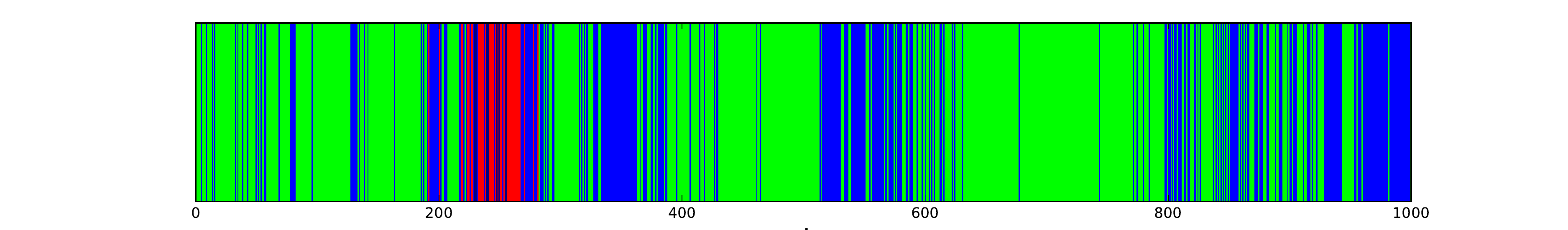}}
	\centerline{\scriptsize{(c) }\includegraphics[width=0.935\linewidth,trim={0.55cm 0 0cm 0},clip]{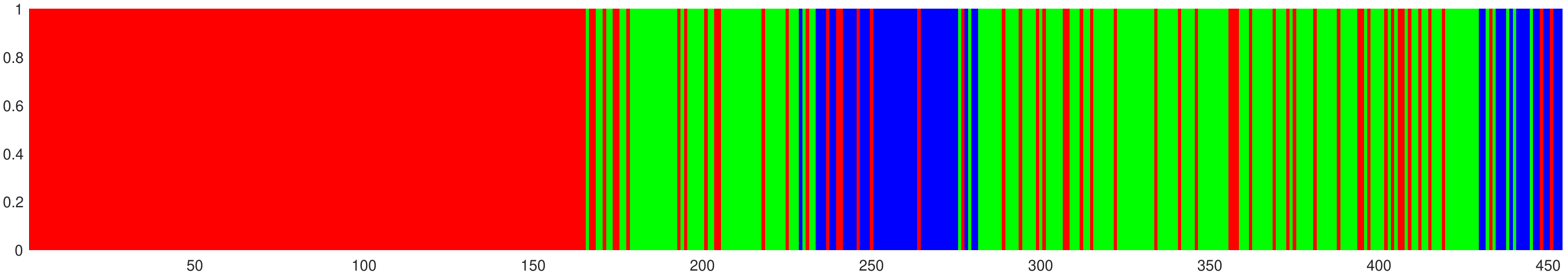}}
	\centerline{\scriptsize{(d)}\includegraphics[width=0.96\linewidth,trim={4cm 0 3.2cm 0},clip]{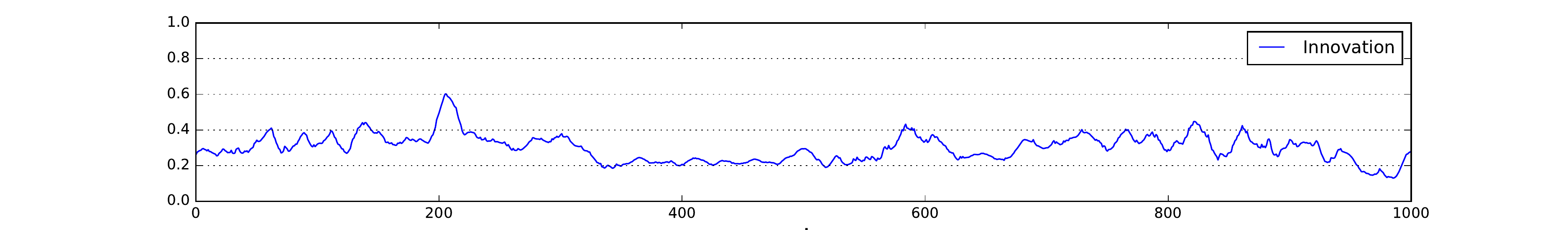}}
	\centerline{\scriptsize{(e)}\includegraphics[width=0.96\linewidth,trim={4.8cm 0 3.5cm 0},clip]{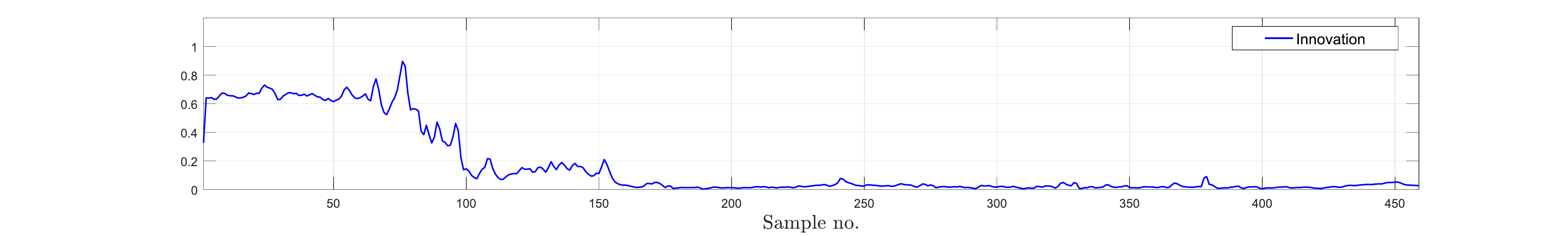}}
	\caption[SA: Abnormality in the U-turn avoiding scenario]{Abnormality in the U-turn avoiding scenario: (a) ground truth labels. (b) and (c) color-coded transition of states $\{C_k\}$ and $\{S_k\}$, respectively. (d) and (e) generated abnormality signal (innovation) from  PL and SL, respectively. The horizontal axis represents the sample number, and the vertical axis shows the innovation values (abnormality signal).}
	\label{fig:testushape}
\end{figure}

\noindent{\textbf{Avoiding a pedestrian by a U-turn action.} } In this scenario, which is illustrated in Fig.~\ref{fig:plans}-b, the vehicle performs an avoidance maneuver over a static pedestrian by a U-turn to continue the standard monitoring afterward. The goal is to detect the abnormality, which is the presence of the pedestrian and consequently the unexpected action of the agent with respect to the learned normality during the perimeter monitoring. In Fig. \ref{fig:testushape} the result of anomaly detection from PL and SL representations is shown. The results are related to the highlighted time slice of the testing scenario II (Fig. \ref{fig:subplans}-b).

In Fig. \ref{fig:testushape}-a the green background means that the vehicle moves on a straight line, the blue bars indicate curving, and red show the presence of an abnormal situation (which in this case is the static pedestrian). The abnormality area starts on first sight of the pedestrian and it continues until the avoiding maneuver finishes (end of U-turn). Similarly, the sequences of states $\{C_k\}$ and $\{S_k\}$ in Fig. \ref{fig:testushape}-(b,c), follow the same pattern. While the situation is normal, the super-states repeat the expected normal pattern, but as soon as the abnormality begins the super-state patterns are changed in the both PL and SL (e.g., dummy super states). Furthermore, in the abnormal supper-states the abnormality signals showing the higher values.  
The abnormality signal generated by SL is shown in Fig. \ref{fig:testushape}-e, which represents the innovation between prediction and the observation. The abnormality produced by the vehicle is higher while it moving thought the path which is indicated by red arrows in Fig. \ref{fig:subplans}-b. This is due to the fact that the observations are outside the domain that superstates are trained on. Namely, during the training such state space configuration is never observed. This means the KF innovation becomes higher in the same time interval due to the opposite velocity compared with the normal behavior of the model, more specific, KF innovation is high due to the difference between prediction (which is predicted higher probability going straight) and the likelihood of the observed behavior in a curved path.

The abnormality signal generated by PL, Fig. \ref{fig:testushape}-d, is computed by averaging over the distance maps between the prediction and the observation score maps: when an abnormality begins this value does not undergo large changes since the observing a compressed local abnormality (see Fig. \ref{fig:visavoiding}-c) can not change the average value significantly. However, as soon as observing a full sight of pedestrian and starting the avoidance action by the vehicle, the abnormality signal becomes higher since both observed appearance and action are presenting an unseen situation. This situation is shown in Fig. \ref{fig:visavoiding}-(d,e). As soon as the agent back to the known situation (e.g., curving) the abnormality signal becomes lower.

\begin{figure}[H]
	\centerline{\scriptsize{(a)}\includegraphics[width=0.94\linewidth,trim={4.35cm 0.6cm 3.2cm 0},clip]{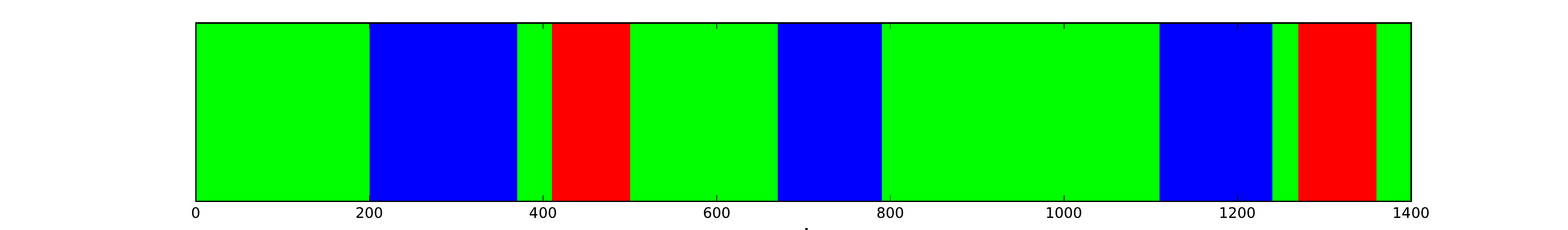}}
	\centerline{\scriptsize{(b)}\includegraphics[width=0.94\linewidth,trim={4.35cm 0 3.2cm 0},clip]{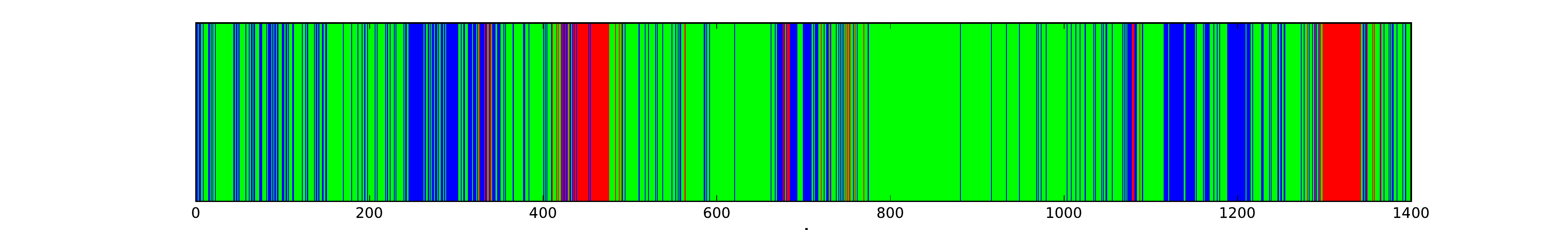}}
	\centerline{\scriptsize{(c)}\includegraphics[width=0.92\linewidth,trim={0.55cm 0 0cm 0},clip]{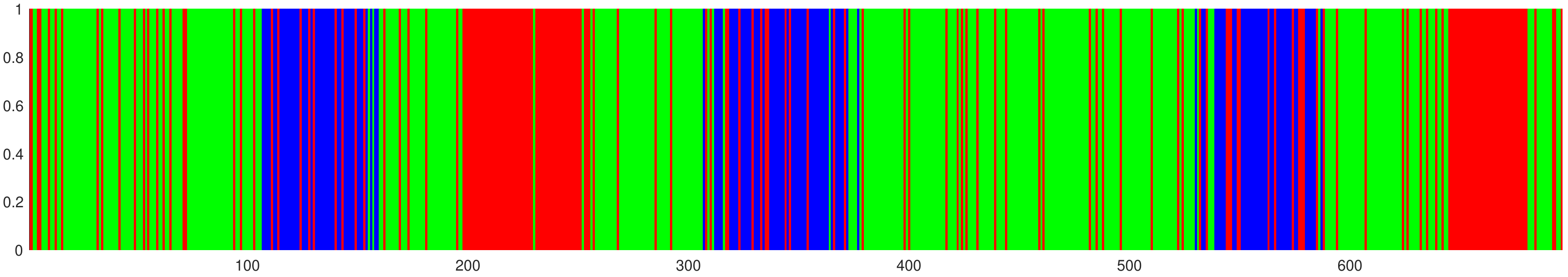}}
	\centerline{\scriptsize{(d)}\includegraphics[width=0.94\linewidth,trim={4cm 0 3.2cm 0},clip]{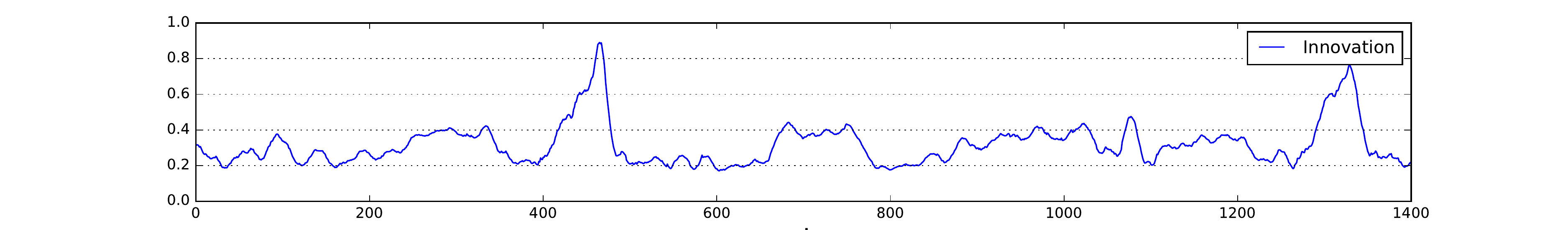}}
	\centerline{\scriptsize{(e)}\includegraphics[width=0.94\linewidth,trim={4.8cm 0 3.5cm 0},clip]{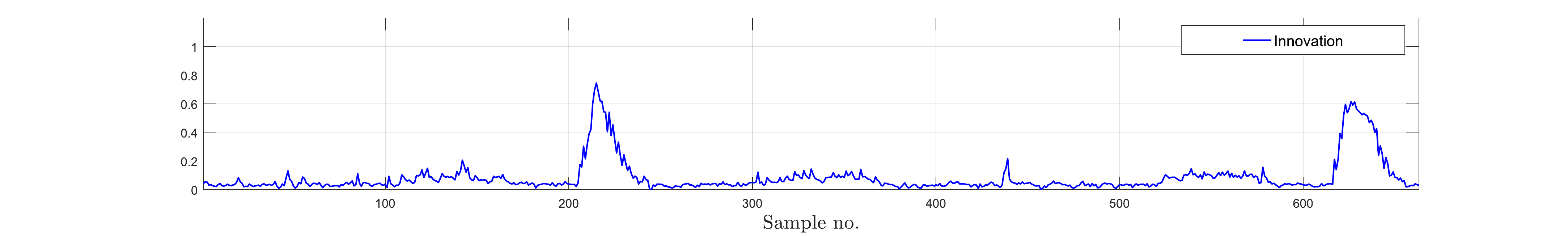}}
	\caption[SA: Abnormality in the emergency stop scenario]{Abnormality in the emergency stop scenario: (a) ground truth labels. (b) and (c) color-coded transition of states from PL and SL, respectively. (d) and (e) generated abnormality signal (innovation) $\{C_k\}$ and $\{S_k\}$, respectively. The horizontal axis represents the sample number, and the vertical axis shows the innovation values (abnormality signal).}
	\label{fig:teststop}
\end{figure}

\noindent{\textbf{Emergency stop maneuver.} } This scenario is shown in Fig. \ref{fig:plans}-c, where the agent performs an emergency stop for a pedestrian to cross. Accordingly, the results of abnormality detection, for the highlighted time slice in Fig. \ref{fig:subplans}-c, are represented in Fig. \ref{fig:teststop}.   
In Fig. \ref{fig:teststop}-a the red bars indicate the abnormality areas, where the agent is stopped and waits until the pedestrian cross. Accordingly, this areas are represented as the dummy super states from PL (light-blue color in Fig. \ref{fig:teststop}-b) with high scores in the abnormality signal (see Fig. \ref{fig:teststop}-d). \\

\begin{figure}[H]
	\centerline{\scriptsize{(a)}\includegraphics[width=0.95\linewidth,height=3.2cm]{images/ped_1}}
	\centerline{\scriptsize{(b)}\includegraphics[width=0.95\linewidth,height=3.2cm]{images/ped_2}}
	\centerline{\scriptsize{(c)}\includegraphics[width=0.95\linewidth,height=3.2cm]{images/ped_3}}
	\centerline{\scriptsize{(d)}\includegraphics[width=0.95\linewidth,height=3.2cm]{images/ped_4}}
	\centerline{\scriptsize{(e)}\includegraphics[width=0.95\linewidth,height=3.2cm]{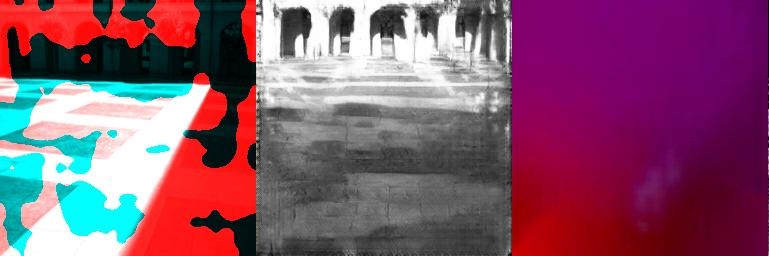}}
	\caption[SA: Visualization of abnormality]{Visualization of abnormality: first column shows the localization over the original frame, second column is the predicted frame, and the last column shows the pixel-by-pixel distance over the optical-flow maps. (a) moving straight, (b) curving, (c) first observation of the pedestrian, (d) and (e) performing the avoiding action.}
	\label{fig:visavoiding}
\end{figure}

The generated abnormality signal from PL increases smoothly as the agent get better visual to the pedestrian, and reaches to the peak when the agent stops and having a full visual of the pedestrian. Once the pedestrian passes and the agent starts to continue its straight path, the signal drops sharply. Similarly, the abnormality signal from SL representation model (see Fig. \ref{fig:teststop}-e) shows two peaks that correspond to high innovation. Those peaks represent the abnormality patterns associated with the emergency stop maneuver. In contrast with the PL signal, the SL signal reaches to the peak sharply, and then smoothly back to the normal level. Such pattern indicates that the vehicle stops immediately and waits for a while, then it starts again to move and increases its velocity constantly to continue the straight path under the normal situations. As a consequence different motion patterns with respect to those predicted are detected by means of innovation. This is also confirmed by the color-coded super states in Fig. \ref{fig:teststop}-c, where the green and dark-blue states are continued longer than what expected with respect to the normal pattern that learned from the previous observations.
\graphicspath{{Chapter3/Figs/a/}}
\subsection{Discussions}
\label{sec:Discussion}
\noindent{\textbf{The cross-modal representation.} }One of our novelties in this work is using GAN for a {\em multi-channel} data representation. Specifically, we use both appearance and motion (optical-flow) information: a two-channel approach which has been proved to be empirically important in previous works. Moreover, we propose to use a cross-channel approach where, we train two networks which respectively transform raw-pixel images in optical-flow representations and vice-versa. The rationale behind this is that the architecture of our conditional generators $G$ is based on an encoder-decoder (see Sec.~\ref{subsec:learn_PL}) and one of the advantage of such channel-transformation tasks is to prevent $G$ learns a trivial identity function and force $G$ and $D$ to construct sufficiently informative internal representations.

\begin{figure*}
	\centering
	\includegraphics[width=\linewidth]{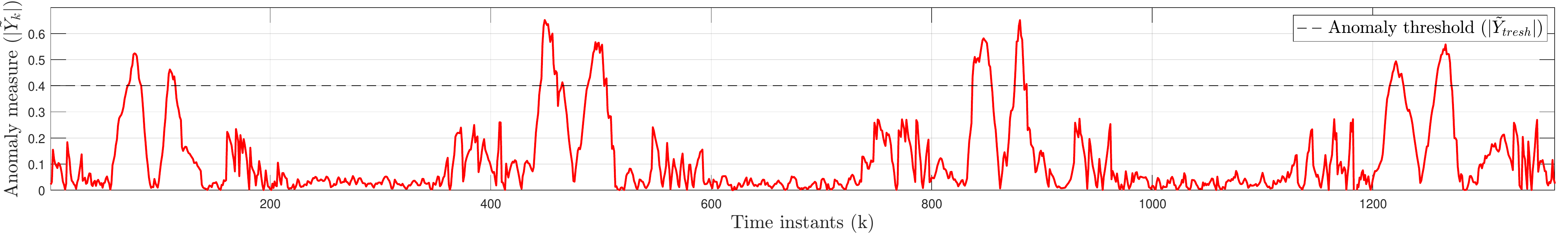}
	(a)
	\includegraphics[width=\linewidth]{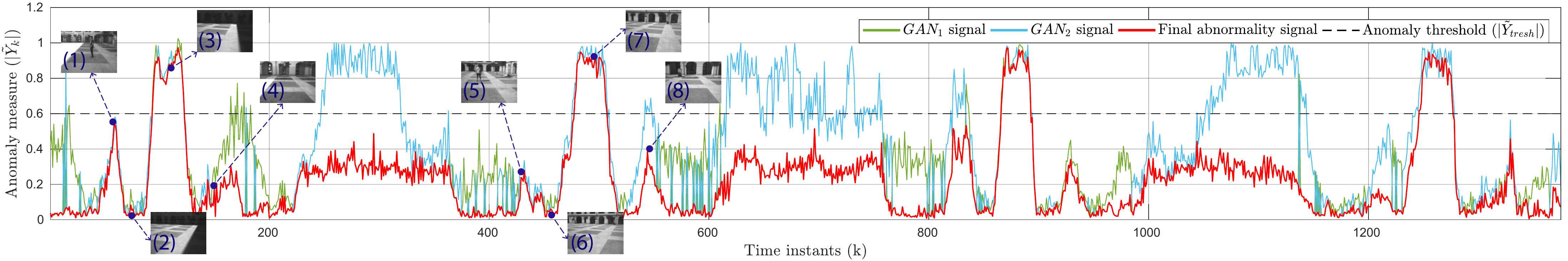}
	
	(b)
	\caption[SA: anomaly measurements from SL and PL]{(a)SL anomaly measurements: perimeter control activity by GP through time with avoidance of static pedestrians. (b)PL anomaly measurements: the distances between the observations and predictions by GANs during the time.} 
	\label{fig:gansig} 
\end{figure*}

\noindent{\textbf{Private layer and shared layer cross-correlation.} }The PL and SL levels are providing complimentary information regarding the situation awareness. As an instance, it has been observed that in case of PL the super-states are invariant to the agent's location, while SL super-states representation is sensitive to such spatial information. In other words, PL representation can be seen as the semantic aspect of agent's situation awareness (e.g., moving straight, curving) regardless to the current location of the agent. Hence, the pattern of superstates sequences is repetitive, see how PL's superstates are repeated in Tab. \ref{tab:pri_ss}, e.g., Zone 1, 3, 5 are 4 correspond to the same superstates. In contrast, the SL representation includes spatial information, which generates more specific superstates for describing each zone, see $\{S_K\}$ in Tab. \ref{tab:pri_ss}. Such pattern is frequently happens for the entire sequences. Fig. \ref{fig:gansig} shows the detailed abnormality signals associated to the PL and SL for the avoiding maneuver scenario. Three signals are shown in Fig. \ref{fig:gansig}-b: The green and blue signals respectively show the computed signals by the base GAN ($GAN_1$ trained on straight path) and the second level GAN ($GAN_2$ trained on curves). The red signal indicates the final abnormality measurement and it is aligned with SL results shown in Fig. \ref{fig:gansig}-a.
In Fig. \ref{fig:gansig} different parts of the curve can be associated and explained by considering the correspondent images acquired from the on-board sensor. Specifically, the small peak identified with number $"1"$ can be justified by the presence of the pedestrian in the field of view of the camera: the vehicle do not start the avoidance maneuver yet, thus, it can be seen as a pre-alarm. The small peak in $"1"$ corresponds to peak in $"5"$, the latter is smaller due to the posture of the pedestrian, see correspondent images $"1"$ and $"7"$. The areas of the curve identified with numbers $"2"$ and $"3"$ or $"6"$ and $"7"$ correspond to the starting point of the abnormal maneuver and the avoiding behavior itself: it can be seen that peaks $"3"$ and $"7"$ are higher than the selected threshold and then correspond to an anomaly. After the small peak $"4"$, that corresponds to the closing part of the avoidance turn, the vehicle goes back to the standard behavior. In particular, at this point of the curve, the vehicle is actually turning. In the wider area (from $220$ to $380$ secs.), the agent keeps moving straight. The slightly higher level of the abnormality curve in straight areas can be explained by a noise related to the vibration of the on-board camera due to the fast movement of the vehicle when increasing its speed.
It is notable that, the signal generated by $GAN_1$ becomes higher in the curving areas since it is only trained on empty straight path for detecting straight paths. Similarly, the $GAN_2$ which is trained on curves, generates higher scores on the straight path. However, both $GAN_1$, and $GAN_2$ can detect the unseen situation (pedestrian avoidance) where both generate a high abnormality score.

In light of the above, these two representations are carrying complimentary information and finding a cross-correlation between PL and SL situation representation (e.g., using coupled Bayesian network) could potentially increase the ability of anomaly detection and consequently boost the entire self-awareness model.

\noindent{\textbf{Modalities alignment.} }In our work, the means of alignment between two modalities localization (in SL) and visual perception (in PL), is provided by the synchronous time stamps assigned from the sensors. Since the time reference is equal for both sensors (odometer and camera) this is possible to collect the aligned multi-modal data. However, another advantage of modeling the cross-correlation between two layers (PL and SL) could be providing an extra reference for finding such alignments. Namely, the cross-correlation of repetitive patterns in PL and SL can be used for the data alignment. In case of having asynchronous sensors, this could be useful to find the exact alignment.

\graphicspath{{Chapter4/}}
\section{Conclusions and Future Works}
\label{ch:concl_future}

This thesis has presented distinct techniques to detect the outliers and learn the new detected unseen situations. Explored strategies have demonstrated that any complex distribution can be understood by breaking it down into a set of small distribution and detect each small distribution as individuals' deviated from the previously learned distributions. Such approach is validated through results of chapters \ref{ch:abnormality}, and \ref{ch:autonomous_car}, where the novel situations are modeled in terms of outliers of the learned situations. 

In particular, in section \ref{sec:gan_anomaly} we presented a GAN-based approach for abnormality detection in crowds. We proposed a generative deep learning method based on two conditional GANs. Since our GANs are trained using only normal data, they are not able to generate abnormal events. At testing time, a local difference between the real and the generated images is used to detect possible abnormalities. We use the mutual supervisory information of the generator and the discriminator GAN networks in order to train a deep network end-to-end with small weakly supervised training sets. Differently from common generation-oriented GANs, after training we directly use the discriminators as the final classifiers to detect the outliers for the task of abnormality detection. In order for this approach to be effective, we designed two non-trivial cross-channel generative tasks for training our networks.

Additionally, as shown in Chapter \ref{ch:autonomous_car}, this thesis also proposes an approach for identifying unseen situations in the environment where simple models are not applicable. Such methodology consists in the unsupervised partitioning a complex distribution into simple components that can be learned later on for further purposes such as classification, prediction or detection of abnormalities. 

We proposed a dynamic incremental self-awareness (SA) model that allow experiences done by an agent to be modeled in a hierarchical fashion, starting from more simple situations to more structured ones. Each situation is learned from subsets of private agent perception data as a model capable to predict normal situations and detect the novel situations in an unsupervised fashion. We show that such model can increase the capability of self-understanding and situational awareness of an autonomous.

Since the identification of novel situations in a given environment based on GANs discriminator networks is the last strategy tackled in this work, advantages of such approach with respect other strategies and possible future paths that lead to a long-life learning self-awareness development and a transfer learning process can be considered as a promising path for further works. In particular, three main levels can be defined for describing the system awareness:
\begin{enumerate}
    \item [\textit{(i)}]\textit{Situation/Self assessment.} It refers to a process to systematically gather, analyze, synthesize and communicate data constantly in a long-life learning framework. this level is performed in order to makes the model capable of making decisions for known situations with the available information, and learning the novel situation by detecting any possible unseen situation constantly.
    \item [\textit{(ii)}]\textit{(Self) Prediction.} An estimation of future states and events in both modules based on learned dynamical models is done. In this stage, it is important to take into account past experiences that facilitate the description of external/internal world dynamics given a particular situation. In the vehicle case, it can be seen how the learning of normal dynamics through a hierarchical structure of GANs is fundamental for prediction of future states given the current situation in which the individual (car) is immersed.
    \item [\textit{(iii)}] \textit{ Novelty Non-stationarity self detection.} A way to measure anomalies or irregularities in observed situations should be defined. Accordingly, it is necessary to determine a deviation measurement from which it is possible to be aware of prediction impressions. In the vehicle case ,the output likelihood scores generated by a bank of GAN discriminators are used as an abnormality/change measurement. Additionally, a fixed threshold value is defined to detect anomalies in the individual's dynamics.     
\end{enumerate}

For future work, in order to complete a full aware system, two main tasks are proposed: \textit{(i)} multimodal awareness system, \textit{(ii)} transferring the agents embedded knowledge to another agent. I explain the importance of each task briefly.

\noindent\textbf{Multimodal awareness system.} We live in a multimodal world, therefore it comes at no surprise that the human brain is tailored for the integration of multi-sensory input. Furthermore, human learning mechanism is base on communication and in its normal state, is \emph{multimodal}~\cite{kress2009}. Inspired by the human brain, the multi sensory data is used in AI for teaching different concepts to computers. However, the proposed awareness system, currently is based on only single modality (e.g., visual data, position), lacks most of the time the robustness and reliability required in several real word applications. To tackle this issue, we propose to take into account the a cross-modalities structure (i.e., internal and external sensory data). That takes place in contexts in the real life problems, where information is combined from various modalities (e.g. vision and language~\cite{martin2016language,shekhar2017vision}), or different domains (e.g. brain and environment~\cite{ghaemmaghami2016sparse,ghaemmaghami2017cross}). Having the perception from the external observer available, the internal body information conveyed by these external observations would be \emph{complementary}.

Based on the limitations of the current model, I propose that future work should focus on including novel \emph{resources} requiring a genuine integration of different sensorial modalities. For generating artificial aware agents, it is proposed to embed the sense of self-awareness (understanding of own states) and situation awareness (comprehension of external surrounding states) in the entity in question. Any of the above-mentioned tasks should necessarily exploit multimodal information to be properly solved. For example, integration of PL and SL resources: it is proposed to use different PL sensory data (as endo-sensors) with the SL information. SL model provides a perception according to an external observer, while the endo-sensors observing the internal parameters of the agent, and allow the system to measure internal parameters of the vehicle, i.e., steering and accelerations exerted by the controlling part of the car. It is clear that both modalities are decisive: If either the SL (e.g., ``spatial position'') or the PL (e.g., visual scene, steering angel) changes, the situation on both model will change accordingly. By doing such procedure, it would be possible consider an full aware autonomous system. The ultimate goal is to develop intelligent awareness systems that are able to concisely summarize their beliefs about the world with diverse predictions, integrate information and beliefs across different components of awareness to extract a holistic view of the world, and explain why they believe what they believe.

\noindent\textbf{Knowledge transfer from agent to agent.} 
Additionally, as future work it is also proposed to transfer the knowledge embedded in an agents' self-awareness model into a totally different body. Transfer learning problem \cite{singh1992transfer} has been studied in many machine learning-related tasks. Transfer learning, in general, is about storing knowledge while solving a problem and then use this knowledge for applying to a different task \cite{tzeng2015simultaneous}, dataset \cite{stamos2015learning} or another agent with a new body \cite{boutsioukis2011transfer}. 

An aware entity should be capable of observing other external entities performing a task and imagine how it can do the same task by using its own body. It should be also capable of representing tasks as a set of interactions with the surroundings, understanding itself as part of a context. 

In our scenario, the same set of rules should apply to other types of entities that intend to replicate observed behaviors of a given individual that accomplishes a given task. Then, when models and relationships between state variables are learned, it is proposed to perform the respective mapping at the level of sensors, actuators, and interfaces in order to make possible a complete awareness transference. Once we obtain a general representation of interactions. This representation is generic enough that can be used to transfer the learned knowledge into other agents. In other words, providing a model based on a set of interactions, not only can be used for training an individual agent but also can provide the means for transferring the knowledge to another agent. Hence, the learning algorithms would be more efficient by incorporating knowledge from previous tasks (done by other agents). However, such methods typically must be provided either a full model of the tasks or an explicit relation mapping one task into the other. We follow the second option, and the mapping would represent a set of interactions. 

This idea can be used to create a methodology in which learned activities by one machine can be transferred into another one that looks at the first one. From that viewpoint, the present work can be potentially used to build a cooperative/coupled framework among machines where they are able to transfer knowledge between them through observation and imitation of accomplished tasks.



\begin{spacing}{0.9}


\bibliographystyle{plainnat} 
\cleardoublepage
\bibliography{references} 



\end{spacing}

\newpage
\thispagestyle{plain} 
\mbox{}
\newpage
\newpage
\thispagestyle{plain} 
\newgeometry{top=0mm, bottom=0mm, left=0mm} 
\includegraphics[]{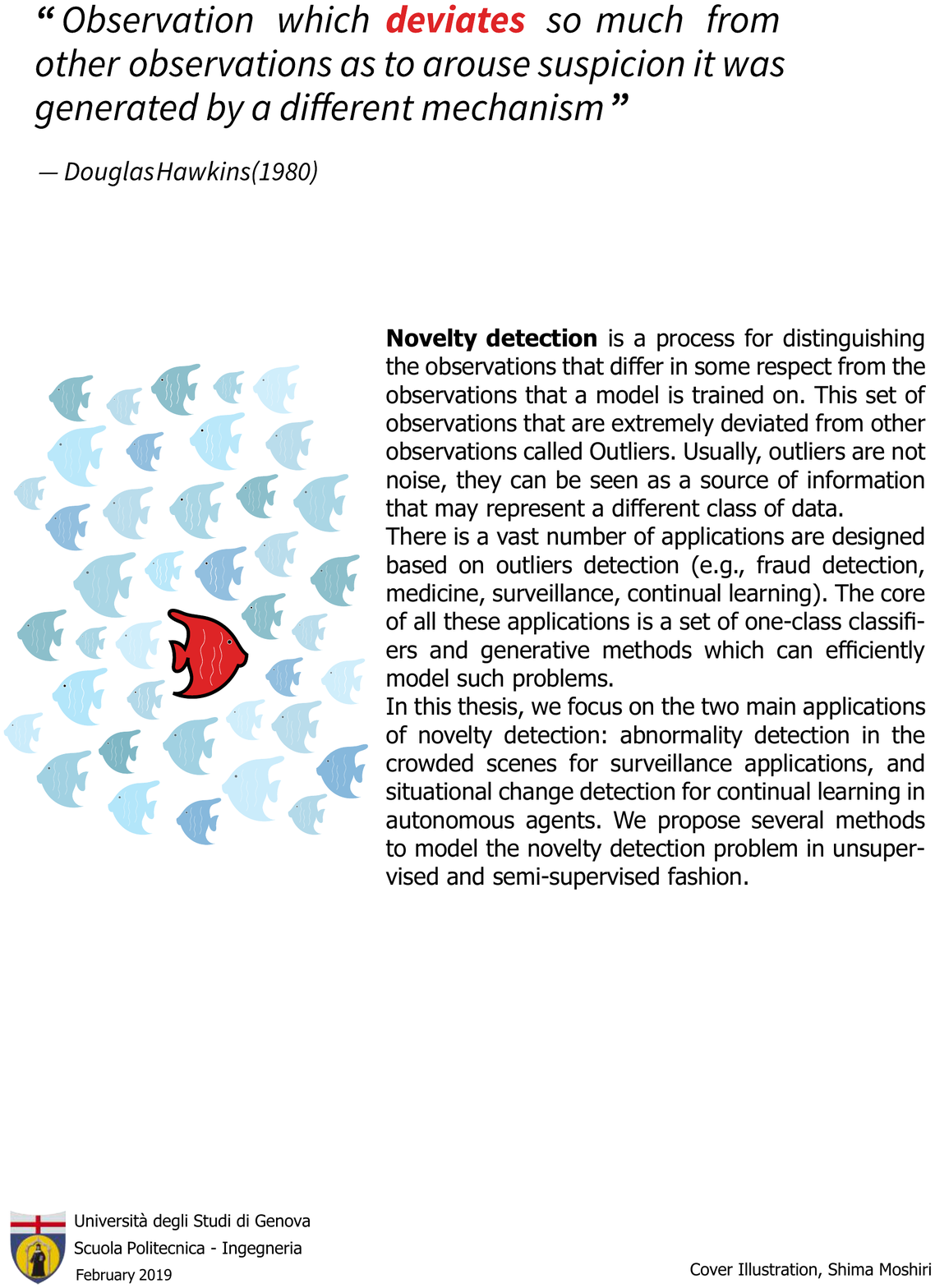}
\restoregeometry




\end{document}